\documentclass[final]{cvpr}

\usepackage{times}
\usepackage{epsfig}
\usepackage{graphicx}
\usepackage{amsmath}
\usepackage{amssymb}
\usepackage{bbm}
\usepackage{enumitem}

\usepackage{tabularx}
\usepackage{threeparttable}
\usepackage{booktabs}
\usepackage{multirow}

\usepackage{microtype}
\usepackage{algorithm}
\usepackage{algorithmicx}
\usepackage[noend]{algpseudocode}

\usepackage[pagebackref=true,breaklinks=true,colorlinks,bookmarks=false]{hyperref}

\def\cvprPaperID{4856} %
\def\confYear{CVPR 2021}

\newcommand{\ourmodel}{TrafficSim} %
\newcommand{\ourmodelshort}{\textsc{TrafficSim}}
\newcommand{\ourdataset}{\textsc{ATG4D}}

\begin{document}

\title{\ourmodel{}: Learning to Simulate Realistic Multi-Agent Behaviors}

\author{Simon Suo$^{1,2}$, 
        Sebastian Regalado$^3$,  
        Sergio Casas$^{1,2}$,  
        Raquel Urtasun$^{1,2}$\\
        $^1$Uber ATG, $^2$University of Toronto, $^3$University of Waterloo \\
        {\tt\small \{suo, sergio, urtasun\}@cs.toronto.edu, sdregala@edu.uwaterloo.ca}
}

\maketitle

\begin{abstract}
    Simulation has the potential to massively scale evaluation of self-driving systems enabling rapid development as well as safe deployment.  
    To close the gap between simulation and the real world,
    we need to simulate realistic multi-agent behaviors.
    Existing simulation environments rely on heuristic-based models that directly encode traffic rules, 
    which cannot capture irregular maneuvers (e.g., nudging, U-turns) and complex interactions (e.g., yielding, merging).
    In contrast, we leverage real-world data to learn directly from human demonstration 
    and thus capture a more diverse set of actor behaviors.
    To this end, we propose \ourmodelshort{}, a multi-agent behavior model for realistic traffic simulation.
    In particular, we leverage an implicit latent variable model to parameterize a joint actor policy that generates socially-consistent plans for all actors in the scene jointly.
    To learn a robust policy amenable for long horizon simulation, we unroll the policy in training and optimize through the fully differentiable simulation across time. 
    Our learning objective incorporates both human demonstrations as well as common sense. 
    We show \ourmodelshort{} generates significantly more realistic and diverse traffic scenarios as compared to a diverse set of baselines.
    Notably, we can exploit trajectories generated by \ourmodelshort{} as effective data augmentation for training better motion planner. 
\end{abstract}

\section{Introduction}

Self-driving has the potential to make drastic impact on our society. One of the key remaining challenges is how to measure progress. 
There are three main approaches for measuring the performance of a self-driving vehicle (SDV):
1) structured testing in the real world,
2) virtual replay of pre-recorded scenarios, and
3) simulation. 
These approaches are complementary, and each has its key advantages and shortcomings. 
The use of a test track enables structured and repeatable evaluation in the physical world.
While this approach is perceptually realistic, testing is often limited to a few scenarios due to the long setup time and high cost for each test. 
Moreover it is hard and often impossible to test safety critical situations, such as unavoidable accidents. 
Virtual replay allows us to leverage diverse scenarios collected from the real world, but it is still limited to what we observe.
Furthermore, since the replay is immutable, actors in the environment do not react when the SDV plan diverges from what happened and the sensor data does not reflect the new viewpoint.
These challenges make simulation a particularly attractive alternative as in a virtual environment we can evaluate against a large number of diverse and dynamic scenarios in a safe, controllable, and cost-efficient manner.

\begin{figure} [t]
    \includegraphics[width=\linewidth]{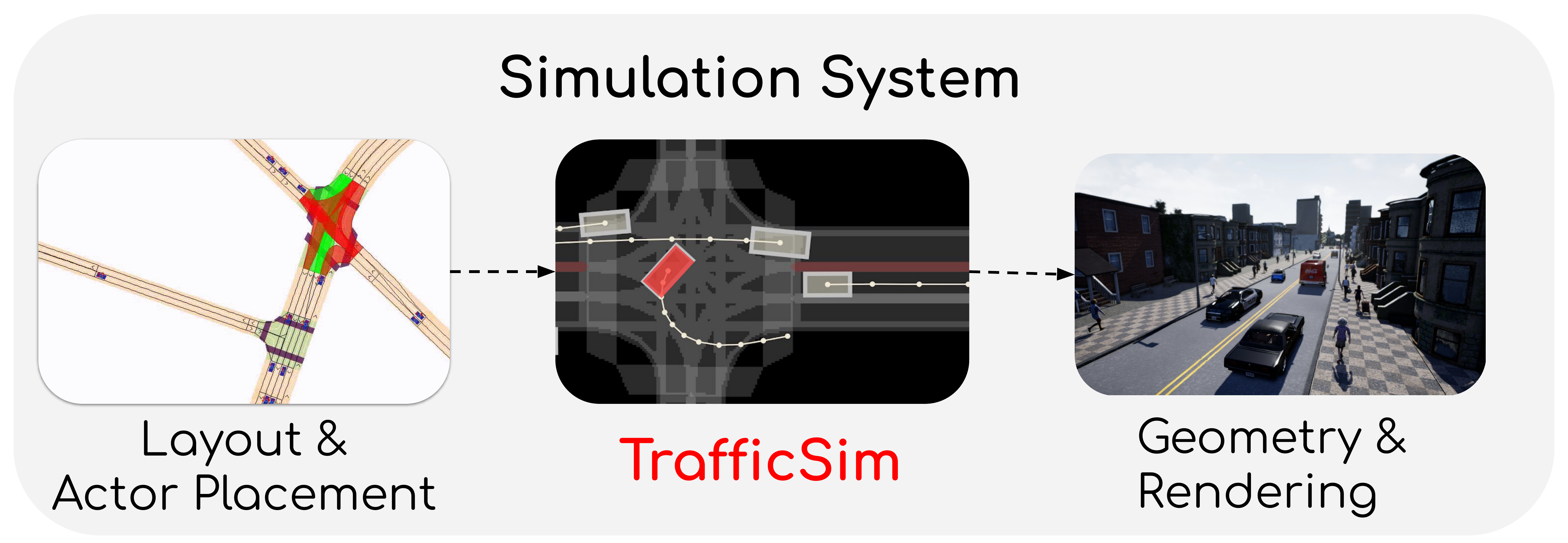}
    \caption{Generating realistic multi-agent behaviors is a key component for simulation}
    \label{fig:teaser}
    \centering
\end{figure}

Simulation systems typically consists of three steps: 1) specifying the scene layout which includes the road topology and actor placement, 2) simulating the motion of dynamic agents forward, and 3) rendering the generated scenario with realistic geometry and appearance, as shown in Figure~\ref{fig:teaser}. 
In this paper, we focus on the second step: generating realistic multi-agent behaviors automatically. 
This can aid simulation design in several important ways: 
it can expedite scenario creation by automating background actors, 
increase scenario coverage by generating variants with emergent behaviors, 
and facilitate interactive scenario design by generating preview of potential interactions.

\begin{figure*}[t]
    \includegraphics[width=\linewidth]{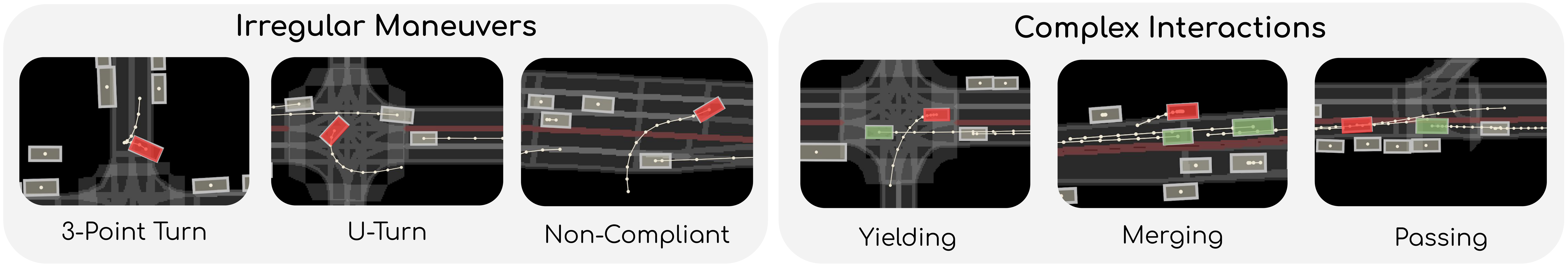}
    \caption{Complex human driving behavior observed in the real world: \textcolor{red}{red} is actor of interest, \textcolor{green}{green} are interacting actors}
    \label{fig:diversity}
    \centering
\end{figure*}

However, bridging the behavior gap between the simulated world and the real world remains an open challenge.
Manually specifying each actor's trajectory is not scalable and results in unrealistic simulations since the actors will not react to the SDV actions.
Heuristic-based models \cite{idm, gipps,kesting2007general} capture basic reactive behavior, but rely on directly encoding traffic rules such as actors follow the road and do not collide.
While this approach generates plausible traffic flow, 
the generated behaviors lack the diversity and nuance of human behaviors and interactions present in real-world urban traffic scenes. For instance, they cannot capture irregular maneuvers that do not follow the lane graph such as U-turns, or complex multi-agent interplays such as nudging past a vehicle stopped in a driving lane, or negotiations at an unprotected left turn.
In contrast, learning-based approaches \cite{casas2020implicit,tang2019multiple, precog} are flexible and can capture a diverse set of behaviors. 
However, they often lack common sense and are generally brittle to distributional shift. 
Furthermore, 
they can also be computationally expensive if not optimized for simulating large numbers of actors over long horizon.

To tackle these challenges, we present \ourmodelshort{}, a multi-agent behavior model for traffic simulation.
We leverage recent advances in motion forecasting, and formulate the joint actor policy with an implicit latent variable model \cite{casas2020implicit},
which can generate multiple scene-consistent samples of actor trajectories in parallel.
Importantly, to learn a robust policy amenable for traffic simulation over long time horizon, we unroll the policy in training
and optimize  through our fully differentiable simulation. 
Furthermore, we propose a time-adaptive multi-task loss that balances between learning from demonstration and common sense at each timestep of the simulation.
Our experiments show that \ourmodelshort{} is able to simulate traffic scenarios that remain realistic over long time horizon, with minimal collisions and traffic rule violations.
In particular, it achieves the lowest scenario reconstruction error in comparison to a diverse set of baselines including heuristic, motion forecasting, and imitation learning models.
We also show that we can train better motion planners by exploiting trajectories generated by \ourmodelshort{}. %
Lastly, we show experiments in trading off simulation quality and computation.
In particular, we can achieve up to 4x speedup with multi-step updates, or further reduce collisions with additional optimization at simulation-time.

\section{Related Work}

\paragraph{Simulation Environments:}
Simulating traffic actors is a ubiquitous task with wide ranging applications in
transportation research, video games, and now training and evaluating self-driving vehicles \cite{traffic-sim-survey}.
Microscopic traffic simulators \cite{sumo} employ heuristic-based models \cite{idm, gipps,kesting2007general} to simulate traffic flow.
These models capture accurate high-level traffic characteristic by directly encoding traffic rules (e.g., staying in lane, avoiding collision).  
However due to rigid assumptions, they are not realistic at the street level even after calibrating with real world data \cite{calibrate-car-following, traffic-sim-agents}. 
In particular, they can not capture irregular maneuvers (e.g., nudging, U-turns) and complex multi-agent interaction (e.g., yielding, merging) that occur in the real world, shown in Figure~\ref{fig:diversity}. 
Progress in game engines greatly advanced the realism of physical simulations. 
Researchers have leveraged racing games \cite{Wymann15torcs, gta-v} and developed 
higher fidelity simulators \cite{dosovitskiy2017carla, Best2018AutonoViSimAV}
to train and evaluate self driving systems.
Real world data is leveraged by \cite{manivasagam2020lidarsim} for realistic sensor simulation.
However, actor behaviors are still very simplistic: 
simulated actors in \cite{dosovitskiy2017carla} are governed by a basic heuristic-based controller that can only follow the lane while respecting traffic rules and avoiding head-on collisions. 
This is insufficient to evaluate SDVs, 
since the one of the main challenge in self-driving is
accurately anticipating and safely planning around diverse and often irregular human maneuvers, 
Thus, this motivates us to learn from real world data to bridge this gap.

\begin{figure*}[t]
    \centering
    \includegraphics[width=\linewidth]{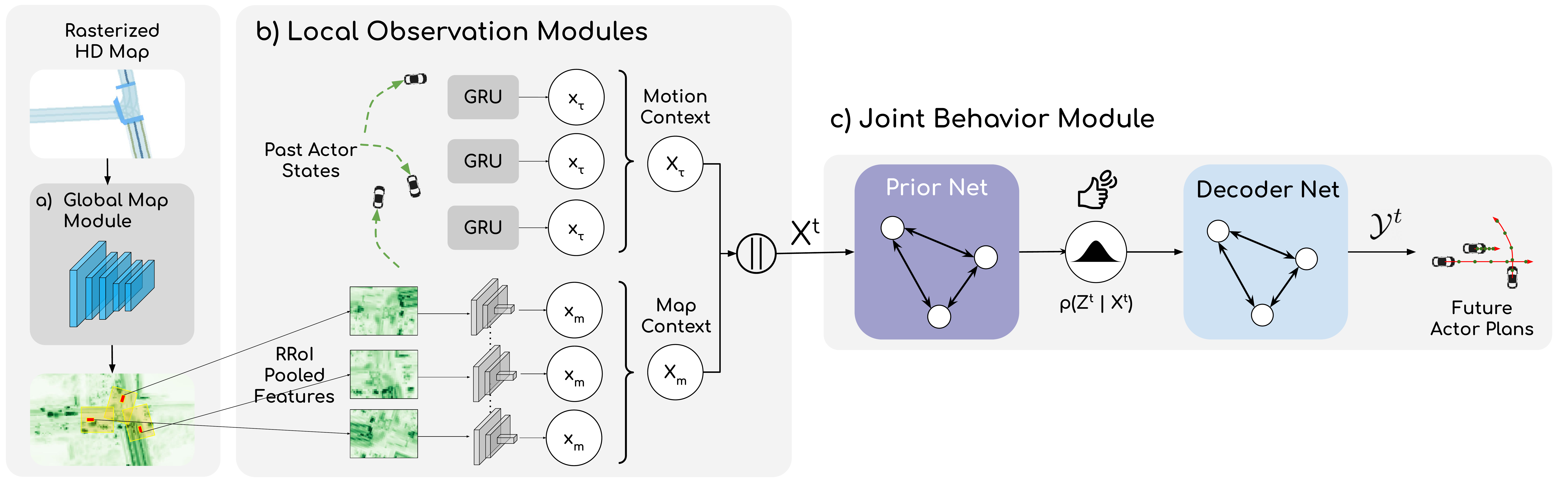}
    \caption{\ourmodelshort{} architecture: global map module (a) is run once per map for repeated simulation runs. At each timestep, local observation module (b) extracts motion and map features, then joint behavior module (c) produces a multi-agent plan.}
    \label{fig:architecture}
    \centering
\end{figure*}

\paragraph{Motion Forecasting:}
Motion forecasting is the task of predicting actor's future motion based on past context, which also requires accurate actor behavior modelling.
Traditional approaches track an object  and propagate its state to predict its future motion (e.g., Unscented Kalman filter \cite{ukf} with kinematic bicycle model \cite{kong2015kinematic}).
More recently, %
deep-learning based models have been developed to capture increasingly more complex behaviors. 
\cite{djuric2018motion} rasterizes an HD map and a history of actor bounding boxes to leverage a CNN to forecast actor behavior.
Since the future is inherently uncertain, 
\cite{cui2018multimodal, chai2019multipath} output multiple trajectories per actor.
\cite{casas2020importance} shows that explicitly incorporating prior knowledge  help learn better predictive distributions.
Several works \cite{alahi2016social, precog, tang2019multiple, casas2020implicit,li2020endtoend} go beyond actor-independent modeling and explicitly reason about interaction among actors as the future unfolds. 
To characterize the behavior of multiple actors jointly, \cite{alahi2016social, precog, tang2019multiple} leverages auto-regressive generation with social mechanisms.
In contrast, \cite{casas2020implicit} employs spatially-aware graph neural networks to model interaction in the latent space, thereby capturing longer range interactions and avoiding slow sequence sampling. 
Importantly, these models can generate multiple socially-consistent samples, where each sample constitute a realistic traffic scenario, thus modeling complex multi-agent dynamics beyond simple pairwise interaction.
Thus, they are particularly amenable for simulating actor behaviors in virtual traffic. 
However, they cannot be directly used for simulation over long time horizon, since they are brittle to distributional shift and cannot recover from compounding error.

\paragraph{Imitation Learning:}
Imitation learning (IL) aims to learn control policies from demonstration.
Behavior cloning \cite{alvinn} treats state-action pairs as i.i.d examples to leverage supervised learning, but suffers from distributional shift due to compounding error \cite{dagger}. 
Intuitively, during offline open-loop training, the policy only observes ground truth past states, but when unrolled in  closed-loop at test time, it encounters novel states induced by its sequence of suboptimal past decisions and fails to recover.
Many approaches have been proposed to mitigate the inevitable deviation from the observed distribution, but each has its drawbacks.  
Online supervision \cite{dagger} require access to an interactive expert 
that captures the full distribution over human driving behaviors. 
Data augmentation \cite{chauffeur-net, end-to-end} depends on manually designed out-of-distribution states and corresponding desired actions, 
which is often brittle and adds bias. 
Uncertainty-based regularization \cite{Brantley2020Disagreement-Regularized,henaff2019modelpredictive} leverages predictive uncertainty to avoid deviating from the observed distribution, but can be challenging and computationally expensive to estimate accurately.
Adversarial IL approaches \cite{gail, ngsim-gail} jointly learn the policy with a discriminator.
However, they are empirically difficult to train (requiring careful reward augmentation \cite{rail} and curriculum design \cite{horizon-gail}), and are generally limited to simulating a small number of actors \cite{ps-gail} in a specific map topology (e.g., NGSIM).
In contrast, we aim to learn a joint actor policy that generalizes to diverse set of urban streets and simulates the behavior for large number of actors in parallel. 
Furthermore, while IL methods typically assume non-differentiable environment, we directly model differentiable state transitions instead. This allow us to directly optimize with back-propagation through the simulation.

\section{Learning  Multi-Agent Traffic Behaviors} %
In this section, we describe our approach for learning realistic multi-agent  behaviors for traffic simulation.
Given a high-definition map $\mathcal{M}$, traffic control $\mathcal{C}$, 
and initial dynamic states of $N$ traffic actors, our goal is to simulate their motion forward.
We use $Y^t = \{y_1^t, y_2^t, ..., y_N^t\}$ to denote a collection of $N$ actor states at time $t$.
More precisely, each actor state is parameterized as a bounding box $y_i^t = (b_x, b_y, b_w, b_h, b_\theta)$ with 2D position, width, height and heading.
In the following,  we first describe how to extract rich context from the simulation environment.
Then, we explain our joint actor policy that explicitly reasons about interaction and generates socially consistent plans. 
Lastly, we present a learning framework that leverages back-propagation through the differentiable simulation, and balances imitation and common sense. 
We illustrate the full architecture in Figure~\ref{fig:architecture}.

\subsection{Extracting  Rich Context from the  Environment}
\label{sec:observation}
Accurately modelling actor behaviors requires rich scene context from past motion and map topology.
Towards this goal, we propose a differentiable observation module that %
takes as input the past actor states $Y^{:t}$, traffic control $\mathcal{C}$ and HD map $\mathcal{M}$, and process them in two stages.
First, we use a CNN-based perception backbone network  inspired by \cite{yang2018pixor, casas2018intentnet} to extract rich geometrical features $\tilde{\mathcal{M}}$ from the raster map $\mathcal{M}$, shown in Figure~\ref{fig:architecture} (a).
Since we are only interested in the region of interest defined by $\mathcal{M}$ and these spatial features are static  across time, we can process each map once, and cached them for repeated simulation runs.

Then we leverage a local observation module with two components: a \emph{map feature extractor} and a \emph{past trajectory encoder} shown in Figure~\ref{fig:architecture} (b). 
Unlike the global map module, these feature extractors are run once per simulation step, and are thus designed to be lightweight. 
To extract local context $X_m^t$ around each actor, we apply Rotated Region of Interest Align \cite{ma2018arbitrary} to the map features $\tilde{\mathcal{M}}$ pre-processed by the map backbone.
To encode the past trajectories of each actor in the scene, we employ a 4-layer GRU with 128 hidden states, yielding $X_\tau^t$.
Finally, we concatenate the map and past trajectory features to form the scene context $X^t = [X_m^t, X_\tau^t]$, which we use as input to the joint actor policy. 

\subsection{Implicit Latent Variable Model for Multi-Agent Reasoning}
\label{sec:joint_actor_policy}
    We use a joint actor policy to explicitly reasons about multi-agent interactions, shown in Figure~\ref{fig:architecture} (c).
    This allows us to sample multiple socially consistent plans for all actors in the scene in parallel.
    Concretely, we aim to characterize the joint distribution over actors' future states $\mathcal{Y}^t = \{Y^{t+1}, Y^{t+2}, ..., Y^{t+T_{\text{plan}}}\}$. 
    This formulation allows us to leverage supervision over the full planning horizon $T_{\text{plan}}$ to learn better long-term interaction.
    To simplify notation, we use $\mathcal{Y}^t$ in subsequent discussions. 

    It is difficult to represent this joint distribution over actors in an explicit form as there is uncertainty over each actor's goal 
    and complex interactions between actors as the future unfolds. 
    A natural solution is to implicitly characterize this distribution via a latent variable model \cite{sohn2015learning,casas2020implicit}:
    \begin{align}
        P(\mathcal{Y}^t | X^t) = \int_{Z} P(\mathcal{Y}^t|X^t, Z^t) P(Z^t | X^t)
    \end{align}
    Following \cite{casas2020implicit}, we use a deterministic decoder $\mathcal{Y}^t = f(X^t,Z^t)$ to encourage the scene latent $Z$ to capture all stochasticity and avoid factorizing $P(\mathcal{Y}^t| X^t, Z^t)$ across time.
    This allow us to generate $K$ scene-consistent samples of actor plans efficiently in one stage of parallel sampling, by first drawing latent samples $Z^t_{(k)} \sim P(Z^t | X^t)$, 
    and then decoding actor plans $\mathcal{Y}^{t}_{(k)} = f(Z^t_{(k)}, X^t)$. 
    Furthermore, we approximate the posterior latent distribution $q(Z^{t}, | X^t , \mathcal{Y}^t)$ to leverage variational inference \cite{kingma2013auto, sohn2015learning} for learning. 
    Intuitively, it learns to map ground truth future $\mathcal{Y}^{t}_{GT}$ to the scene latent space for best reconstruction.

    We leverage the graph neural network (GNN) based scene interaction module introduced by \cite{casas2019spatiallyaware} to parameterize the prior network $p_{\gamma}(Z^t | X^t)$, posterior network $q_\phi(Z^{t}, | X^t , \mathcal{Y}^t)$, and the deterministic decoder $\mathcal{Y}^t = f_{\theta}(X^t, Z^t)$, for encoding to and decoding from the scene-level latent variable $Z^t$. 
    By propagating messages across a fully connected interaction graph with actors as nodes, the latent space learns to capture not only individual actor goals and style, but also multi-agent interactions.
    More concretely, we partition the latent space to learn a distributed representation $Z^t = \{z_1, z_2, ..., z_N\}$ of the scene, where  $z_n$ is spatially anchored to actor $n$ and captures unobserved dynamics most relevant to that actor.
    This choice enables effective relational reasoning across a large and variable number of actors and diverse map topologies
    (i.e., to deal with the complexity of urban traffic).
    Additional implementation details can be found in the supplementary.

    \begin{figure}[t]
        \includegraphics[width=\linewidth]{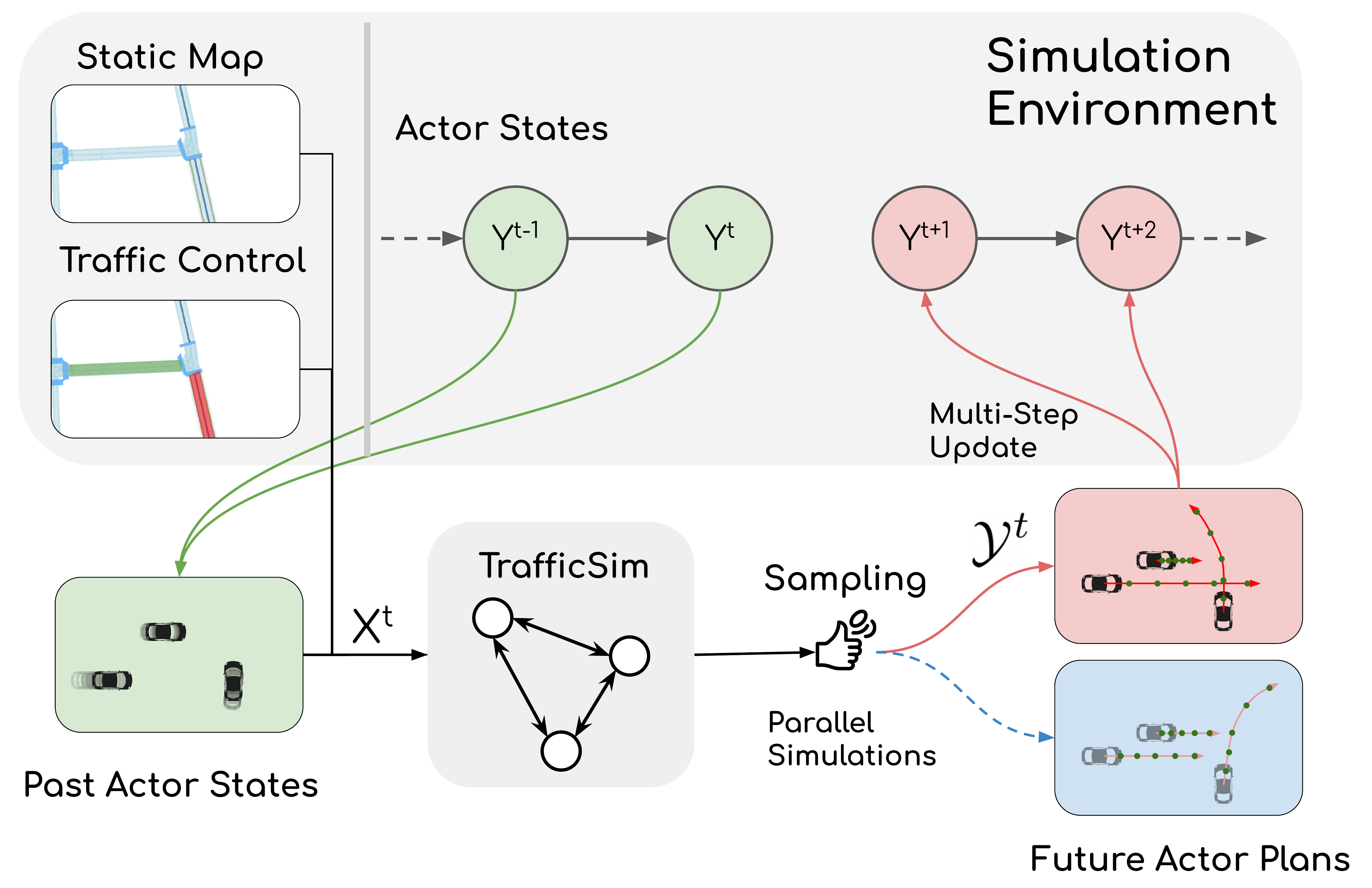}
        \caption{\ourmodelshort{} models all actors jointly to simulate realistic traffic scenarios through time. We can sample at each timestep to obtain parallel simulations.}
        \label{fig:model}
        \centering
    \end{figure}

    \begin{figure*}[t]
        \includegraphics[width=\linewidth]{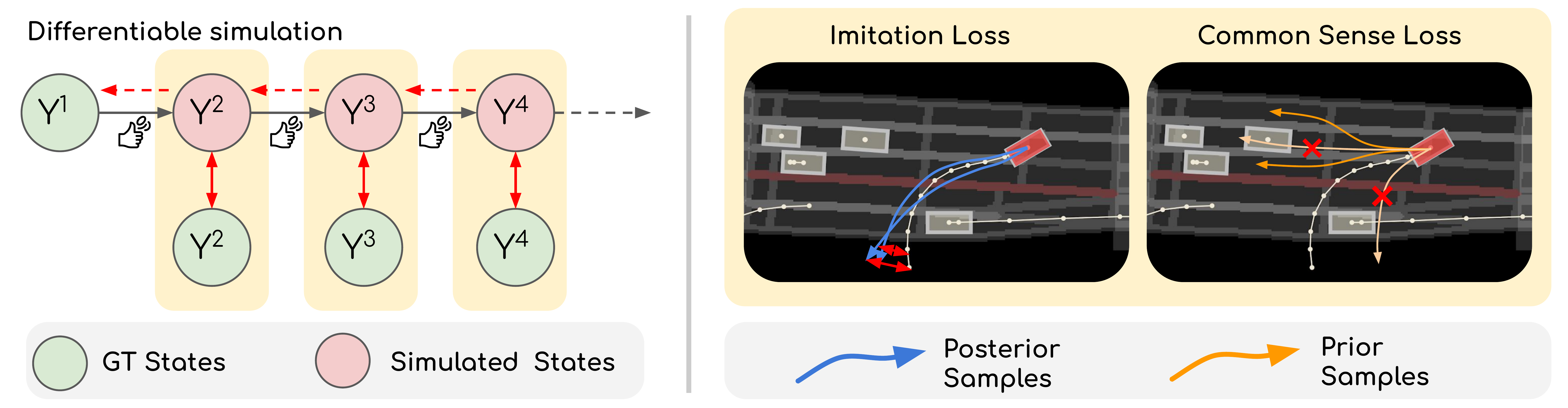}
        \caption{We optimize our policy with back-propagation through the differentiable simulation (\textbf{left}), and apply imitation and common sense loss at each simulated state (\textbf{right}).}
        \label{fig:learning}
        \centering
    \end{figure*}

    \subsection{Simulating Traffic Scenarios}
    \label{sec:traffic_scenario}
    We model each traffic scenario as a sequential process where traffic actors interact and plan their behaviors at each timestep.
        Leveraging the differentiable observation module and joint actor policy, we can generate traffic scenarios by starting with an initial history of the actors $Y^{-H:0}$ and simulating their motion forward for T steps.        
        Concretely, at each timestep t, we first extract scene context $X^t$, then sample actor plans $\mathcal{Y}^t \sim P_{\theta, \gamma}(\mathcal{Y}^t | X^t)$ from our joint actor policy, shown in Figure~\ref{fig:model}.
        Since our policy produces a $T_{\text{plan}}$ step plan of the future $\mathcal{Y}^t = \{Y^{t+1}, ..., Y^{t+T_{\text{plan}}}\}$, we can either use the first timestep of the joint plan $Y^{t+1}$ to update the simulation environment at the highest frequency, 
        or take multiple steps $\{Y^{t+1}, ..., Y^{t+\kappa}\}$ for faster simulation with minimal loss in simulation quality:
        \begin{align}
            P(Y^{1:T} | Y^{-H:0}, \mathcal{M}, \mathcal{C}) = \prod_{t\in \mathcal{T}} P(Y^{t+1:t+\kappa} | X^t)
        \end{align}
        We provide further discussion on trading off simulation quality and computation in Section \ref{sec:experiment}.

 \subsection{Learning from Examples and Common Sense}
        In this section, we describe our approach for learning multi-agent behaviors by leveraging large-scale datasets of human driving behaviors.
        We train by unrolling our policy (i.e., in closed-loop)
        and exploiting our fully differentiable formulation to directly optimize with back-propagation through the simulation over time.
        Furthermore, we propose a multi-task loss that balances between learning from demonstration and injecting common sense.

        \paragraph{Backpropagation through Differentiable Simulation:}
        Learning from demonstration via behavior cloning yields good open-loop behaviors (i.e., accurate $P(\mathcal{Y}^t|X^t)$ when $X^t$ comes from the observation distribution), but can suffer from compounding error in closed-loop execution \cite{dagger} (i.e., when $X^t$ is induced by the policy).
        To bridge this gap, we propose to unroll the policy for closed-loop training and compute the loss $\mathcal{L}^t$ at each simulation step $t$, as shown in Figure~\ref{fig:learning} (left).
        Since we model state transitions in a fully differentiable manner, we can directly optimize the total loss  
        with back-propagation through the  simulation across time.
        In particular, the gradient is back-propagated through action sampled from the policy at each timestep via reparameterization.
        This gives a direct signal for how current decision influences future states.
        
        \paragraph{Augmenting Imitation with Common Sense:}
        Pure imitation suffers from
        poor supervision when a stochastic policy inevitably deviates from the observed realization of the scenario.
        Furthermore, inherent bias in the collected data (e.g., lack of safety critical scenarios) means pure imitation can not reason about the danger of collision.
        Thus, we augment imitation with an auxiliary common sense objective, and use an time-adaptive multi-task loss to balance the supervision.
        Through the simulation horizon, we anneal $\lambda(t)$ to favour supervision from common sense over imitation.
        \begin{align}
            \mathcal{L} = \sum_t \lambda(t) \mathcal{L}^t_{\text{imitation}} + (1- \lambda(t)) \mathcal{L}^t_{\text{collision}}
        \end{align}
 
        Furthermore, we unroll the model in two distinct segments during training. 
        First for $t \le T_{\text{label}}$, we unroll with posterior samples from the model $\mathcal{Y}_{\text{post}}^t = f(X^t, Z^t_{\text{post}})$ where $Z^t_{post}$ is conditioned on ground truth future $\mathcal{Y}^t_{\text{GT}}$.
        Subsequently for $T_{\text{label}} < t \le T$, we use $\mathcal{Y}_{\text{prior}}^t  = f(X^t, Z^t_{\text{prior}})$ instead.
        Intuitively, posterior samples reconstruct the ground truth future, whereas prior samples cover diverse possible futures.
        We now describe both objectives in details, also shown in Figure~\ref{fig:learning} (right).

        \paragraph{Imitation Objective:}
        To learn from demonstrations, 
        we adapt the variational learning objective of the CVAE framework \cite{sohn2015learning} and optimize the evidence-based lower bound (ELBO) of the log likelihood  $\log P(\mathcal{Y}^t| X^t)$ at each timestep $t \le T_{\text{label}}$. 
        Concretely, the imitation loss consists of a reconstruction component and a KL divergence component:
        \begin{align}
        & \mathcal{L}^{t}_{\text{imitation}} (\mathcal{Y}^t_{\text{posterior}}, \mathcal{Y}^t_{\text{GT}})  = \mathcal{L}^{t}_{\text{recon}} + \beta \cdot \mathcal{L}^{t}_{\text{KL}} \\
        & L_{recon}^t = \sum_a^N \sum_{\tau = t + 1}^{t+P} L_{\delta}(y_a^{\tau} - y_{a, GT}^{\tau})\\
        & L_{KL}^t =  \text{KL}\left(q_{\phi}\left(Z^t | X^t, \mathcal{Y}_{GT}^t \right)|| p_{\gamma}\left(Z^t | X^t \right ) \right)
        \end{align}
        We use Huber loss $L_{\delta}$ for reconstruction and reweight the KL term with $\beta$ as proposed by \cite{higgins2017beta}.
       
        \newcolumntype{s}{>{\centering\arraybackslash}X}

\begin{table*}[t]
    \centering
	\begin{threeparttable}
        \begin{tabularx}{\linewidth}{
                       l
                        >{\centering\arraybackslash} l@{\hskip .5cm}|  %
                        s s s s s s s
                        }
		    \toprule
                \multirow{2}{*}{Model} & {} &
                SCR$_{12s}$ (\%) & TRV$_{12s}$ (\%) & 
                minSFDE (m) & minSADE (m) & 
                meanSFDE (m) & meanSADE (m) & 
                MASD$_{12s}$ (m)\\ 
            \midrule
                \multirow{1}{1.5cm}{Heuristic} 
                & IDM \cite{idm}   				        & 1.19          & \textbf{0.25} & 4.97          & 3.03          & 5.39          & 3.48          & 4.01 \\ 
            \midrule
                \multirow{3}{1.5cm}{Motion Forecasting} 
                & MTP \cite{cui2018multimodal}  	    & 11.00         & 9.67          & 2.19          & 1.47          & 2.77          & 2.19          & 7.15 \\    	     
                & ESP \cite{precog} 	    	        & 4.08          & 4.79          & 3.42          & 1.56          & 3.52          & 1.60          & 0.29 \\   
                & ILVM \cite{casas2020implicit}         & 2.90          & 4.37          & 2.56          & 1.33          & 2.92          & 1.50          & 1.17 \\ 			 
            \midrule
                \multirow{2}{1.5cm}{Imitation Learning}
                & AdversarialIL                         & 10.05         & 8.34          & 2.89          & 1.19          & 3.87          & 1.51          & 4.89\\      
                & DataAug                               & 3.78          & 8.23          & 2.04          & 1.22          & 2.62          & 1.56          & 2.29\\      
            \midrule
                Ours 
                & \ourmodelshort{}                      & \textbf{0.50} & 2.77          & \textbf{1.13} & \textbf{0.57} & \textbf{1.75} & \textbf{0.85} & 2.50\\ 
	        \bottomrule
		\end{tabularx}
    \end{threeparttable}
    \caption{\textbf{[\ourdataset{}]} Comparison against existing approaches ($S=15$ samples, $T=12$ seconds, $T_{\text{label}}=8$ seconds)}
	\label{table:atg4d_main}
\end{table*}

        \paragraph{Common Sense Objective:}
        We use a pair-wise collision loss
        and design a efficient differentiable relaxation to ease optimization.
        In particular, we approximate each vehicle with 5 circles, and compute L2 distance between centroids of the closest circles of each pair of actors.
        We apply this loss on prior samples from the model $\mathcal{Y}^t_{prior}$ to directly regularize $P(\mathcal{Y}^t |X^t)$.
        More concretely, we define the loss as follows:
        \begin{align}
            \mathcal{L}^t_{\text{collision}}(\mathcal{Y}^t_{\text{prior}}) = \frac 1 {N^2} \sum_{i\neq j} \max(1, \sum_{\tau=t+1}^{t+P}\mathcal{L}_{\text{pair}}(y_i^\tau, y_j^\tau))\\
            \mathcal{L}_{\text{pair}}(y_i^\tau, y_j^\tau) = 
            \begin{cases}
                1 - \frac d {r_i + r_j}, & \text{if } d \le r_i + r_j\\
                0,                       & \text{otherwise}
            \end{cases}
        \end{align}

\section{Experimental Evaluation}
\label{sec:experiment}
In this section, we first describe the simulation setup and propose a suite of metrics for measuring  simulation quality.
We show our approach generates more realistic and diverse traffic scenarios as compared to a diverse set of baselines.
Notably, training an imitation-based motion planner on synthetic data generated by \ourmodelshort{} outperforms in planning L2 as compared to using same amount of real data.
This shows there's minimal behavior gap between \ourmodelshort{} and the real world.
Lastly, we study how to tradeoff between simulation quality and computation.

\paragraph{Dataset:} We benchmark our approach on a large-scale self driving dataset \ourdataset{}, which %
contains more than one million frames collected over several cities in North America with a 64-beam, roof-mounted LiDAR. 
Our labels are very precise 3D bounding box tracks. %
There are 6500 snippets in total, each 25 seconds long. 
In each city, we have access to high definition maps capturing the geometry and the topology of each road network.
We consider a rectangular region of interest centered around the self-driving vehicle that spans 140 meters along the direction of its heading and 80 meters across.
The region is fixed across time for each simulation.

\paragraph{Simulation Setup:} 
In this work, we use real traffic states from \ourdataset{} as initialization for the simulations.
This give us realistic actor placement and dynamic state,
thus controlling for domain gap that might arise from initialization.
We subdivide full snippets into 11s chunks, using the first 3s as the initial states $Y^{-H:0}$, 
and the subsequent $T_{\text{label}} = 8s$ as expert demonstration for training.
We run the simulation forward for $T=12$ seconds for both training and evaluation.
We use $\delta_t = 0.5s$ as the duration for a simulation tick (i.e., simulation frequency of $2Hz$).
We use observed traffic light states from the log snippets for simulation.

\paragraph{Baselines:} We use a wide variety of baselines. 
The Intelligent Driver Model (IDM) \cite{idm} is a heuristic car-following model that explicitly encode traffic rules.
  We adapt three state-of-the-art motion forecasting models for traffic simulation. 
    MTP \cite{cui2018multimodal} models multi-modal futures, but assume independence across actors.
    ESP \cite{precog} models interaction at the output level, via social auto-regressive formulation.
    ILVM \cite{casas2020implicit} models interaction using a scene-level latent variable model. 
    Finally, we consider imitation learning techniques that have been applied to learning driving behaviors.
    Following \cite{end-to-end, codevilla2018endtoend, chauffeur-net}, DataAug  adds perturbed trajectories to help the policy learn to recover from mistakes.
    Inspired by \cite{gail, ps-gail, horizon-gail}, AdversarialIL learns a discriminator as supervision for the policy.
    We defer implementation details to the supplementary.

\begin{figure*}[t]
    \centering
    \begin{tabular} {@{}c@{\hspace{.1em}}c@{\hspace{.1em}}c@{\hspace{.1em}}c@{\hspace{.1em}}c}
        {} & \textbf{T=0s} & \textbf{4s} & \textbf{8s} & \textbf{12s} \\
        \rotatebox[origin=c]{90}{\textbf{scenario 1}} &
        \raisebox{-0.5\height}{\includegraphics[width=0.248\linewidth]{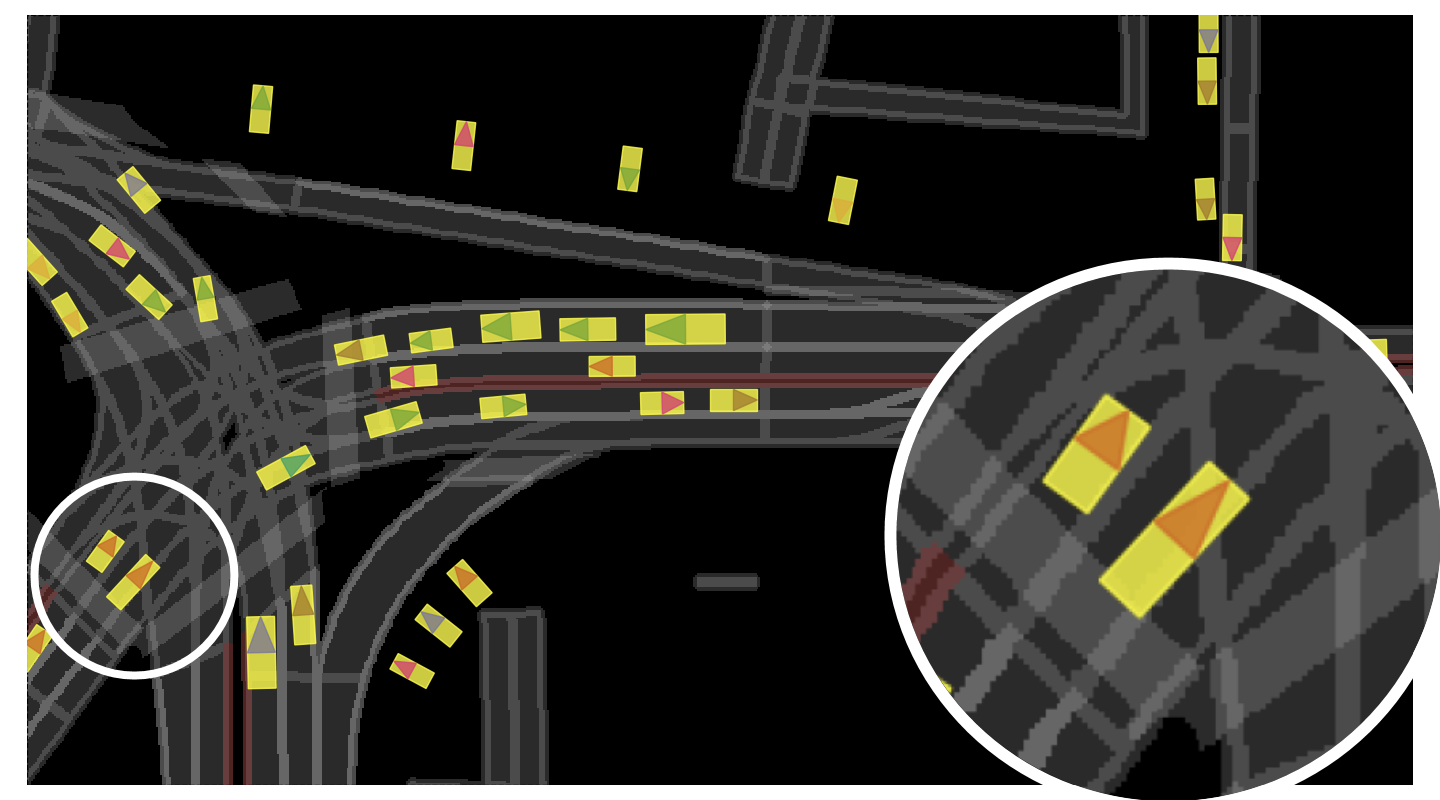}} &
        \raisebox{-0.5\height}{\includegraphics[width=0.248\linewidth]{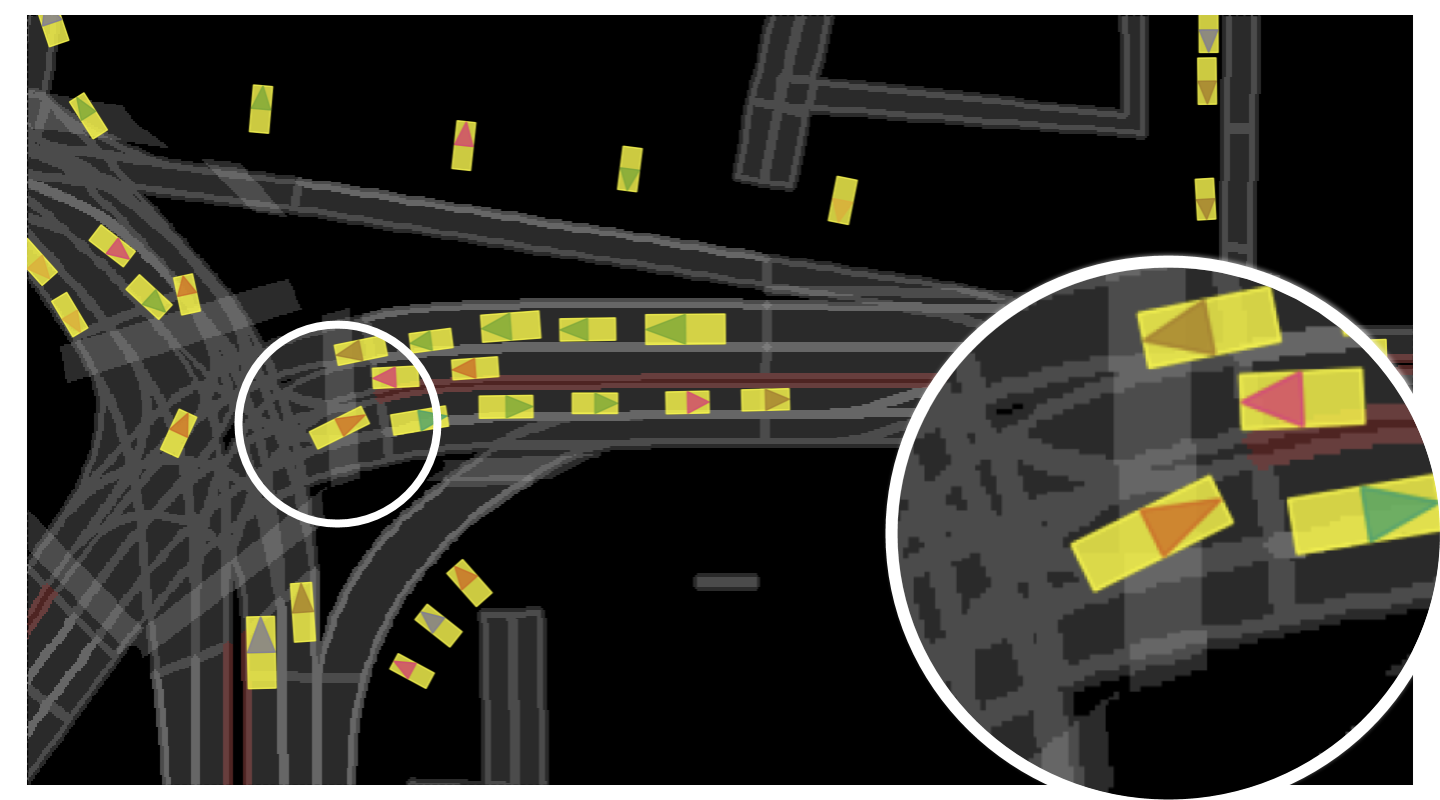}} &
        \raisebox{-0.5\height}{\includegraphics[width=0.248\linewidth]{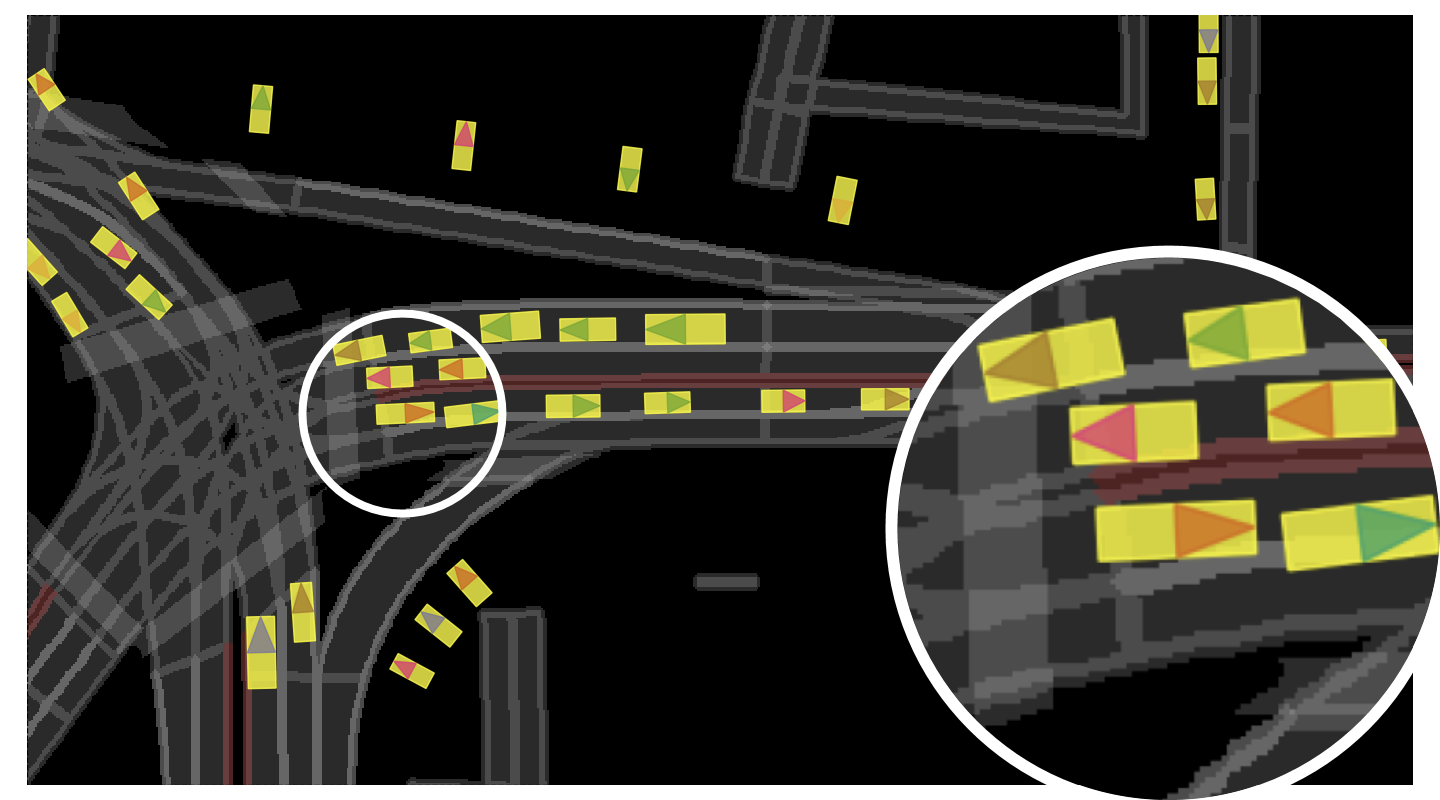}} &
        \raisebox{-0.5\height}{\includegraphics[width=0.248\linewidth]{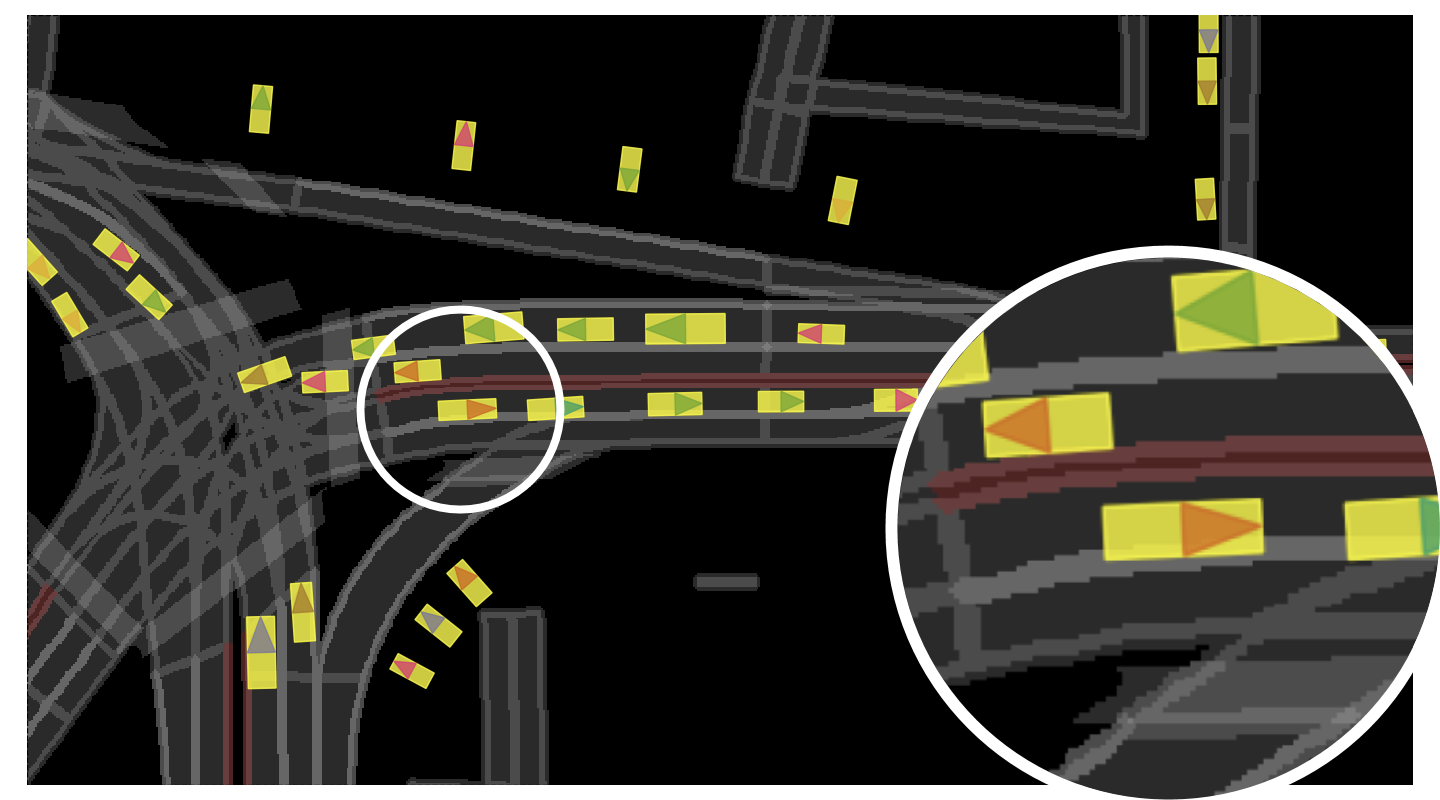}} \vspace{.1em} \\
        \rotatebox[origin=c]{90}{\textbf{Scenario 2}} &
        \raisebox{-0.5\height}{\includegraphics[width=0.248\linewidth]{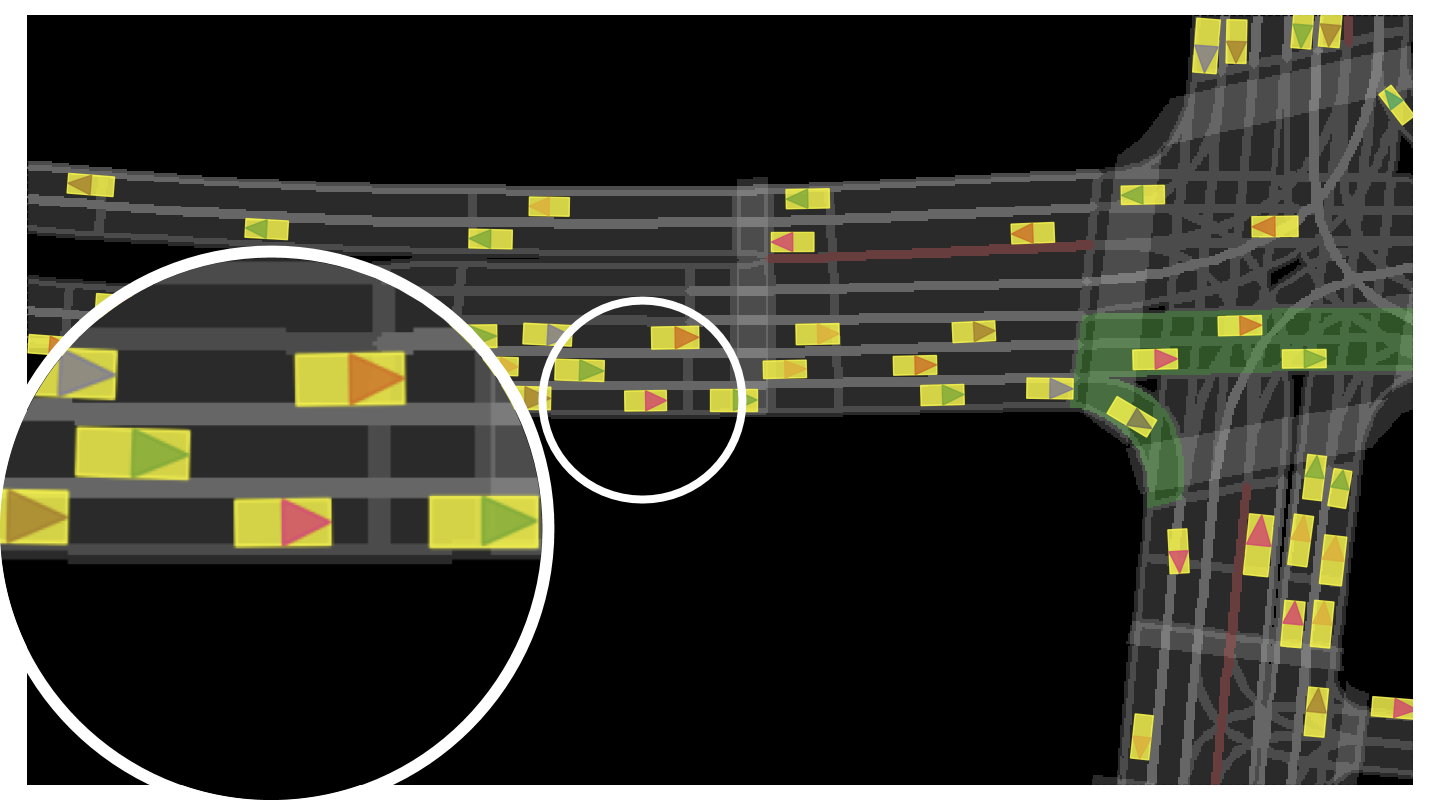}} &
        \raisebox{-0.5\height}{\includegraphics[width=0.248\linewidth]{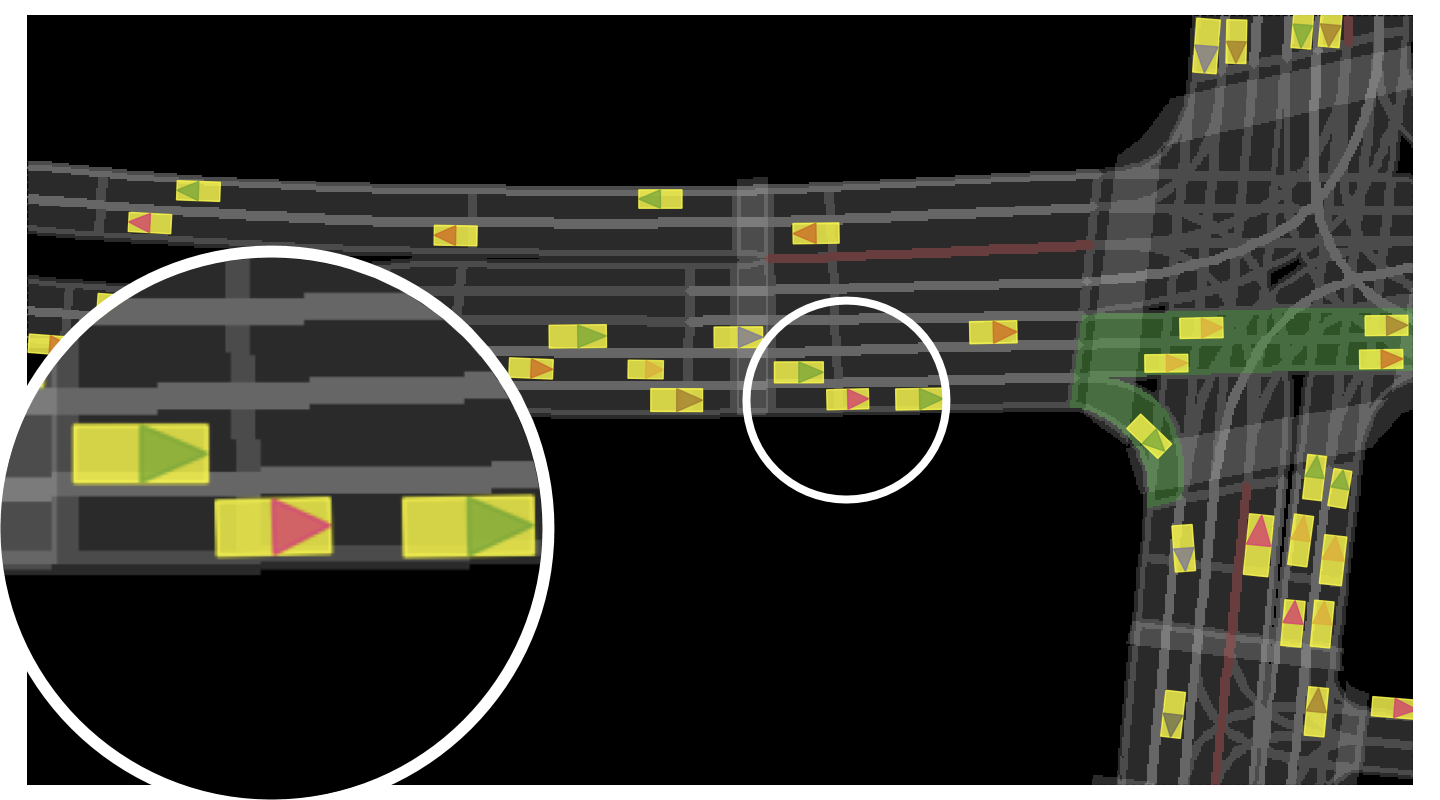}} &
        \raisebox{-0.5\height}{\includegraphics[width=0.248\linewidth]{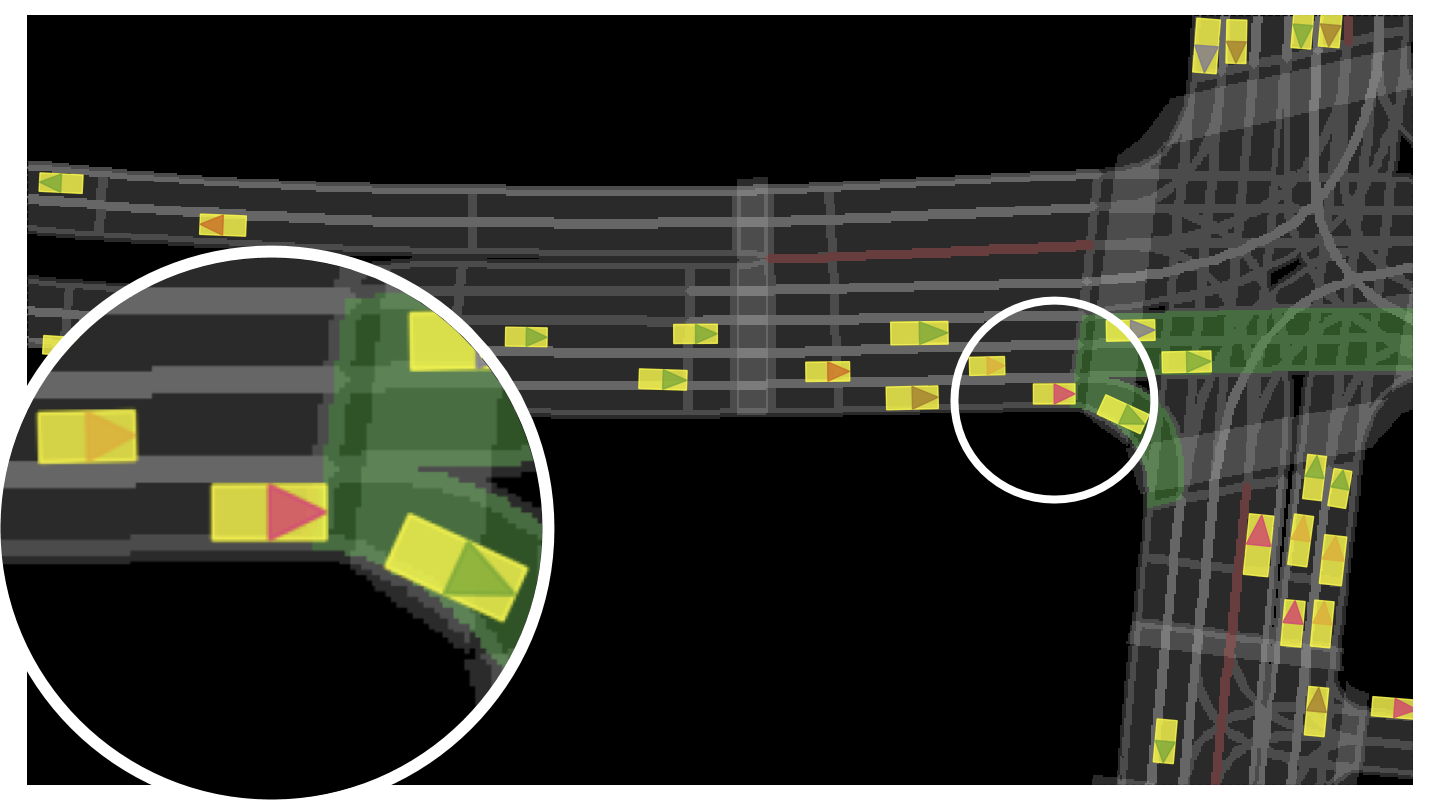}} &
        \raisebox{-0.5\height}{\includegraphics[width=0.248\linewidth]{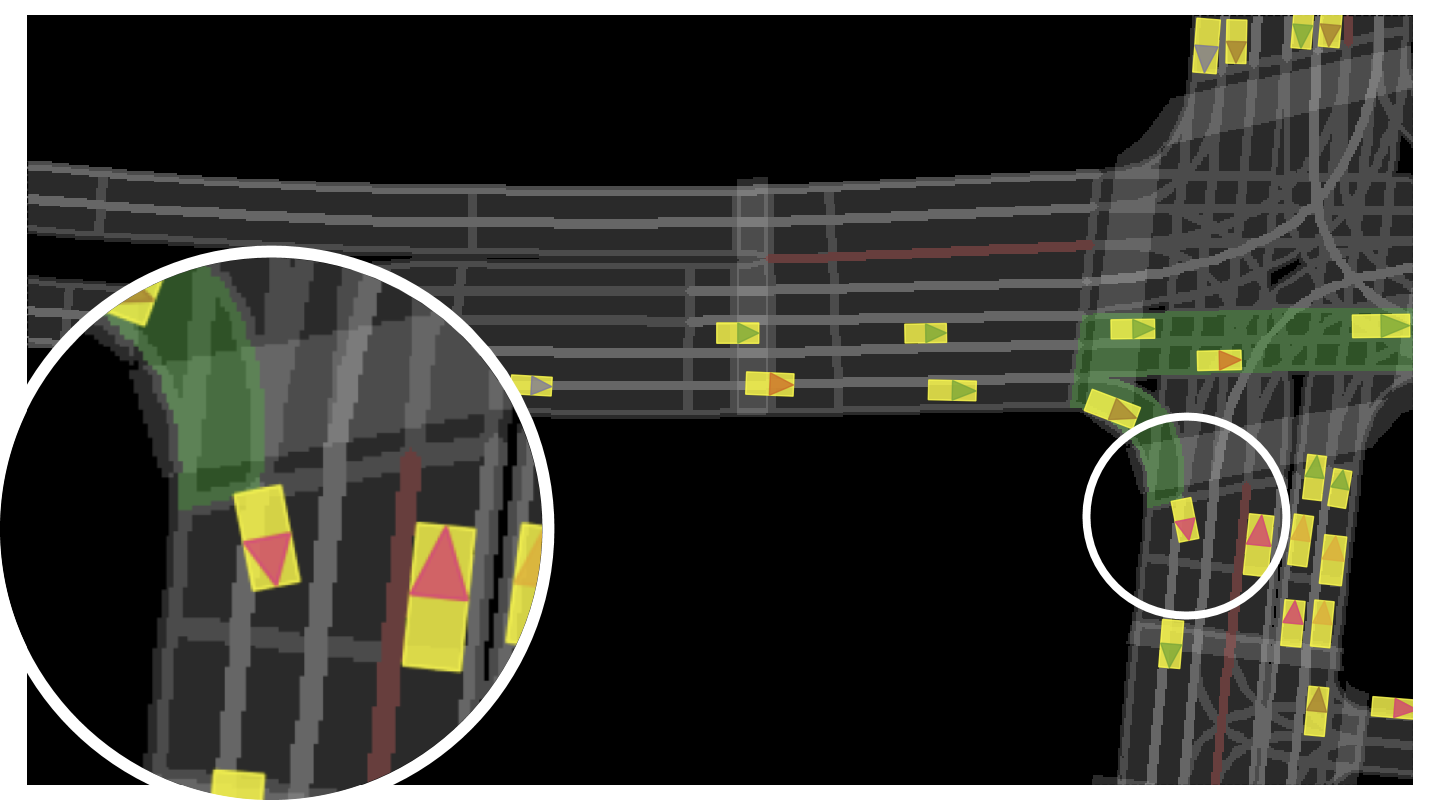}} \\
    \end{tabular}
    \caption{Simulated traffic scenarios sampled from \ourmodelshort{}: colored triangle shows heading and tracks instances across time}
    \label{fig:simulated_scenarios}
\end{figure*}

\newcolumntype{s}{>{\centering\arraybackslash}X}

\begin{table*}[t]
    \centering
    \small
	\begin{threeparttable}
        \begin{tabularx}{\linewidth}{
                        s |  %
                        s s s |
                        s s s s s s s
                        }
		    \toprule
                \multirow{2}{*}{Model} & $T_{\text{plan}}$ (timesteps) & Unroll in Training & Common Sense &
                SCR$_{12s}$ (\%) & TRV$_{12s}$ (\%) & 
                minSFDE (m) & minSADE (m) & meanSFDE (m) & meanSADE (m) & 
                MASD$_{12s}$ (m) \\
            \midrule
                $\mathcal{M}_0$     &1        &           &               &5.92           & 10.19         &2.04           &0.88           &2.50           & 1.04          & 0.80\\ 
                $\mathcal{M}_1$     &10       &           &               &2.32           & 3.43          &1.72           &0.99           &2.09           &  1.29         & 2.40\\ 
                $\mathcal{M}_2$     &1        &\checkmark &               &1.28           & 3.30          &\textbf{1.02}  &\textbf{0.54}  &\textbf{1.70}  & 0.88          & 3.57 \\ 
                $\mathcal{M}_3$     &10       &\checkmark &               &0.60           & 3.02          &1.21           &0.58           &1.70           & \textbf{0.84} & 2.16\\ 
            \midrule
                $\mathcal{M}^*$     &10       &\checkmark &\checkmark      &\textbf{0.50}  & \textbf{2.77} & 1.13          &0.57           &1.75           & 0.85          & 2.50 \\ 
	        \bottomrule
		\end{tabularx}
    \end{threeparttable}
    \caption{\textbf{[\ourdataset{}]} Ablation study ($S=15$ samples, $T=12$ seconds, $T_{\text{label}}=8$ seconds)}
	\label{table:atg4d_ablation}
\end{table*}

\subsection{Metrics}
Evaluating traffic simulation is challenging since there is no singular metric that can fully capture the quality of the generated traffic scenarios.
Thus we propose a suite of metrics for measuring the diversity and realism, with a particular focus on \emph{coverage} of real world scenarios.
We provide implementation details in the supplementary.
For all evaluations, we sample $K=15$ scenarios from the model given each initial condition.
More concretely, we create batches of $K$ scenarios with the same initialization.
Then at each timestep, we sample a single $\mathcal{Y}^t_{(k)}$ from $P(\mathcal{Y}^t_{(k)} | X^t_{(k)})$ for each scenario $(k)$, all in parallel. 
After unrolling for $\frac T {\delta_t}$ steps, we obtain the full scenarios.

\medskip\noindent\textbf{Interaction Reasoning:}
To evaluate the consistency of the actors' behaviors, we propose to measure the scenario collision rate ($\mathrm{SCR}$): 
the average percentage of actors in collision in each sampled scenario (thus lower being better). 
Two actors are considered in collision if the overlap between their bounding boxes at any time step is higher than a small IOU threshold. %

\medskip\noindent\textbf{Traffic Rule Compliance:}
Traffic actors should comply with traffic rules.
Thus, we propose to measure traffic rule violation (TRV) rate, and  %
 focus on two specific traffic rules: 1) staying within drivable areas, and 2) obey traffic light signals.

\medskip\noindent\textbf{Scenario Reconstruction:} 
We use distance-based scenario reconstruction metric to evaluate the model's ability to sample a scenario close to the ground truth.
(i.e., recovering irregular maneuvers and complex interactions collected from the real world). 
For each scenario sample, we calculate average distance error (ADE) across time, and final distance error (FDE) at the last labeled timestep.
We calculate minSADE/minSFDE by selecting the best matching scenario sample, and meanSADE/meanSFDE by averaging over all scenario samples.

\medskip\noindent\textbf{Diversity:}
Following \cite{yuan2019diverse},  
we use a map-aware average self distance (MASD) metric to measure the diversity of the sampled scenarios.
In particular, we measure the average distance between the two most distinct sampled scenarios that do no violate traffic rules.
We note that this metric can be exploited by models that generate diverse but unrealistic traffic scenarios.

\subsection{Experimental Results}

\begin{figure*}[t]
    \centering
    \begin{tabular} {@{}c@{\hspace{.5em}}c@{\hspace{.5em}}c@{\hspace{.5em}}c}
        \textbf{Nudge} & \textbf{U-Turn} & \textbf{Yield} & \textbf{Swerve}\\
        \raisebox{-0.5\height}{\includegraphics[width=0.245\linewidth]{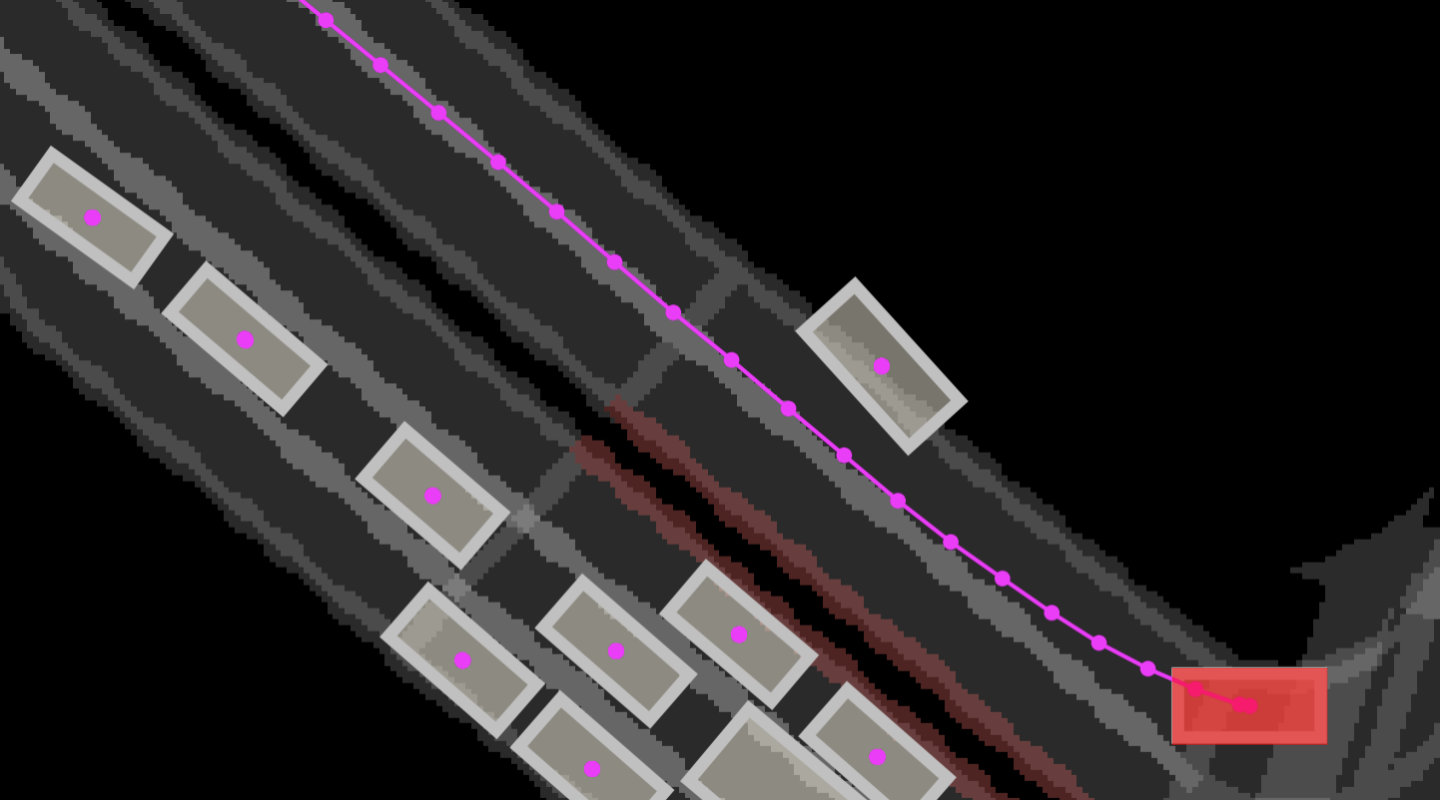}} &
        \raisebox{-0.5\height}{\includegraphics[width=0.245\linewidth]{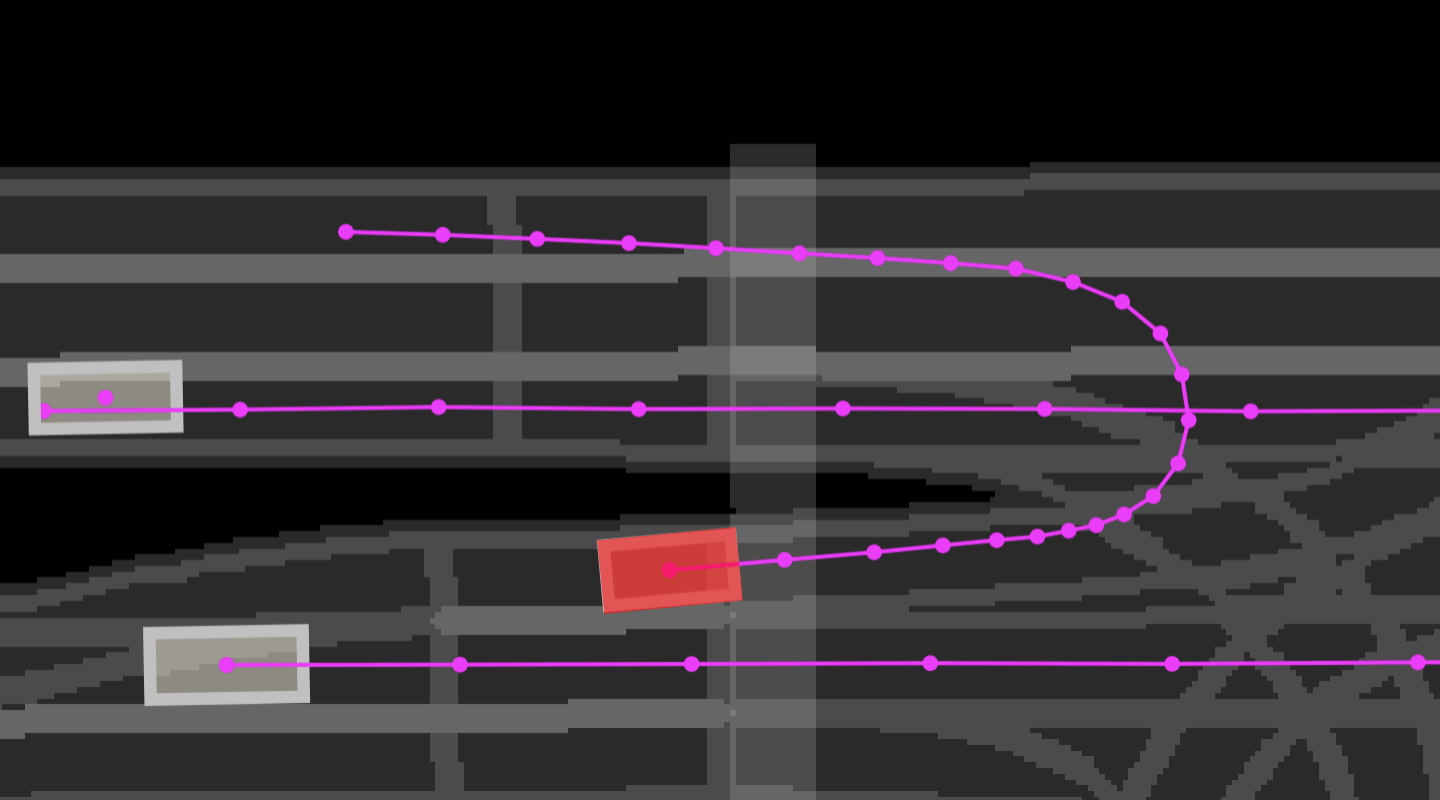}} &
        \raisebox{-0.5\height}{\includegraphics[width=0.245\linewidth]{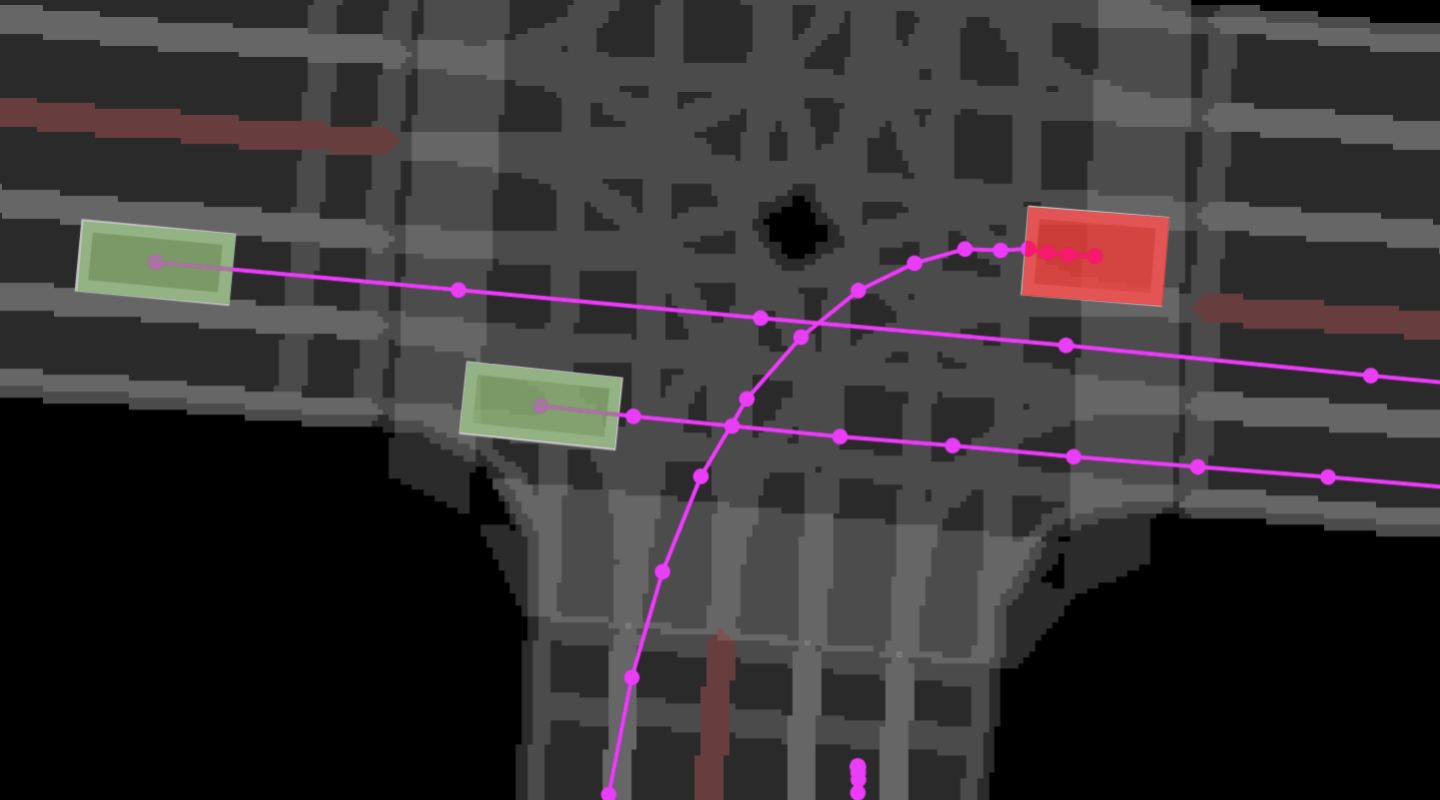}} &
        \raisebox{-0.5\height}{\includegraphics[width=0.245\linewidth]{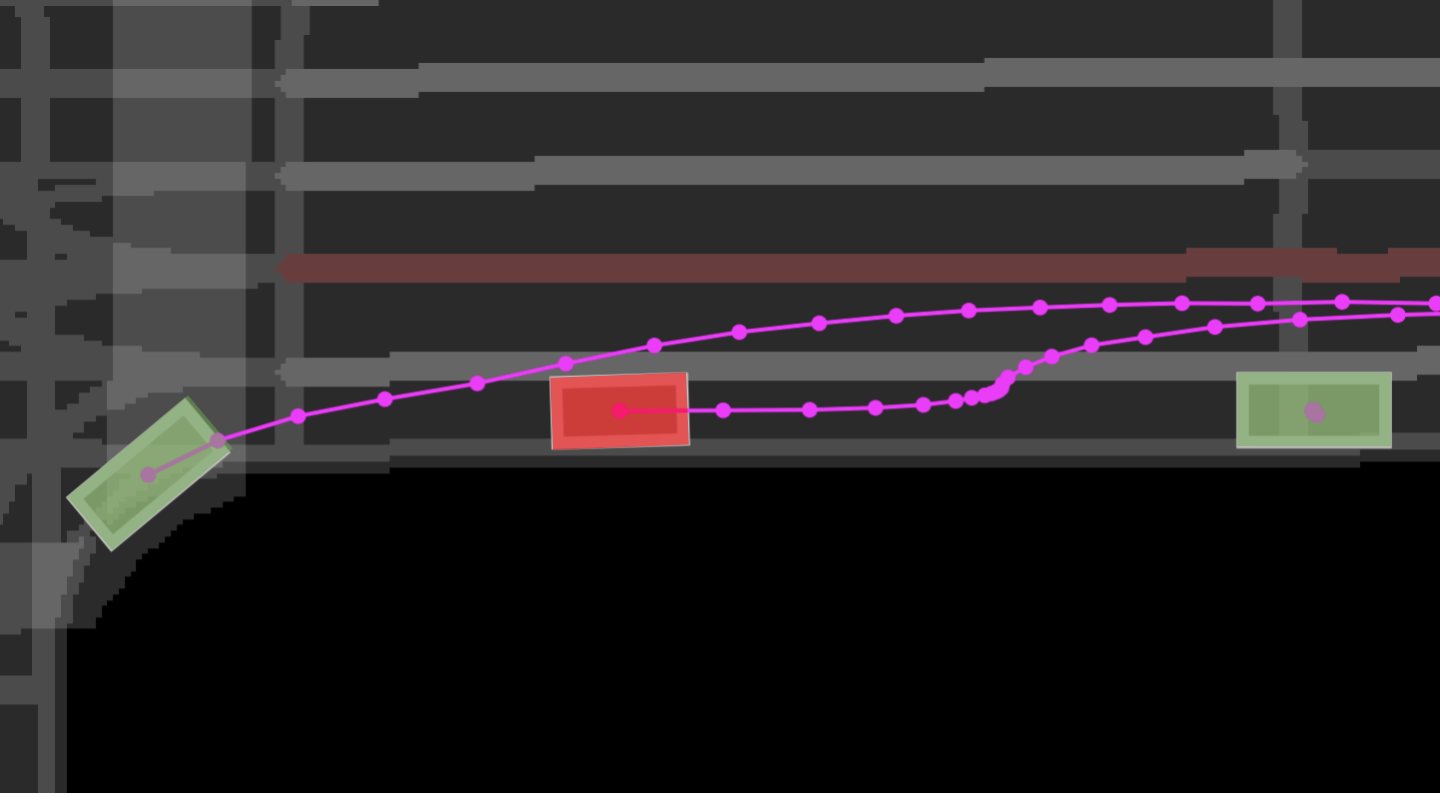}} \\
    \end{tabular}
    \caption{Irregular actor maneuvers and complex interactions sampled from \ourmodelshort{}} 
    \label{fig:simulated_behaviors}
\end{figure*}

\paragraph{Comparison Against Existing Approaches:}
Table~\ref{table:atg4d_main} shows quantitative results. 
Car following models generate collision free behavior that strictly follows traffic rules, but do not recover naturalistic driving, and thus scores poorly on scenario reconstruction metrics.
Motion forecasting models recover accurate traffic behavior, but exhibit unrealistic interactions and traffic rule violations when unrolled for a long simulation horizon.
Imitation learning techniques attempt to bridge the gap between train and test, and thus results in marginally better scenario reconstruction as compared to motion forecasting baselines. However, they inject additional bias that results in worse collision rate and traffic rule violation.
Our \ourmodelshort{} achieves the best of both worlds: best results on scenario reconstruction and interaction, and similar to IDM in traffic rule violation, without directly encoding the rules.
We note that the ground truth TRV rate is 1.26\%, since human exhibit non-compliant behaviors. 
Figure~\ref{fig:simulated_scenarios} shows qualitative visualization of traffic scenarios generated from \ourmodel{}.
Figure~\ref{fig:simulated_behaviors} shows that \ourmodelshort{} can generate samples with irregular maneuvers and complex interactions, which cannot be captured by heuristic models like IDM.

\paragraph{\ourmodelshort{} for Data Augmentation:}
    \ourmodelshort{} can be used to generate synthetic training data for learning better motion planners.
    More concretely, we generate $5s$ scenario snippets and train a planner to imitate behaviors of all actors in the scenario.
    As shown in Table~\ref{table:data_augmentation}, the planner trained with synthetic data generated from \ourmodelshort{} significantly outperforms
    baselines in open-loop planning metrics when evaluated against real scenarios. 
    Most notably, we achieve \emph{lower} planning L2 error, while matching collision rate and progress of planner trained with the same amount of real data.
    This shows that the scenarios generated from \ourmodelshort{} are realistic and have minimal gap from behaviors observed in the real world, and can be used as effective data augmentation.
    We show more details on this experiment in the supplementary. 

\newcolumntype{s}{>{\centering\arraybackslash}X}

\begin{table}[t]
    \centering
    \small
	\begin{threeparttable}
        \begin{tabularx}{\linewidth}{
                        >{\centering\arraybackslash} l@{\hskip .5cm}|  %
                        s s s %
                        }
		    \toprule
                \multirow{2}{2cm}{Training Data} & Collision Rate (\%) & Planning L2 (m) & Progress (m) \\
            \midrule
                Real                             &10.56 & 4.85 & 31.05 \\ 
            \midrule
                IDM \cite{idm}                   &22.05 & 10.49 & 29.17 \\ 
                MTP \cite{cui2018multimodal}     &19.54 & 9.89 & 26.46 \\ 
                ESP \cite{precog}                &19.38 & 8.76 & 24.73 \\ 
                ILVM \cite{casas2020implicit}    &14.82 & 7.16 & 26.39 \\ 
                AdversarialIL                    &13.83 & 5.19 & 29.02 \\ 
                DataAug                          &15.75 & 6.88 & 26.88 \\ 
                \ourmodelshort{}                 &\textbf{10.73} & \textbf{4.52} & 29.44 \\ 
	        \bottomrule
		\end{tabularx}
    \end{threeparttable}
    \caption{Imitation planner trained with synthetic data from \ourmodelshort{} outperforms real data in planning L2 error.}
    \label{table:data_augmentation}
\end{table}

\paragraph{Ablation Study:}
    We show the importance of each component of our model and training methodology in Table~\ref{table:atg4d_ablation}.
    Open-loop training ($\mathcal{M}_0$ \& $\mathcal{M}_1$) performs poorly due to compounding error at test-time. 
    Closed-loop training with back-propagation through simulation ($\mathcal{M}_2$) is the most important component in learning a robust policy.
    Explicitly modelling longer horizon plan (i.e., $\mathcal{Y}^t = \{Y^{t+1}, ..., Y^{t+T_{\text{plan}}}\}$ instead of $Y^{T+1}$) ($\mathcal{M}_3$) improves interaction reasoning.
    Augmenting imitation with common sense ($\mathcal{M}^*$) further reduces collision and traffic rule violation rates.

\paragraph{Multi-Step Update for Fast Simulation:}

    We can achieve faster simulation by running model inference once per $\kappa$ ticks of the simulation.
    This is possible since \ourmodelshort{} explicitly models the actor plans $\mathcal{Y}^t = \{Y^{t+1}, Y^{t+2}, ..., Y^{t+T_{\text{plan}}}\}$ and accurately captures future interactions in the planning horizon $T_{\text{plan}}$ even without extracting scene context at the highest simulation frequency.
    In particular, we can choose the desired tradeoff between simulation quality and speed by modulating $\kappa$ at simulation-time without retraining, as long as $\kappa \le T_{\text{plan}}$.
    Table~\ref{table:tradeoff} shows we can effectively achieve 4x speedup with minimal degradation in simulation quality.
    Runtime is profiled on a single Nvidia GTX 1080 Ti.

\paragraph{Incorporating Constraints at Simulation-Time:}
    Explicitly modelling actor plans $\mathcal{Y}^t$ at each timestep also makes it easy to incorporate additional constraints at simulation time.
    In particular, we can define constraints such as avoiding collision and obeying traffic rules over the actor plans $\mathcal{Y}^t$, to anticipate and prevent undesired behaviors in the future.
    Concretely, we evaluate two optimization methods for avoiding collision: 1) rejection sampling which discard actor plans that collide and re-sample, and 2) gradient-based optimization of the scene latent $Z^t$ to minimize the differentiable relaxation of collision.
    Table~\ref{table:test_constraint} shows that both methods are effective in reducing collision while keeping the simulation realistic.
    More  details in supplementary.

\paragraph{\ourmodelshort{} for interactive Simulation}
    We create an interactive simulation tool to showcase how simulation designers can leverage \ourmodelshort{} to construct and preview interesting traffic scenarios.
    In particular, they can alter traffic light states and add, modify, or remove actors during simulation.
    In response, \ourmodelshort{} generates realistic variants of the traffic scenario.
    We show visual demonstrations in the supplementary.

    \newcolumntype{s}{>{\centering\arraybackslash}X}

\begin{table}[t]
    \centering
    \small
	\begin{threeparttable}
        \begin{tabularx}{\linewidth}{
                        >{\centering\arraybackslash} l@{\hskip .5cm}|  %
                        s s s s s
                        }
		    \toprule
                \multirow{2}{1cm}{Inference Frequency}  & \multirow{2}{1cm}{Runtime (s)} & \multirow{2}{1cm}{SCR$_{12s}$ (\%)}  & \multirow{2}{1cm}{TRV$_{12s}$ (\%)} & min SFDE~(m) & mean SFDE~(m) \\
                \midrule
                2Hz                                     & 0.83          & \textbf{0.50}     & \textbf{2.77} & 1.13 & 1.75 \\ 
                1Hz                                     & 0.45           & 0.85             & 3.17 & \textbf{1.12} & \textbf{1.73} \\ 
                0.5Hz                                   & \textbf{0.24}  & 0.96             & 3.64 & 1.16 & \textbf{1.73} \\ 
	        \bottomrule
		\end{tabularx}
    \end{threeparttable}
    \caption{Multi-step update achieves up to 4x speedup with minimal degradation in simulation quality}
	\label{table:tradeoff}
\end{table}
    \newcolumntype{s}{>{\centering\arraybackslash}X}

\begin{table}[t]
    \centering
    \small
	\begin{threeparttable}
        \begin{tabularx}{\linewidth}{
                        >{\centering\arraybackslash} l@{\hskip .1cm}|  %
                        s s s s%
                        }
		    \toprule
                \multirow{2}{2.1cm}{Post-Processing} & 
                \multirow{2}{1cm}{SCR$_{12s}$ (\%)} & 
                \multirow{2}{1cm}{TRV$_{12s}$ (\%)} & 
                min SFDE~(m) & 
                mean SFDE~(m) \\
                \midrule
                None                                    & 0.50          & \textbf{2.77} & \textbf{1.13} & \textbf{1.75} \\
                Rejection Sampling                      & 0.33          & 3.01          & \textbf{1.13} & \textbf{1.75}\\ 
                Gradient Optimization                   & \textbf{0.12} & 3.00          & \textbf{1.13} & \textbf{1.75}\\

	        \bottomrule
		\end{tabularx}
    \end{threeparttable}
    \caption{Additional optimization at simulation-time further reduces collision rate}
	\label{table:test_constraint}
\end{table}

\section{Conclusion}
In this work, we have proposed a novel method for generating diverse and realistic traffic simulation.
\ourmodelshort{} is a multi-agent behavior model that generates socially-consistent plans for all actors in the scene jointly.
It is learned using back-propagation through the fully differentiable simulation, 
by imitating trajectory observations from a real-world self driving dataset and incorporating common sense.
\ourmodelshort{} enables exciting new possibilities in data augmentation, interactive scenario design, and safety evaluation.
For future work, we aim to extend this work to learn controllable actors where we can specify attributes such as goal, route, and style.
\pagebreak

{\small
\bibliographystyle{ieee_fullname}
\bibliography{cvpr}
}
\clearpage

\appendix

\vspace*{0.2cm}
{ \noindent \Large \textbf{Appendix}} \\

\noindent
In this supplementary material, we provide the following:
additional details of our method in Section~\ref{sec:supp_method}, 
implementation details of baselines and metrics in Section~\ref{sec:supp_experiment},
and lastly, additional qualitative results in Section~\ref{sec:supp_result}.

\section{Additional \ourmodelshort{} Details}
\label{sec:supp_method}
\noindent
In this section, we describe additional details of input parameterization, model architecture, and learning methodology for \ourmodelshort{}.

\paragraph{Input Parameterization:}
We use a rasterized map representation that encodes traffic elements into different channels of a raster.
There are a total of 13 map channels consisting of intersections, lanes, roads, etc.
We encode traffic control as additional channels, by rasterizing the lane segments controlled by the traffic light.
We initialize each scenario with 3s of past actor states, with each actor history represented by 7 bounding boxes across time, each 0.5s apart.
When an actor does not have the full history, we fill the missing states with NaNs.

\paragraph{Global Map Module:}
We use a multi-scale backbone to extract map features at different resolution levels to encode both near and long-range map topology.
The architecture is adapted from \cite{yang2018pixor}:
it consists of a sequence of 4 blocks, each with a single convolutional layer of kernel size 3 and [8, 16, 32, 64] channels.
After each block, the feature maps are down-sampled using max pooling with stride 2.
Finally, feature maps from each block are resized (via average-pooling or bilinear sampling) to a common resolution of 0.8m, concatenated, 
and processed by a header block with 2 additional convolutional layers with 64 channels.

\paragraph{Local Observation Module:}
We design the local observation modules to be lightweight and differentiable.
This enables the simulation to be fast and allows us to backpropagate gradient through the simulation.
Works in motion forecasting (e.g., \cite{cui2018multimodal} typically rasterize the bounding boxes of each actor,
use a convolutional network to extract motion features, 
and rely on a limited receptive field to incorporate influences from its neighbors.
In contrast, we directly encode the numerical values parameterizing the bounding boxes using a 4-layer GRU with 128 hidden states, 
and rely on the graph neural network based module for interaction reasoning.
More concretely, we fill NaNs with zeros, and also pass in binary mask indicating missing values to the GRU.
For extracting local map features from the pre-processed map features, we use Rotated Region of Interest Align with 70m in front, 10m behind, and 20m on each side. The extracted local features are further processed by a 3-layer CNN, and then max-pooled across the spatial dimensions. The final local context $x_i$ for each actor $i$ is a 192 dimensional vector formed by concatenating the map and motion features.

\paragraph{Scene Interaction Module:}
We leverage a graph neural network based scene interaction module to parameterize our joint actor policy.
In particular, our scene interaction module is inspired by \cite{casas2019spatiallyaware,casas2020implicit}, and is used in our Prior, Posterior, and Decoder networks. 
We provide an algorithmic description in Algorithm~\ref{alg:sim}.
The algorithm is written with for loops for clarity, but in practice our implementation is fully vectorized, since the only loop that is needed is that of the $K$ rounds of message passing, but in practice we observe that $K=1$ is sufficient. 
Our edge function $\mathcal{E}^{(k)}$ consists of a 3-layer MLP that takes as input the hidden states of the 2 terminal nodes at each edge in the graph at the previous propagation step as well as the projected coordinates of their corresponding bounding boxes.
We use feature-wise max-pooling as our aggregate function $\mathcal{A}^{(k)}$. 
To update the hidden states we use a GRU cell as $\mathcal{U}^{(k)}$. Finally, to output the results from the graph propagations, we use another MLP as readout function $\mathcal{O}$.
\begin{algorithm*}[t]
    \caption{SIM: Scene Interaction Module} \label{alg:sim}
    \textbf{Input:}
    Initial hidden state for all of the actors in the scene $H^0 = \begin{Bmatrix}h_0^0, h_1^0, \cdots, h_N^0\end{Bmatrix}$. 
    BEV coordinates of the actor bounding boxes $\begin{Bmatrix} c_0,c_1, ..., c_N \end{Bmatrix}$.
    Number of message propagations $K$ (defaults to $K=1$).
    
    \textbf{Output:} 
    Output vector per node $\begin{Bmatrix}o_0, o_1, \cdots, o_N\end{Bmatrix}$.
    
    \begin{algorithmic}[1]
    \State Construct actor interaction graph $G = (V, E)$
    \State Compute pairwise coordinate transformations $\mathcal{T} (c_{u}, c_{v})$, $\forall (u, v) \in E $
    \For{$k=1, ..., K$} \Comment{Loop over graph propagations}
        \For {$ (u, v) \in E $} \Comment{Compute message for every edge in the graph}
            \State $m_{u \rightarrow v}^{(k)} = \mathcal{E}^{(k)}\left(h_{u}^{k-1}, h_{v}^{k-1}, \mathcal{T} (c_{u}, c_{v}) \right)$
        \EndFor
        \For {$ v \in V $} \Comment{Update node states}
            \State $a_{v}^{(k)}=\mathcal{A}^{(k)}\left(\left\{m_{u \rightarrow v}^{(k)} : u \in \mathbf{N}(v)\right\}\right)$ \Comment{Aggregate messages from neighbors}
            \State $h_{v}^{(k)}=\mathcal{U}^{(k)}\left(h_{v}^{(k-1)}, a_{v}^{(k)}\right)$ \Comment{Update the hidden state}
        \EndFor
    \EndFor
    \For {$ v \in V $}
        \State $o_{v}=\mathcal{O}\left(h_{v}^{(K)}\right)$ \Comment{Compute outputs}
    \EndFor
    \State\Return $\begin{Bmatrix}o_0, o_1, \cdots, o_N\end{Bmatrix}$
    \end{algorithmic}
\end{algorithm*}

\paragraph{Simulating Traffic Scenarios:}
We provide an algorithmic description of the overall inference process of simulating traffic scenarios in Algorithm~\ref{algo:traffic_sim}.
While the algorithm describes the process of sampling a single scenario, and loop over actors in the scene, we can sample multiple scenarios with arbitrary number of actors in parallel by batching over samples and actors.
Note that we do not directly regress heading of the actors. 
Instead, we approximate headings in a post-processing step by taking the tangent of segments between predicted waypoints.
This ensures the headings are consistent with the predicted motion.

\begin{figure}[t]
    \centering
    \includegraphics[width=\linewidth]{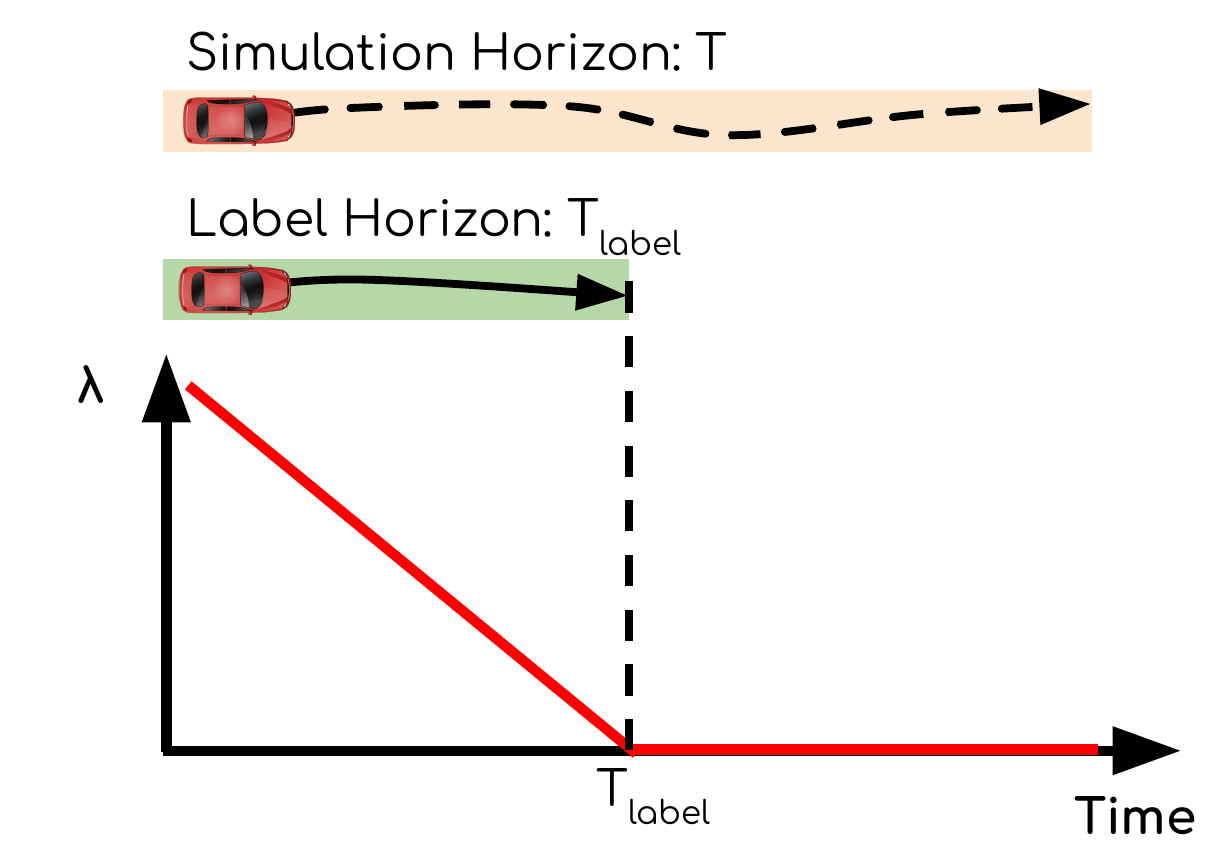}
    \label{fig:adaptive_loss}
    \caption{Our adaptive weight is a decreasing function of simulation timestep.}
\end{figure}

\paragraph{Time-Adaptive Multi-Task Loss:}
We use a multi-task loss to balance supervision from imitation and common sense:
\begin{align}
    \mathcal{L} = \sum_t \lambda(t) \mathcal{L}^t_{\text{imitation}} + (1- \lambda(t)) \mathcal{L}^t_{\text{collision}}
\end{align}
Concretely, we define the time-adaptive weight as:
\begin{align}
    \lambda(t) = \min(\frac {T_{\text{label}} - t} {T_{\text{label}}}, 0)
\end{align}
where $T_{\text{label}}$ is the label horizon (i.e., latest timestep which we have labels).
Figure~\ref{fig:adaptive_loss} illustrates this function.
Intuitively, the weight on imitation must drop to zero at $T_{\text{label}}$ as we no longer have access to labels.
We note that our method is not sensitive to the choice of $\lambda(t)$. Experiments with other decreasing function of simulation timestep yields similar results.

\begin{figure}[t]
    \centering
    \includegraphics[width=\linewidth]{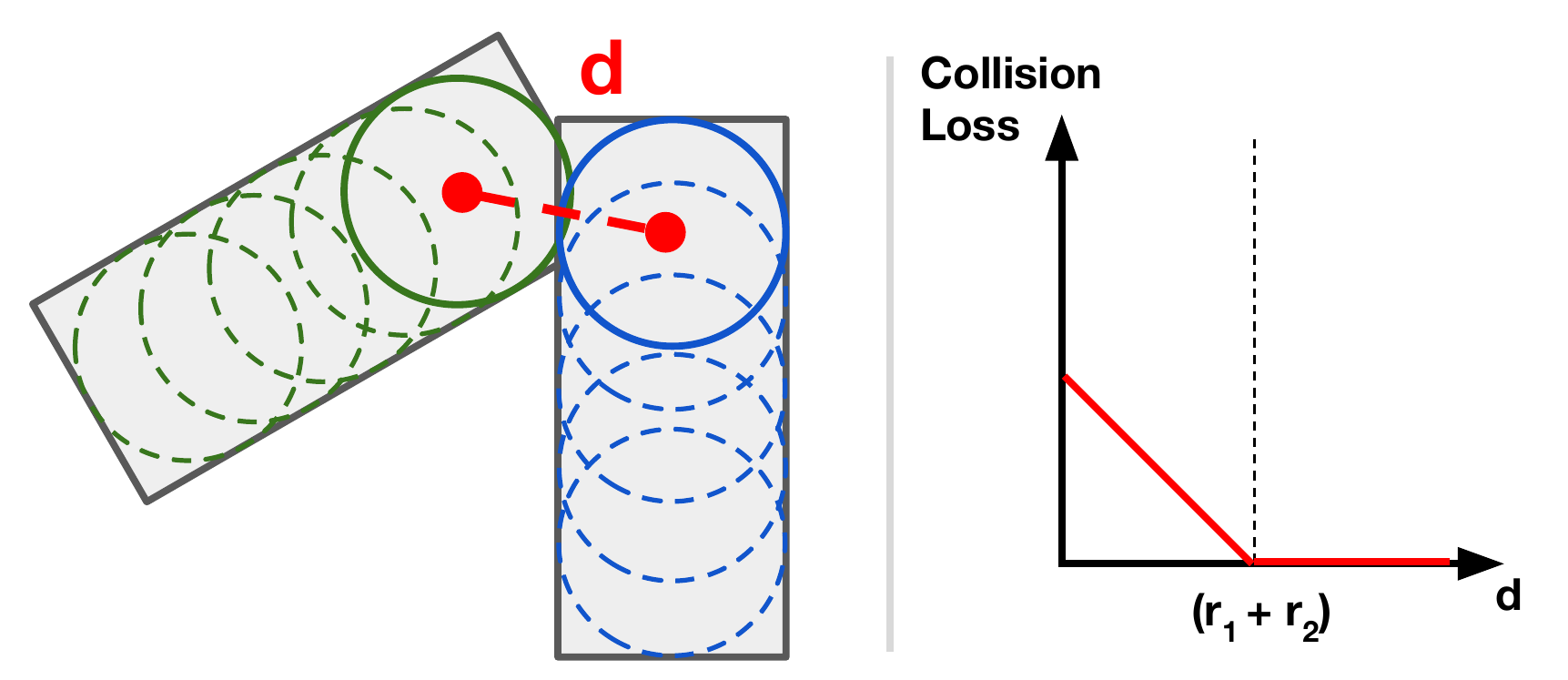}
    \label{fig:collision_loss}
    \caption{Differentiable relaxation of collision loss approximates each vehicle as 5 circles and considers distance between closest centroids.}
\end{figure}

\paragraph{Differentiable Relaxation of Collision:}
Figure~\ref{fig:collision_loss} illustrates our proposed differentiable relaxation of collision. More concretely, the loss is defined as follows:
{\small
\begin{align} \mathcal{L}^t_{\text{collision}}(\mathcal{Y}^t_{\text{prior}}) = \frac 1 {N^2} \sum_{i\neq j} \max(1, \sum_{\tau=t+1}^{t+P}\mathcal{L}_{\text{pair}}(y_i^\tau, y_j^\tau))\\
    \mathcal{L}_{\text{pair}}(y_i^\tau, y_j^\tau) = 
    \begin{cases}
        1 - \frac d {r_i + r_j}, & \text{if } d \le r_i + r_j\\
        0,                       & \text{otherwise}
    \end{cases}
\end{align}
}
Intuitively, if there's no overlap between any circles, the collision loss is 0. If two circles completely overlap, the collision loss is 1.
We further reweight the collision loss with a factor of $0.01$ to be on similar scale as the imitation loss.

\begin{algorithm*}[t]
    \caption{\ourmodelshort{}: Simulating Traffic Scenarios} \label{algo:traffic_sim}
    \textbf{Input:} 
    Rasterized high definition map $M$. 
    Initial actor states $Y^{-H:0} = \begin{Bmatrix}Y^{-H}, \cdots, Y^{0}\end{Bmatrix}$ where
    each $Y^t = \begin{Bmatrix} y_1^{t}, y_2^{t}, ..., y_N^{t}\end{Bmatrix}$ for the $N$ actors in the scene.

    \textbf{Output:} 
    Simulated actor states $Y^{1:T} = \begin{Bmatrix}Y^1, Y^2, \cdots, Y^{T}\end{Bmatrix}$ for $T$ simulation timesteps. 
    
    \begin{algorithmic}[1]
    \State $ \tilde{M} \gets \operatorname{MapBackbone}(M)$ \Comment{Extract global map feature once per environment}
    \For{$t=1, ..., T$} \Comment{Simulate for requested number of timesteps}
        \For{$i=1, ..., N$} \Comment{Extract local context for each actor at each timestep}
            \State $r_i \gets \operatorname{RRoiAlign}(y_i^t, \tilde{M})$ 
            \State $x_i^{\text{map}} \gets \operatorname{MaxPooling}(\operatorname{CNN}(r_i))$
            \State $x_i^{\text{motion}} \gets \operatorname{GRU}(y^{-H:t}_i)$  
            \State $x_i \gets x_i^{\text{map}} \oplus x_i^{\text{motion}}$
        \EndFor
        \State $X = \{x_i: \forall i \in 1 \dots N\}$
        \State $\begin{Bmatrix}Z_\mu, Z_\sigma \end{Bmatrix} \gets \text{Prior}_{\gamma}(X)$ \Comment{Use SIM modules to output latent prior distribution}
        \State $Z  \sim \mathcal{N}\left(\begin{Bmatrix}Z_\mu, Z_\sigma \cdot I \end{Bmatrix}\right)$ \Comment{Sample a scene latent from diagonal Gaussian}
    \State $H = \begin{Bmatrix}\text{MLP}(x_i \oplus z_i): \forall i \in 1 \dots N \end{Bmatrix}$
        \State $\mathcal{Y} \gets \operatorname{Decoder}_\theta (H)$ \Comment{Use SIM module to decode actor plans}
        \State $Y^{t+1: t+\kappa} \gets \mathcal{Y}$ \Comment{Update environment by taking the first $\kappa$ steps of the actor plans}

    \EndFor
    \State\Return $Y^{1:T} = \begin{Bmatrix}Y^1, Y^2, \cdots, Y^{T}\end{Bmatrix}$
    \end{algorithmic}
\end{algorithm*}
\section{Additional Experiment Details}
\label{sec:supp_experiment}
\noindent
In this section, we provide additional details on baselines, metrics, and experimental setup.

\subsection{Baselines}

\paragraph{IDM \cite{idm}:}
The Intelligent Driver Model (IDM) is a heuristic car following model that implement reactive keep lane behavior by following a specific headway vehicle.
We resolve headway vehicle based on lane association, and also a narrow field of view in a 30$^\circ$ sector in front of each actor, with visibility up to 10 meters.
We implement traffic control as a phantom actor that has zero size.
We use a simulation frequency of 2.5Hz (0.4s per timestep), 
max deceleration of $3$ m/s$^2$,
reaction time of $0.1$s,
time headway of $1.5$s. 
To generate diverse simulations, we sample 
max acceleration in the range of $[0.6, 2.5]$ m/s$^{2}$, and 
desired speed in the range of $[10, 20]$ m/s.
Since the motion of IDM actors are paramterized against lane centerlines,
it can trivially avoid traffic rule violations.
However, it does suffer from an inherent limitation: 
unlike learned models,
it cannot infer traffic flow from the initial actors states when given partially observed traffic light states,
and thus results in occasional collisions at intersections.

\paragraph{MTP \cite{cui2018multimodal}:}
Multiple Trajectory Prediction (MTP) models uncertainty over actors' future trajectories with mixture of Gaussians at each prediction timestep.
It does not explicitly reason about interaction between actors as the future unrolls, and makes conditional independence assumption across actors.
We use a mixture of Gaussian with 16 modes. Following the training methodology described in the original paper, we select the closest matching mode to compute loss,
instead of directly optimizing the mixture density.

\paragraph{ESP \cite{precog}:}
ESP models multi-agent interaction by leveraging an autoregressive formulation, where actors influence each other as the future unrolls.
Due to memory constraints, we limit the radii of the whiskers to [1, 2, 4] m while keeping the seven angle bins.
We implement the social context condition with a minor modification.
The original paper specifies a fixed number of actors (since Carla has a small number of actors).
, but this is not possible in \ourdataset{} since traffic scenes contain many more actors.
Thus, we use k-nearest neighbors to select $M=4$ neighbors to gather social features.

\paragraph{ILVM \cite{casas2020implicit}:}
We adapt ILVM from the joint perception and prediction setting by replacing voxelized LiDAR input by rasterized actor bounding boxes.
Since processing noise-free actor bounding boxes require less model capacity than performing LiDAR perception,
we reduce the number of convolutional layers in the backbone to improve inference speed.
We noticed no degredation in performance in reducing the model capacity.

\paragraph{DataAug:}
We follow data augmentation technique described in \cite{chauffeur-net}, 
since it also leverages large-scale self driving datasets and is closest to our setting.
To factor out effects of model architecture, we use the best motion forecasting model ILVM as the base architecture.
In particular, for each eligible trajectory snippet, we perturb the waypoint at the current timestep with a probability of 50\%.
we consider a trajectory to be eligible if has moved more than 16m in the 2s window around the perturbation (i.e. speed higher than 8m/s).
We uniformly sample perturbation distance in the range of [0, 0.5] m, and sample a random direction to perturb the waypoint.
Finally, we fit quadratic curves for both x and y as function of time, to smooth out the past and future trajectory. 
We only alter waypoints up to 2 seconds before and after the perturbation point.

\paragraph{AdversarialIL:}
Inspired by \cite{gail,ngsim-gail, ps-gail,horizon-gail, rail}, we implement an adversarial imitation learning baseline by jointly learning a discriminator as the supervision for the policy. Similarly, to factor our the effects of model architecture, we leverage our differentiable observation modules and scene interaction module to parameterize the policy. In particular, the extracted scene context is directly fed to the scene interaction module to output a bivariate Gaussian of the next waypoint for each actor in the scene.
The discriminator has a similar architecture with spectral normalization.
By leveraging our differentiable components, we enable our adversarial IL baseline to also leverage backpropagation through simulation.
This allows it to sidestep the challenge of policy optimization, and enables a more direct comparison with our method.
For optimization, we use a separate Adam optimizer for the policy, and RMSProp optimizer for the discriminator.
Furthermore, we found it's necessary to periodically use behavior cloning loss to stabilize training.
We use a replay buffer size of 200 and batch size 8 for optimizing the policy and the discriminator.
Furthermore, following \cite{horizon-gail}, we use a curriculum of increasing simulation horizon to ease optimization. 
More concretely, we first pre-train with behavior cloning for 25k steps, then 
follow a schedule of [2,4,6,8] simulation timesteps increasing every 25k steps. 
We found increasing simulation horizon further does not improve the model.

\subsection{Metrics}

\paragraph{Scenario Reconstruction:}
For each simulation environment (i.e., with a given map, traffic control, and initial actor states), we define the following scenario reconstruction metrics over the $K$ traffic scenarios sampled from the model:
{\small
\begin{align*}
    \mathrm{minSADE} & = \min_{ k \in 1 \dots K} \frac{1}{N T_{\text{label}}} \sum_{n=1}^{N} 
    \sum_{t=1}^{T_\text{label}}||y^t_{n, GT} - y^t_{n, (k)}||^2\\
    \mathrm{minSFDE} & = \min_{ k \in 1 \dots K} \frac{1}{N} \sum_{n=1}^{N} 
    ||y^{T_{\text{label}}}_{n, GT} - y^{T_{\text{label}}}_{n, (k)}||^2\\
    \mathrm{meanSADE} & =  \frac{1}{K N T_{\text{label}}} \sum_{k=1}^K \sum_{n=1}^{N} 
    \sum_{t=1}^{T_\text{label}}||y^t_{n, GT} - y^t_{n, (k)}||^2\\
    \mathrm{meanSFDE} & =  \frac{1}{K N} \sum_{k=1}^K \sum_{n=1}^{N} 
    ||y^{T_{\text{label}}}_{n, GT} - y^{T_{\text{label}}}_{n, (k)}||^2
\end{align*}
}
In particular, we only evaluate up to $T_{\text{label}} = 8s$ due to the availability of ground truth actor trajectories.

\paragraph{Interaction Reasoning:}
Our scenario level collision rate metric is implemented as follows:
{\small
\begin{align*}
    \mathrm{SCR} = \frac{1}{NS} \sum_{s=1}^S
    \frac{}{}\sum_{i=1}^{N} \min \left(1, \sum_{j>i}^{N} 
    \sum_{t=1}^T \mathbbm{1} \left[IoU(b^t_{i, s}, b^t_{j, s}) > \varepsilon\right] \right)
\end{align*}
}
In particular, we consider two actors to be in collision if their bounding boxes overlap each other with IOU greater than a small $\epsilon$ of 0.1.
This threshold is necessary since the labeled bounding boxes are slightly larger than true vehicle shape, thus sometimes resulting in collisions even in ground truth scenarios.
Furthermore, we count a maximum of 1 collision per actor. In other words, we count number of actors in collision, rather than number of total collisions between pairs of actors.

\paragraph{Traffic Rule Compliance:}
We leverage our high definition map with precise lane graph annotations to evaluate traffic rule compliance. 
More concretely, we first obtain the drivable areas of each actor in the scenario by first performing lane association. 
Then, we traverse the lane graph to derive a set of reachable lane segments from the initial location, including neighbours and successors.
Furthermore, we cut off any connection influenced by traffic control (i.e. red traffic light).
We rasterize the drivable area with binary values: 1 for drivable and 0 for violation.
This allows us to efficiently index and calculate traffic rule violations. 
To handle actors that begin initially outside of mapped region (i.e. parked vehicles on the side of the road or in a parking lot), 
we ignore actors that do not have initial lane associations.

\paragraph{Diversity:}
To calculate our map-aware diversity metric, we leverage the same drivable area raster employed by the traffic rule compliance metric.
More concretely, we first filter out actor trajectory samples that violate traffic rule, then we measure the average distance (across time) 
between the two most distinct trajectory samples for each actor.
{\small
\begin{align*}
    \mathrm{MASD} = \max_{k, k' \in 1...K} \frac 1 {NT} \sum_{n=1}^N\sum_{t=1}^T || y_{n, (k)}^t - y_{n, (k')}^t||^2
\end{align*}
}

\subsection{Experimental Setup}

\paragraph{\ourmodelshort{} for Data Augmentation:}
For synthetic data generation, we generate approximately 15k examples by initializing from a subset of training scenarios used to train the traffic simulation models. We use the same amount of real data for fair comparison.
We use a simple imitation planner, which takes rasterized map and actor bounding box history as input, and directly regresses the future plan.
The planning horizon is 5s, with 0.5s per waypoint, for a total of 10 waypoints.
The imitation planner is learned with supervision from all actor trajectories in the synthetic scenarios.
For evaluation, we test on the same set of ground truth scenarios that are used for evaluating traffic simulation metrics.

\paragraph{Incorporating Constraints at Simulation-Time:}
For rejection sampling at simulation-time, we resample at most 10 times to keep the runtime bounded.
If we cannot generate enough collision-free plans after 10 re-sampling steps, we sort all the generated samples by collision loss,
then return the minimum cost plans.
To reason about potential collision in the future, we evaluate our collision loss on the first 5 timesteps of the sampled actor plans (i.e. 2.5s into the future).
For gradient-based optimization, we leverage our differentiable relaxation of collision loss.
Similarly, we evaluate the collision loss on the first 5 timesteps of the actor plans (i.e. 2.5s into the future).
While keeping our model frozen, we backpropagate the gradient to optimize the latent samples $Z^t$.
Performing the optimization in the latent space allows us to influence the actor plans while remaining 
in the model distribution.
More concretely, we take 5 gradient steps with a learning rate of 1e-2.

\section{Additional Qualitative Results}
\label{sec:supp_result}
\begin{figure*}[t]
    \centering
    \begin{tabular} {@{}c@{\hspace{.1em}}c@{\hspace{.1em}}c@{\hspace{.1em}}c@{\hspace{.1em}}c}
        {} & \textbf{T=0s} & \textbf{4s} & \textbf{8s} & \textbf{12s} \\
        \rotatebox[origin=c]{90}{\textbf{Scenario 1}} &
        \raisebox{-0.5\height}{\includegraphics[width=0.248\linewidth]{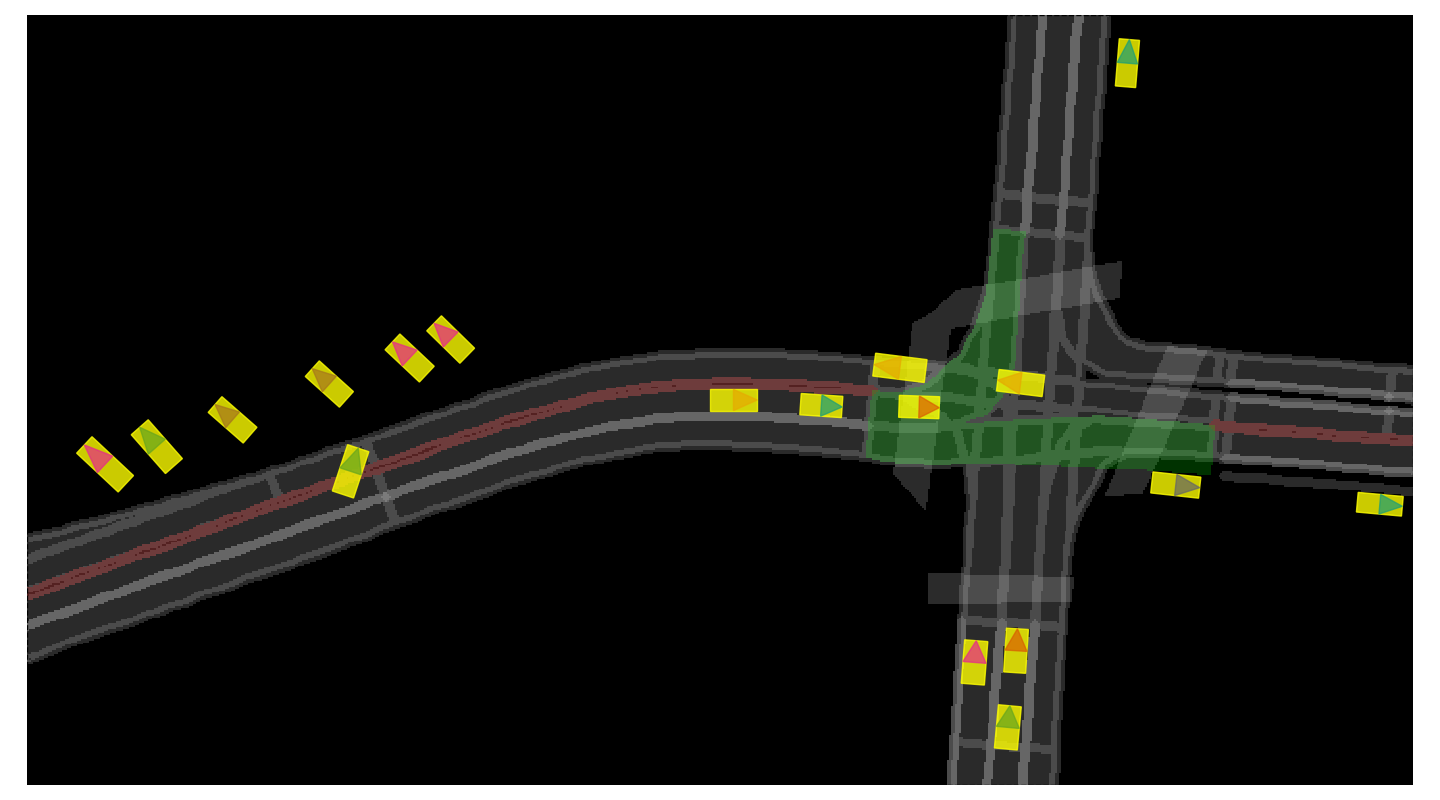}} &
        \raisebox{-0.5\height}{\includegraphics[width=0.248\linewidth]{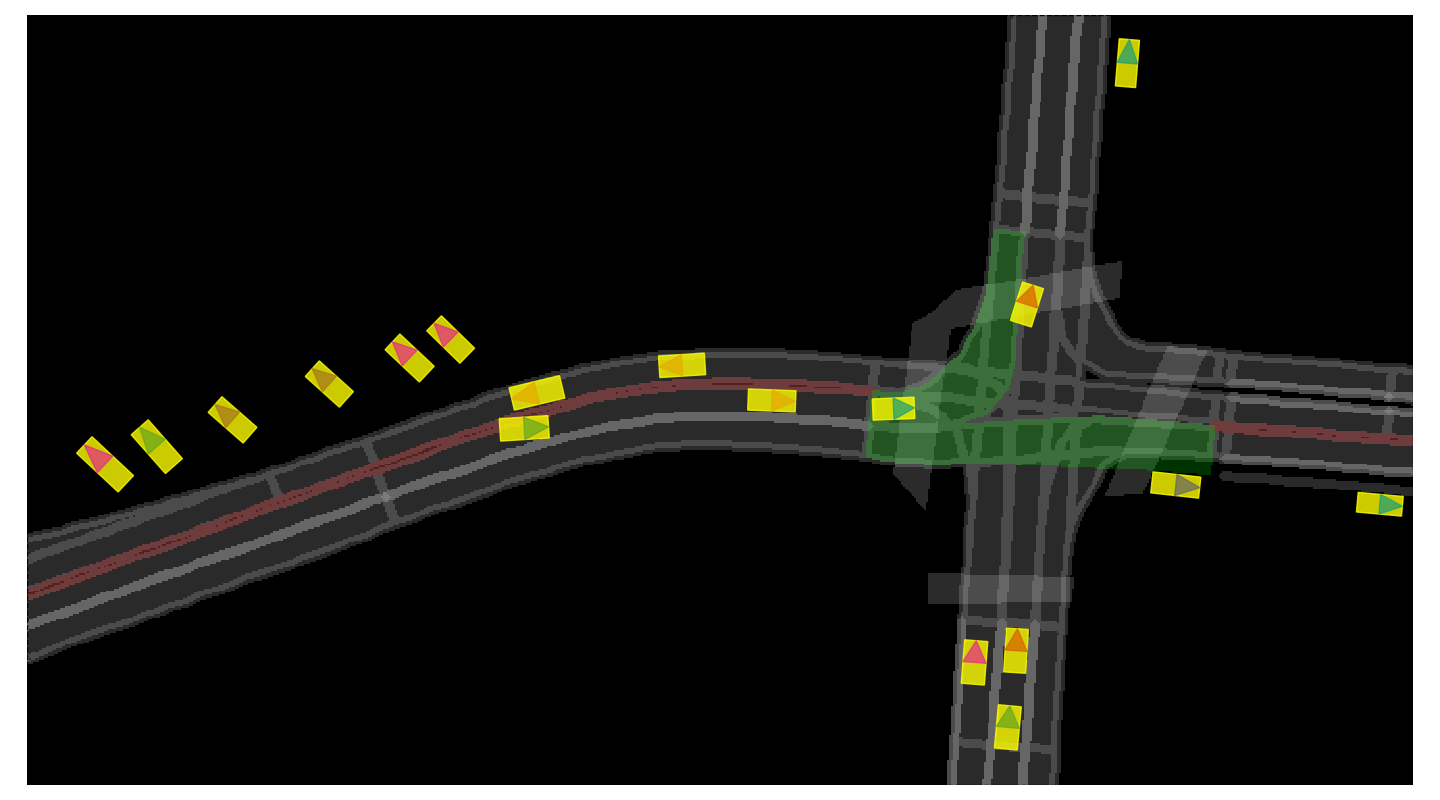}} &
        \raisebox{-0.5\height}{\includegraphics[width=0.248\linewidth]{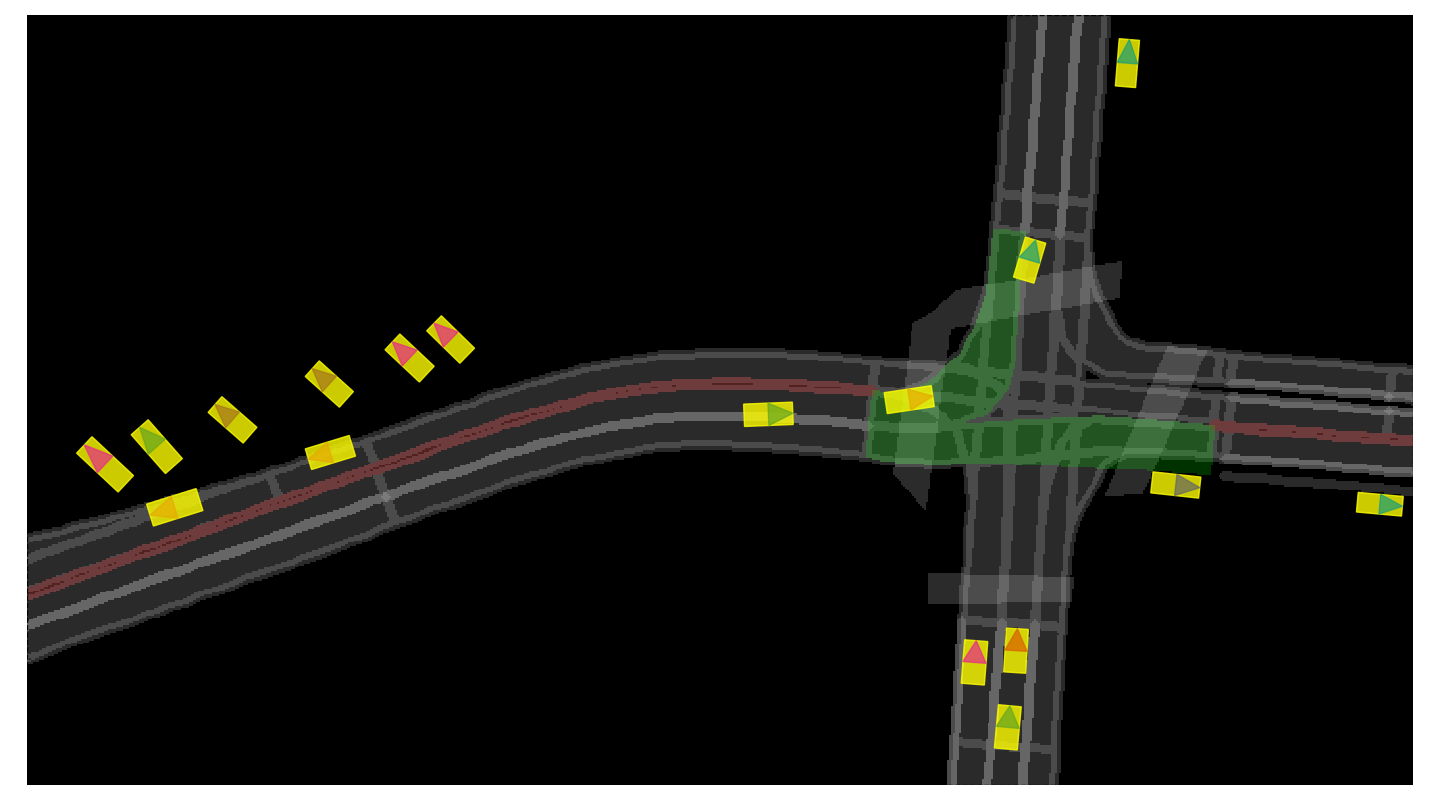}} &
        \raisebox{-0.5\height}{\includegraphics[width=0.248\linewidth]{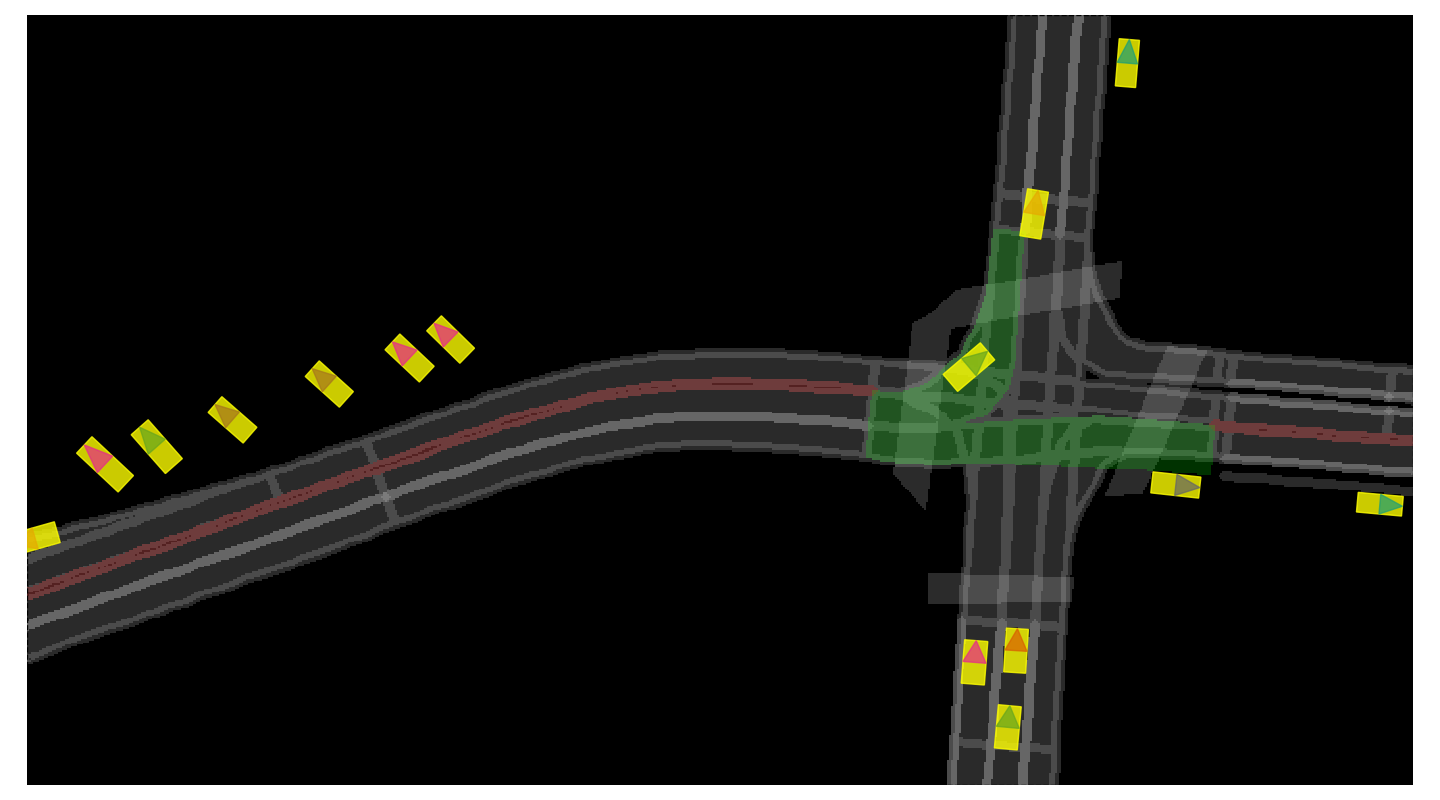}} \vspace{.1em} \\
        \rotatebox[origin=c]{90}{\textbf{Scenario 2}} &
        \raisebox{-0.5\height}{\includegraphics[width=0.248\linewidth]{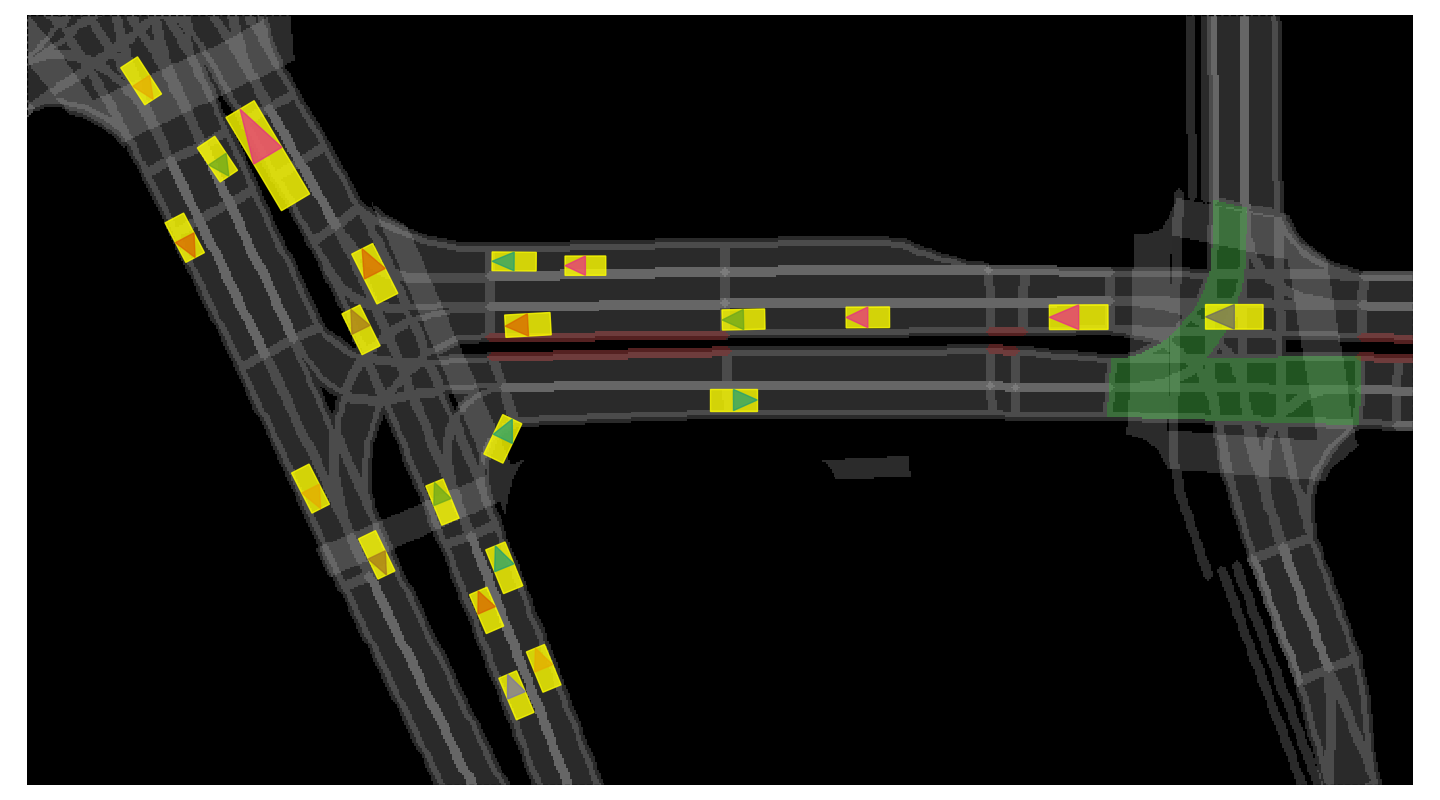}} &
        \raisebox{-0.5\height}{\includegraphics[width=0.248\linewidth]{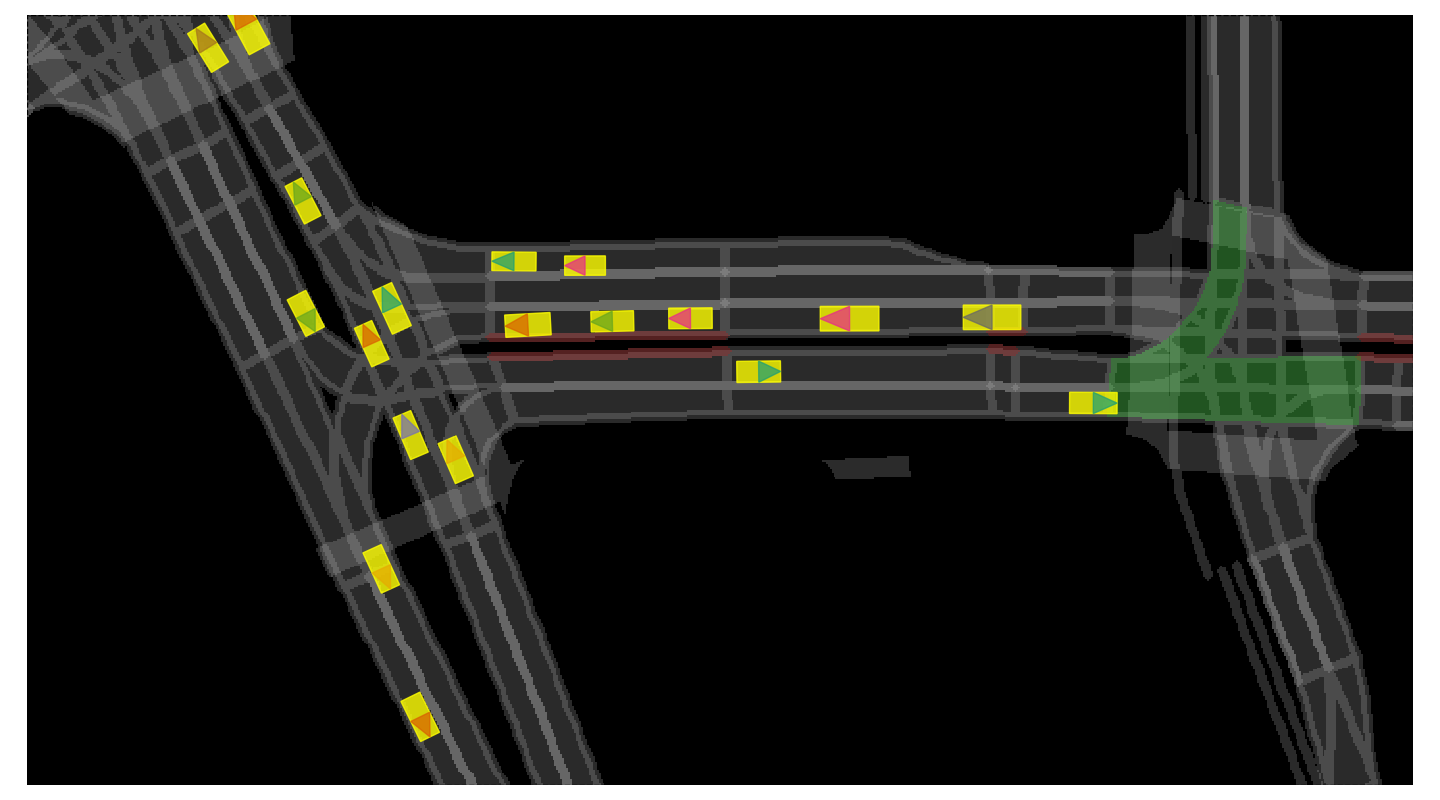}} &
        \raisebox{-0.5\height}{\includegraphics[width=0.248\linewidth]{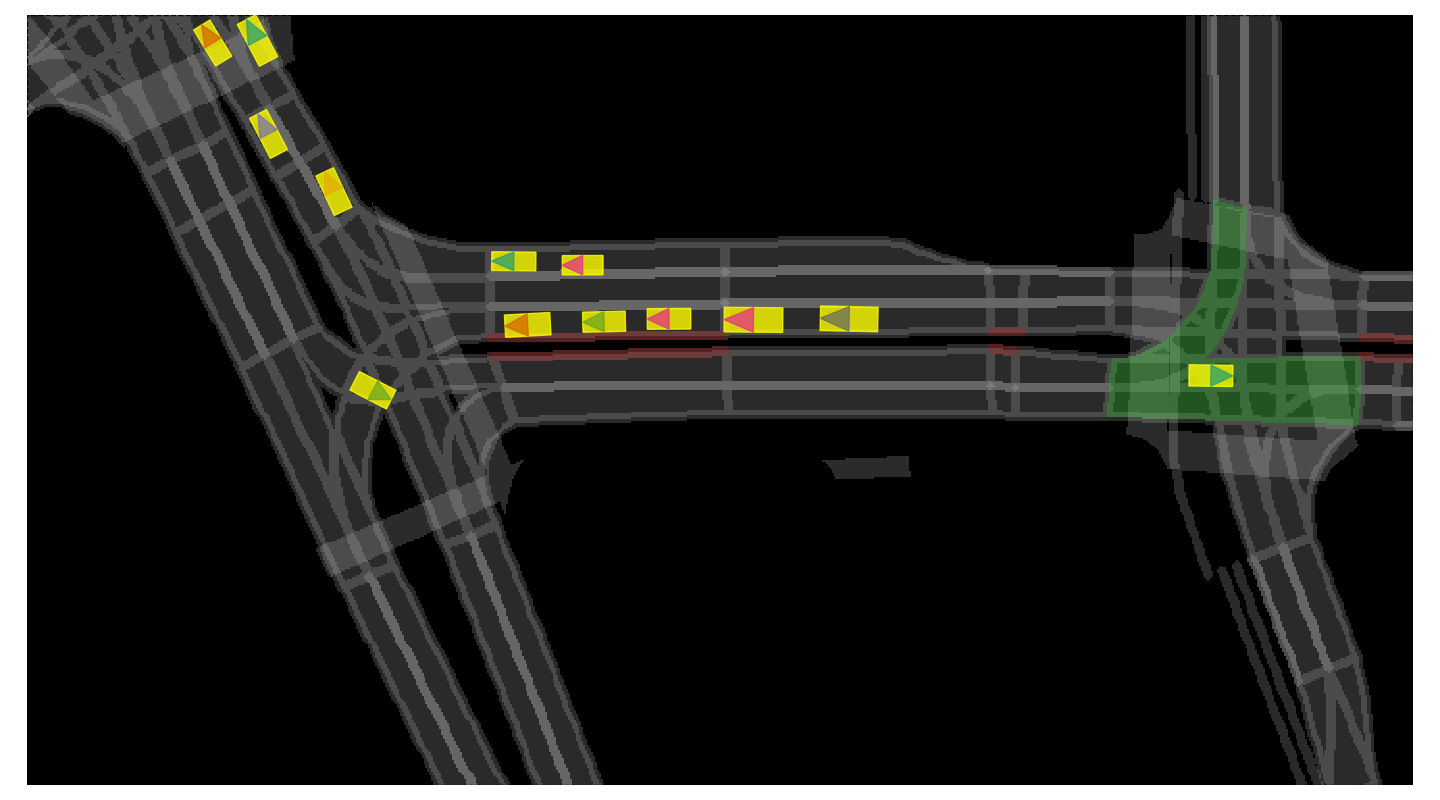}} &
        \raisebox{-0.5\height}{\includegraphics[width=0.248\linewidth]{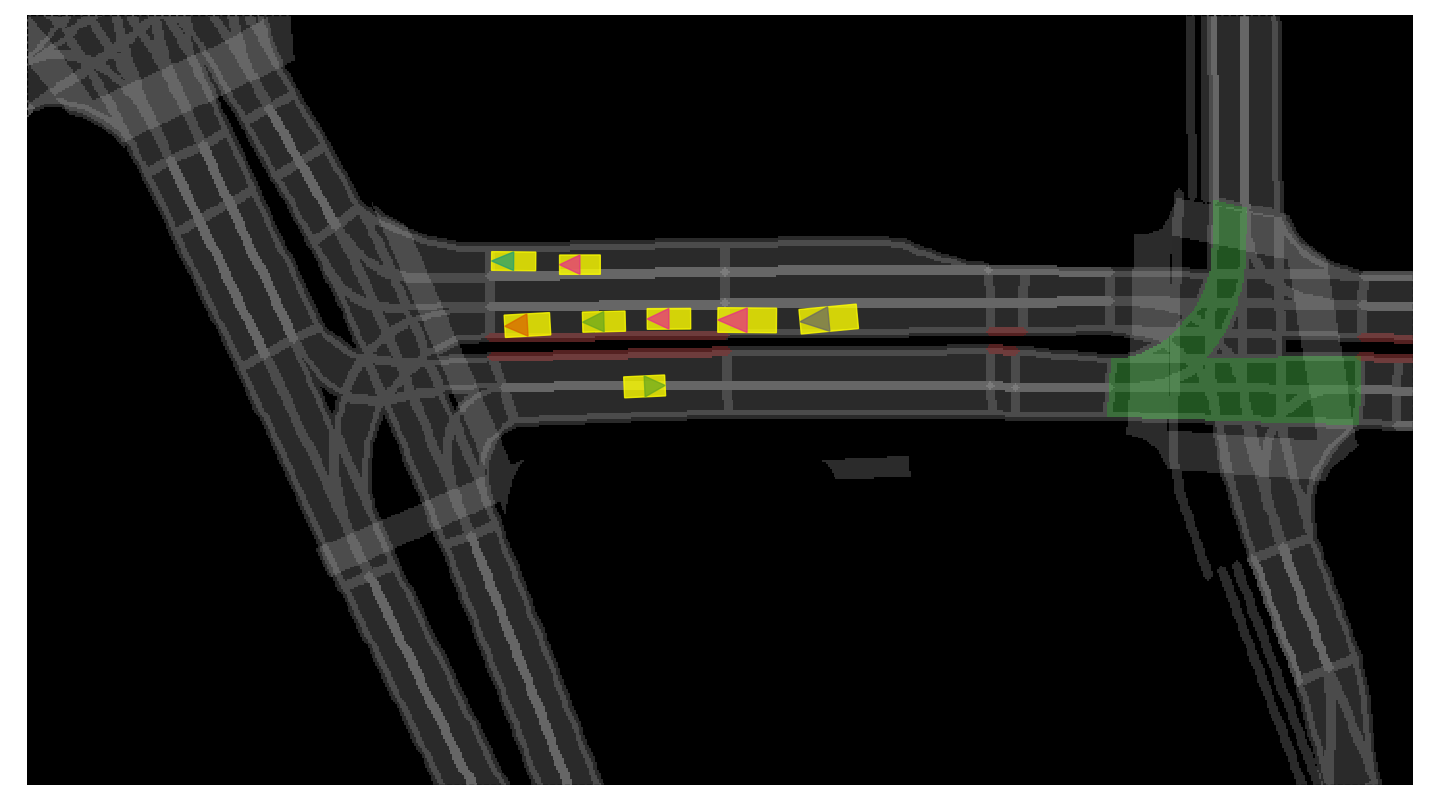}} \vspace{.1em} \\
        \rotatebox[origin=c]{90}{\textbf{Scenario 3}} &
        \raisebox{-0.5\height}{\includegraphics[width=0.248\linewidth]{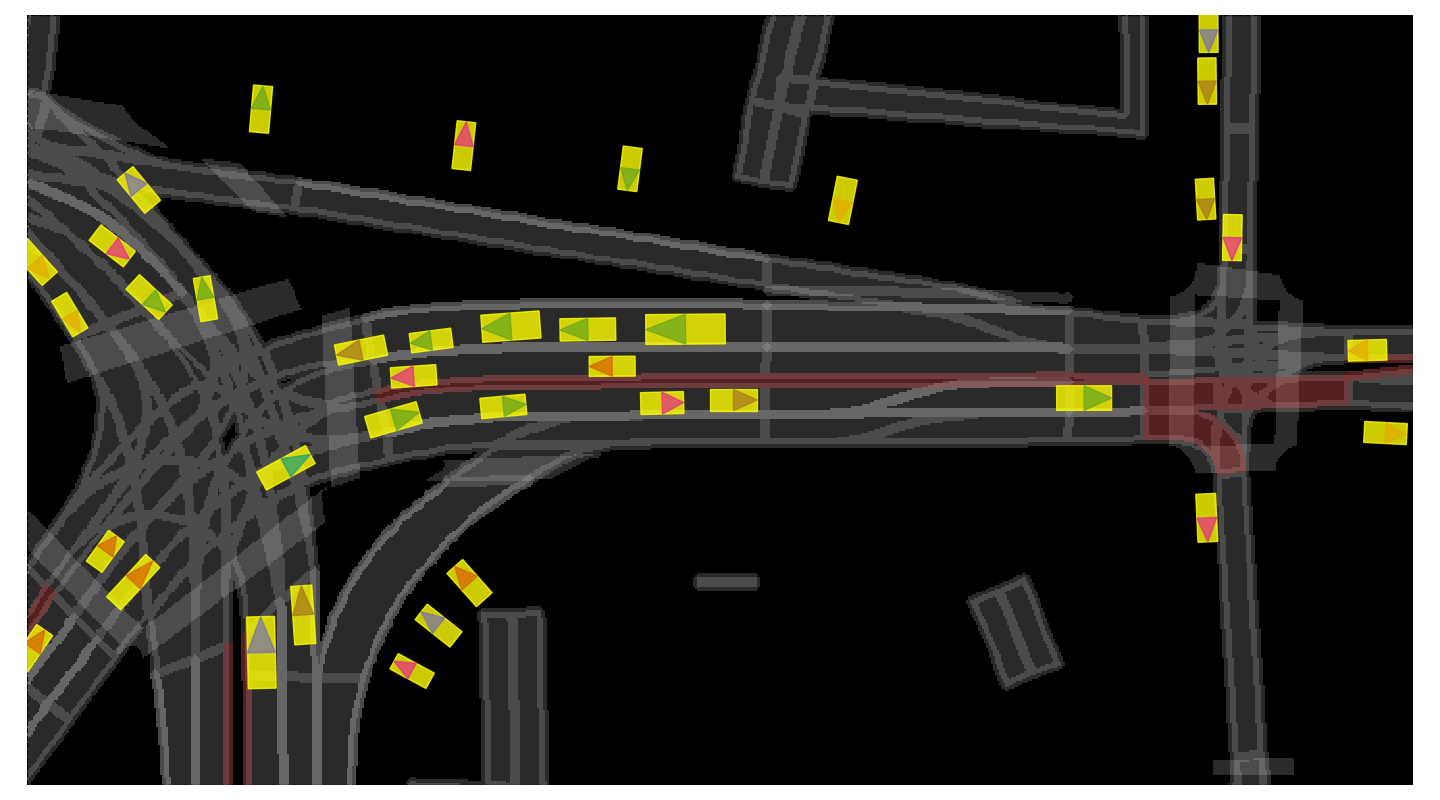}} &
        \raisebox{-0.5\height}{\includegraphics[width=0.248\linewidth]{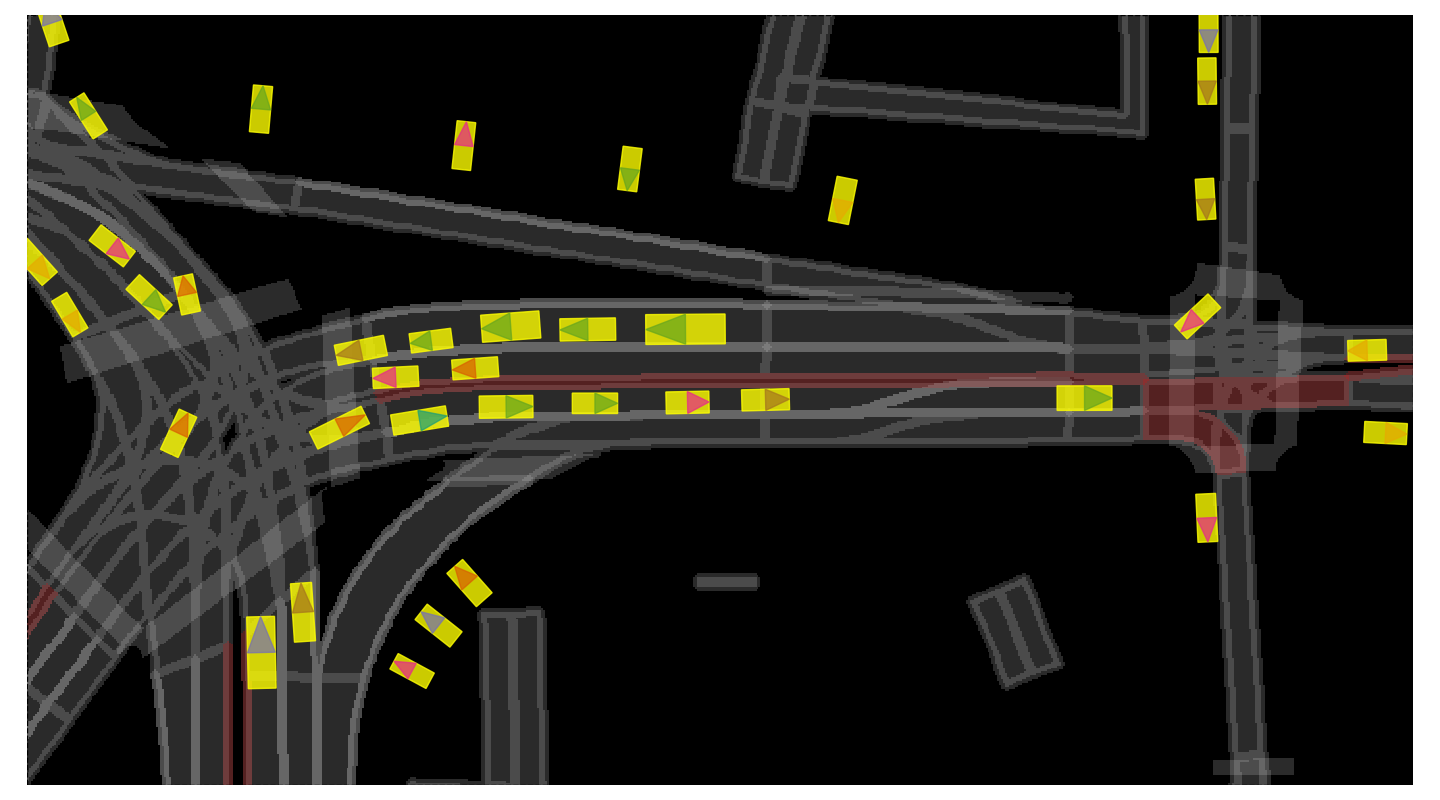}} &
        \raisebox{-0.5\height}{\includegraphics[width=0.248\linewidth]{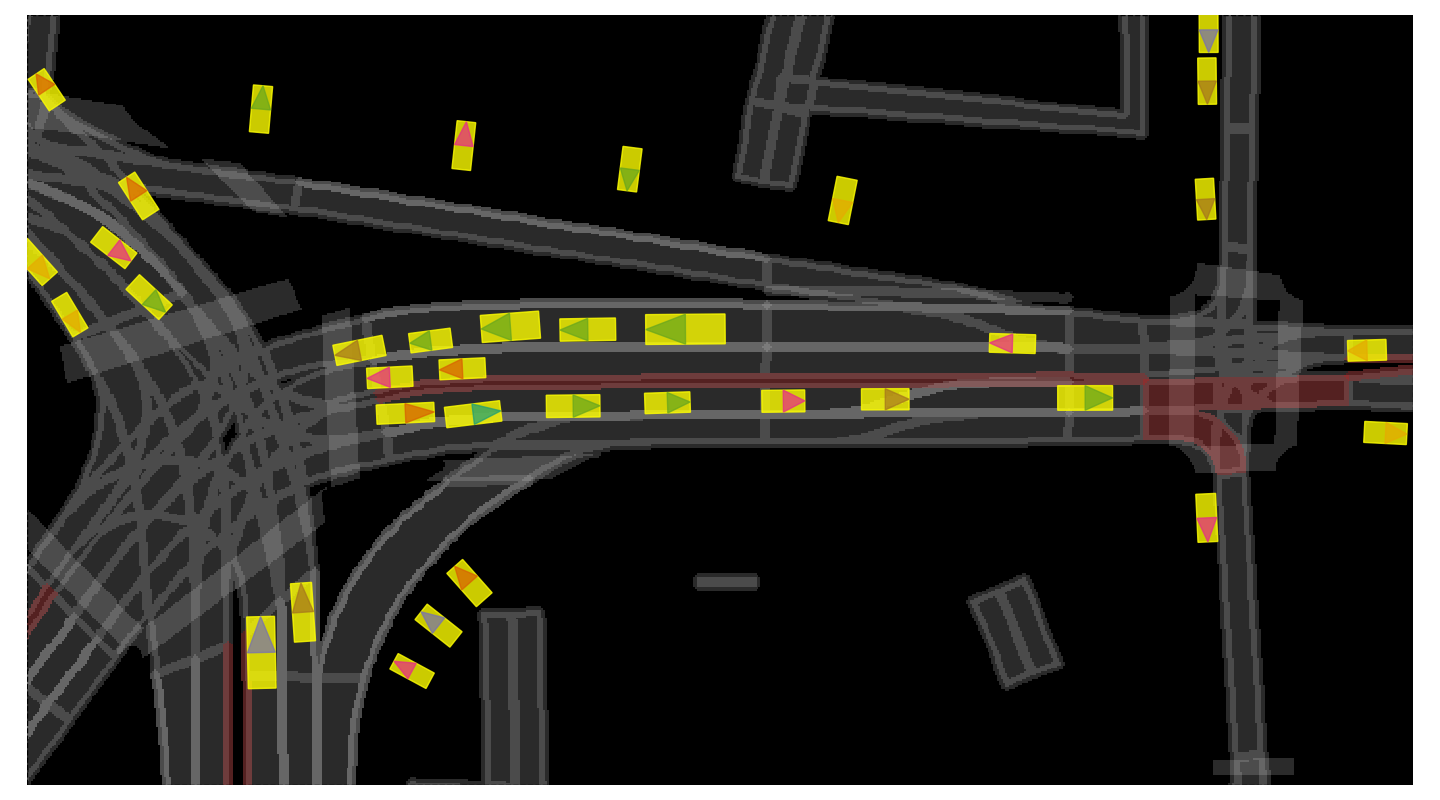}} &
        \raisebox{-0.5\height}{\includegraphics[width=0.248\linewidth]{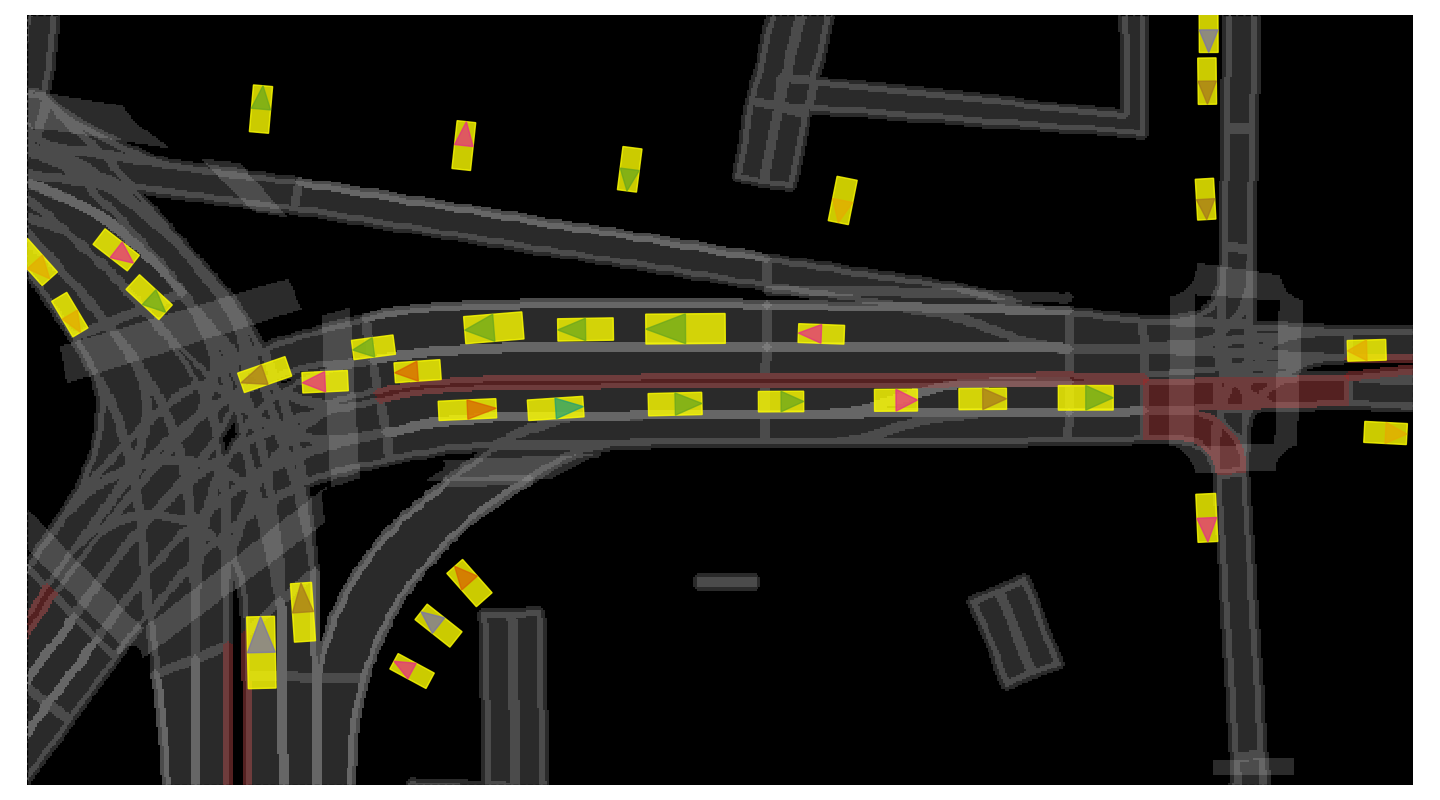}} \vspace{.1em} \\
        \rotatebox[origin=c]{90}{\textbf{Scenario 4}} &
        \raisebox{-0.5\height}{\includegraphics[width=0.248\linewidth]{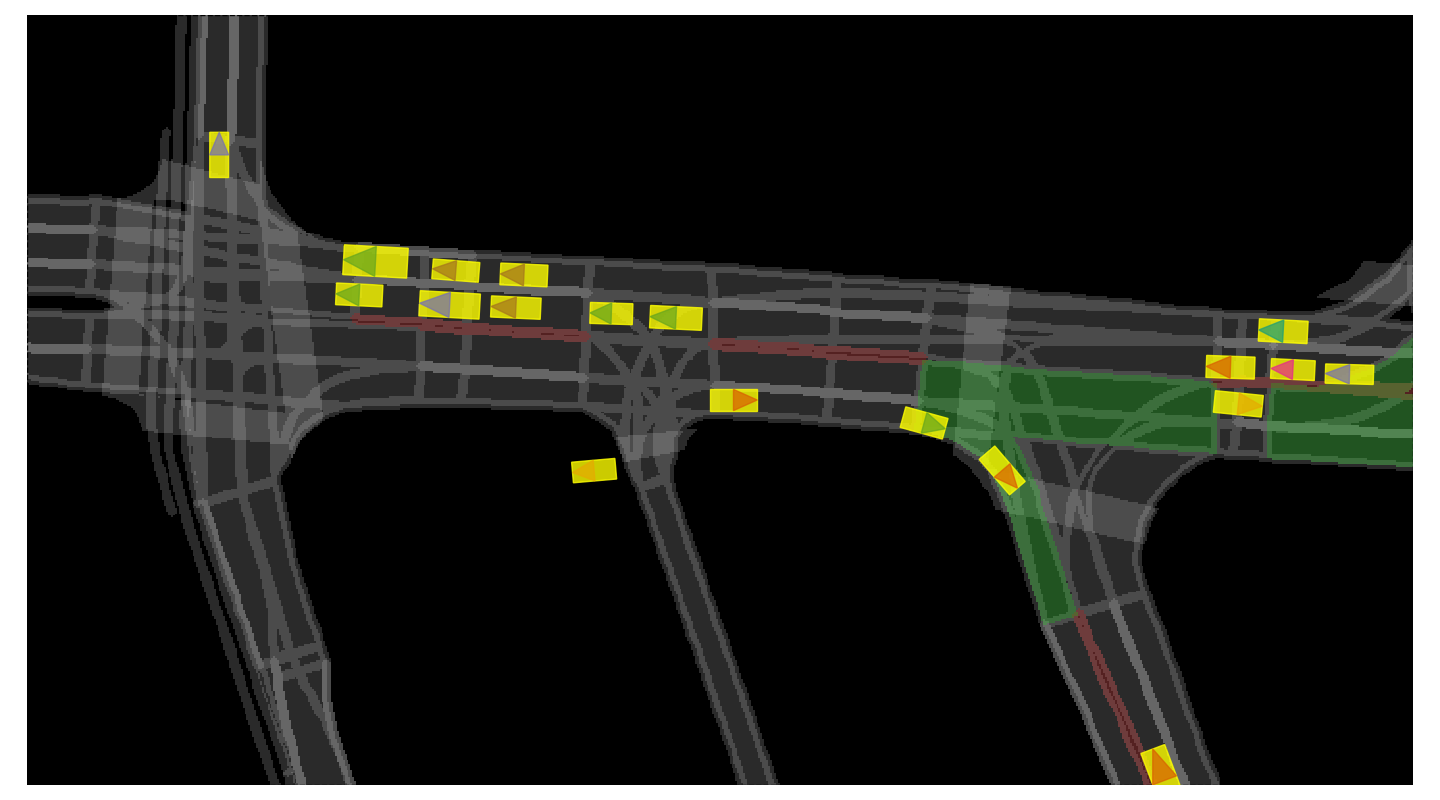}} &
        \raisebox{-0.5\height}{\includegraphics[width=0.248\linewidth]{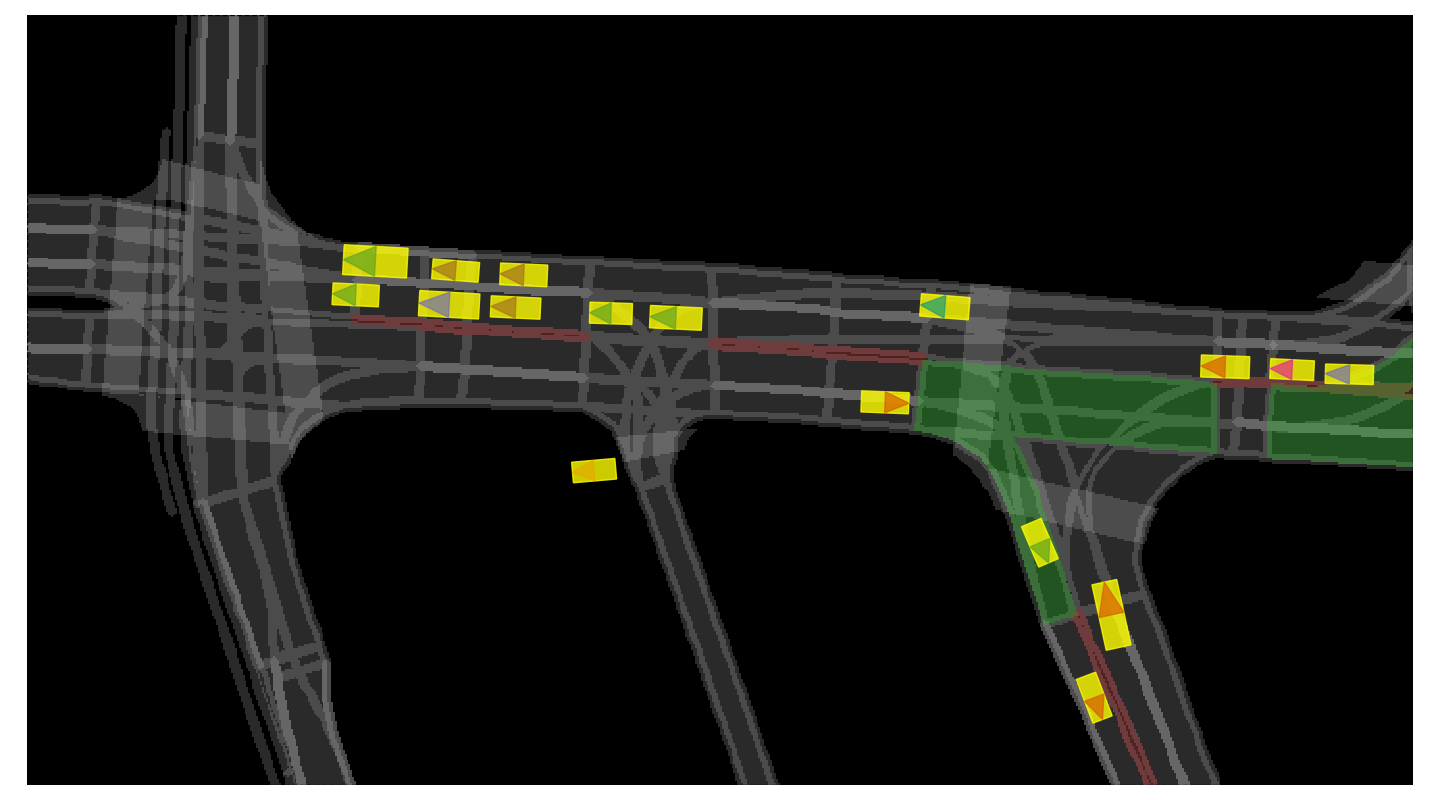}} &
        \raisebox{-0.5\height}{\includegraphics[width=0.248\linewidth]{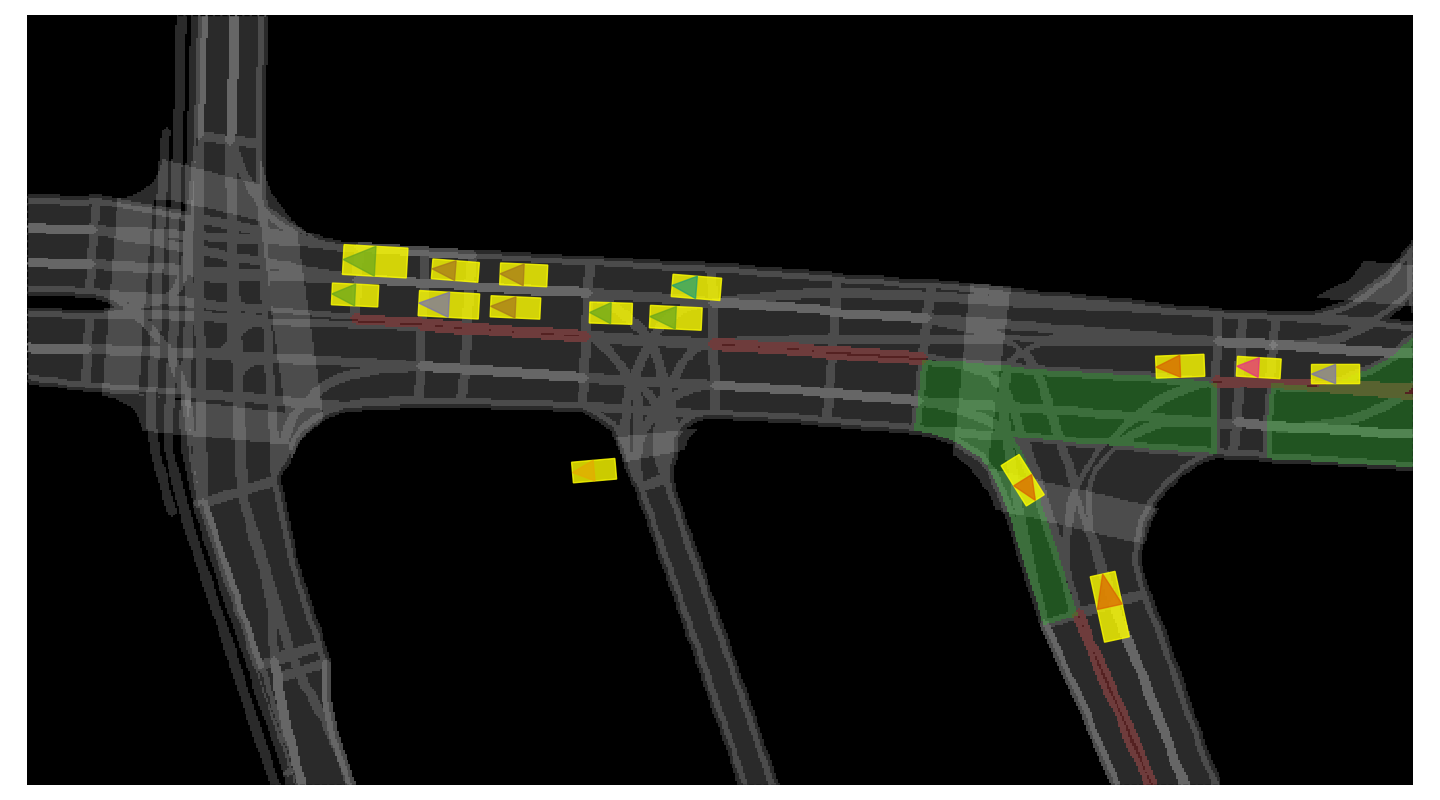}} &
        \raisebox{-0.5\height}{\includegraphics[width=0.248\linewidth]{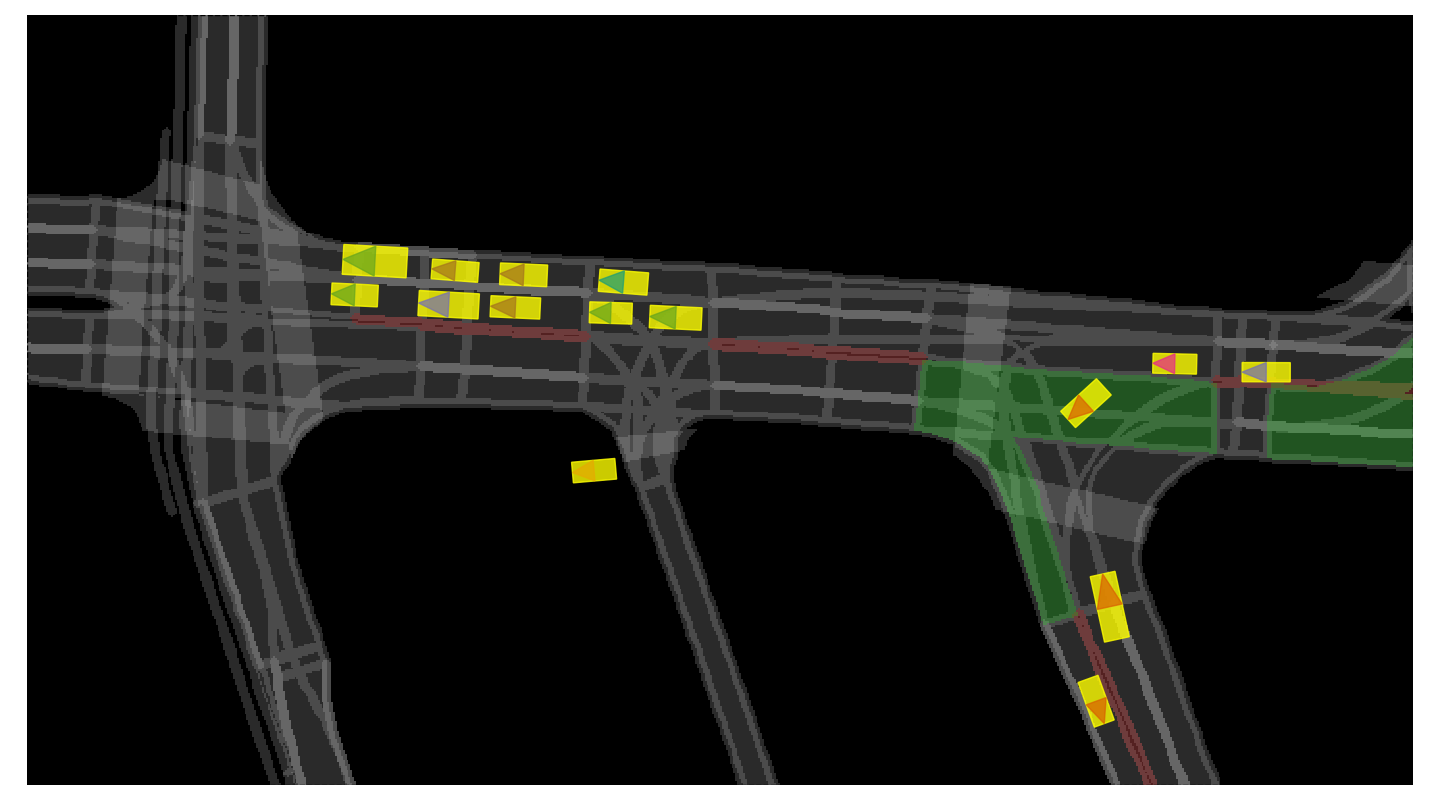}} \vspace{.1em} \\
        \rotatebox[origin=c]{90}{\textbf{Scenario 5}} &
        \raisebox{-0.5\height}{\includegraphics[width=0.248\linewidth]{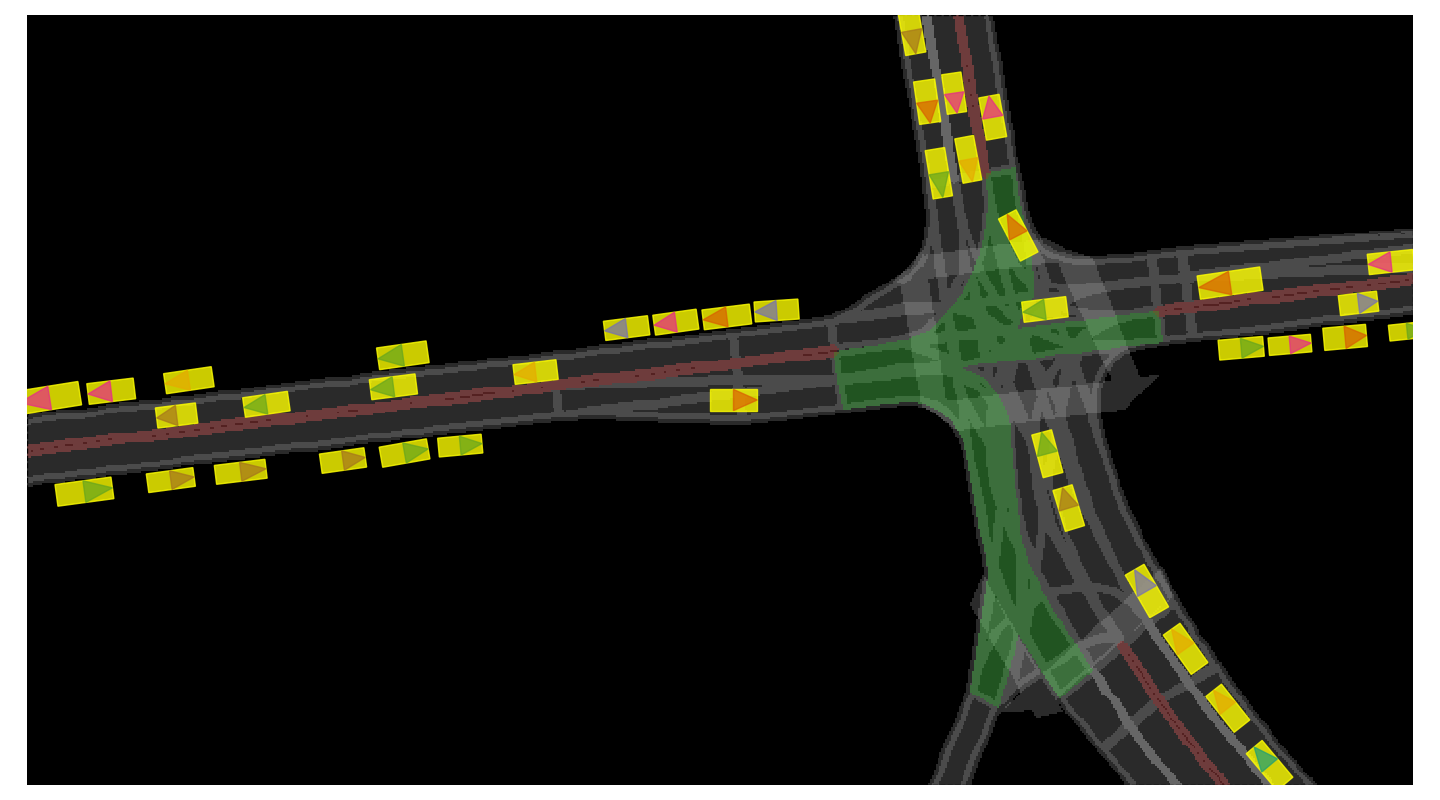}} &
        \raisebox{-0.5\height}{\includegraphics[width=0.248\linewidth]{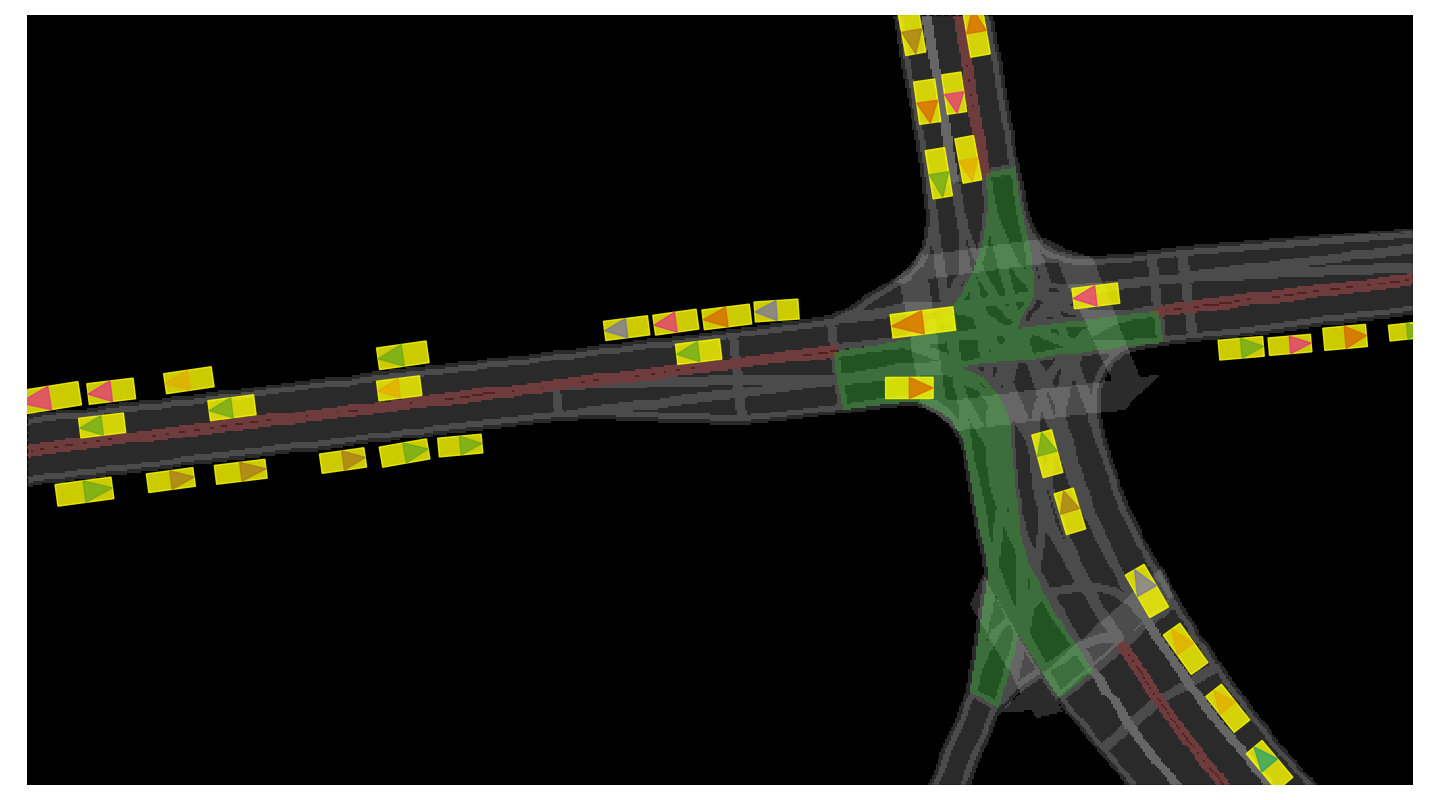}} &
        \raisebox{-0.5\height}{\includegraphics[width=0.248\linewidth]{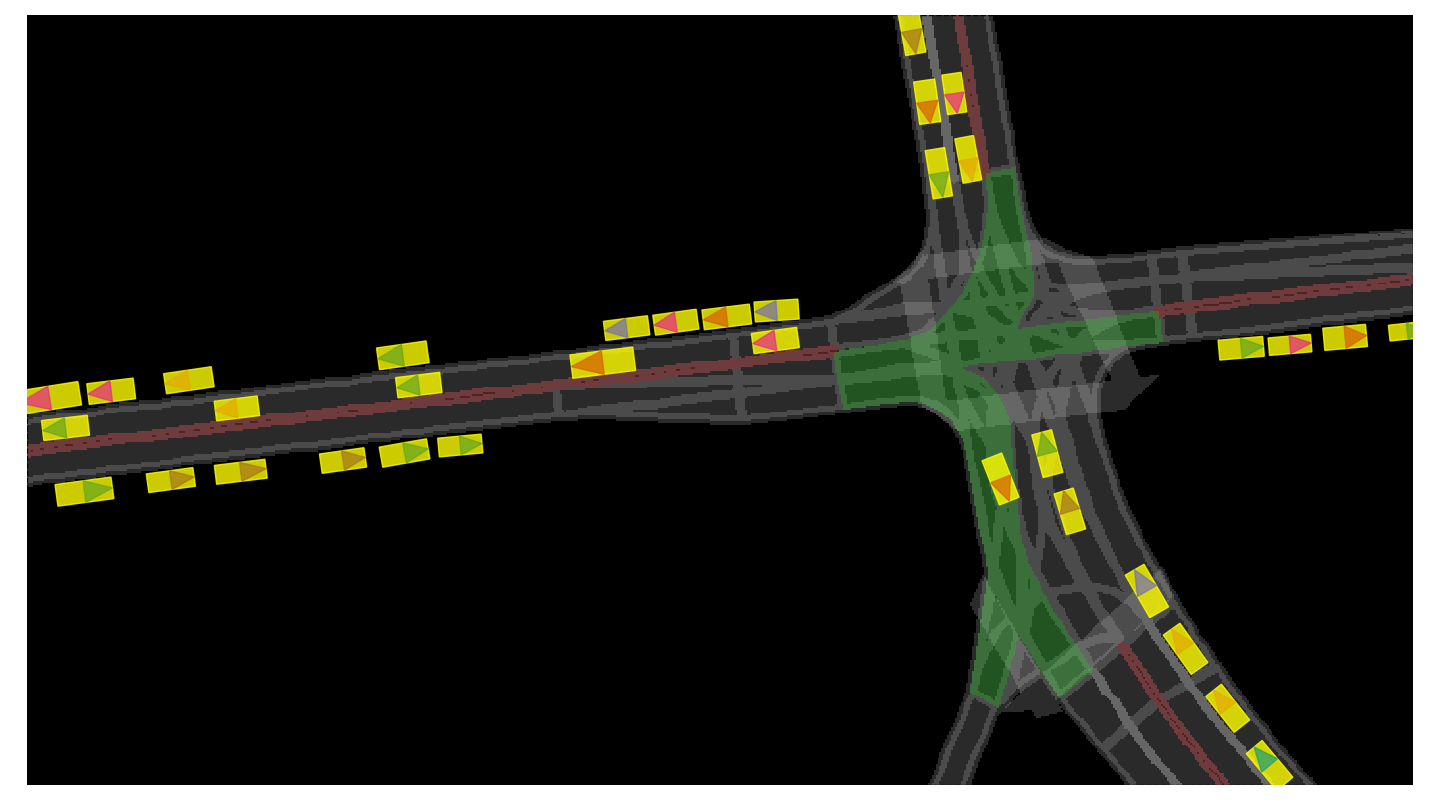}} &
        \raisebox{-0.5\height}{\includegraphics[width=0.248\linewidth]{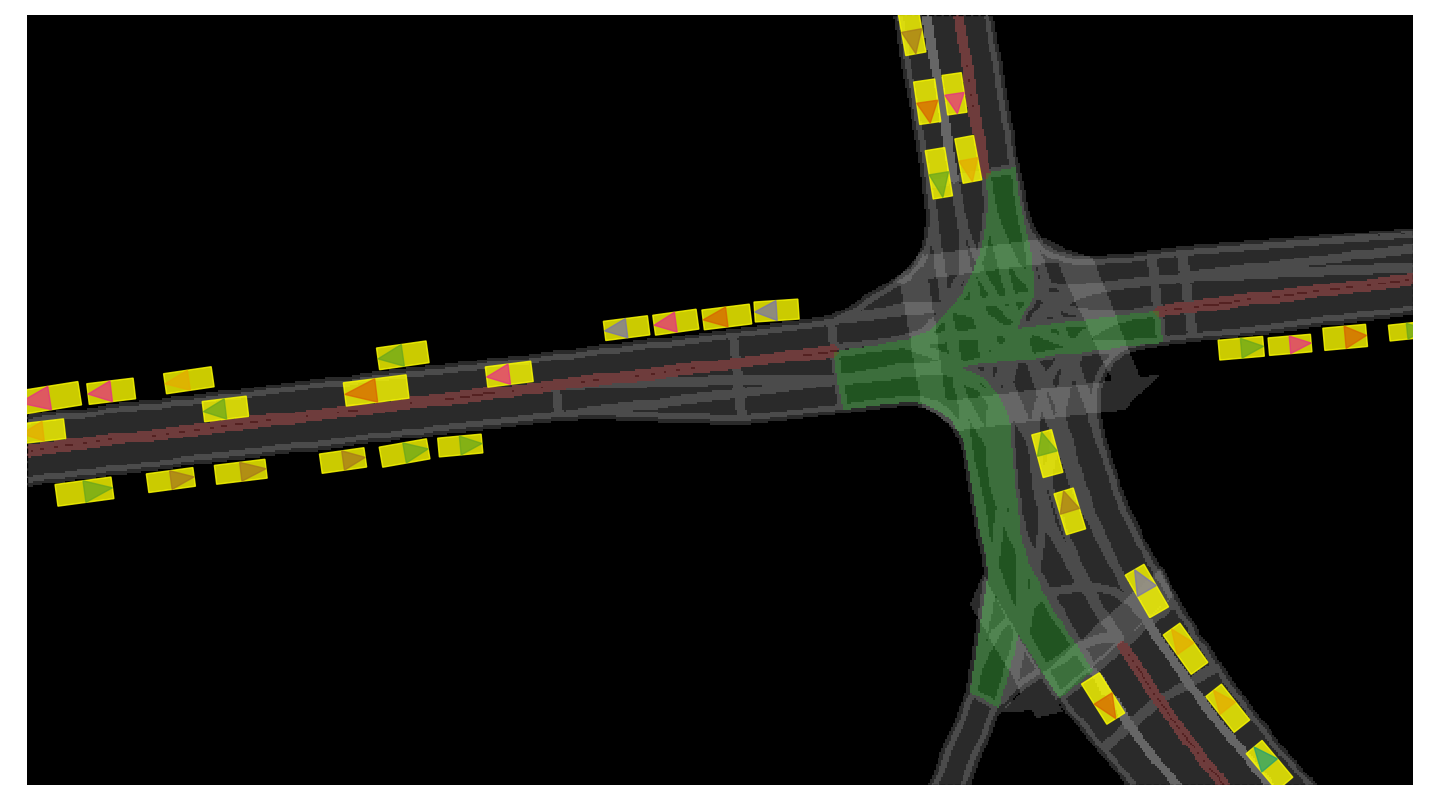}} \vspace{.1em} \\
        \rotatebox[origin=c]{90}{\textbf{Scenario 6}} &
        \raisebox{-0.5\height}{\includegraphics[width=0.248\linewidth]{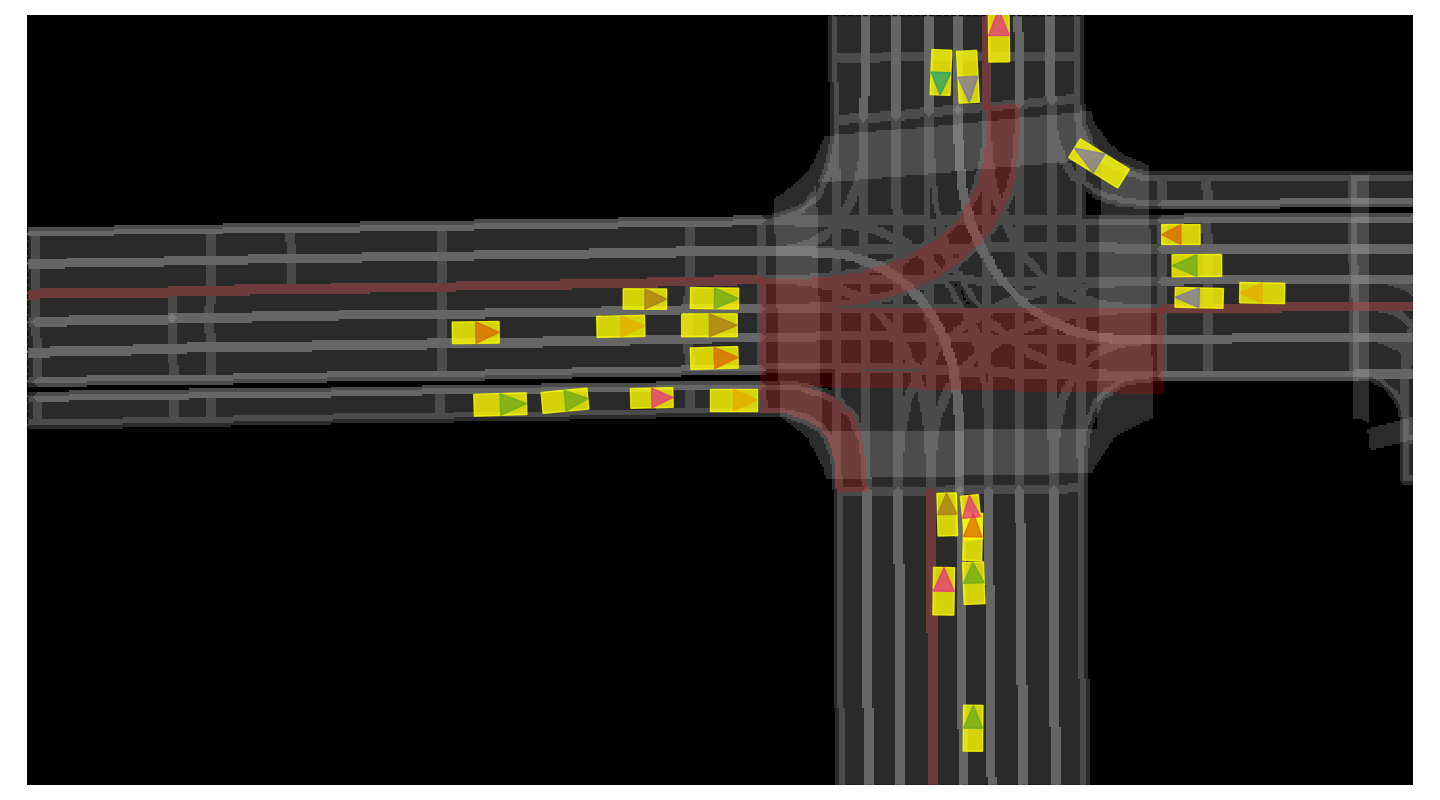}} &
        \raisebox{-0.5\height}{\includegraphics[width=0.248\linewidth]{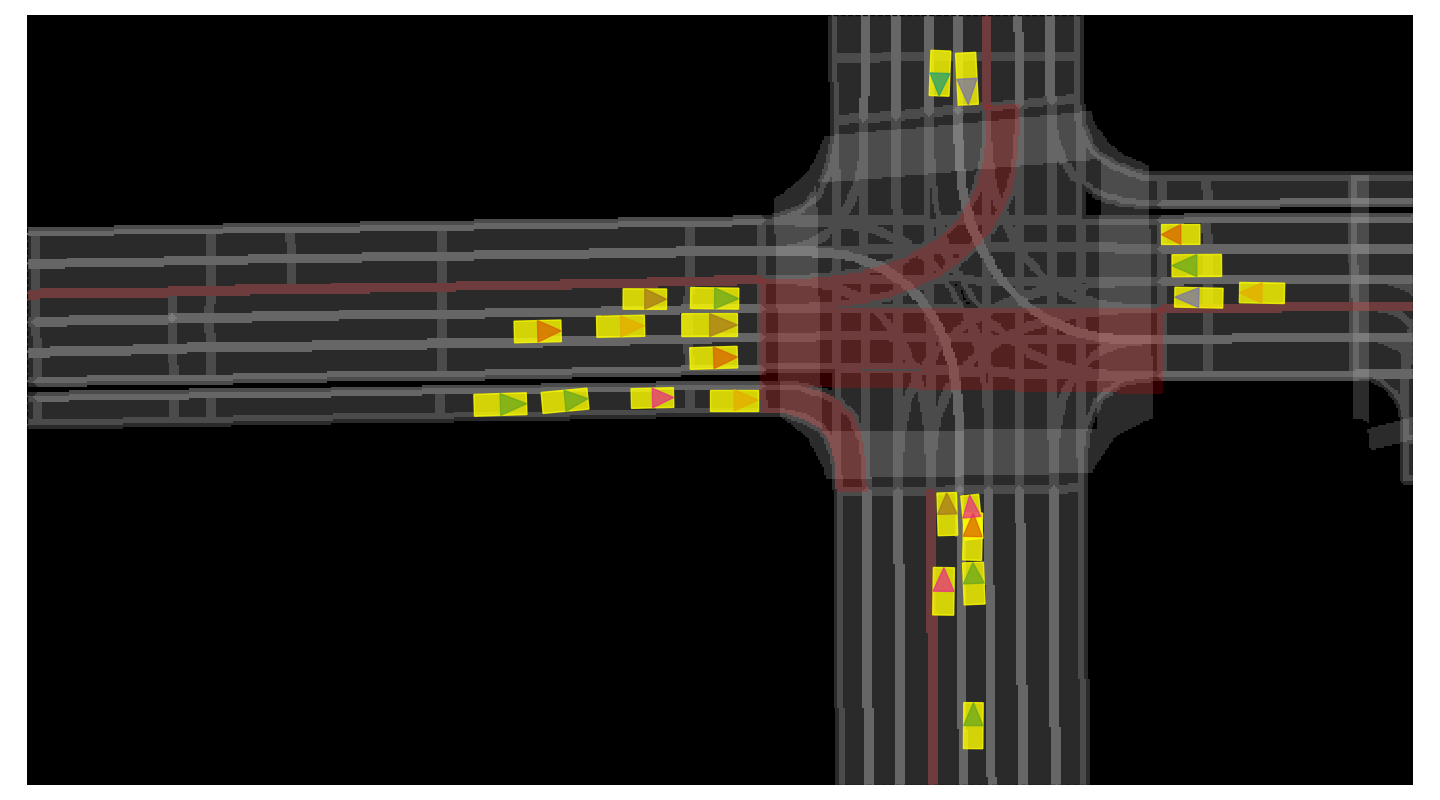}} &
        \raisebox{-0.5\height}{\includegraphics[width=0.248\linewidth]{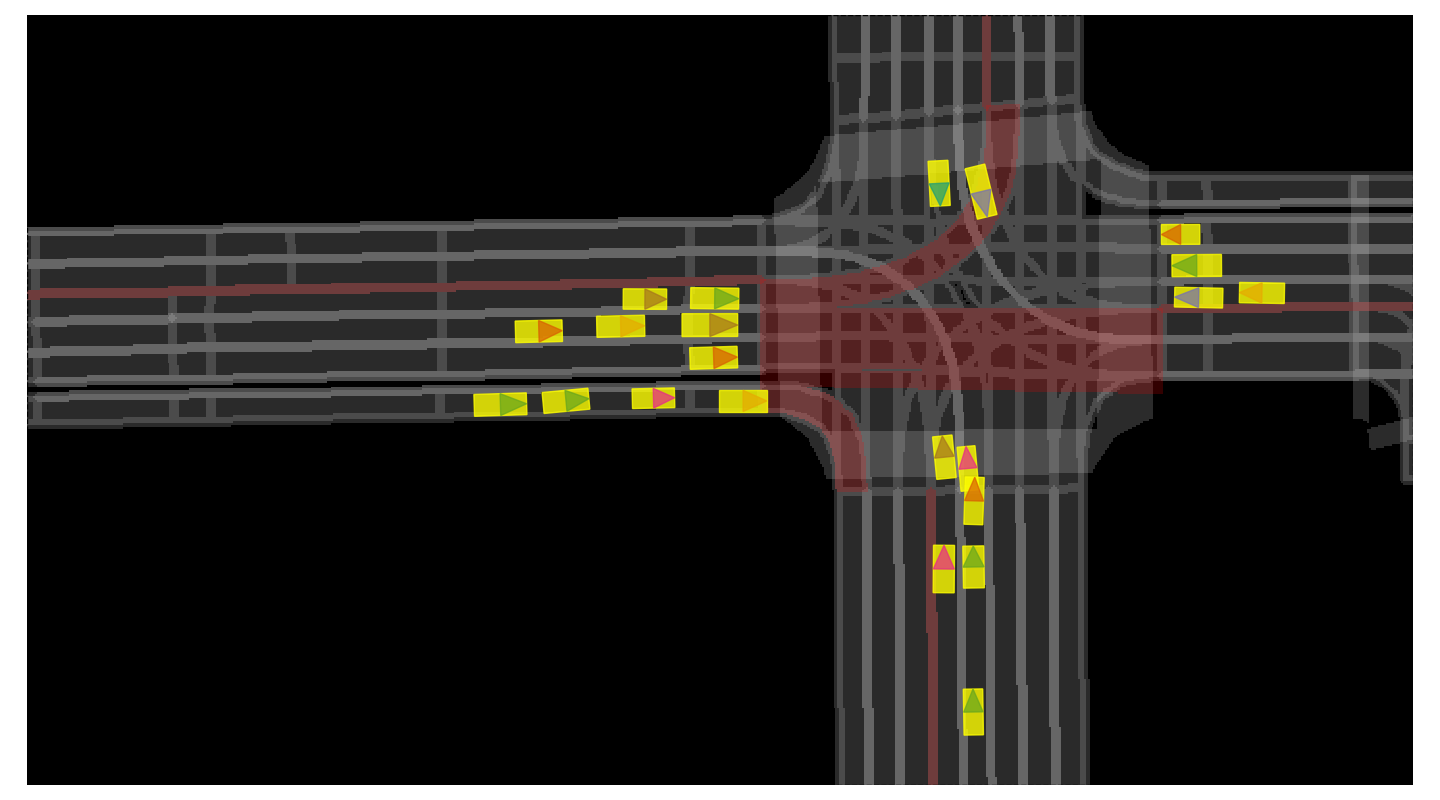}} &
        \raisebox{-0.5\height}{\includegraphics[width=0.248\linewidth]{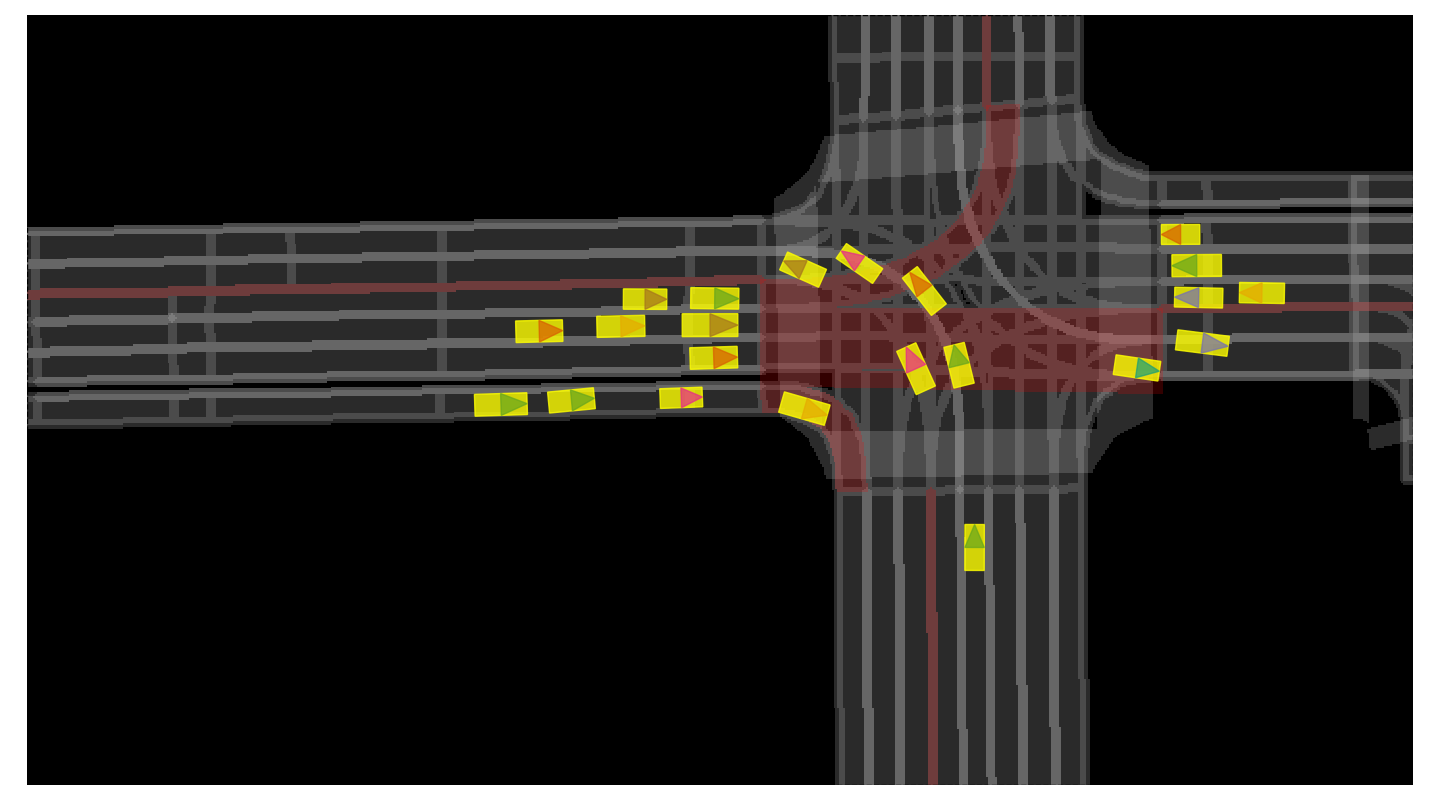}} \vspace{.1em} \\
        \rotatebox[origin=c]{90}{\textbf{Scenario 7}} &
        \raisebox{-0.5\height}{\includegraphics[width=0.248\linewidth]{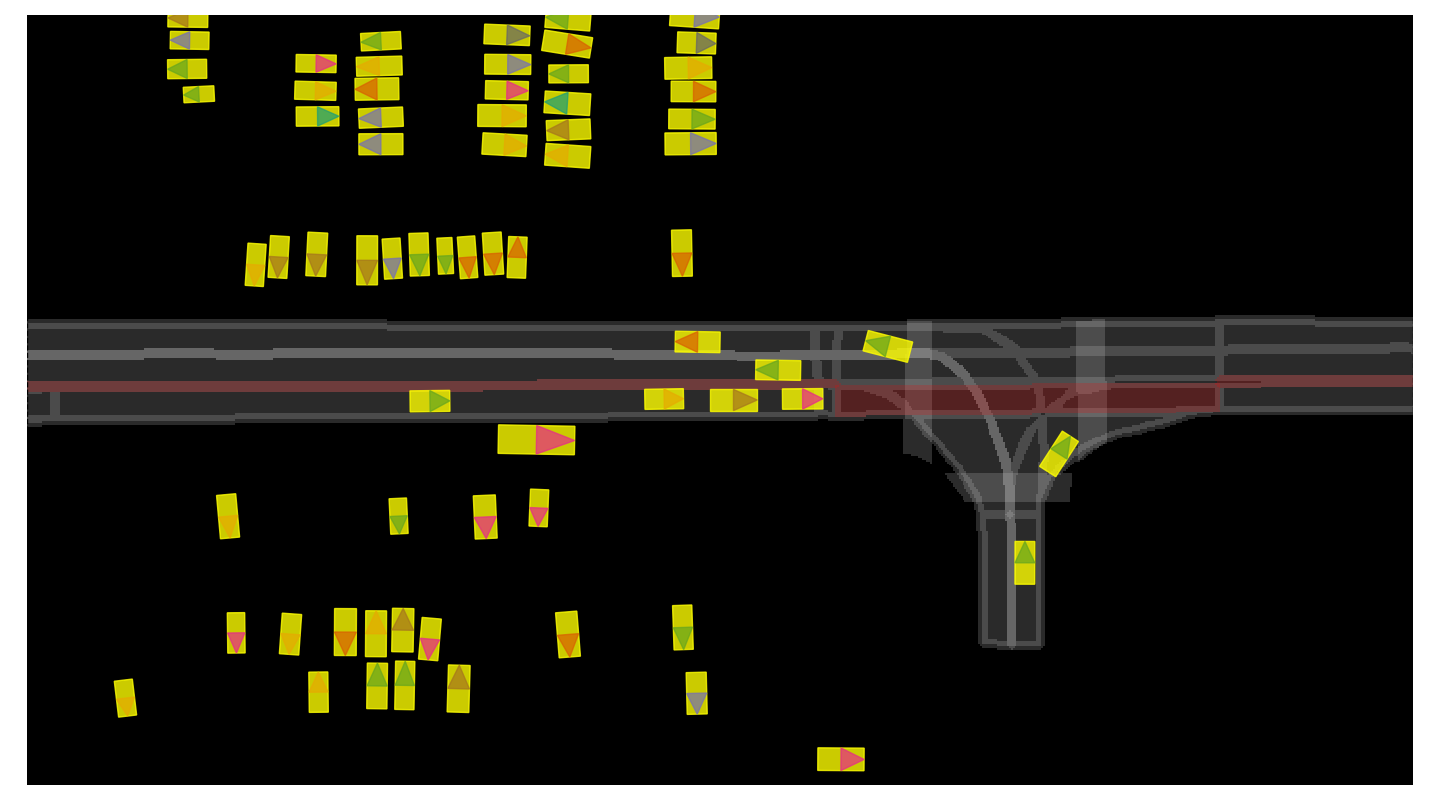}} &
        \raisebox{-0.5\height}{\includegraphics[width=0.248\linewidth]{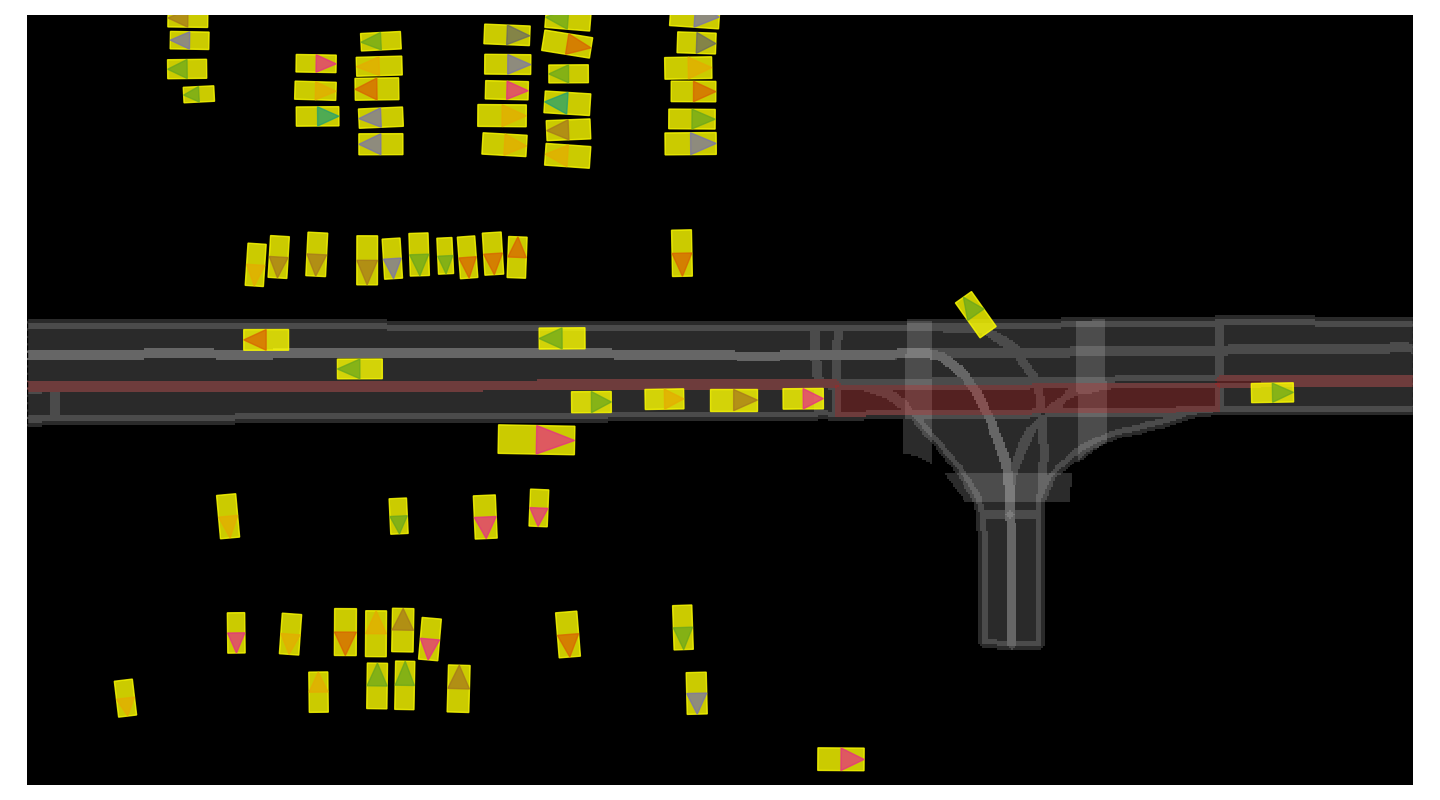}} &
        \raisebox{-0.5\height}{\includegraphics[width=0.248\linewidth]{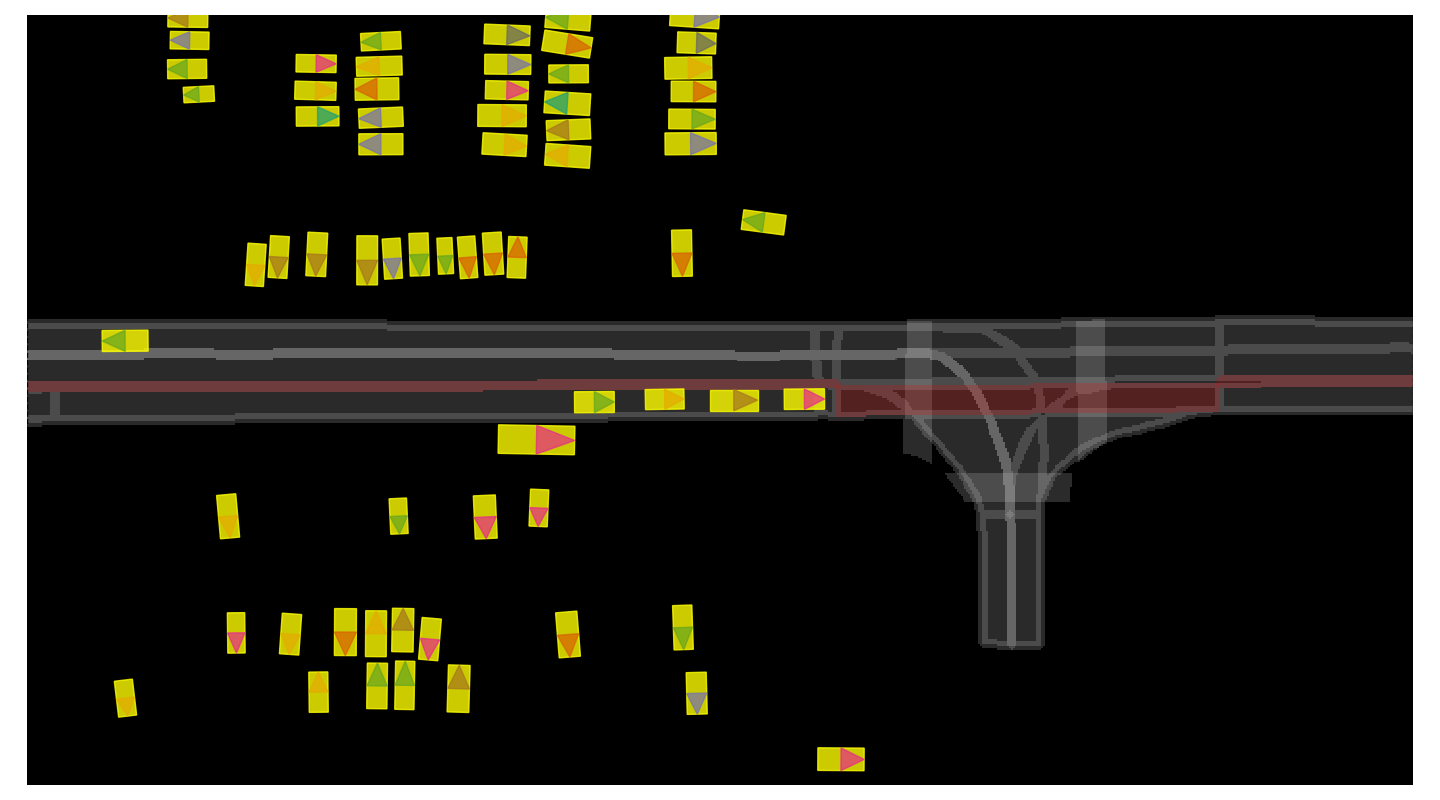}} &
        \raisebox{-0.5\height}{\includegraphics[width=0.248\linewidth]{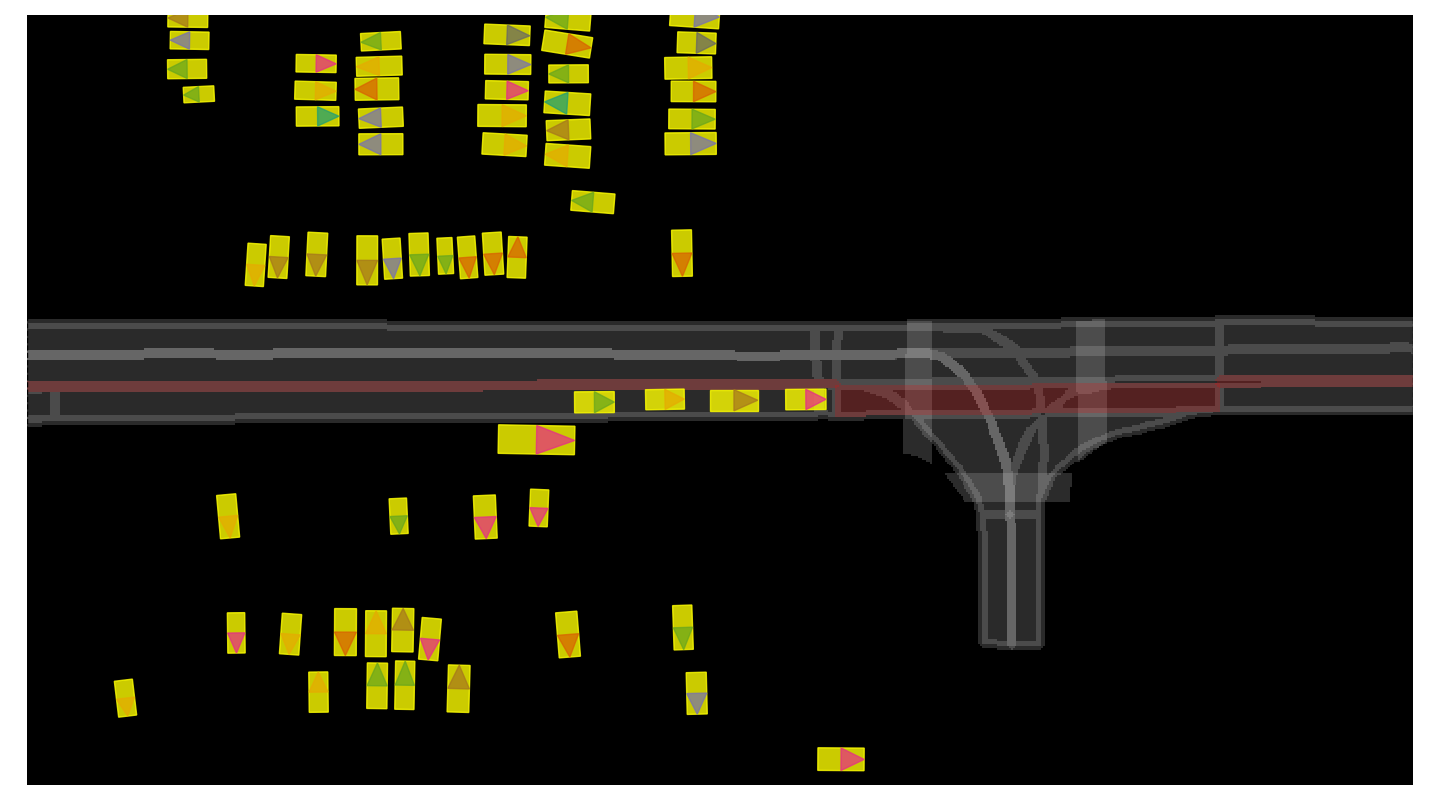}} \vspace{.1em} \\
        \rotatebox[origin=c]{90}{\textbf{Scenario 8}} &
        \raisebox{-0.5\height}{\includegraphics[width=0.248\linewidth]{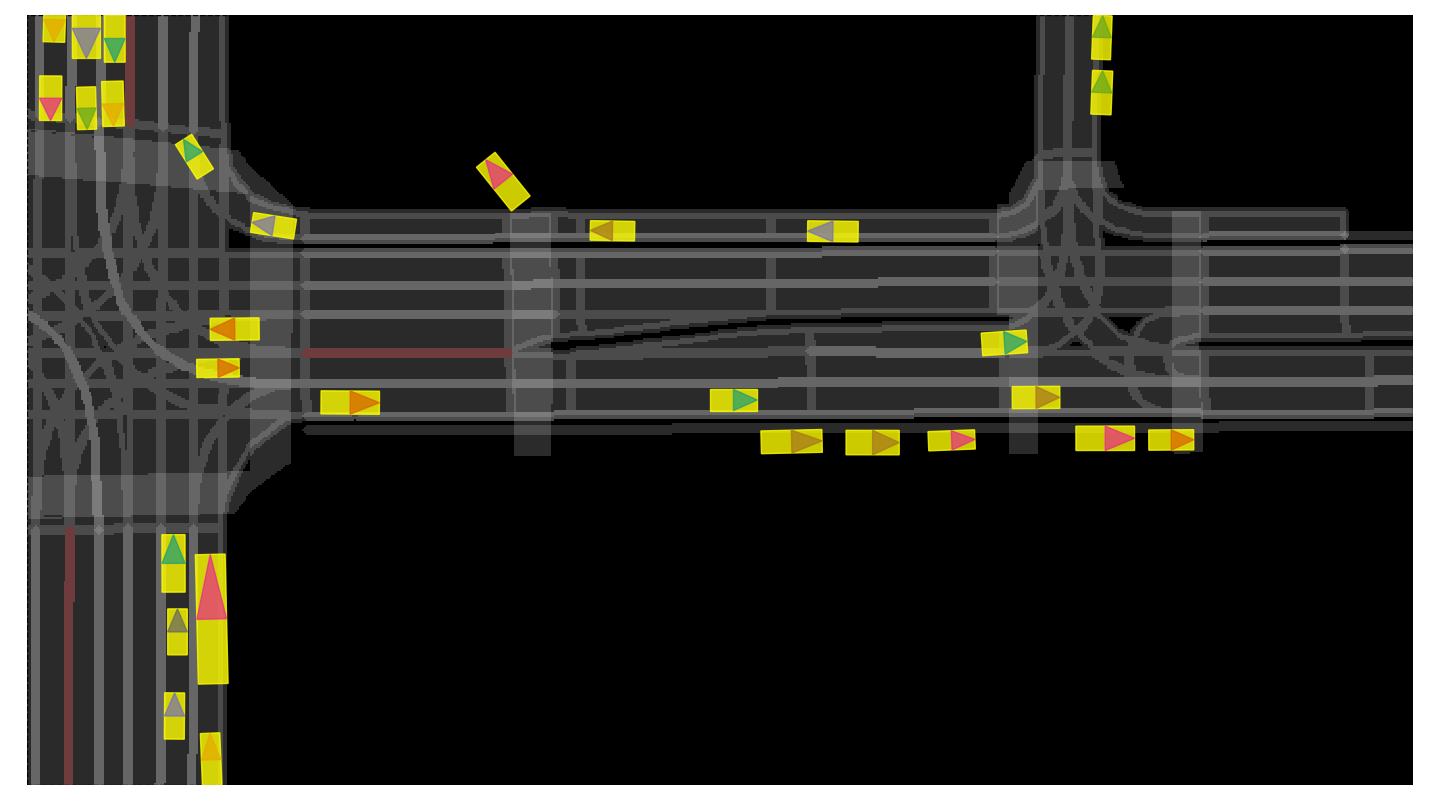}} &
        \raisebox{-0.5\height}{\includegraphics[width=0.248\linewidth]{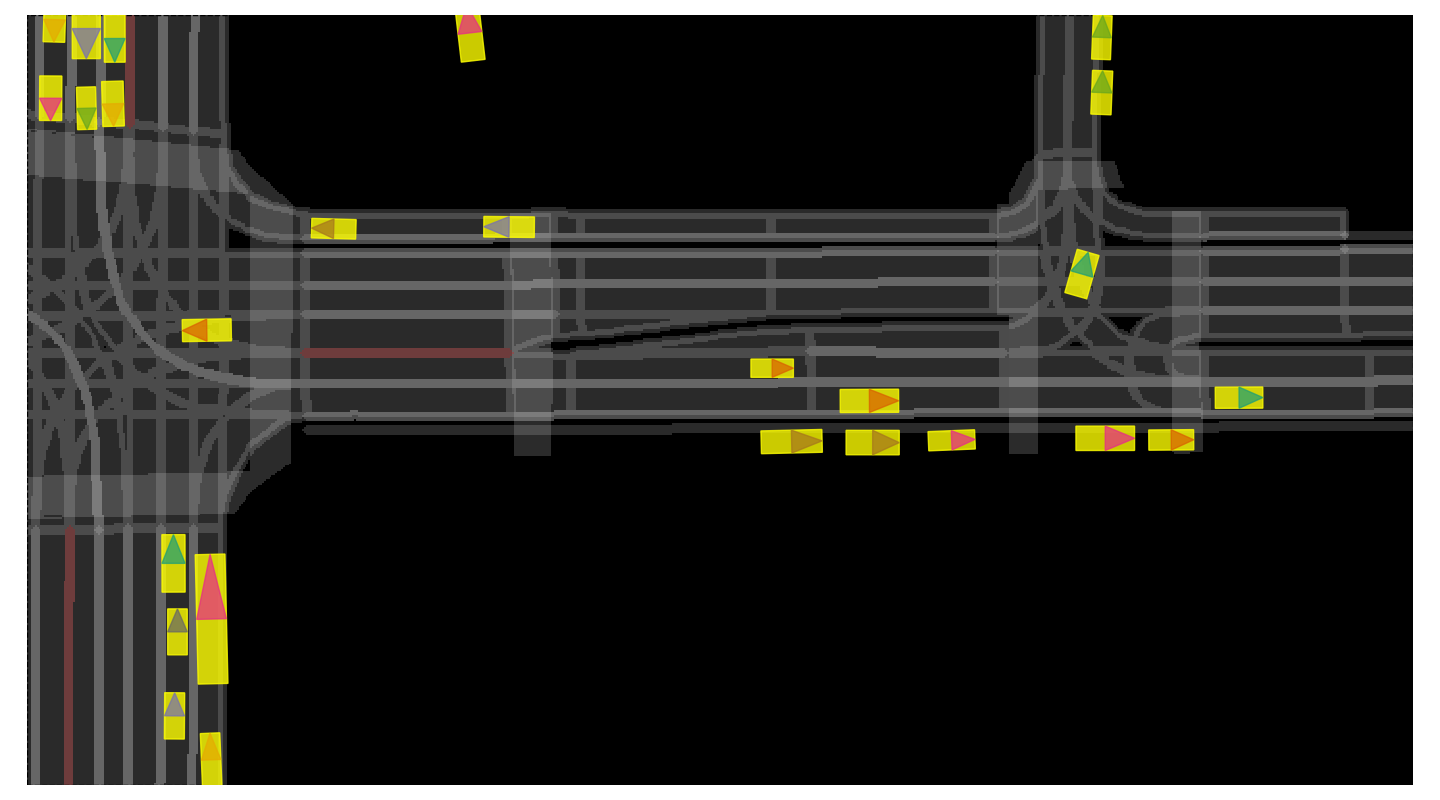}} &
        \raisebox{-0.5\height}{\includegraphics[width=0.248\linewidth]{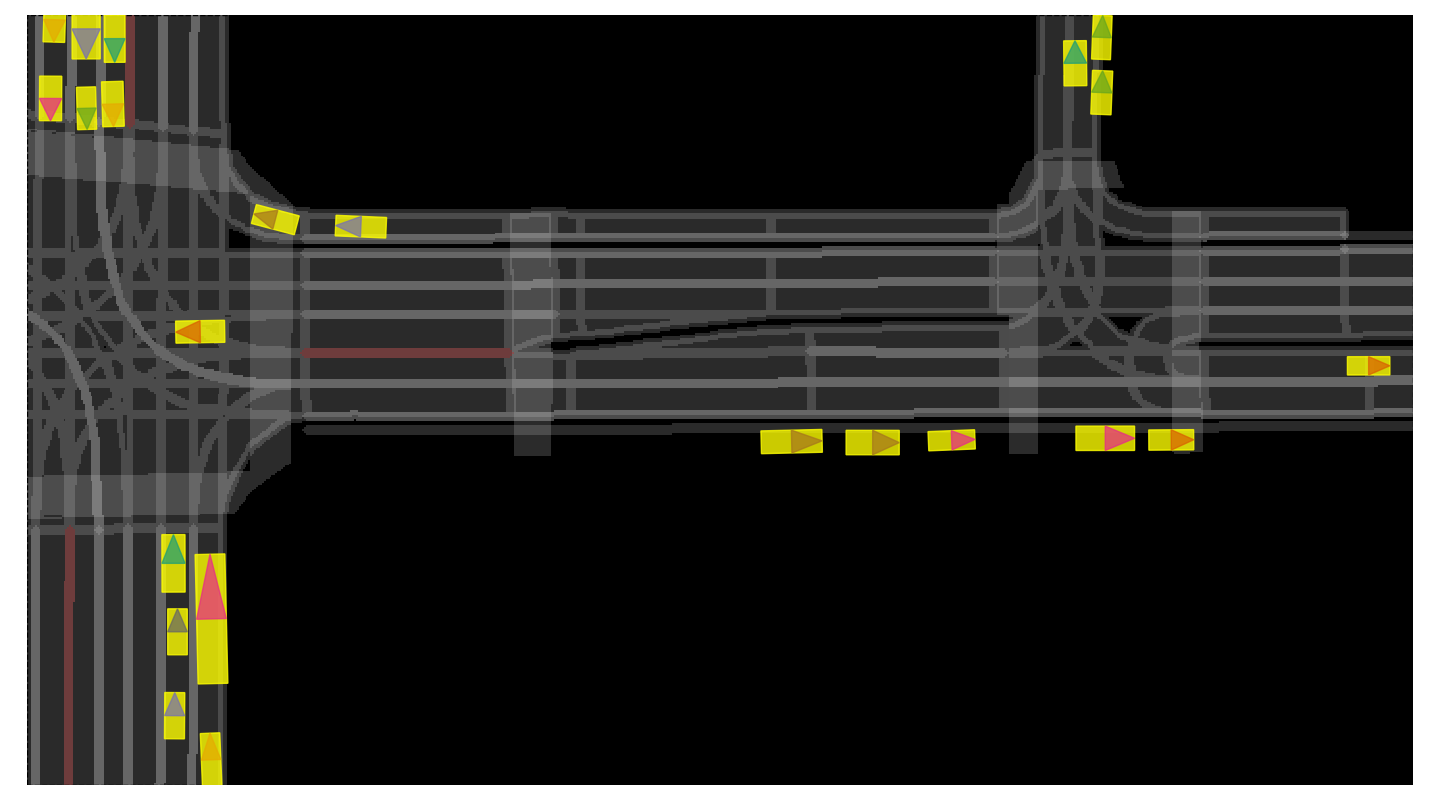}} &
        \raisebox{-0.5\height}{\includegraphics[width=0.248\linewidth]{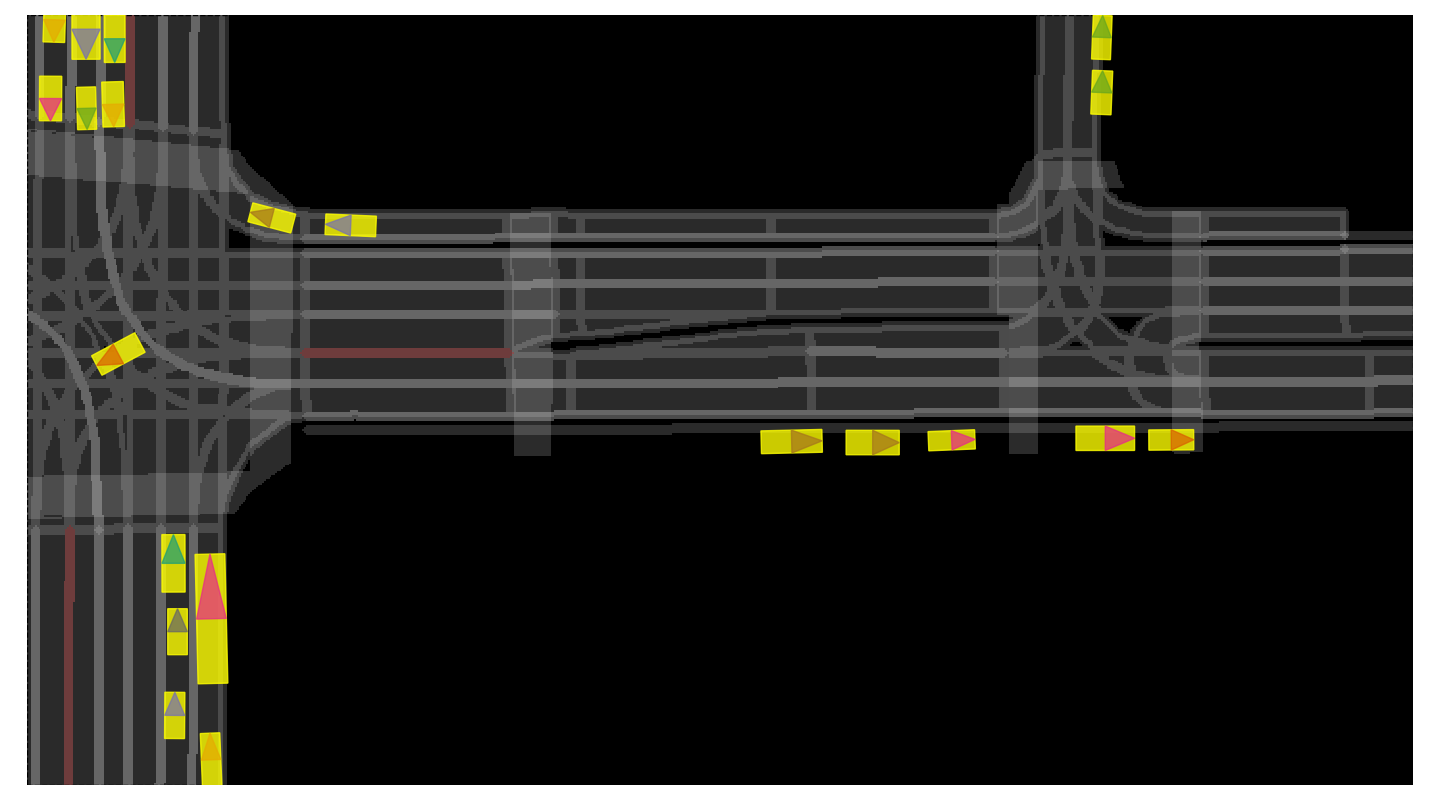}} \vspace{.1em} \\
        \rotatebox[origin=c]{90}{\textbf{Scenario 9}} &
        \raisebox{-0.5\height}{\includegraphics[width=0.248\linewidth]{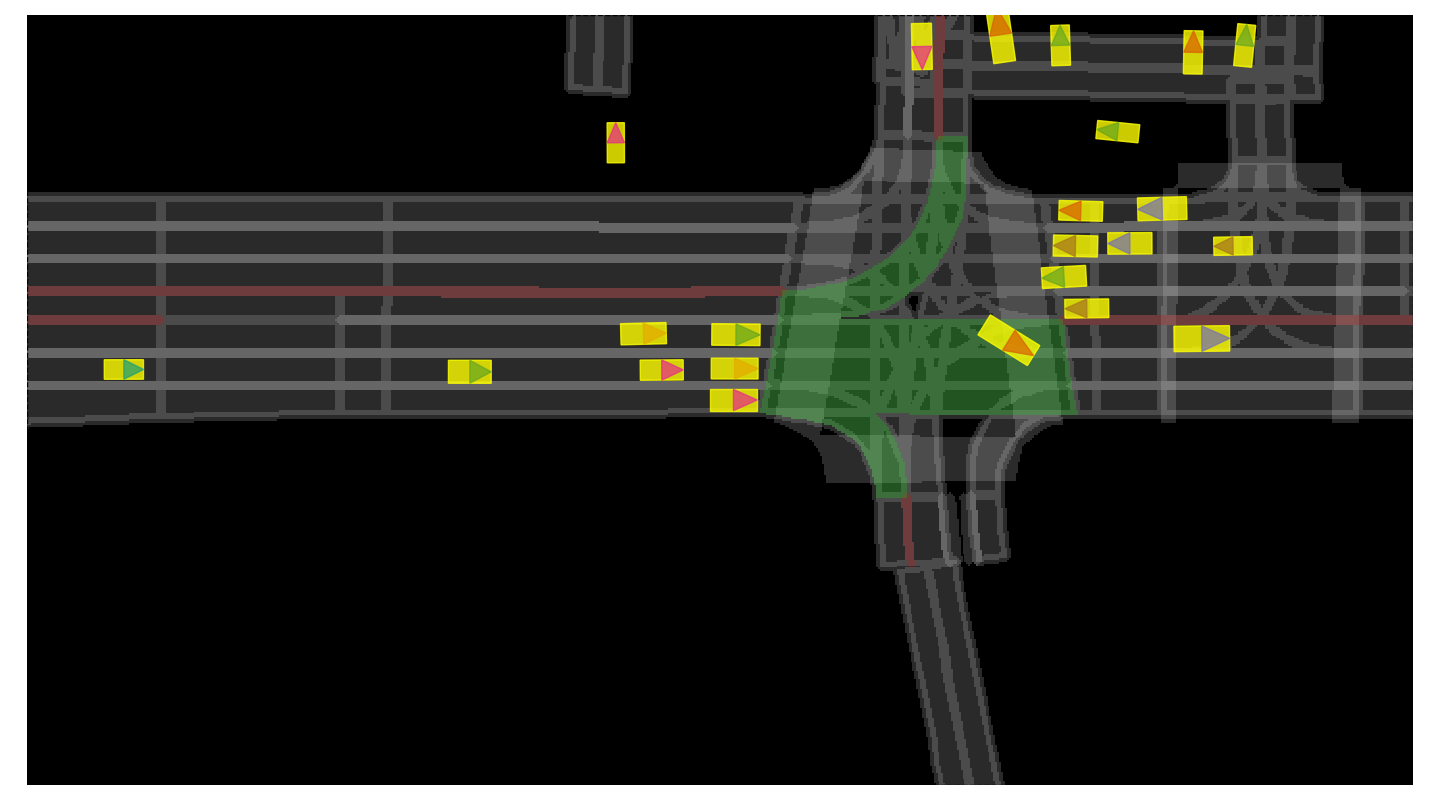}} &
        \raisebox{-0.5\height}{\includegraphics[width=0.248\linewidth]{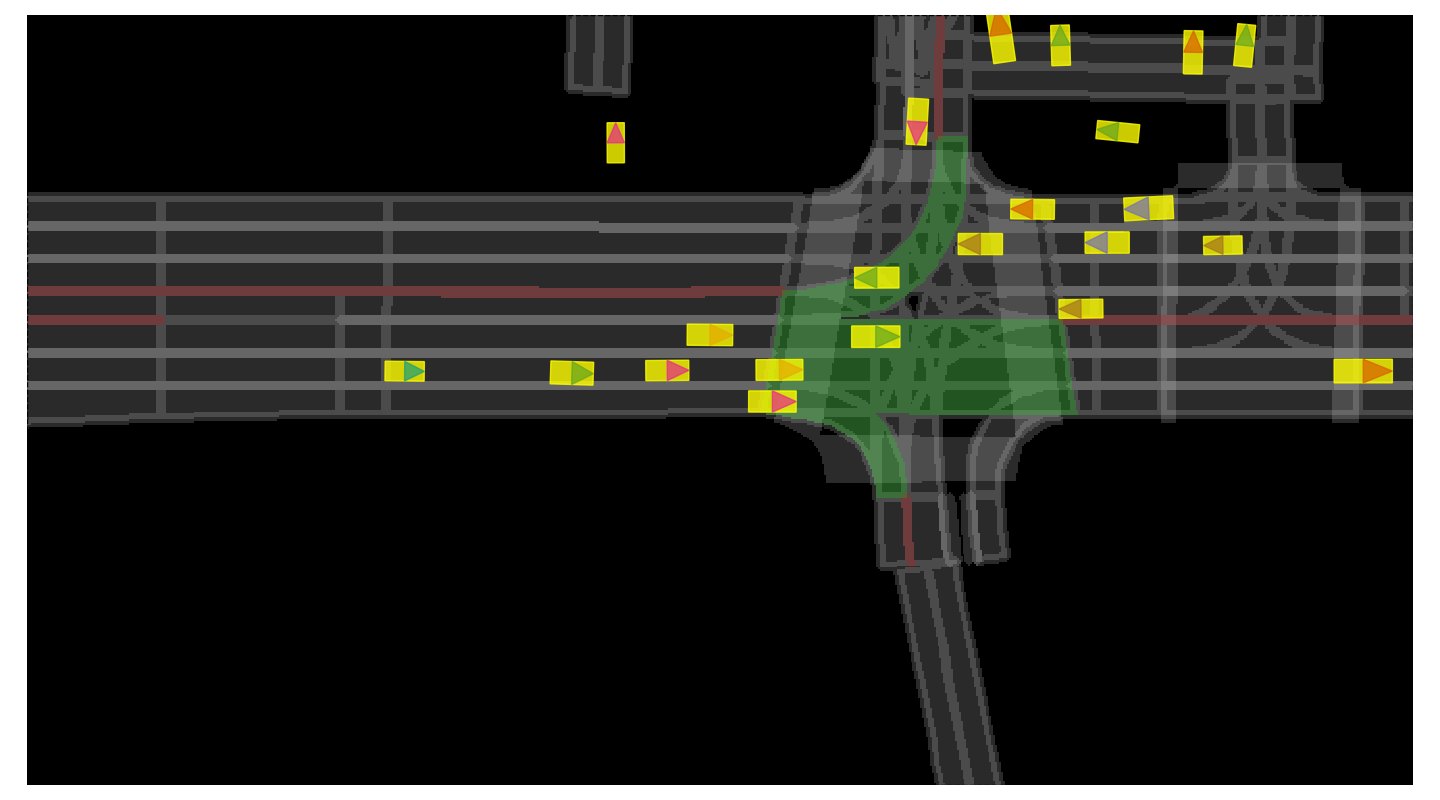}} &
        \raisebox{-0.5\height}{\includegraphics[width=0.248\linewidth]{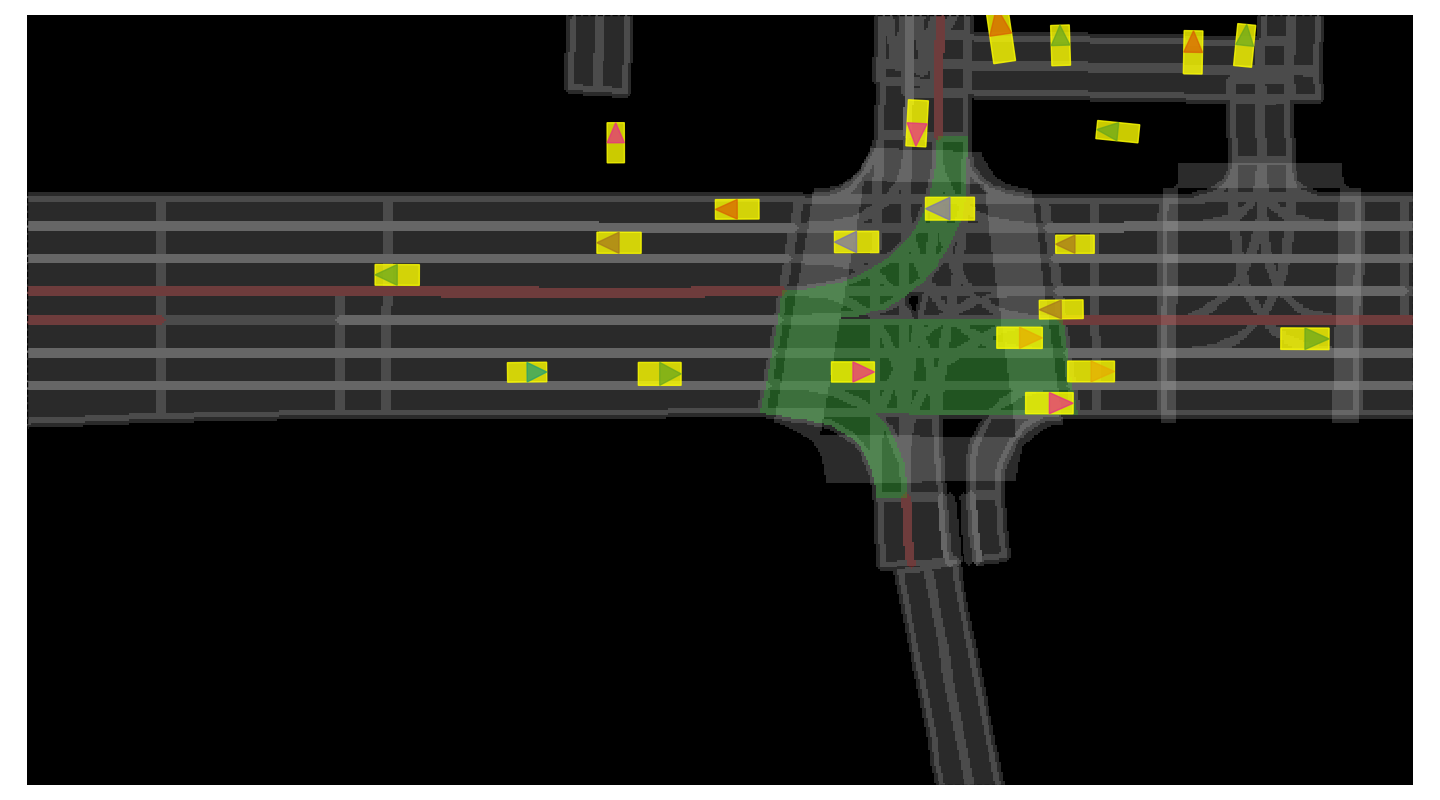}} &
        \raisebox{-0.5\height}{\includegraphics[width=0.248\linewidth]{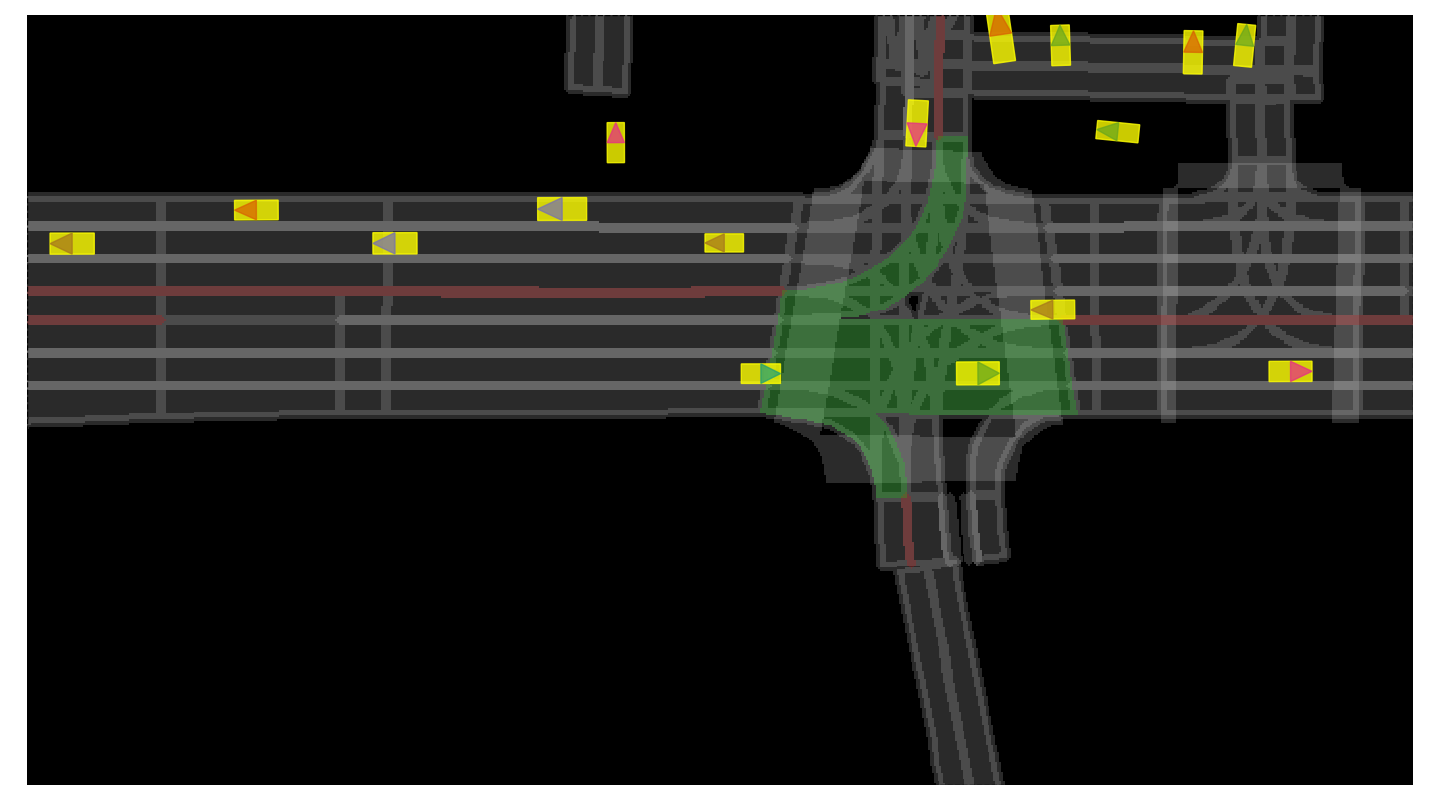}} \\
    \end{tabular}
    \caption{Simulated traffic scenarios sampled from \ourmodelshort{}: colored triangle shows heading and tracks instances across time}
    \label{fig:supp_qualitative_our}
\end{figure*}

\begin{figure*}[t]
    \centering
    \begin{tabular} {@{}c@{\hspace{.1em}}c@{\hspace{.1em}}c@{\hspace{.1em}}c@{\hspace{.1em}}c}
        {} & \textbf{T=0s} & \textbf{4s} & \textbf{8s} & \textbf{12s} \\
        \rotatebox[origin=c]{90}{\textbf{Scenario 10}} &
        \raisebox{-0.5\height}{\includegraphics[width=0.248\linewidth]{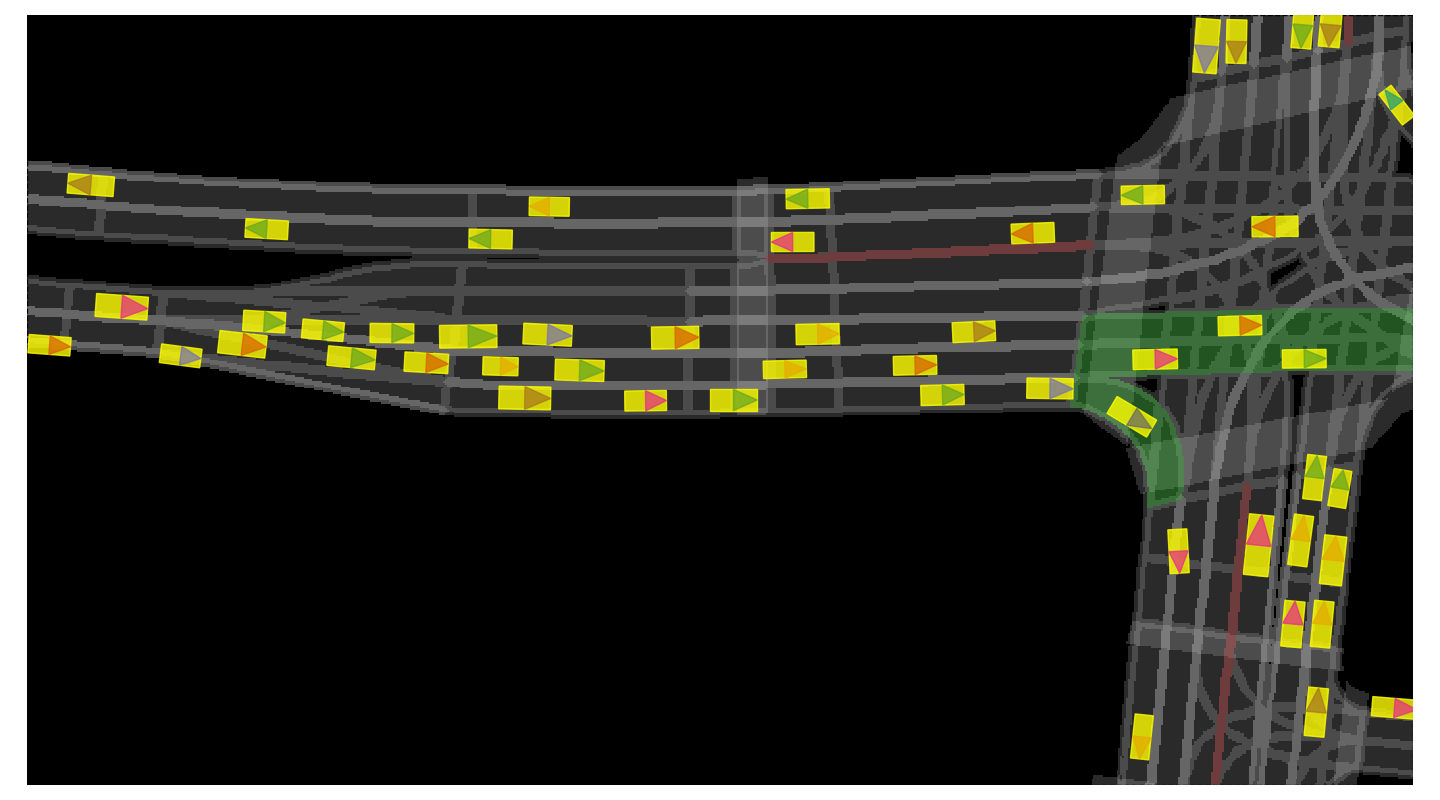}} &
        \raisebox{-0.5\height}{\includegraphics[width=0.248\linewidth]{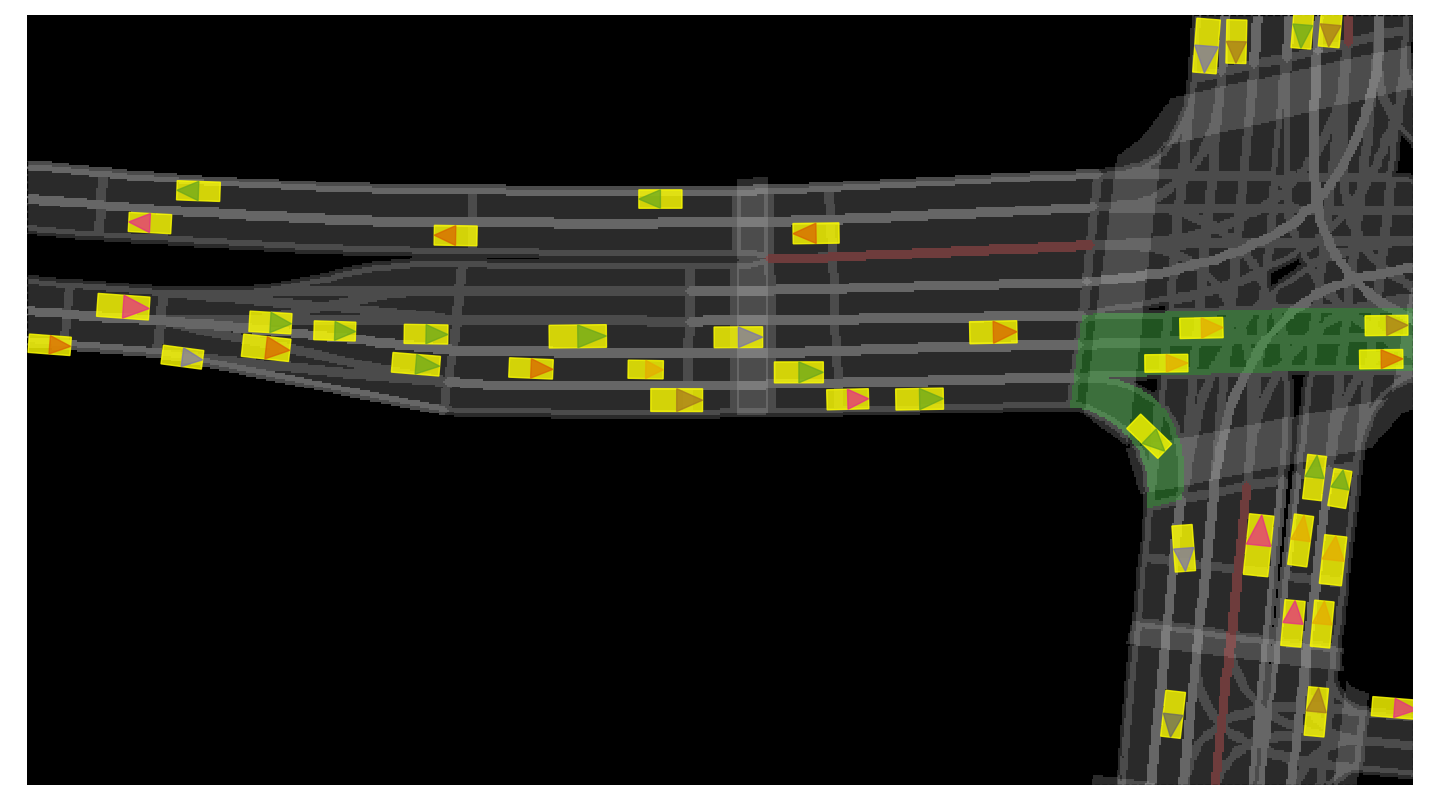}} &
        \raisebox{-0.5\height}{\includegraphics[width=0.248\linewidth]{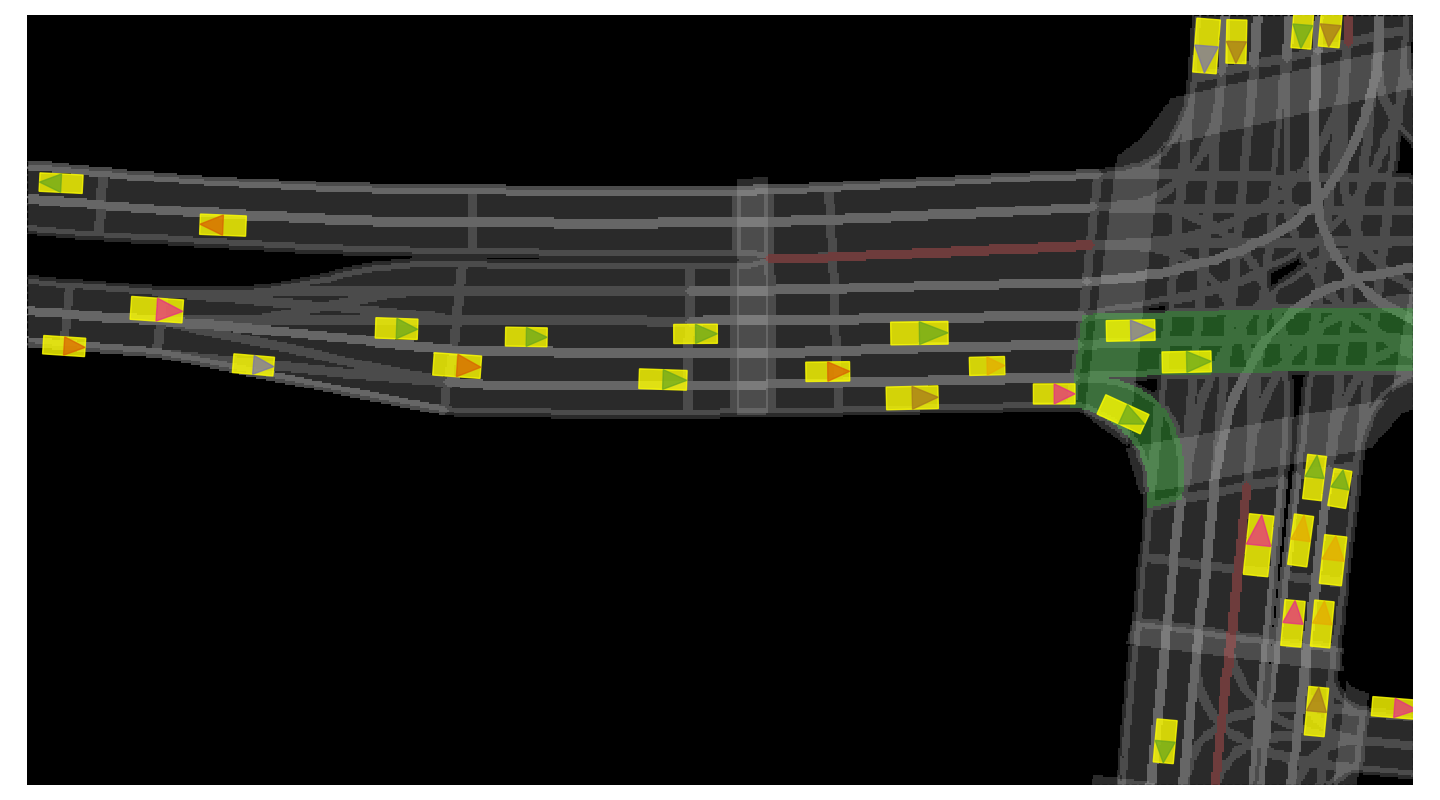}} &
        \raisebox{-0.5\height}{\includegraphics[width=0.248\linewidth]{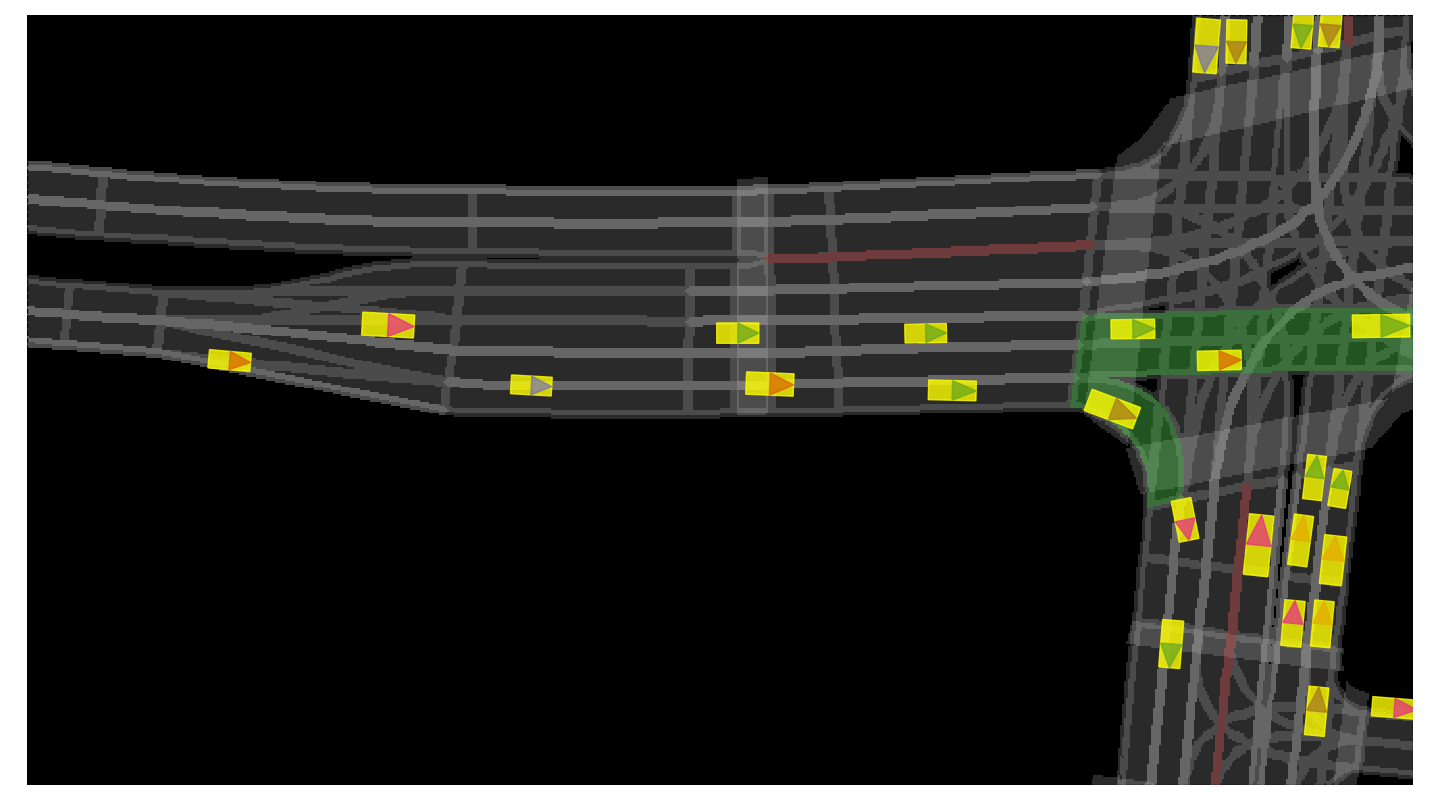}} \vspace{.1em} \\
        \rotatebox[origin=c]{90}{\textbf{Scenario 11}} &
        \raisebox{-0.5\height}{\includegraphics[width=0.248\linewidth]{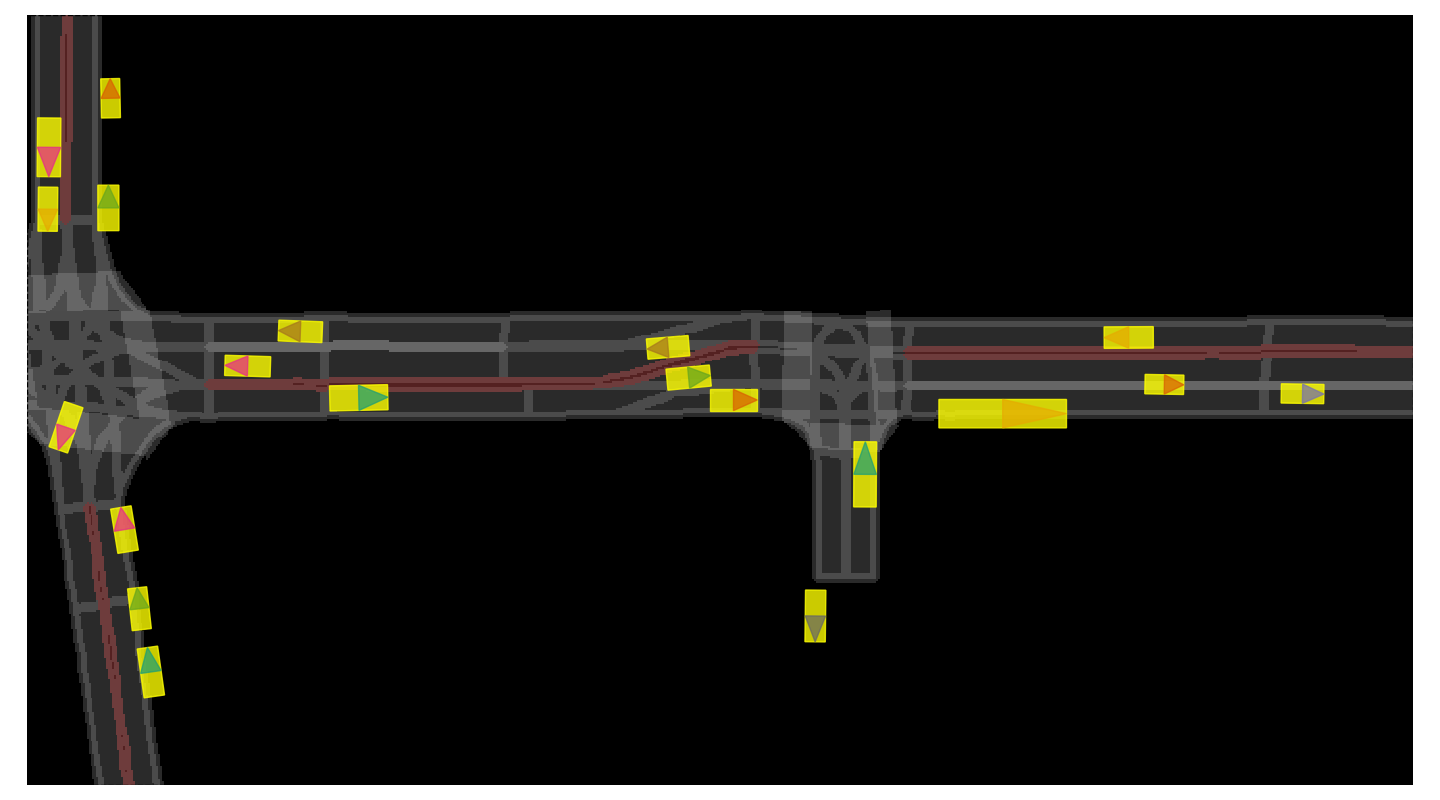}} &
        \raisebox{-0.5\height}{\includegraphics[width=0.248\linewidth]{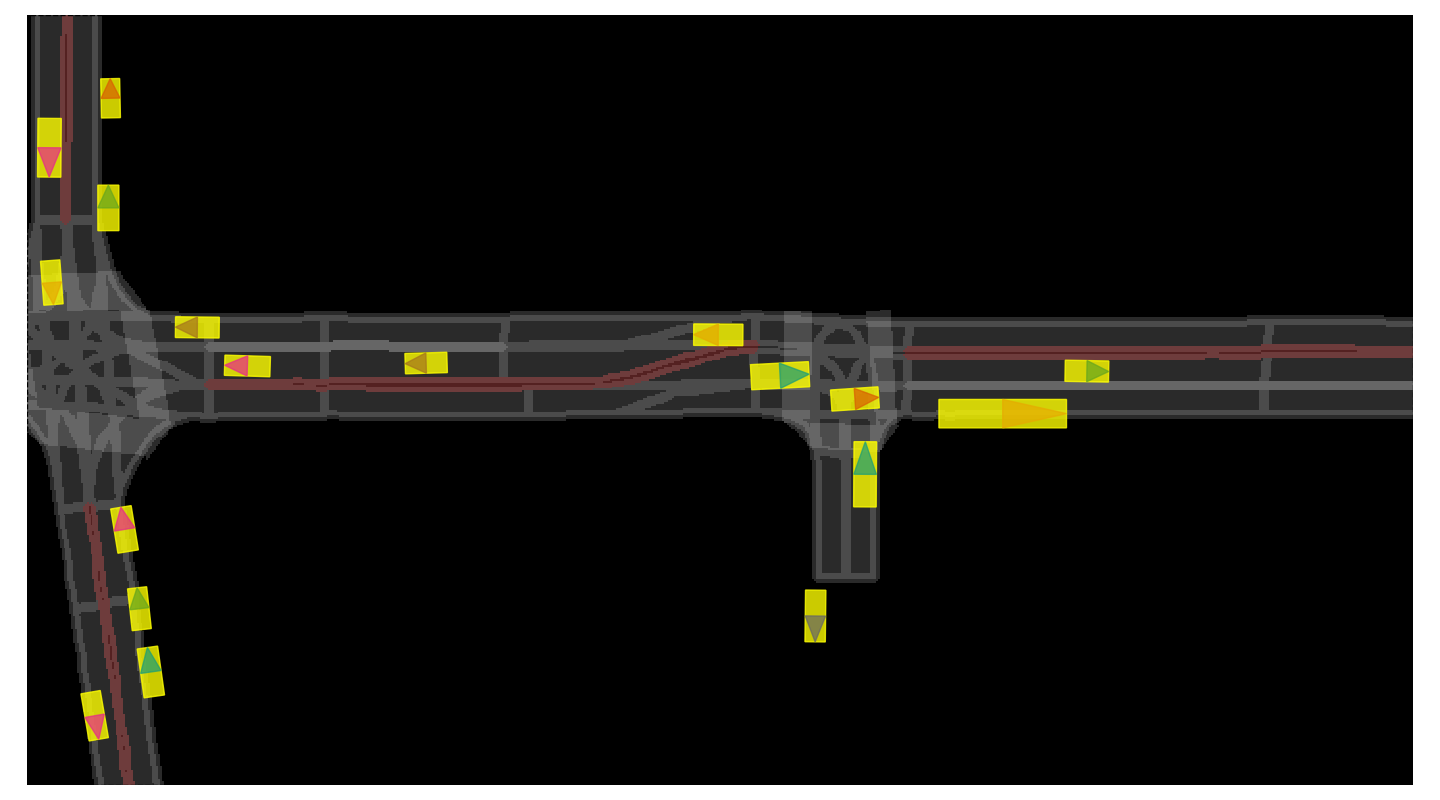}} &
        \raisebox{-0.5\height}{\includegraphics[width=0.248\linewidth]{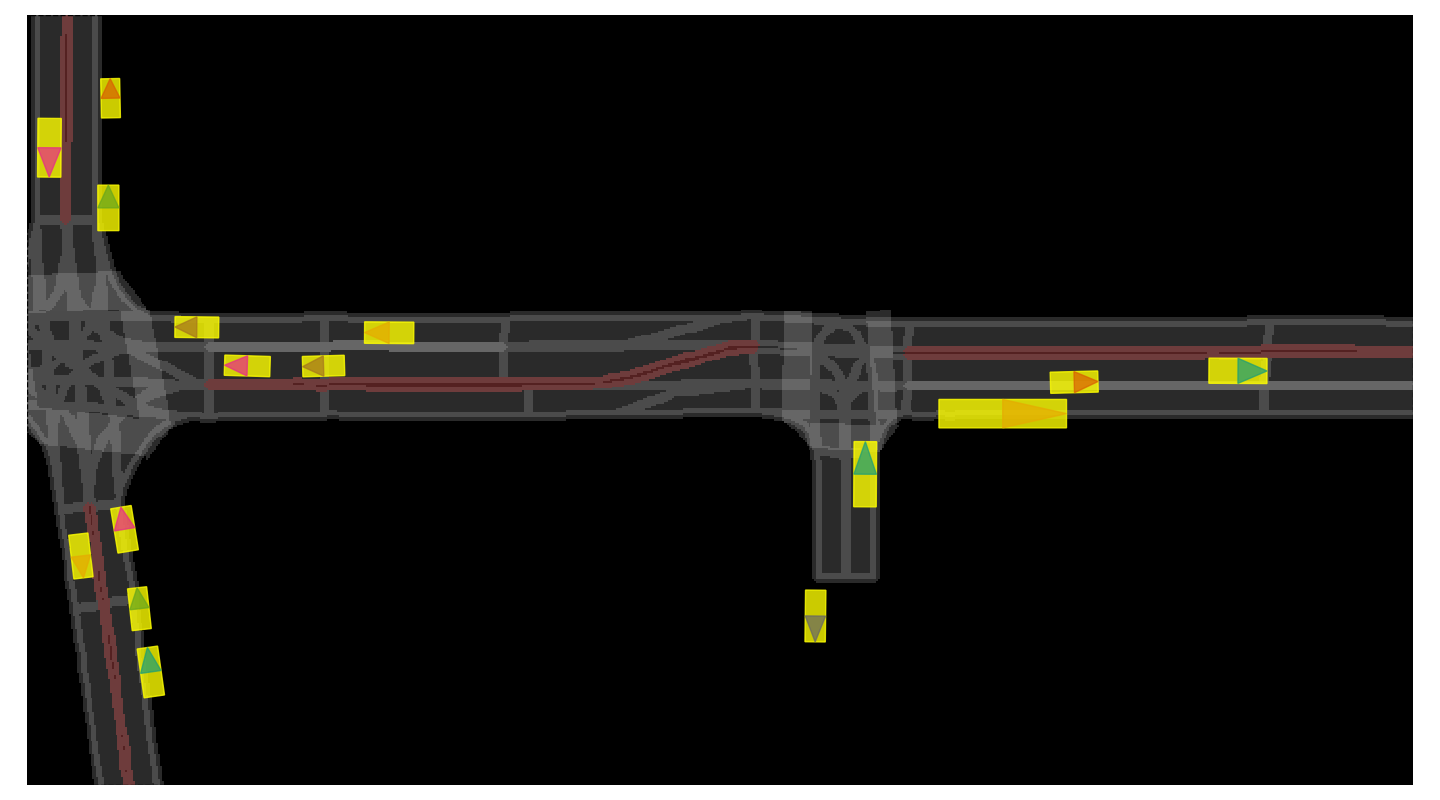}} &
        \raisebox{-0.5\height}{\includegraphics[width=0.248\linewidth]{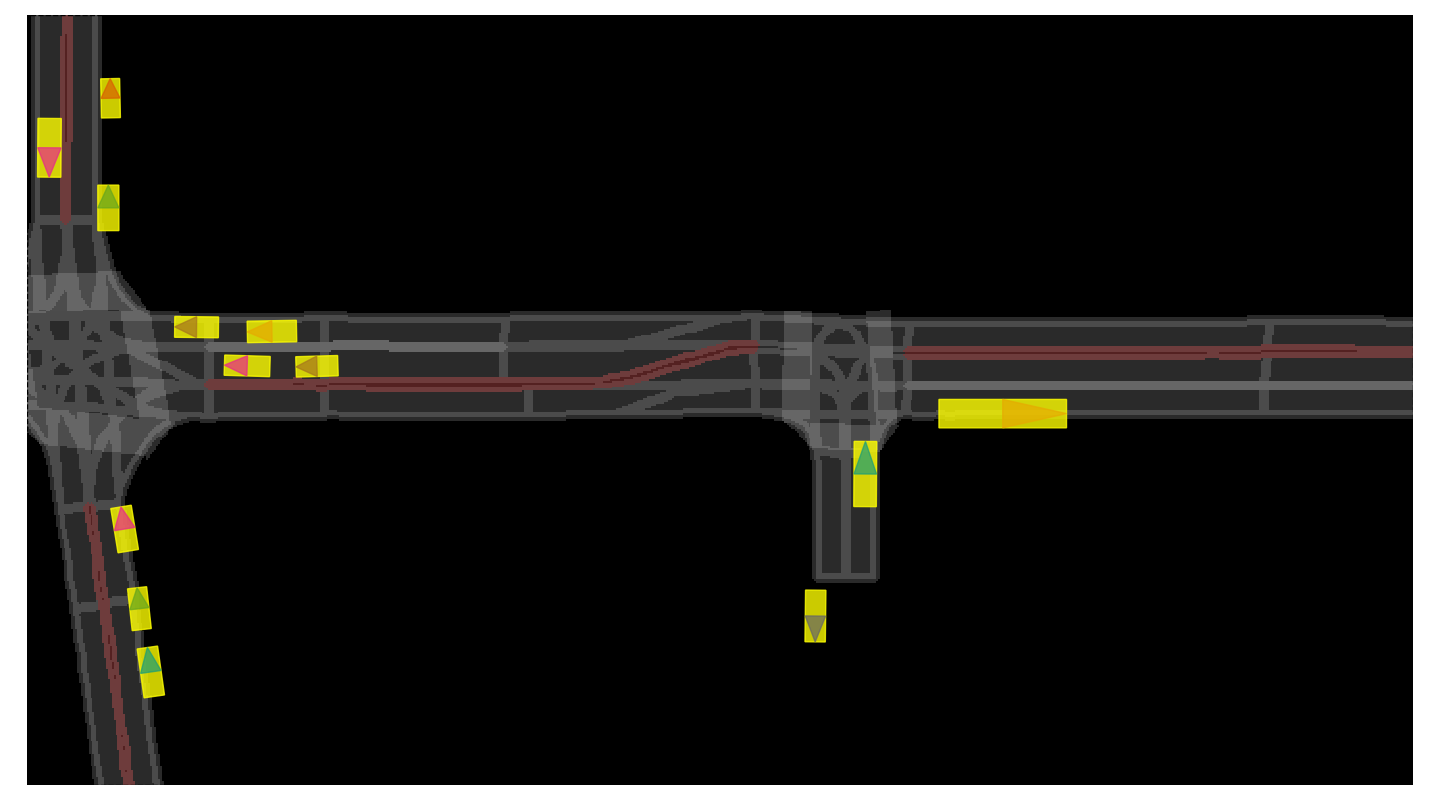}} \vspace{.1em} \\
        \rotatebox[origin=c]{90}{\textbf{Scenario 12}} &
        \raisebox{-0.5\height}{\includegraphics[width=0.248\linewidth]{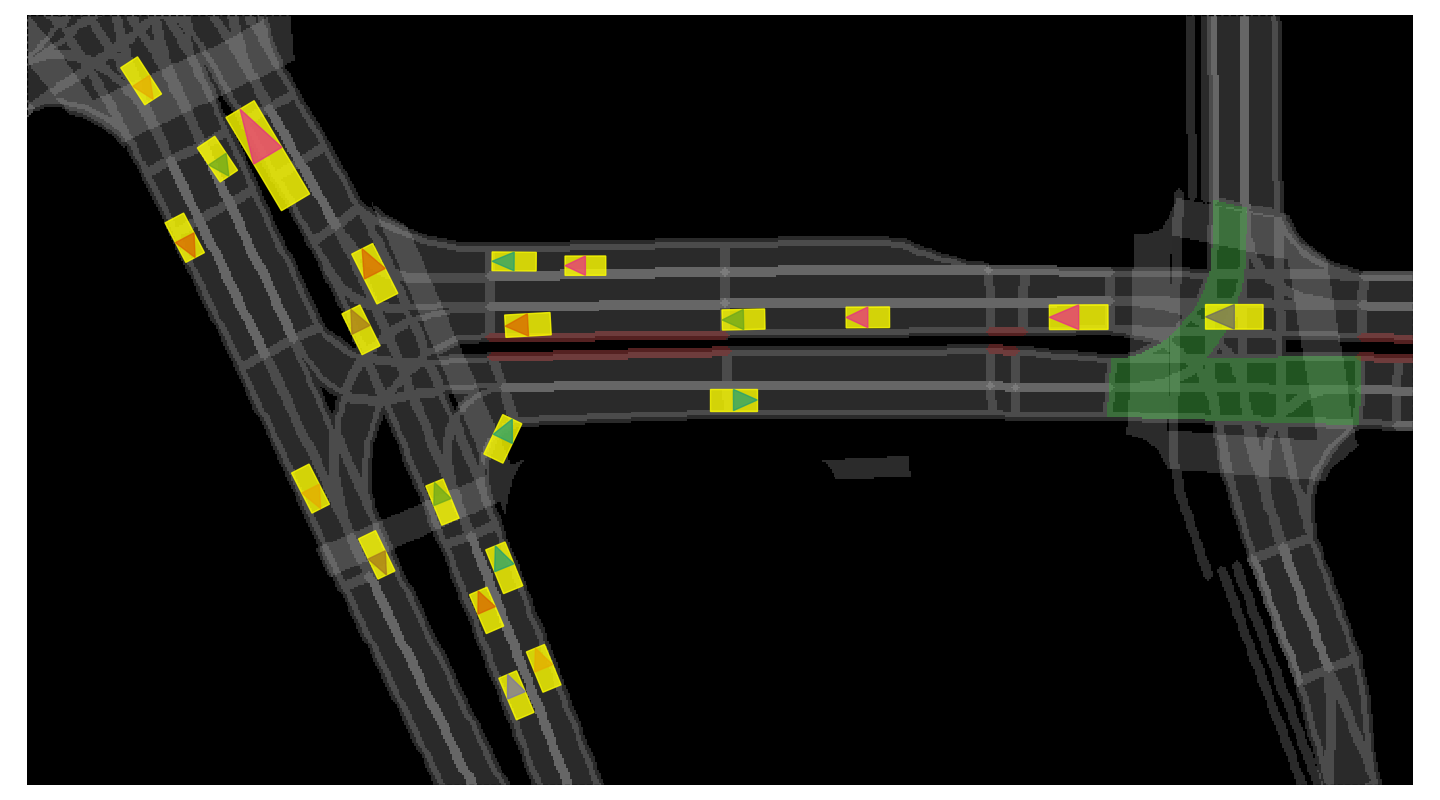}} &
        \raisebox{-0.5\height}{\includegraphics[width=0.248\linewidth]{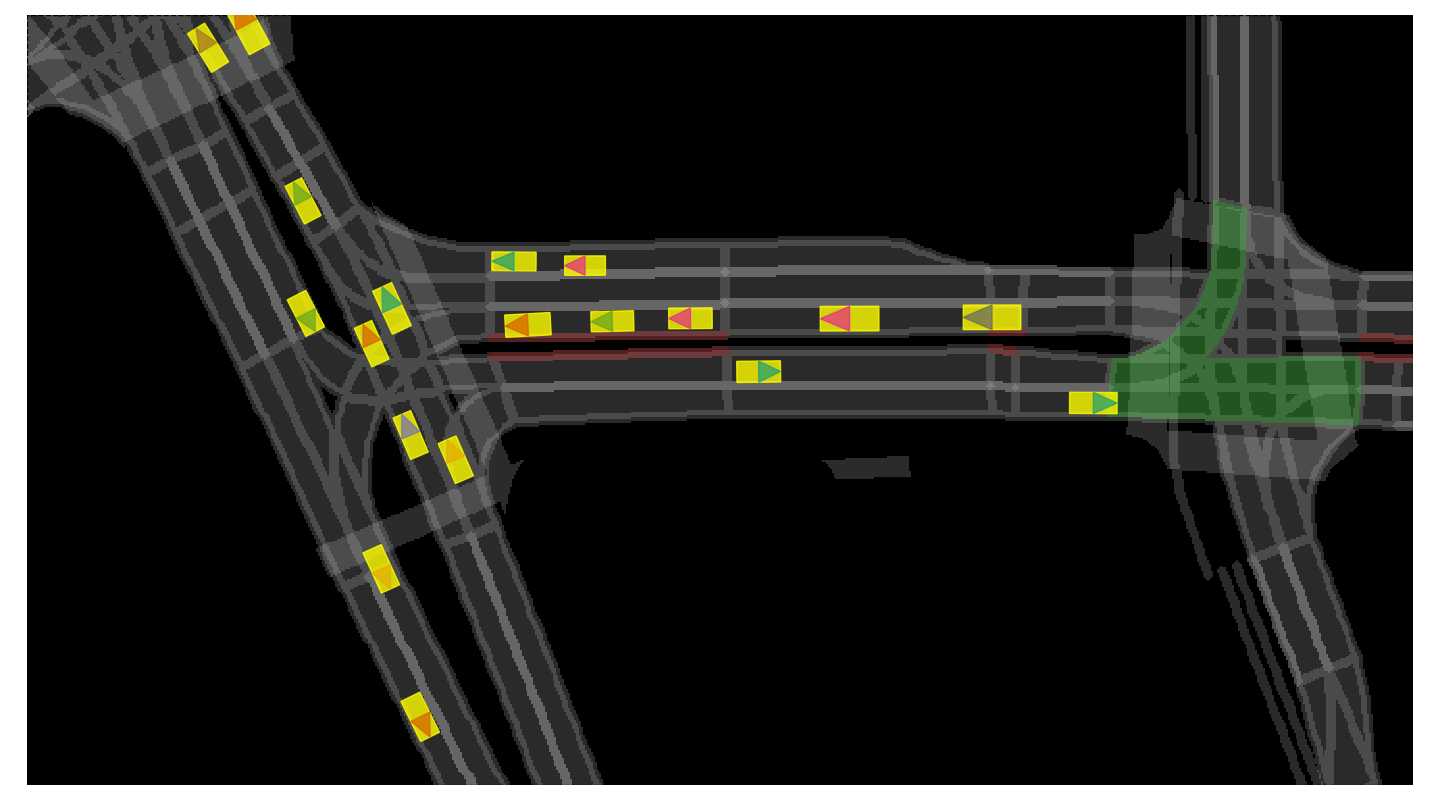}} &
        \raisebox{-0.5\height}{\includegraphics[width=0.248\linewidth]{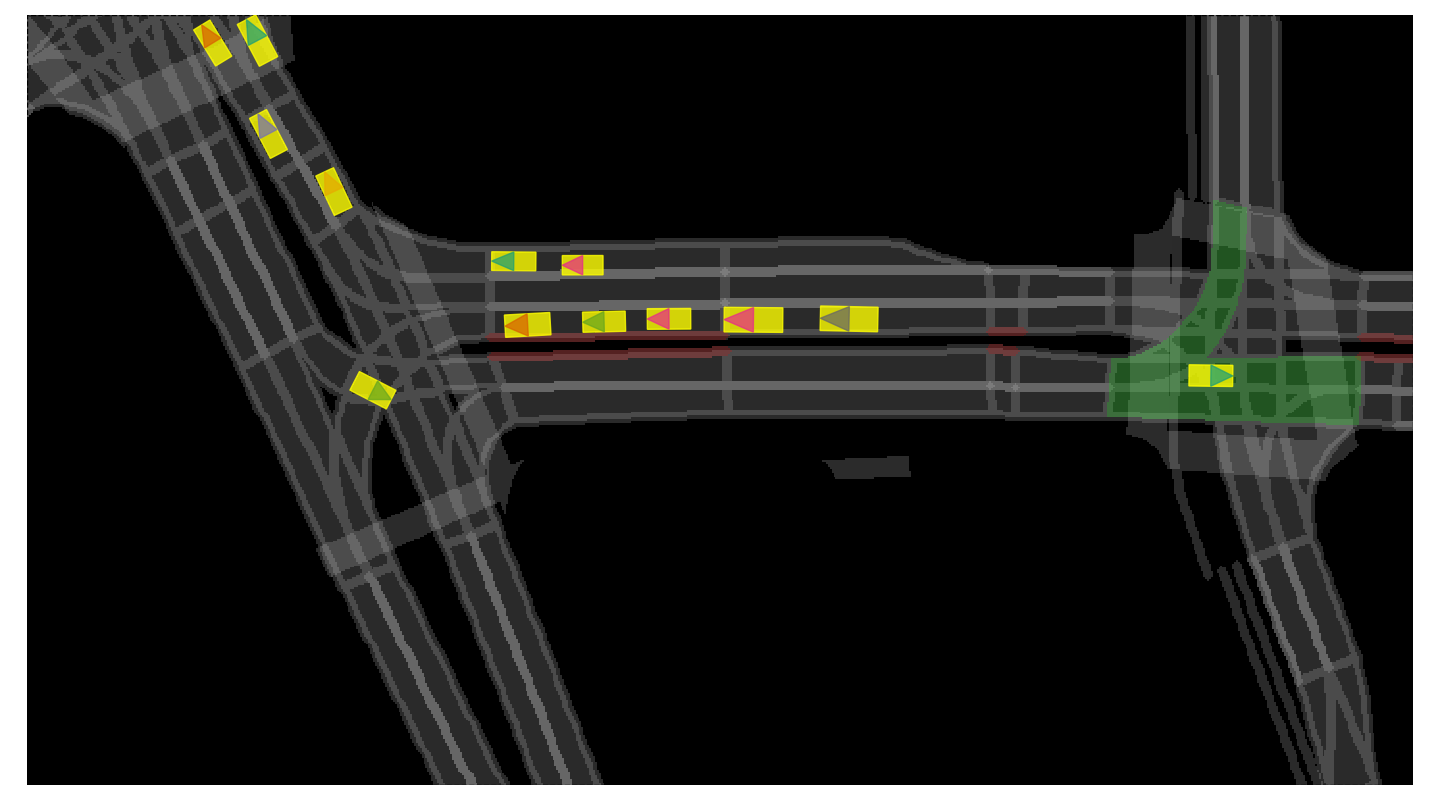}} &
        \raisebox{-0.5\height}{\includegraphics[width=0.248\linewidth]{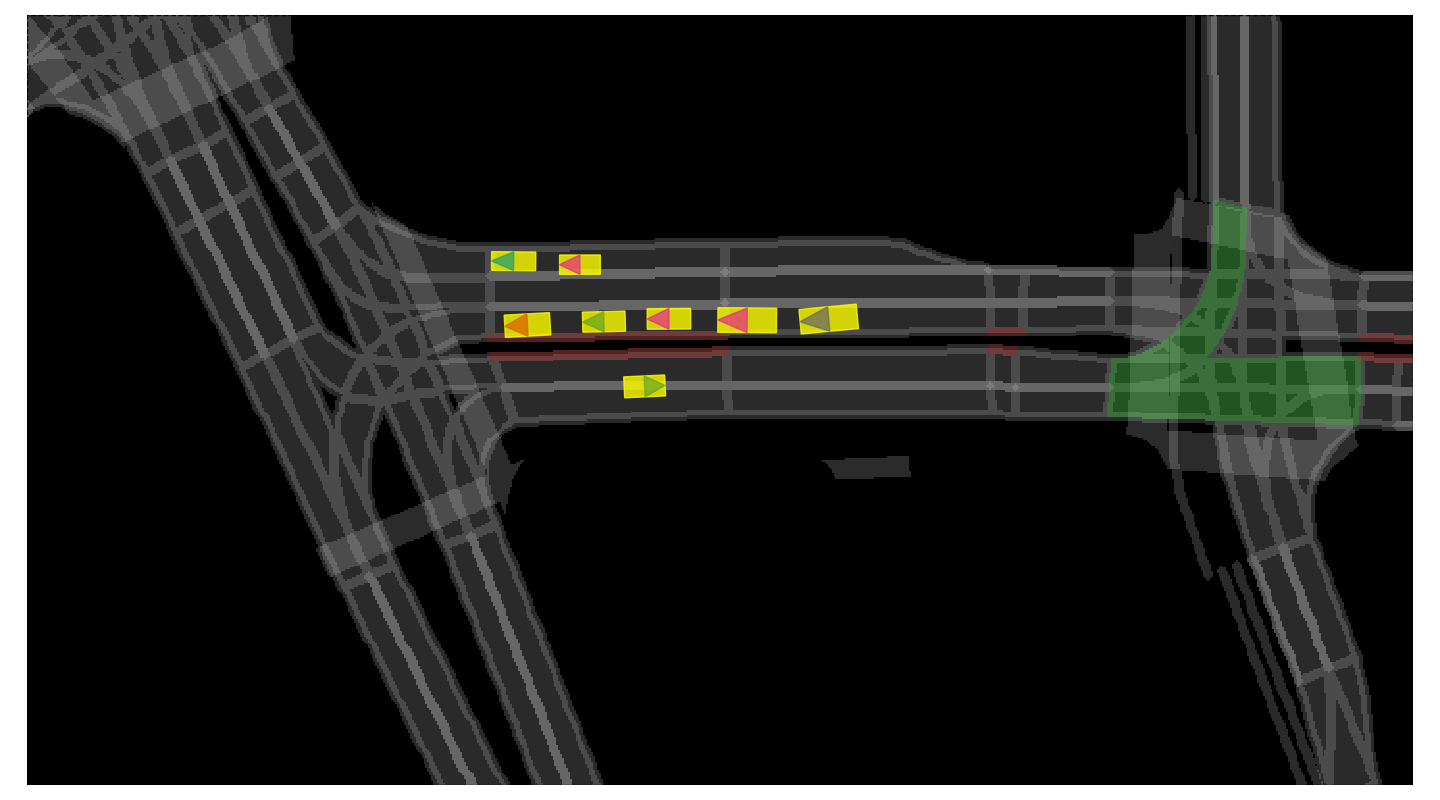}} \vspace{.1em} \\
        \rotatebox[origin=c]{90}{\textbf{Scenario 13}} &
        \raisebox{-0.5\height}{\includegraphics[width=0.248\linewidth]{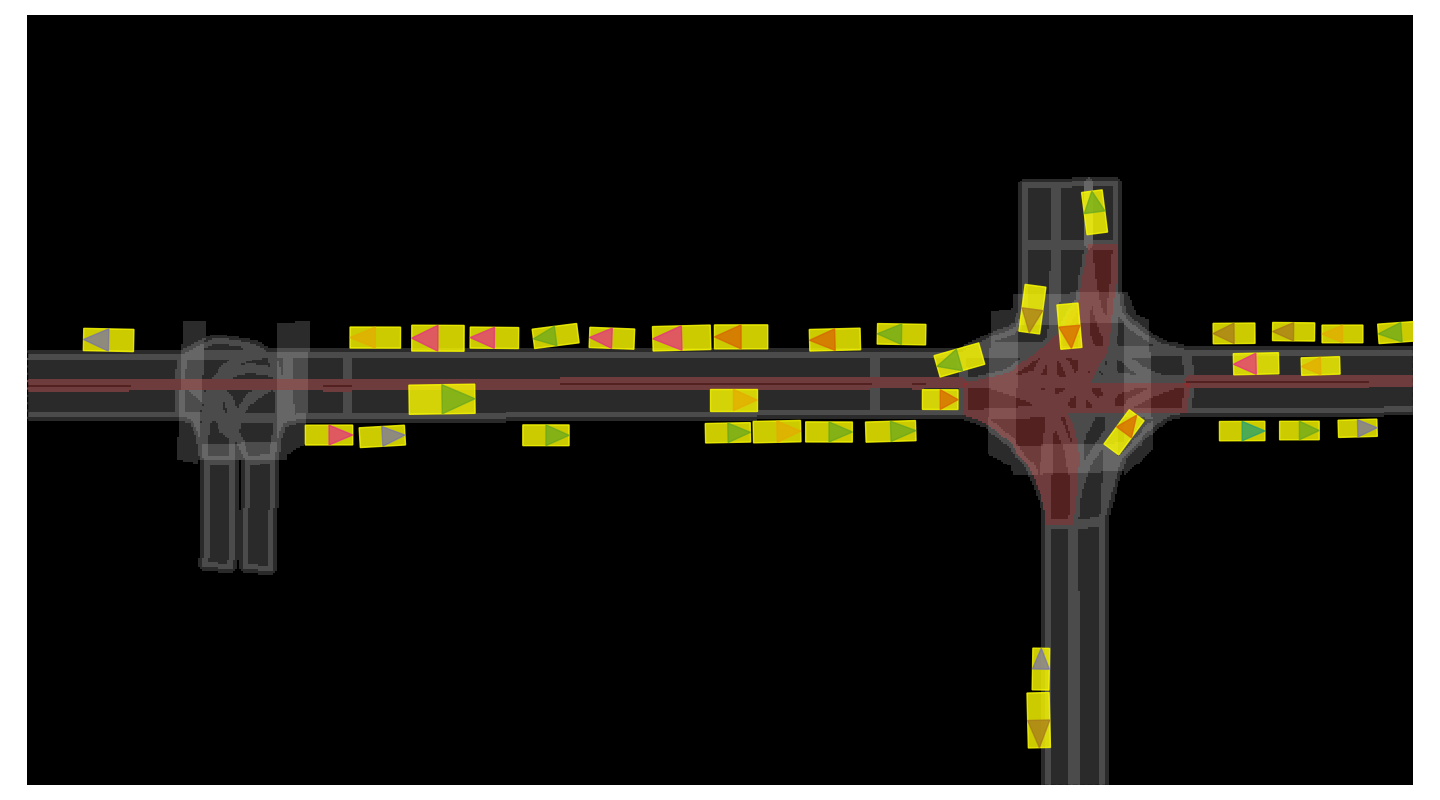}} &
        \raisebox{-0.5\height}{\includegraphics[width=0.248\linewidth]{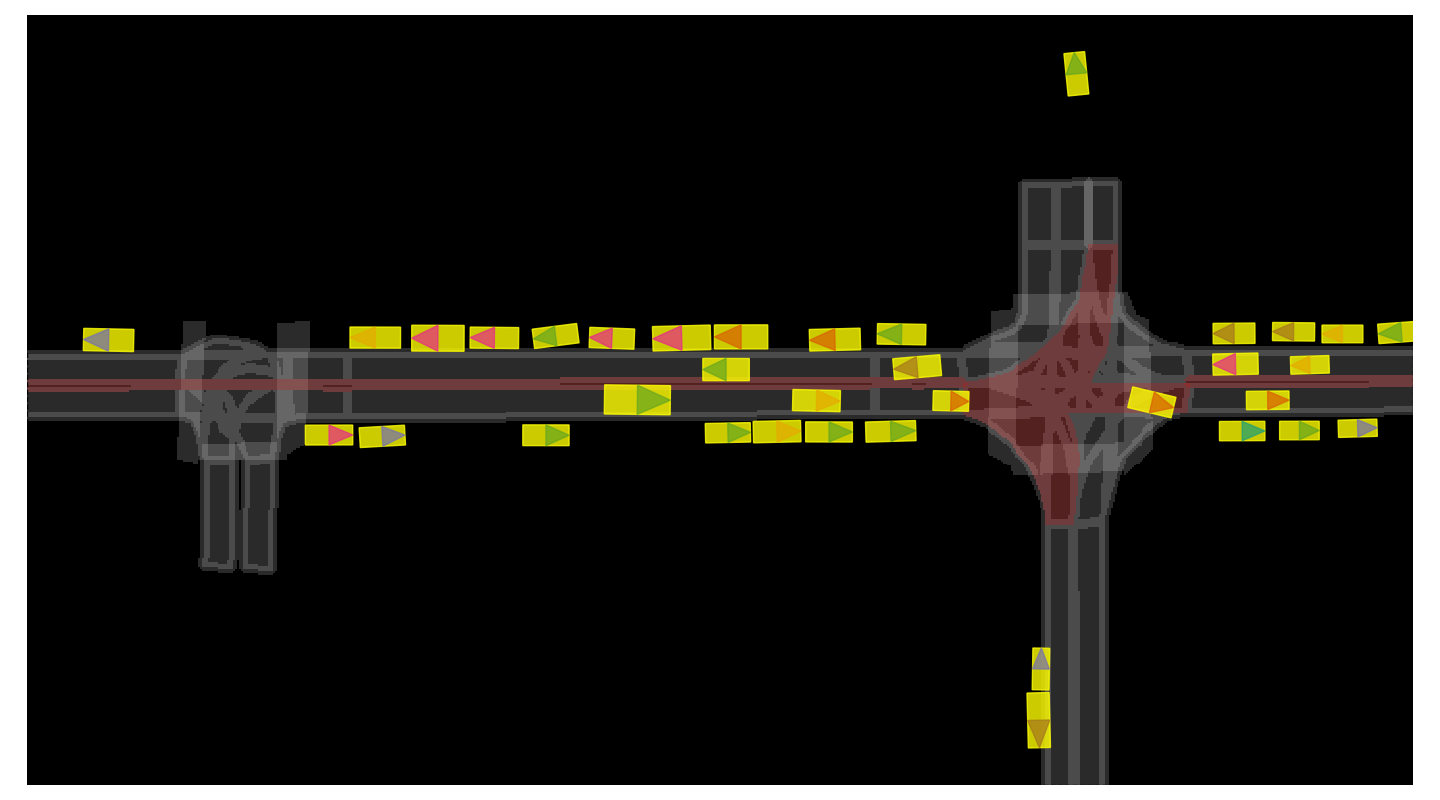}} &
        \raisebox{-0.5\height}{\includegraphics[width=0.248\linewidth]{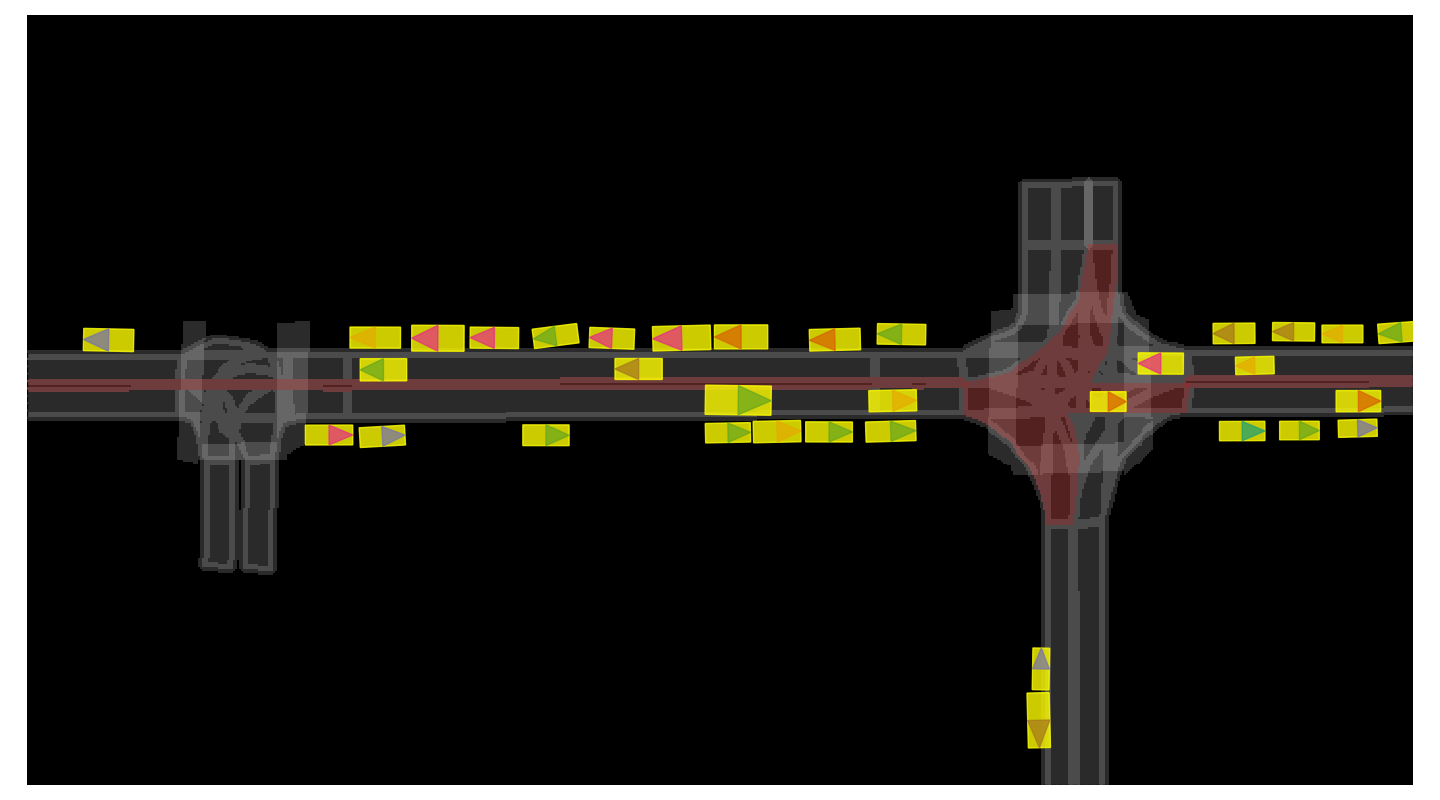}} &
        \raisebox{-0.5\height}{\includegraphics[width=0.248\linewidth]{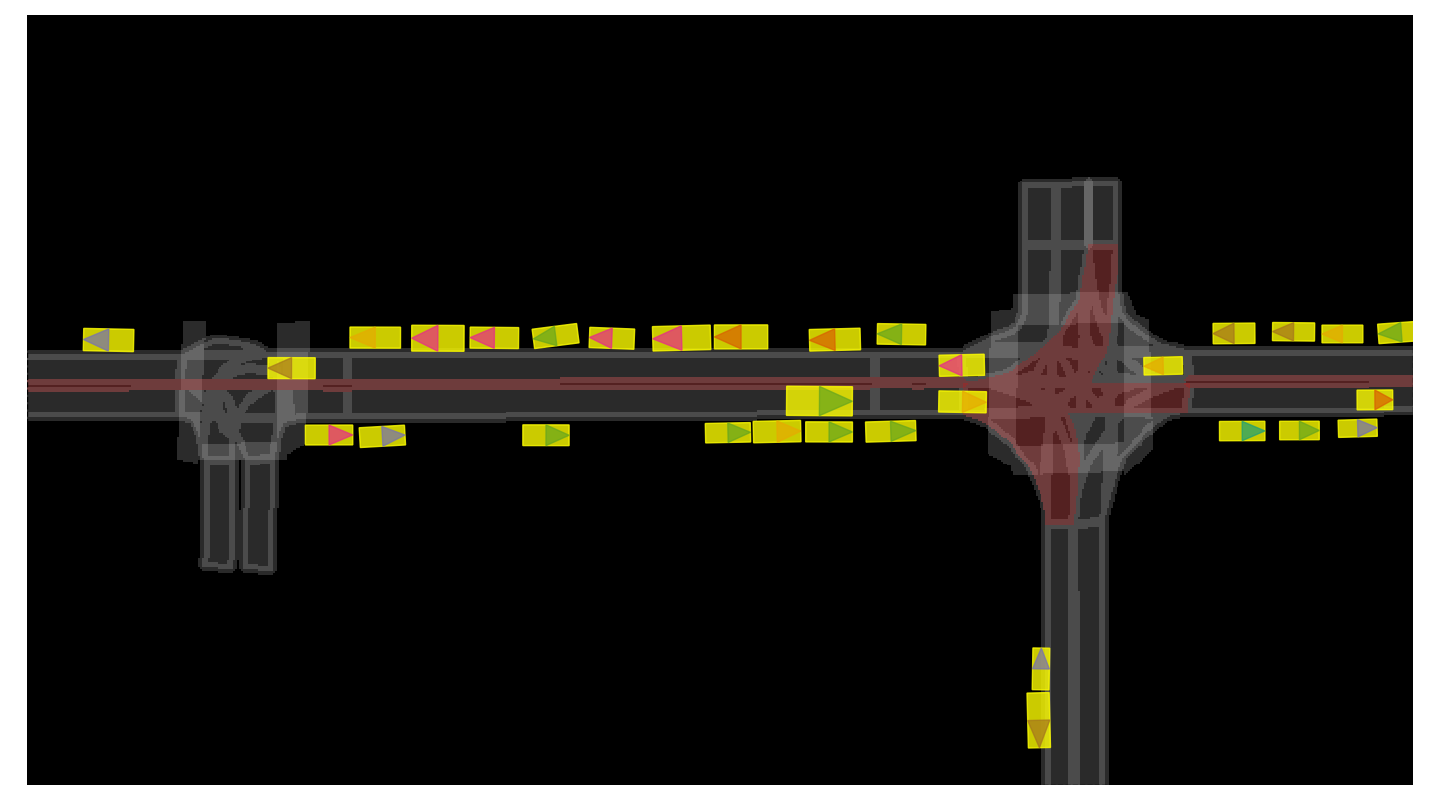}} \vspace{.1em} \\
        \rotatebox[origin=c]{90}{\textbf{Scenario 14}} &
        \raisebox{-0.5\height}{\includegraphics[width=0.248\linewidth]{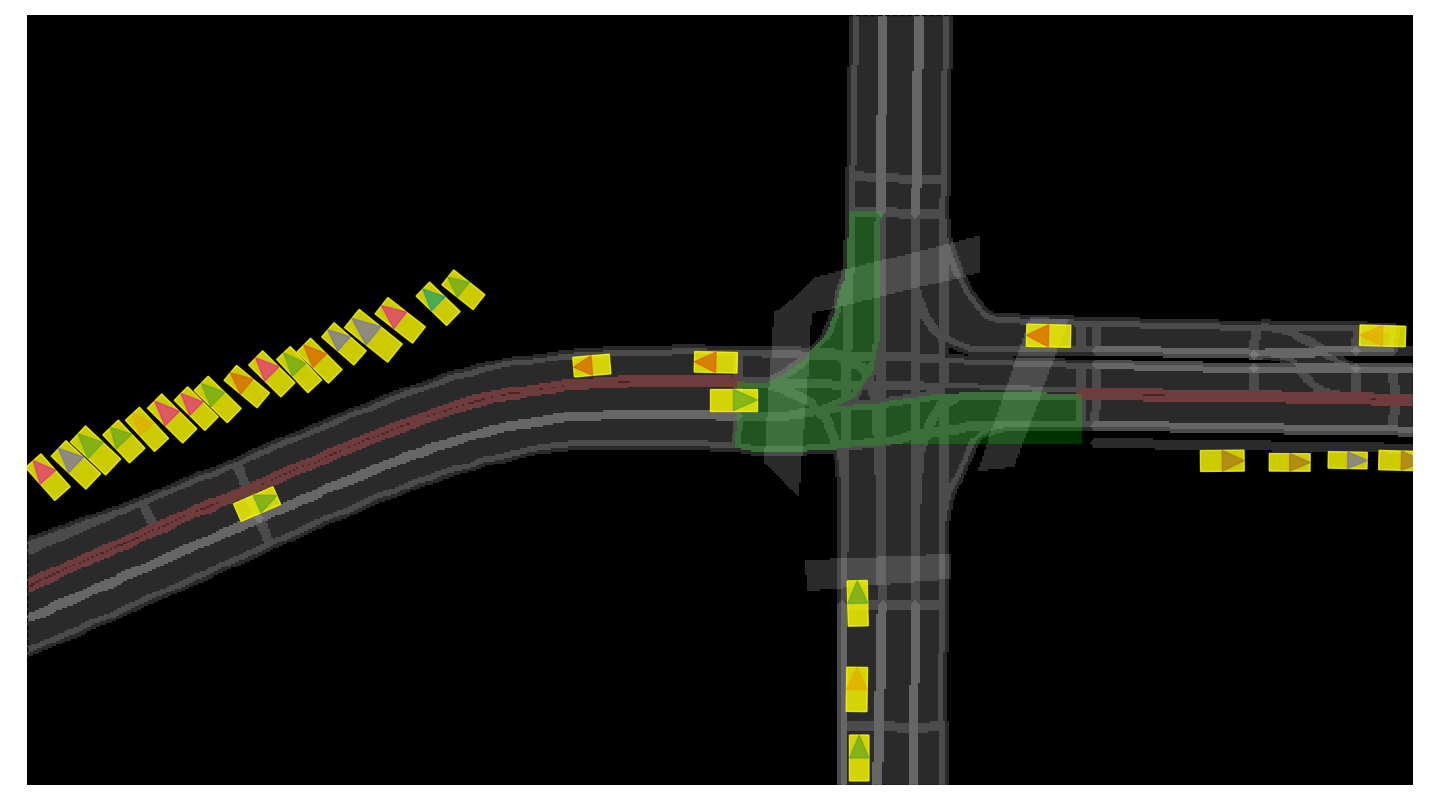}} &
        \raisebox{-0.5\height}{\includegraphics[width=0.248\linewidth]{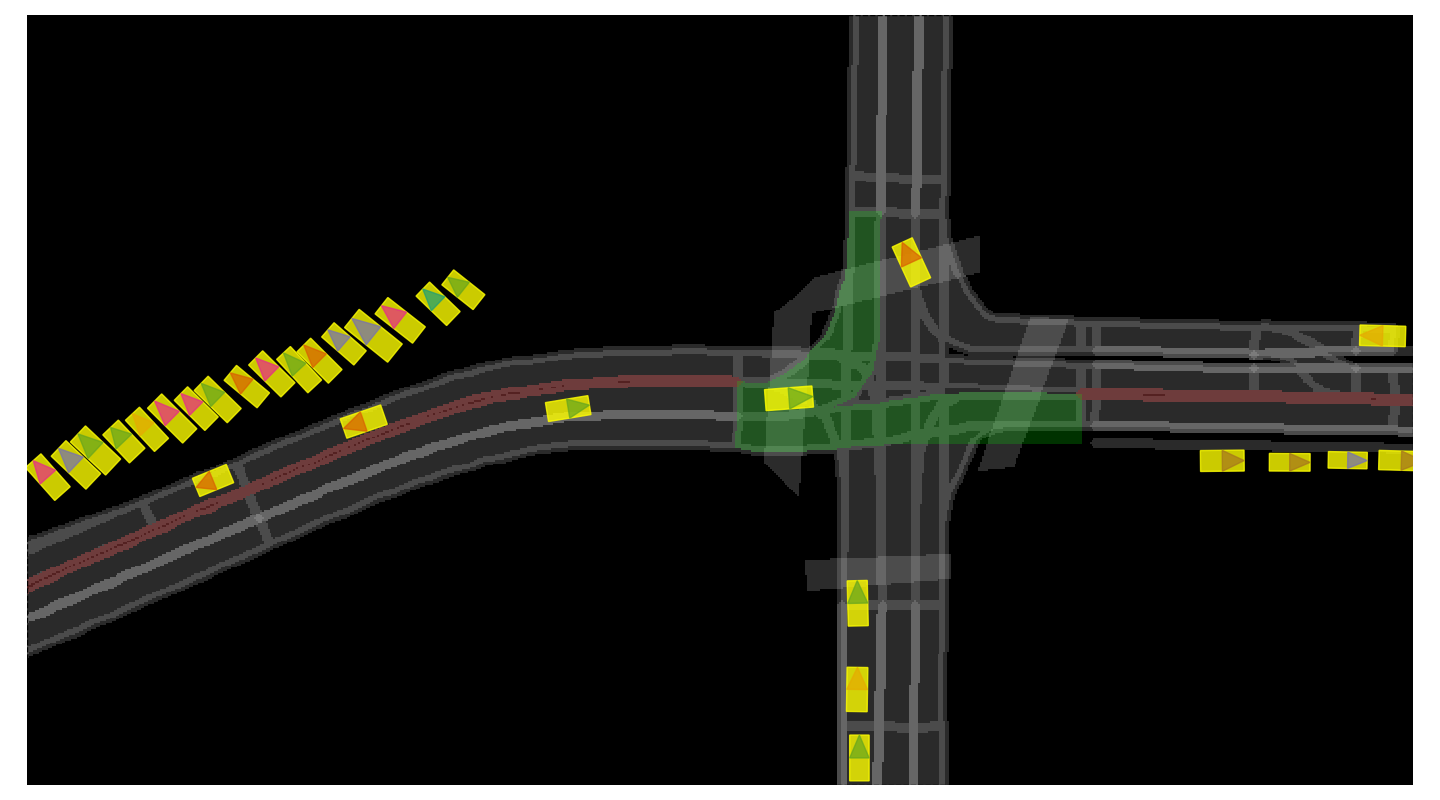}} &
        \raisebox{-0.5\height}{\includegraphics[width=0.248\linewidth]{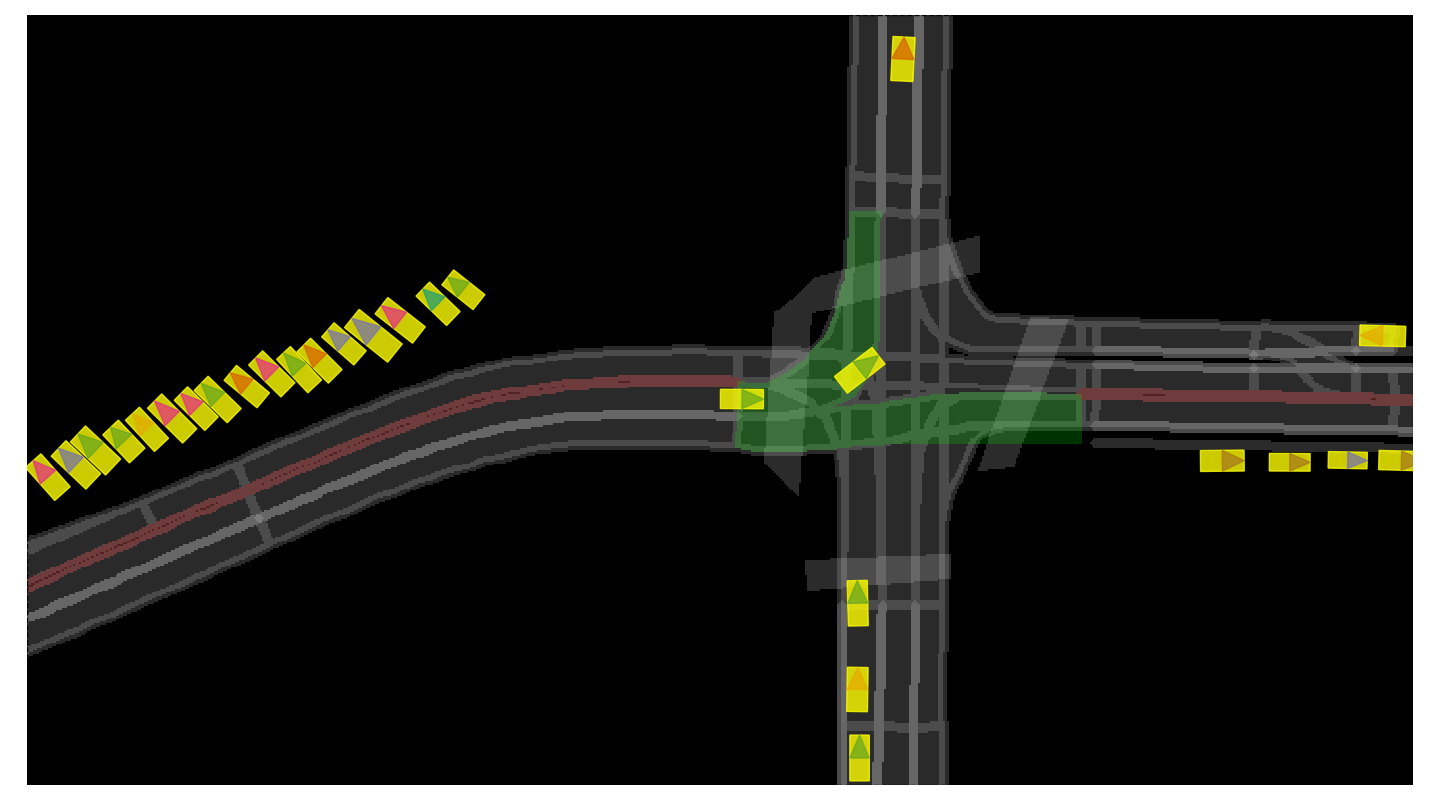}} &
        \raisebox{-0.5\height}{\includegraphics[width=0.248\linewidth]{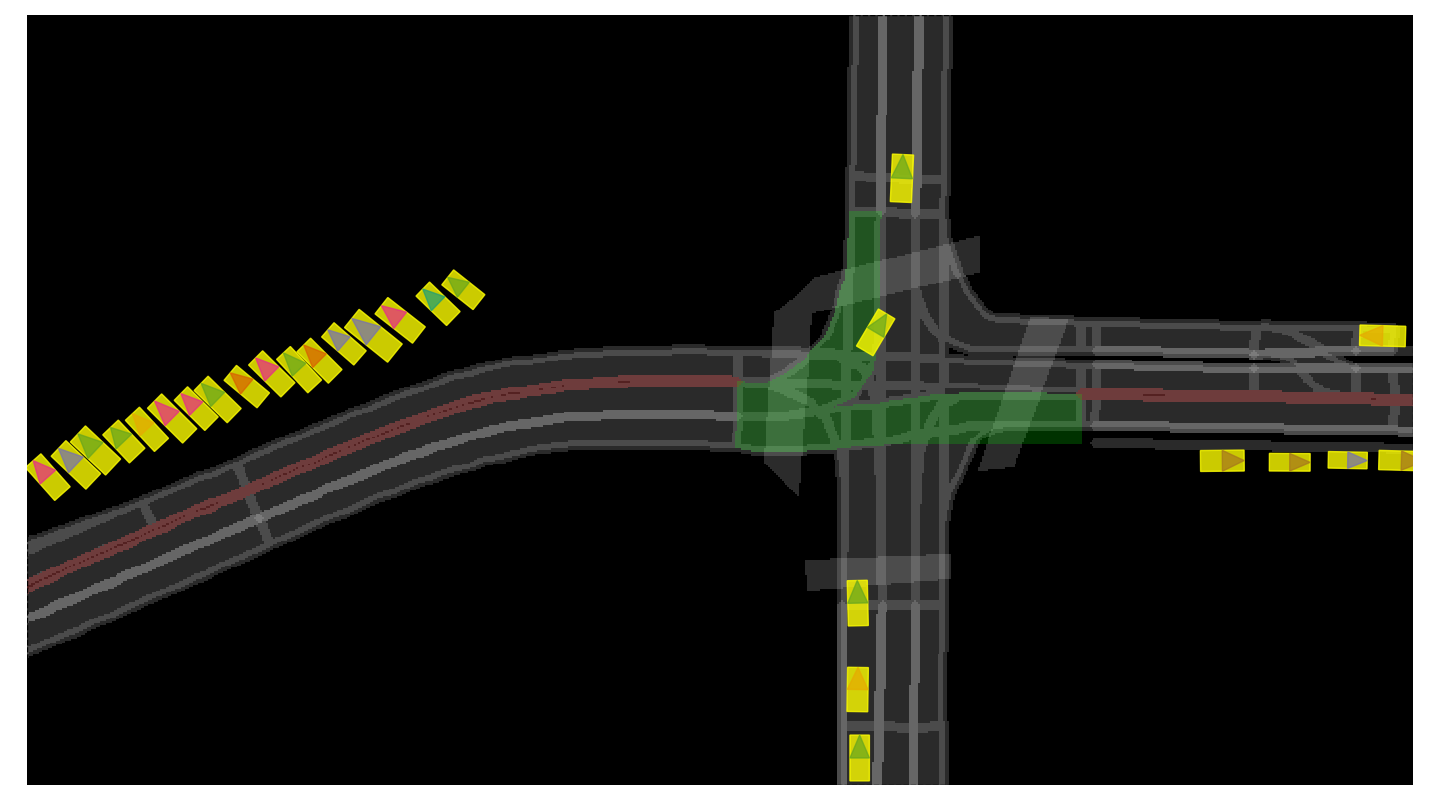}} \vspace{.1em} \\
        \rotatebox[origin=c]{90}{\textbf{Scenario 15}} &
        \raisebox{-0.5\height}{\includegraphics[width=0.248\linewidth]{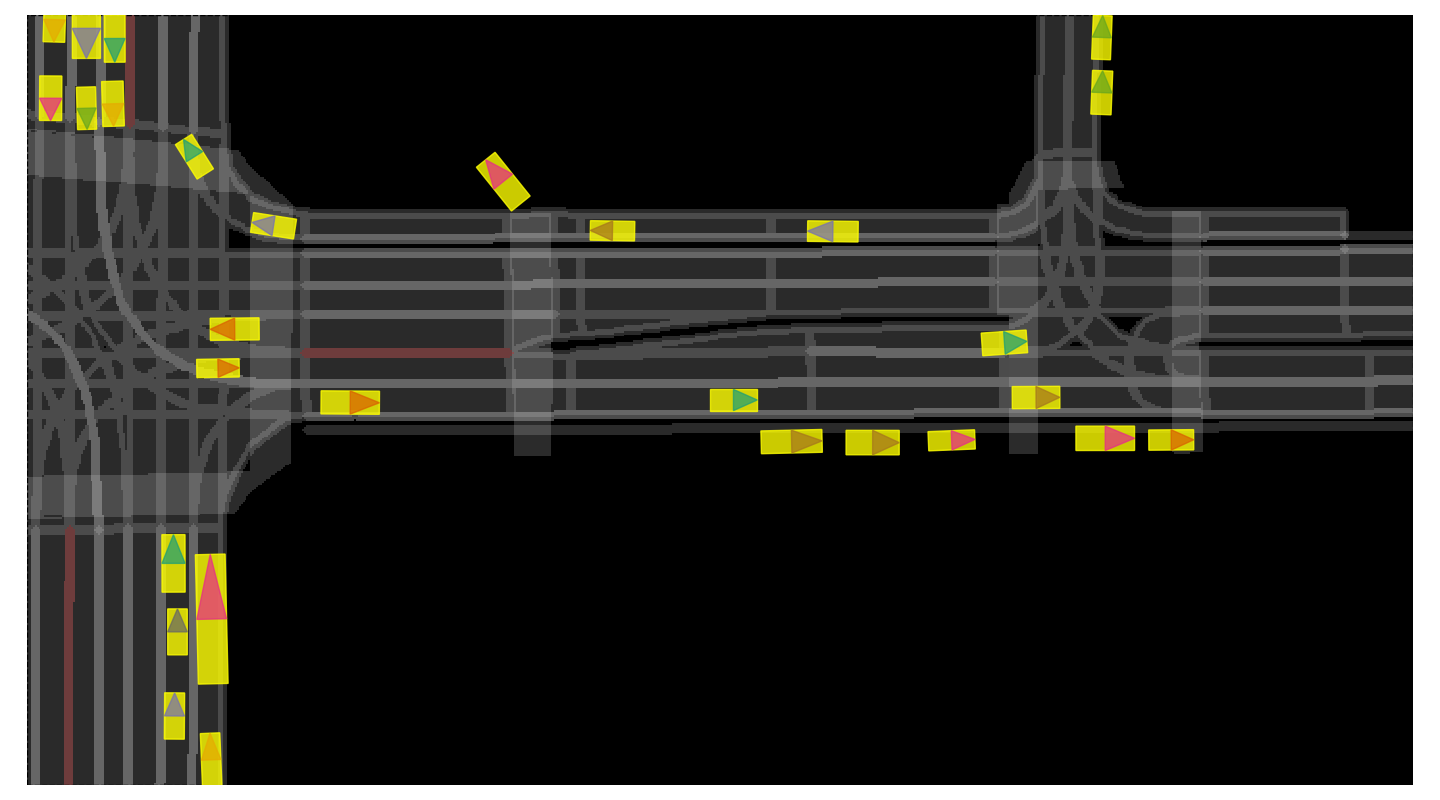}} &
        \raisebox{-0.5\height}{\includegraphics[width=0.248\linewidth]{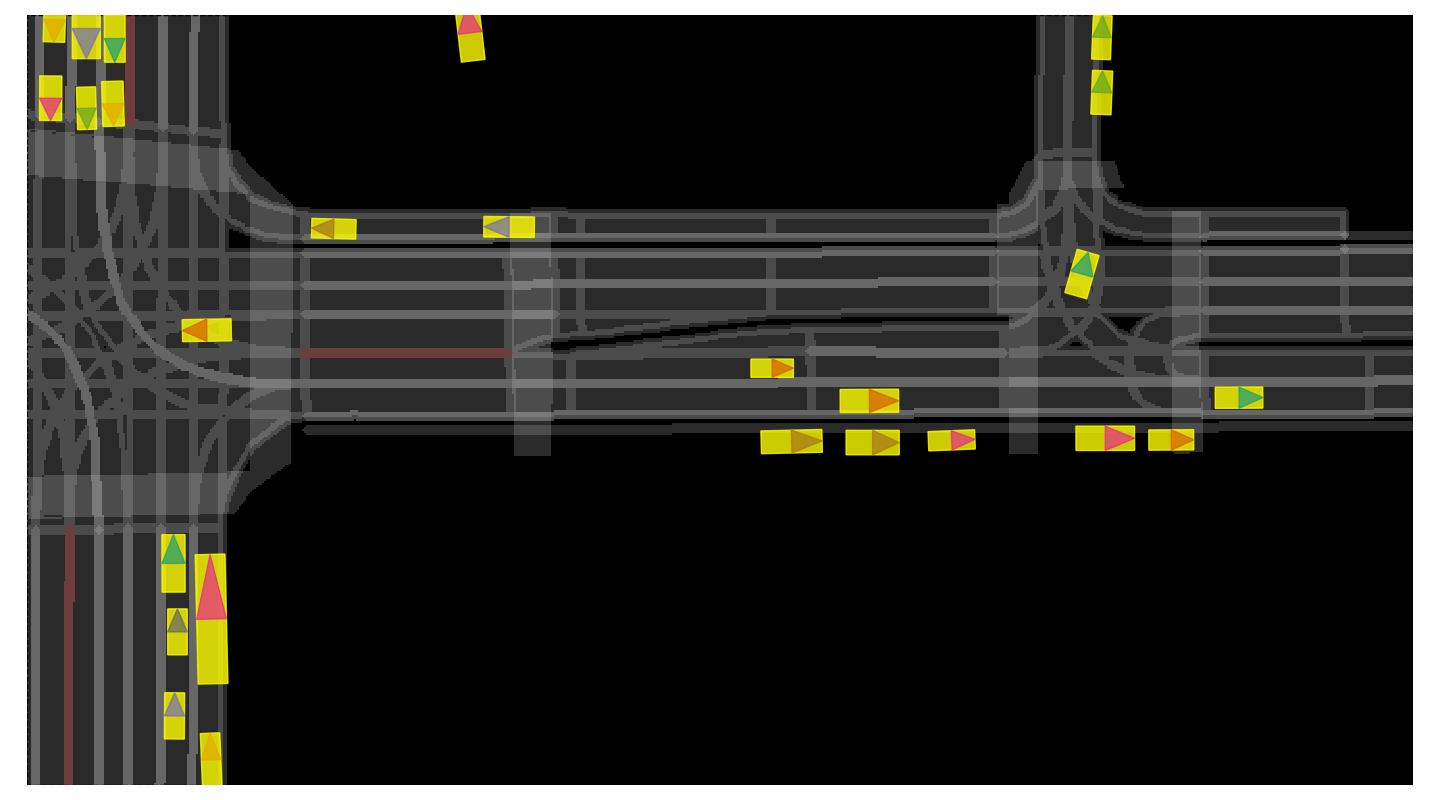}} &
        \raisebox{-0.5\height}{\includegraphics[width=0.248\linewidth]{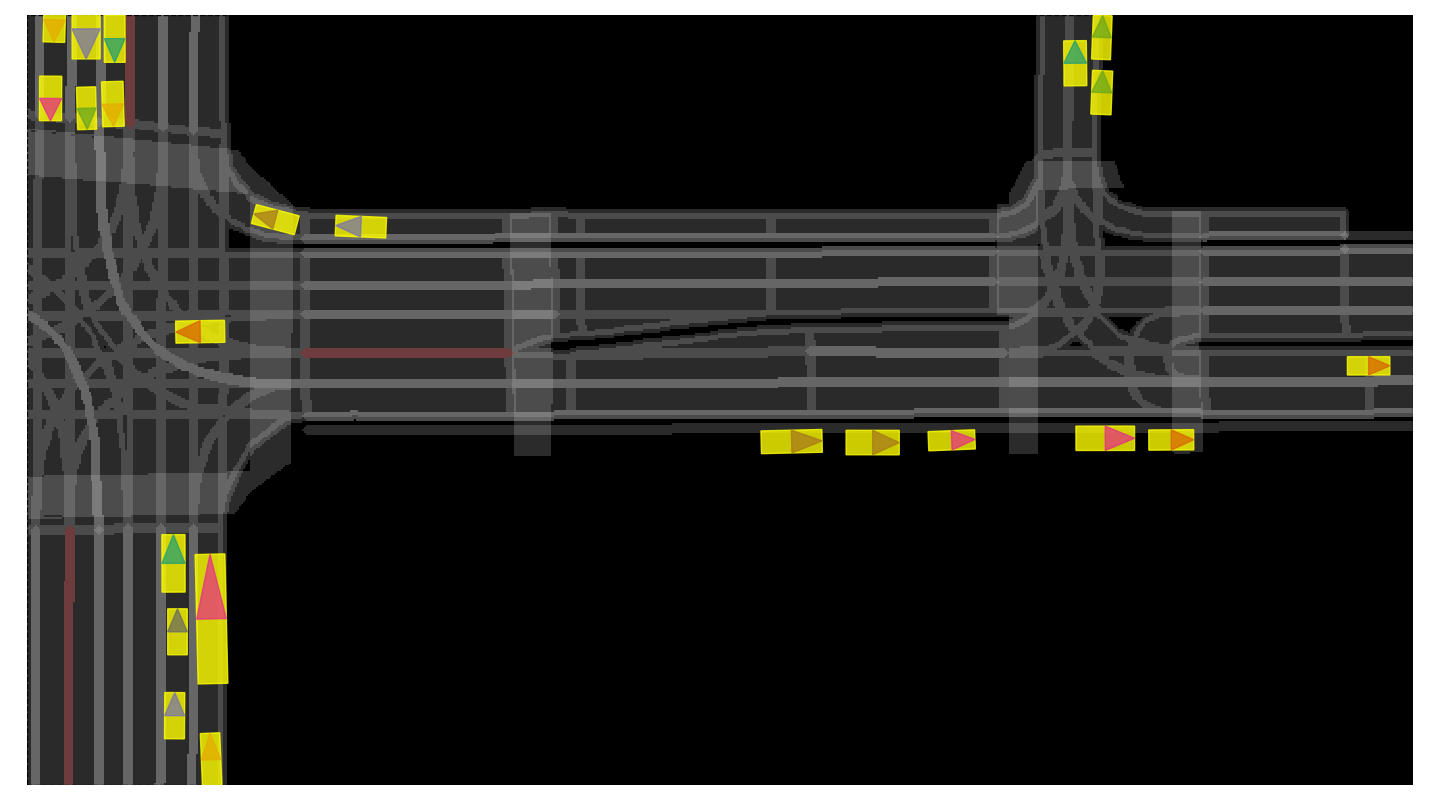}} &
        \raisebox{-0.5\height}{\includegraphics[width=0.248\linewidth]{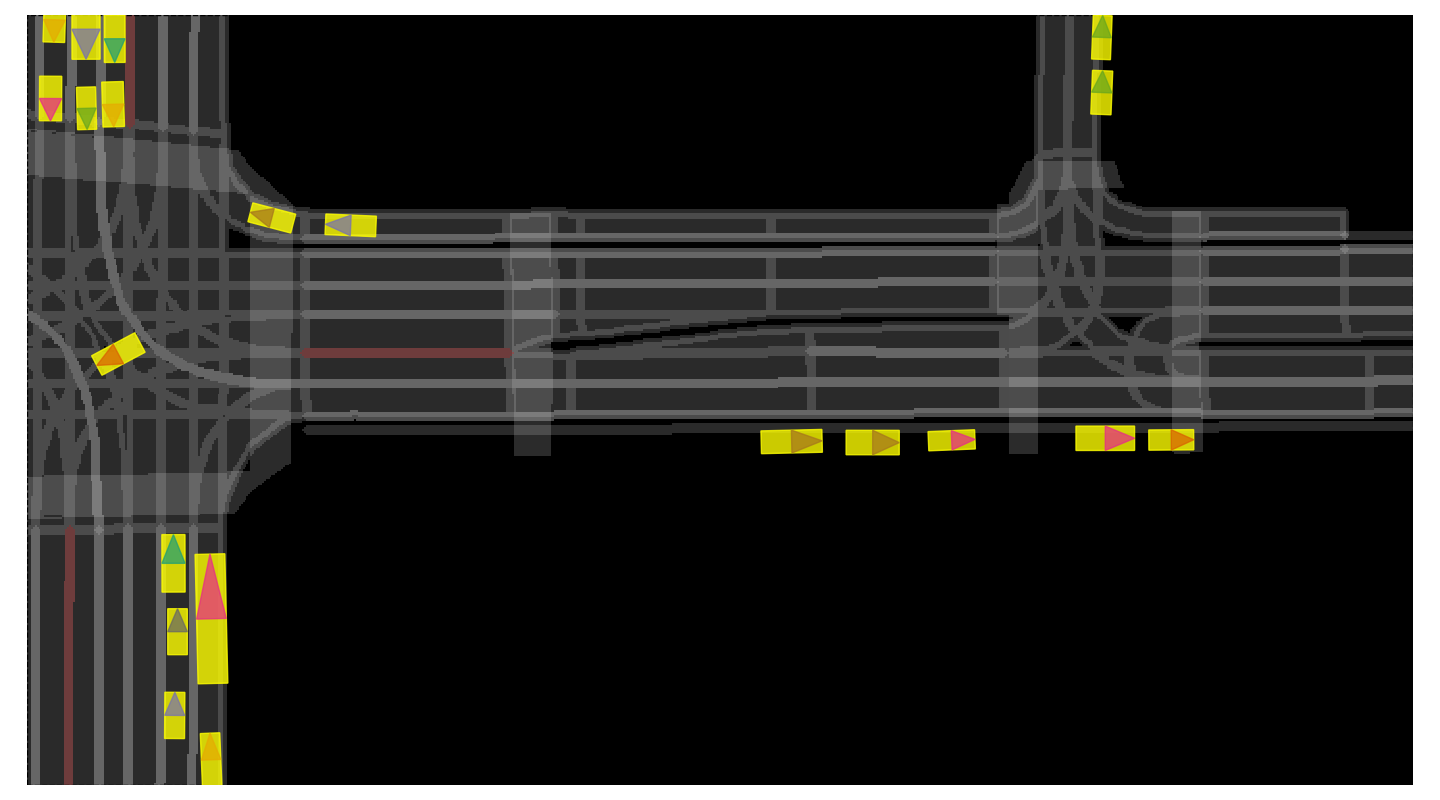}} \vspace{.1em} \\
        \rotatebox[origin=c]{90}{\textbf{Scenario 16}} &
        \raisebox{-0.5\height}{\includegraphics[width=0.248\linewidth]{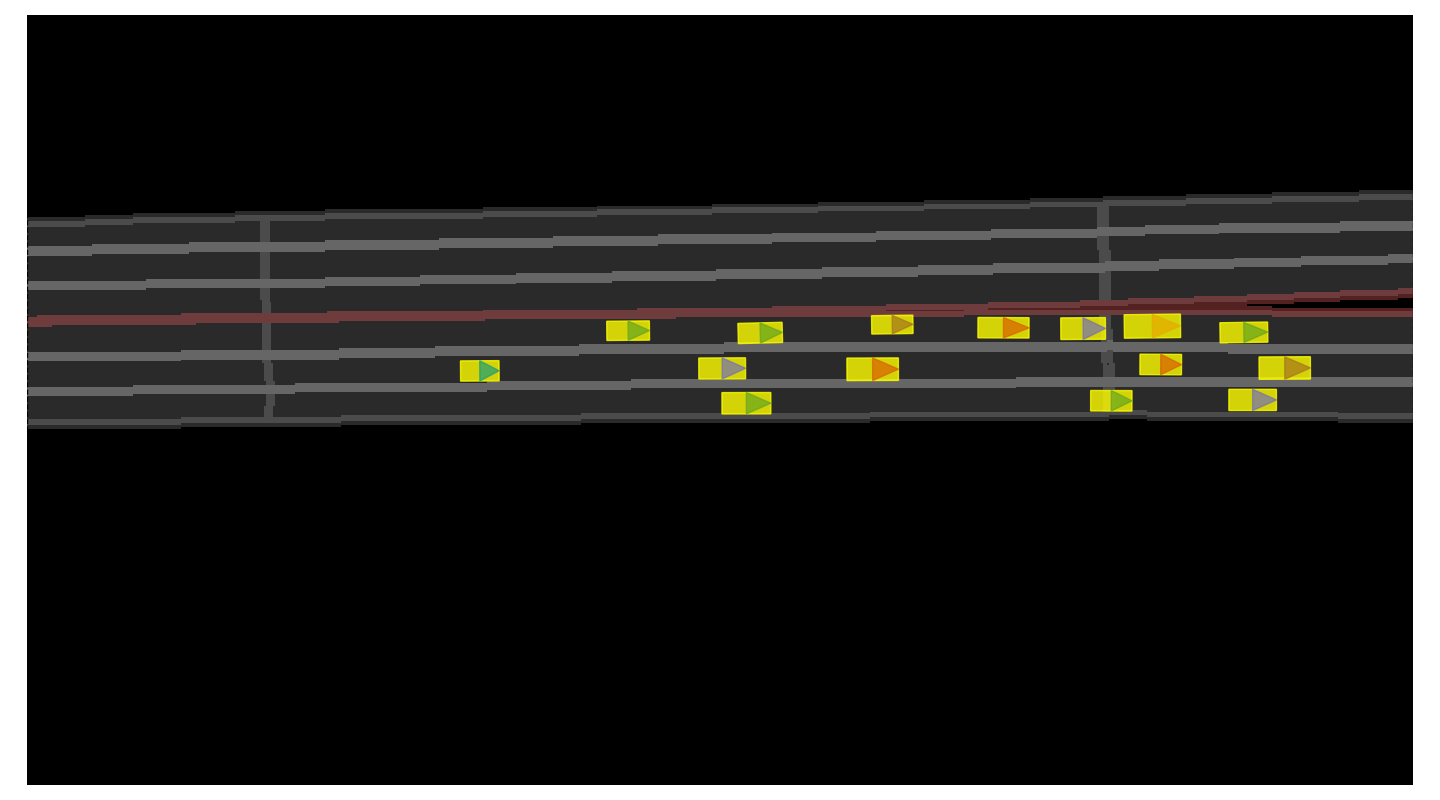}} &
        \raisebox{-0.5\height}{\includegraphics[width=0.248\linewidth]{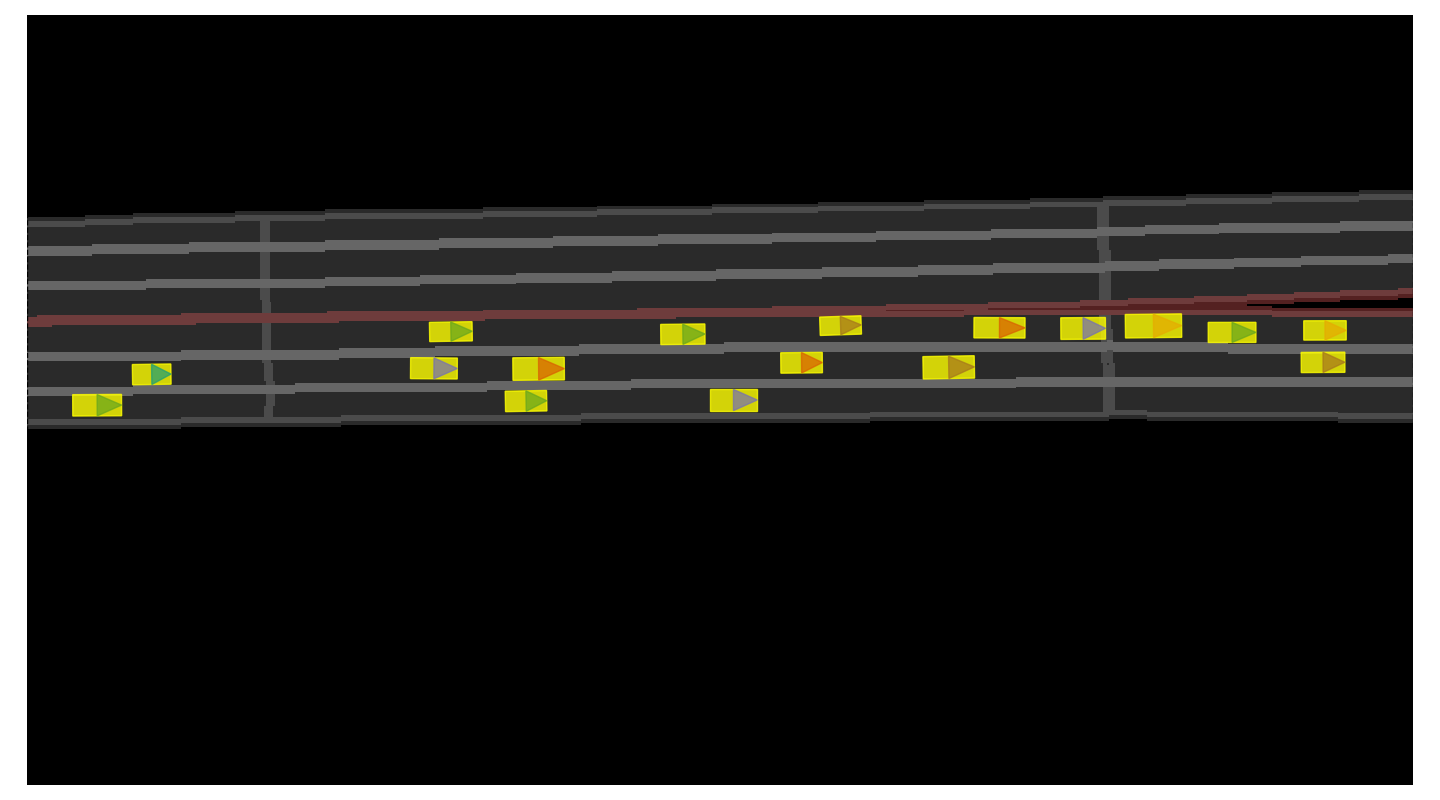}} &
        \raisebox{-0.5\height}{\includegraphics[width=0.248\linewidth]{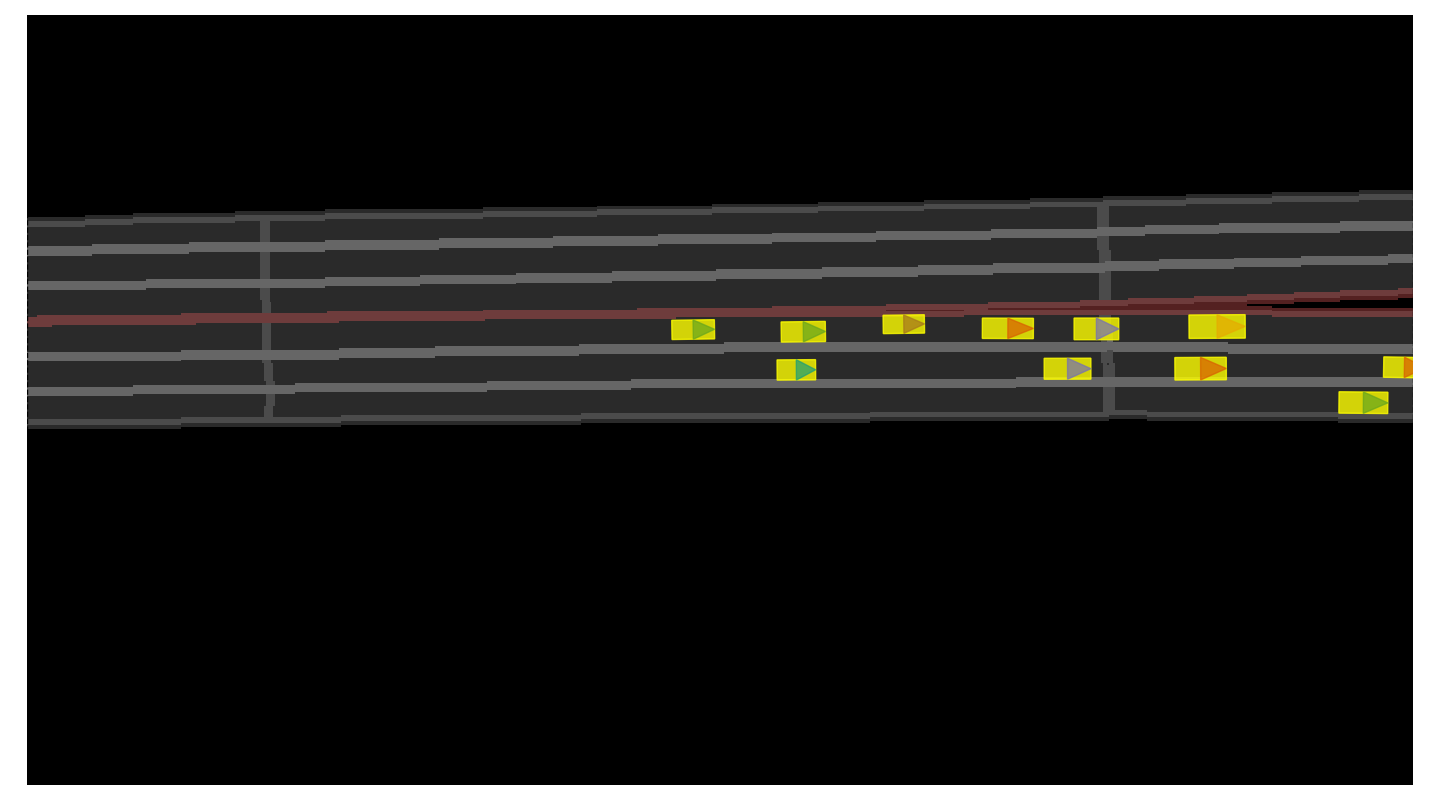}} &
        \raisebox{-0.5\height}{\includegraphics[width=0.248\linewidth]{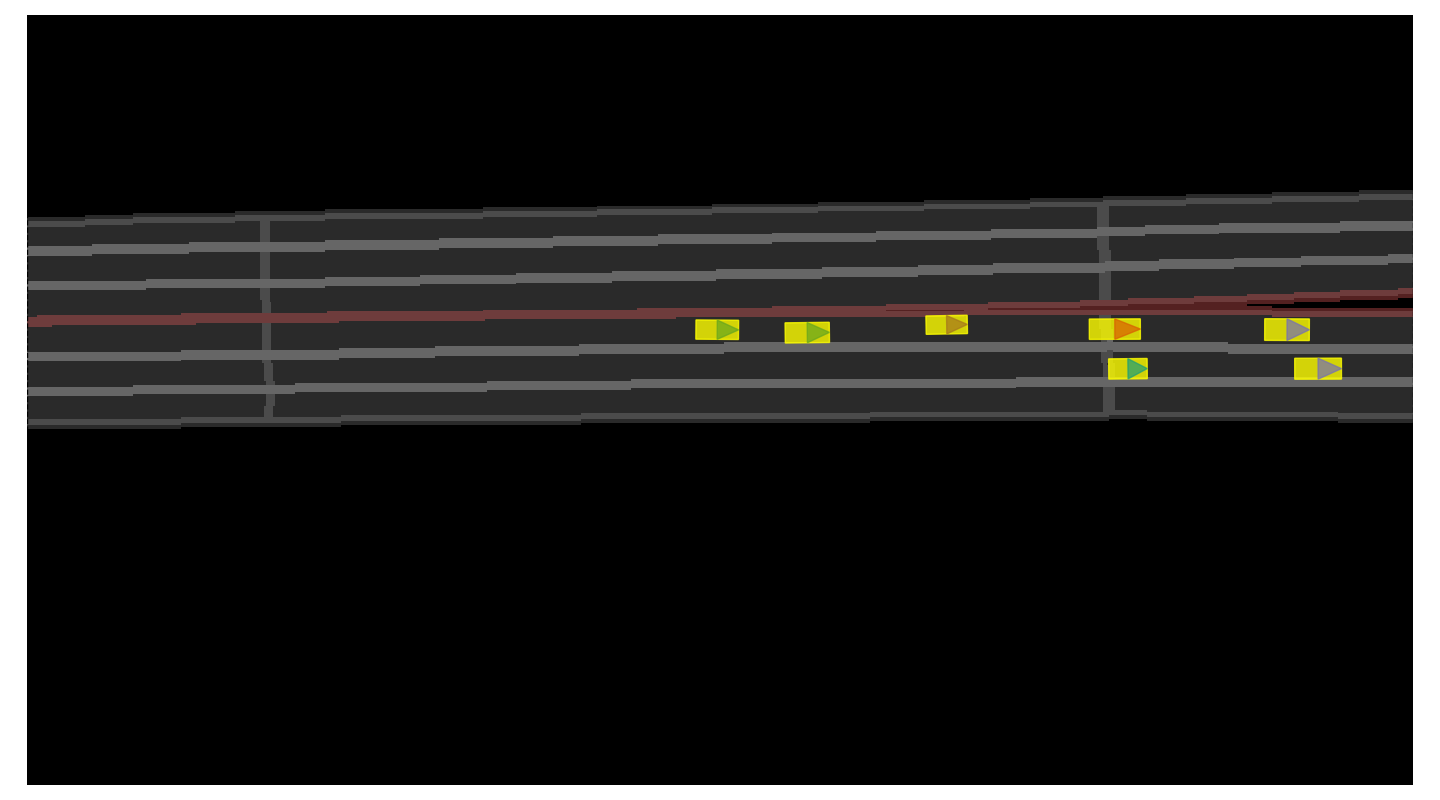}} \vspace{.1em} \\
        \rotatebox[origin=c]{90}{\textbf{Scenario 17}} &
        \raisebox{-0.5\height}{\includegraphics[width=0.248\linewidth]{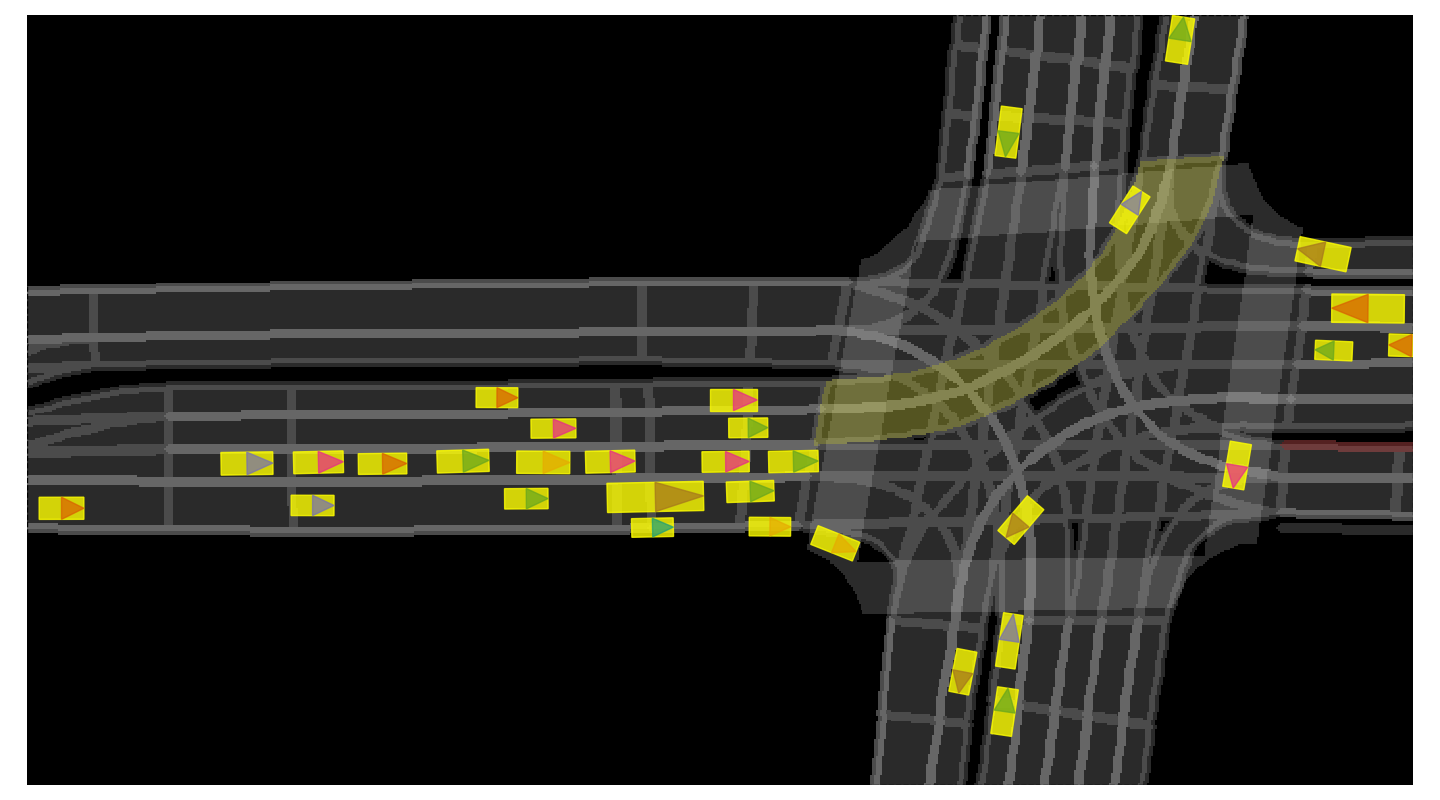}} &
        \raisebox{-0.5\height}{\includegraphics[width=0.248\linewidth]{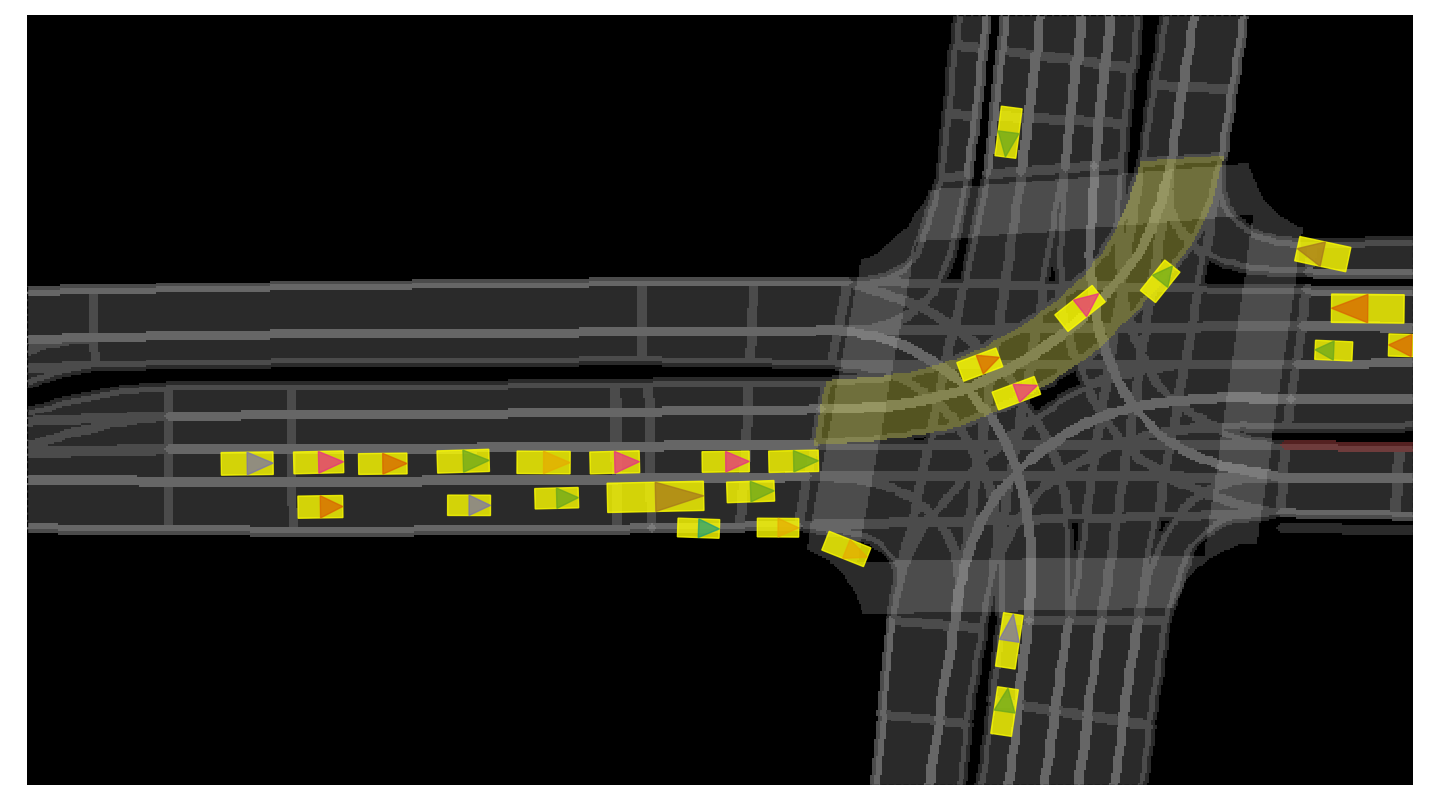}} &
        \raisebox{-0.5\height}{\includegraphics[width=0.248\linewidth]{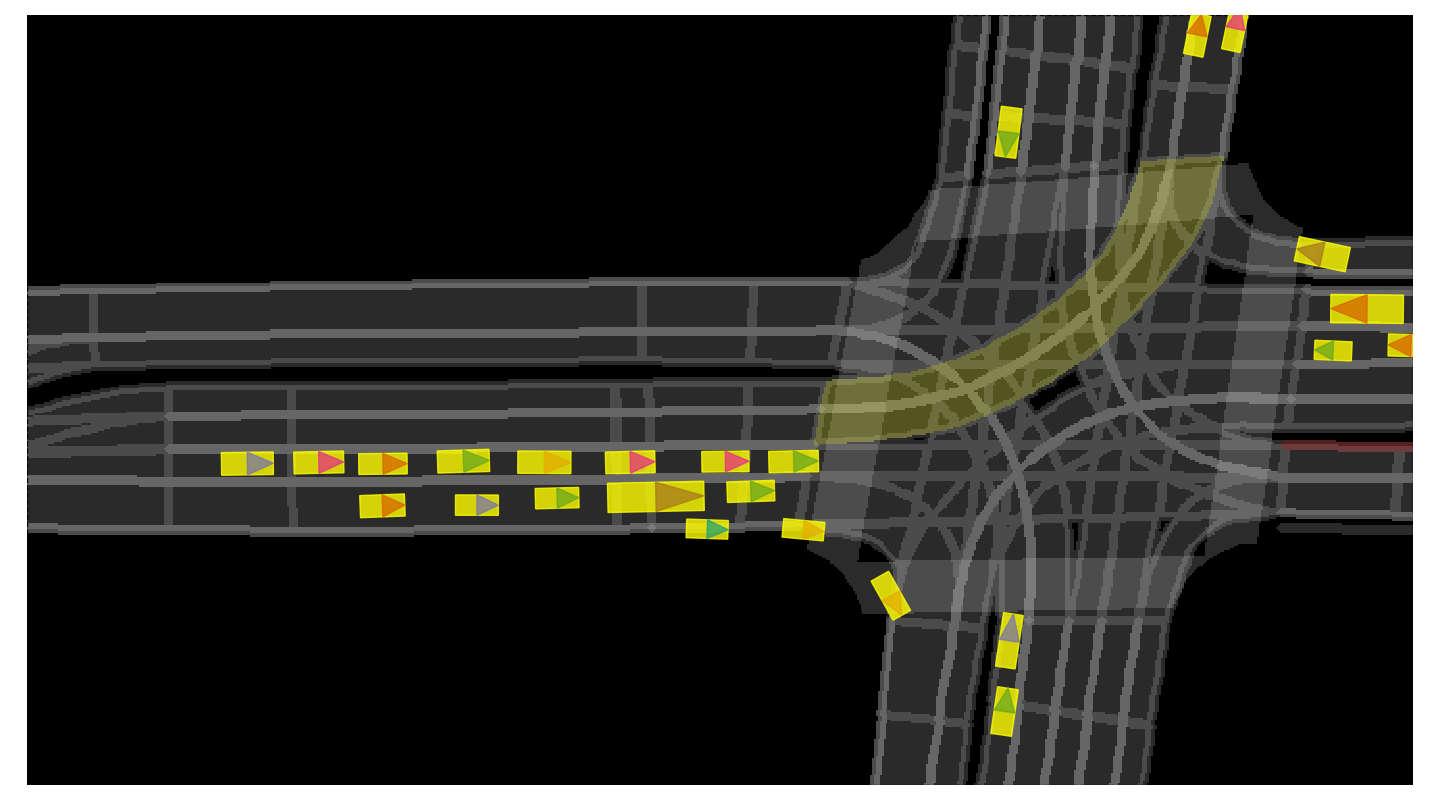}} &
        \raisebox{-0.5\height}{\includegraphics[width=0.248\linewidth]{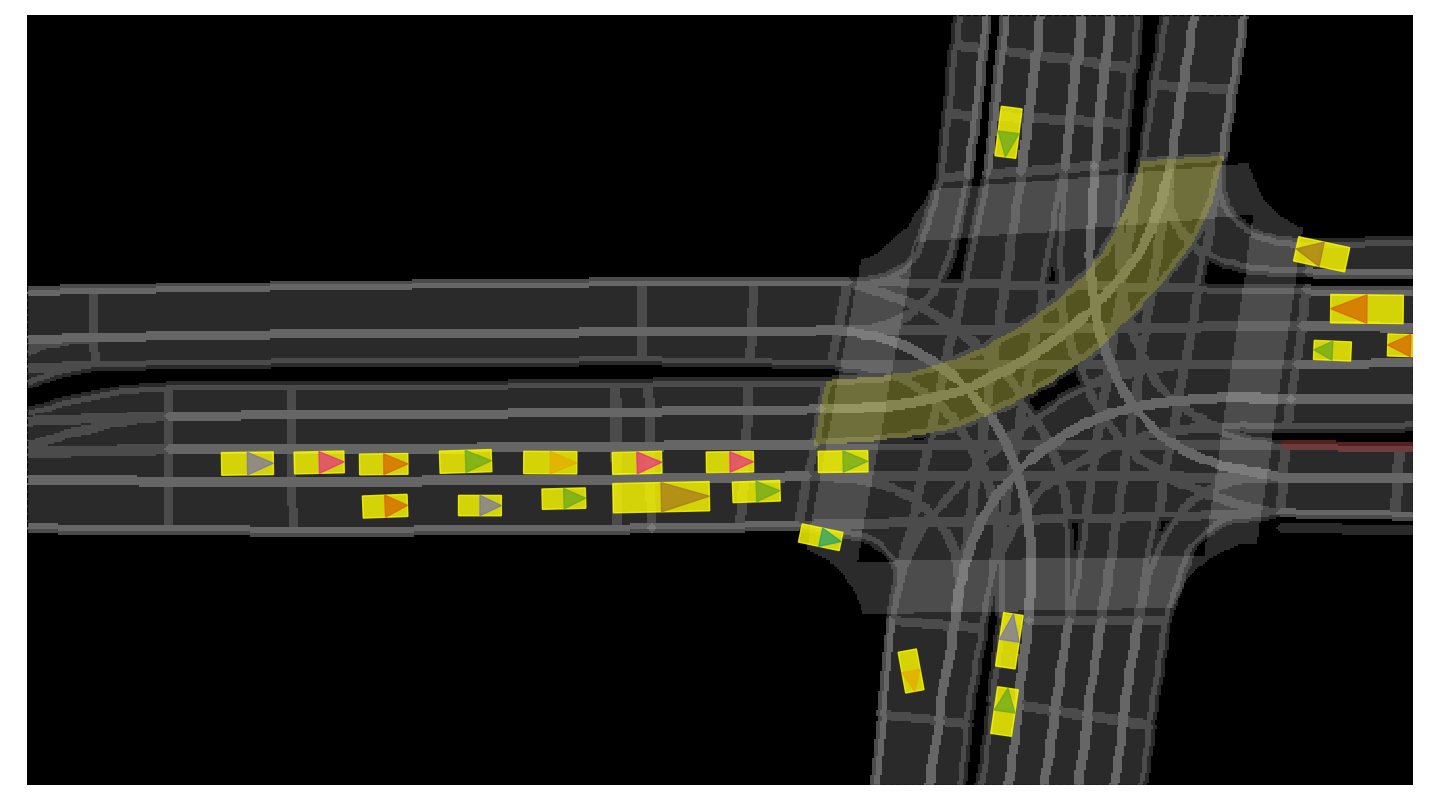}} \vspace{.1em} \\
        \rotatebox[origin=c]{90}{\textbf{Scenario 18}} &
        \raisebox{-0.5\height}{\includegraphics[width=0.248\linewidth]{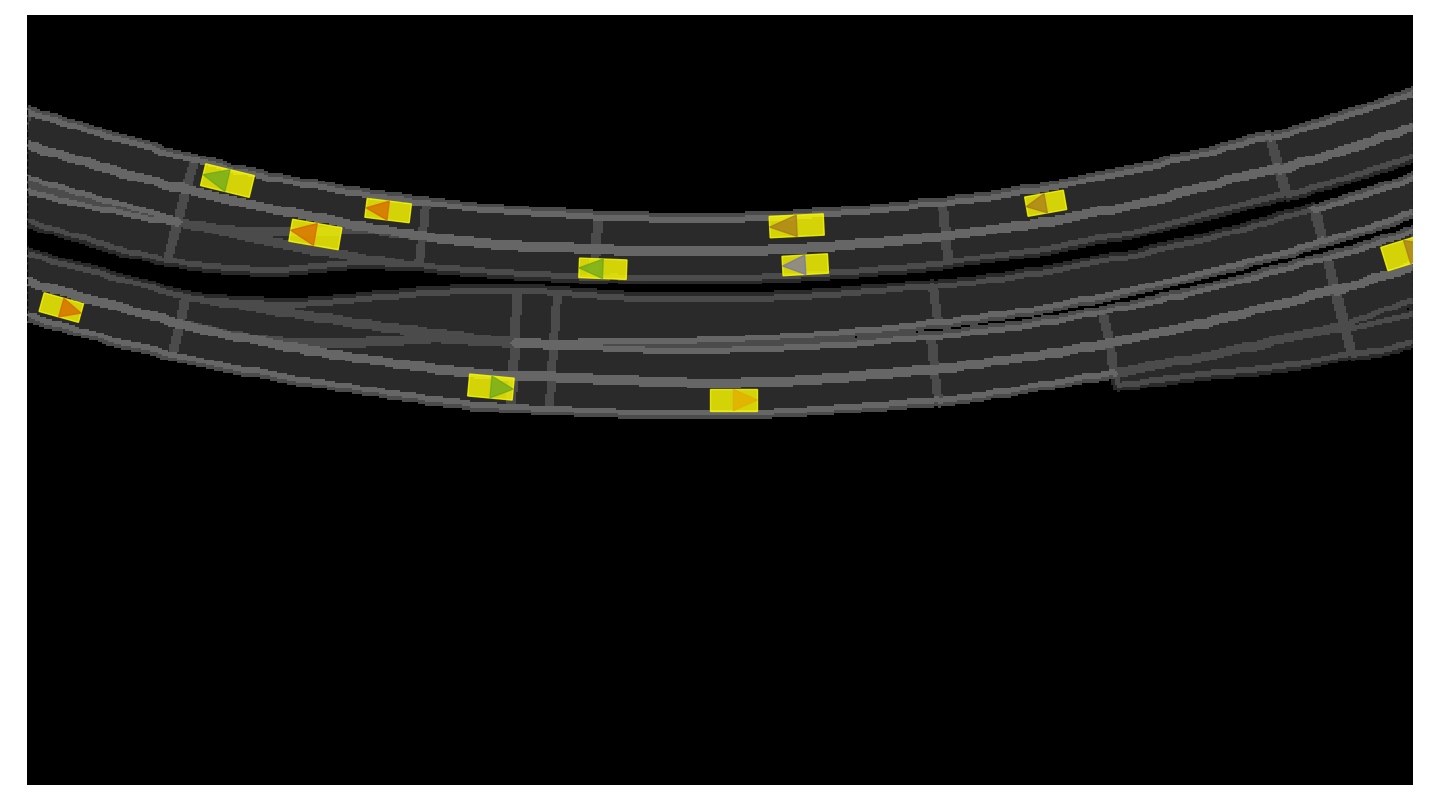}} &
        \raisebox{-0.5\height}{\includegraphics[width=0.248\linewidth]{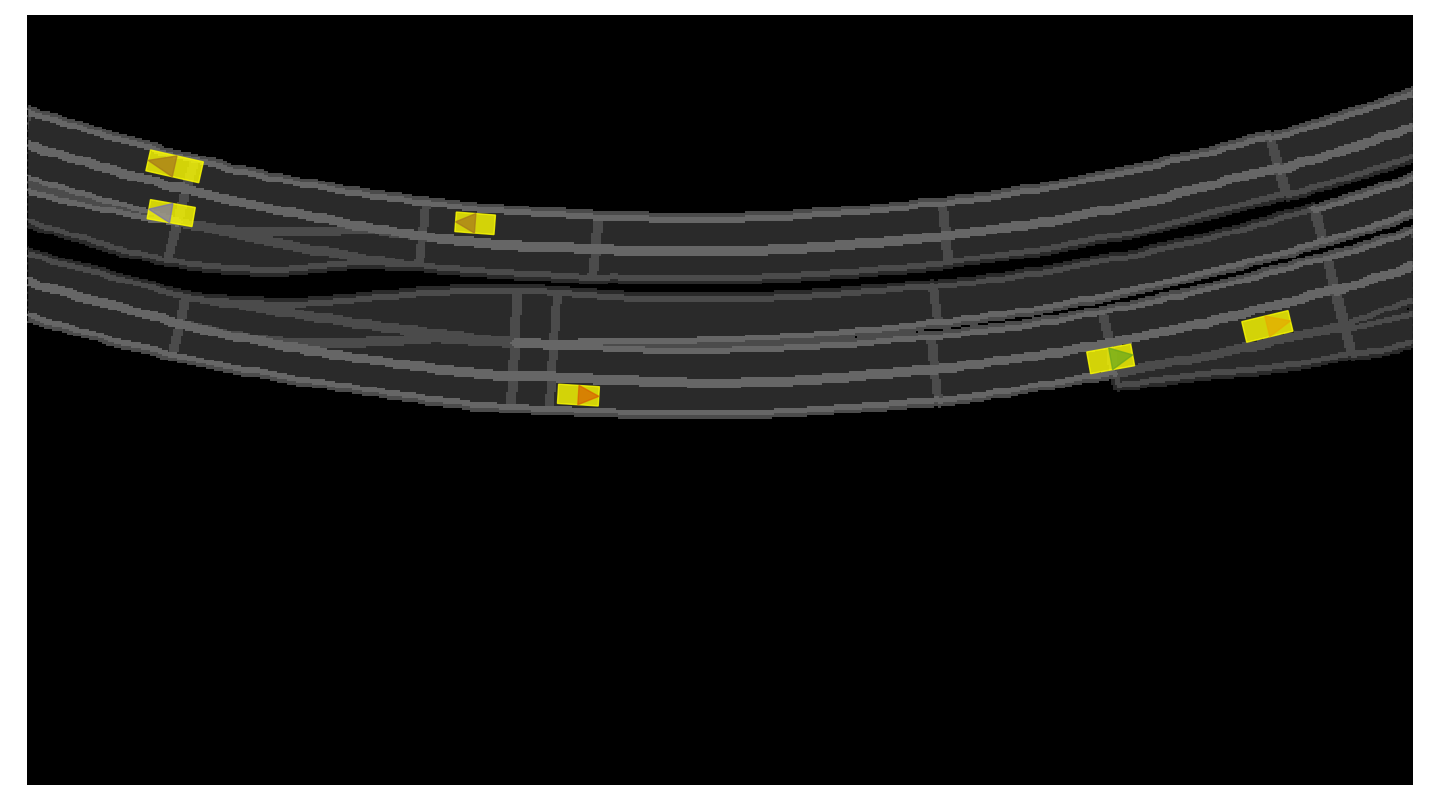}} &
        \raisebox{-0.5\height}{\includegraphics[width=0.248\linewidth]{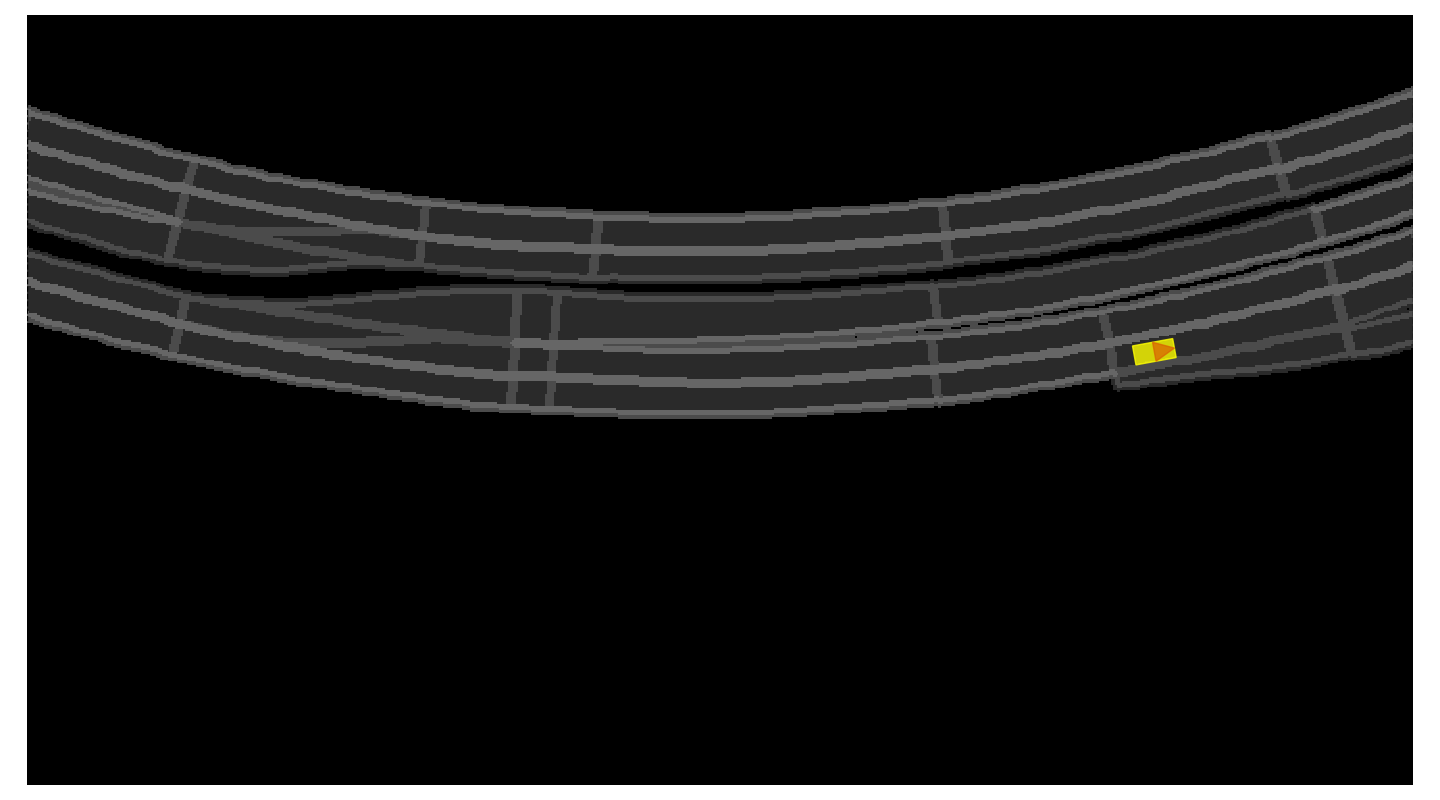}} &
        \raisebox{-0.5\height}{\includegraphics[width=0.248\linewidth]{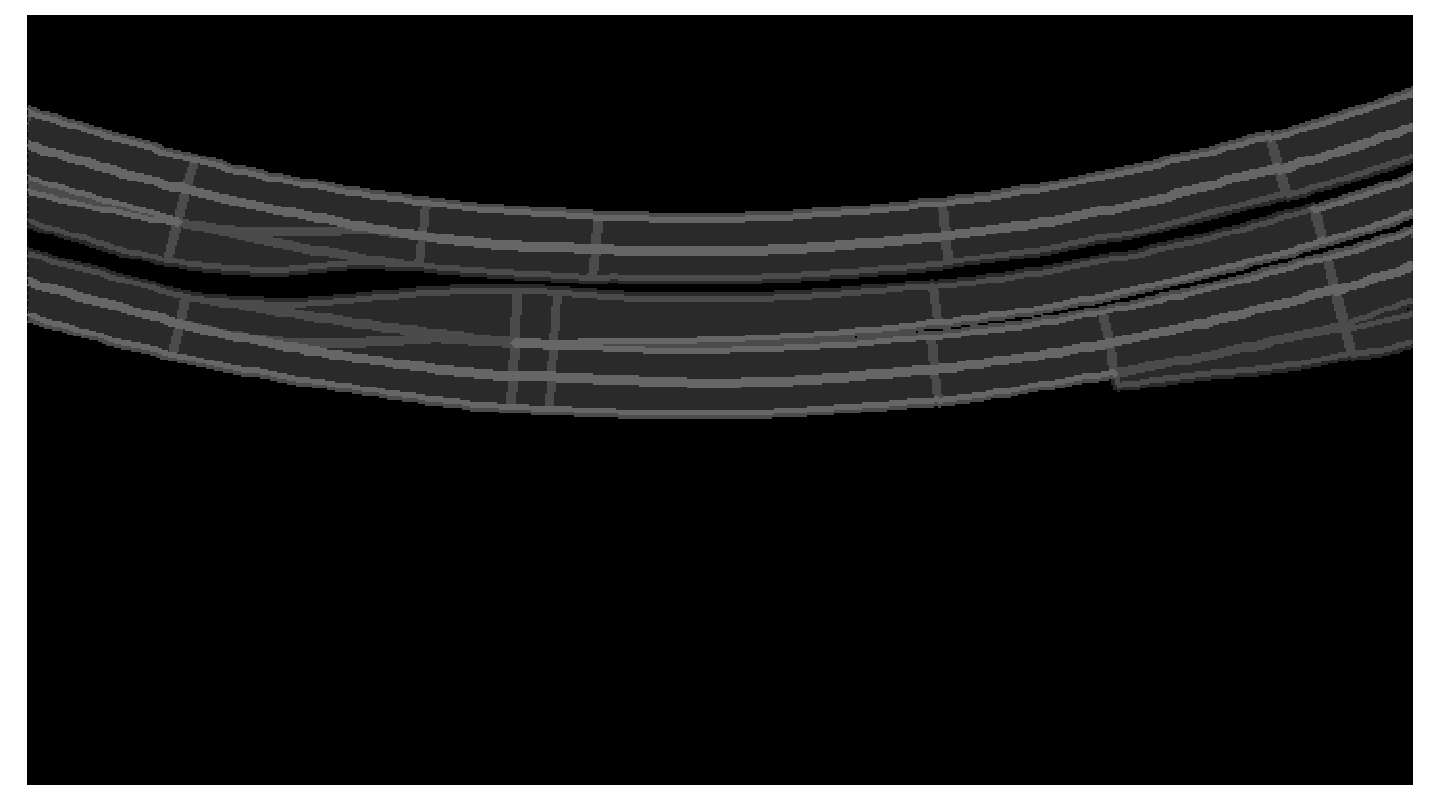}} \\
    \end{tabular}
    \caption{Simulated traffic scenarios sampled from \ourmodelshort{}: colored triangle shows heading and tracks instances across time}
    \label{fig:supp_qualitative_our_2}
\end{figure*}

\begin{figure*}[t]
    \centering
    \begin{tabular} {@{}c@{\hspace{.1em}}c@{\hspace{.5em}}c@{\hspace{.5em}}c}
        {} & \textbf{Scenario 1} & \textbf{Scenario 2 } & \textbf{Scenario 3} \\
        \rotatebox[origin=c]{90}{\textbf{IDM}} &
        \raisebox{-0.5\height}{\includegraphics[width=0.325\linewidth, trim={3cm, 3cm, 3cm, 1cm}, clip]{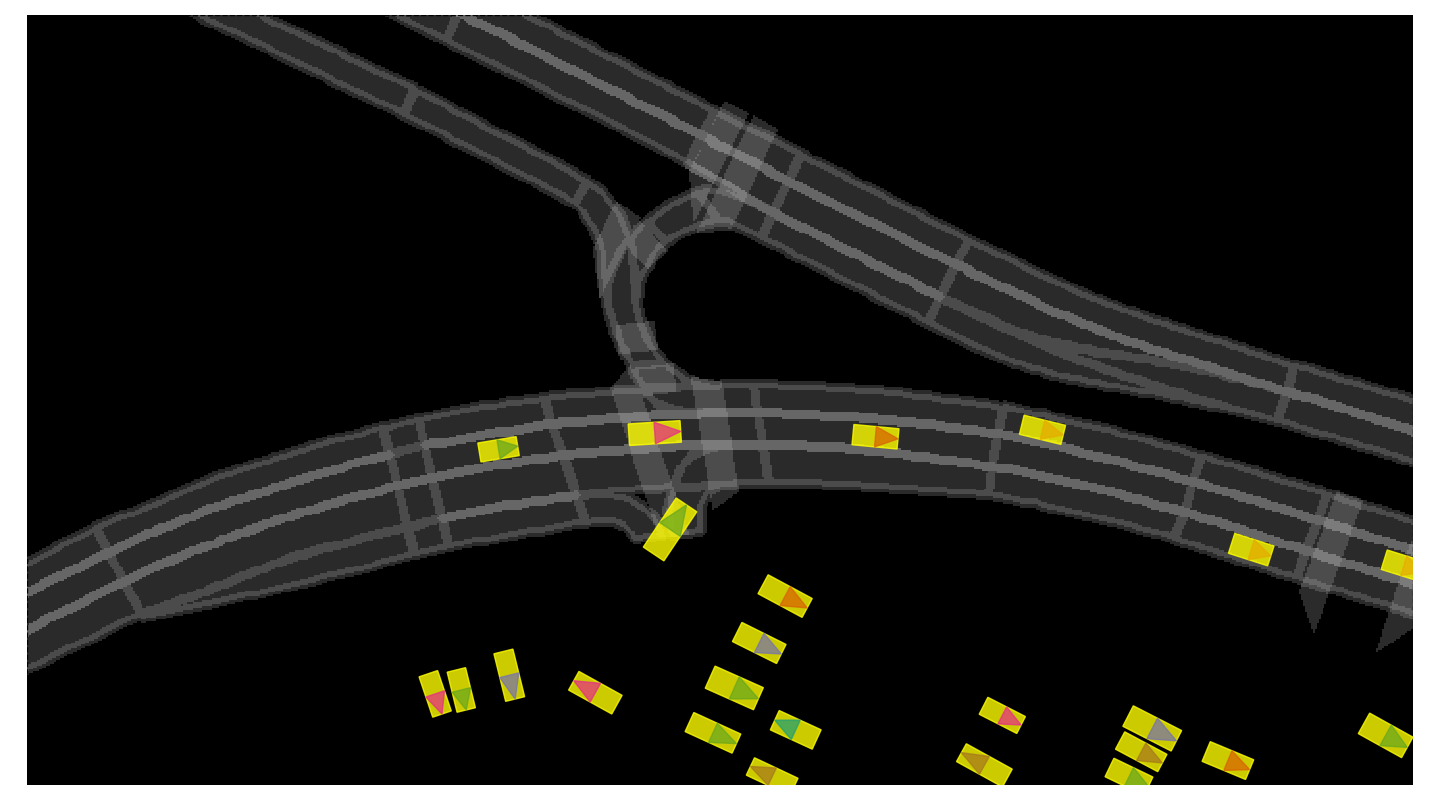}} &
        \raisebox{-0.5\height}{\includegraphics[width=0.325\linewidth, trim={3cm, 3cm, 3cm, 1cm}, clip]{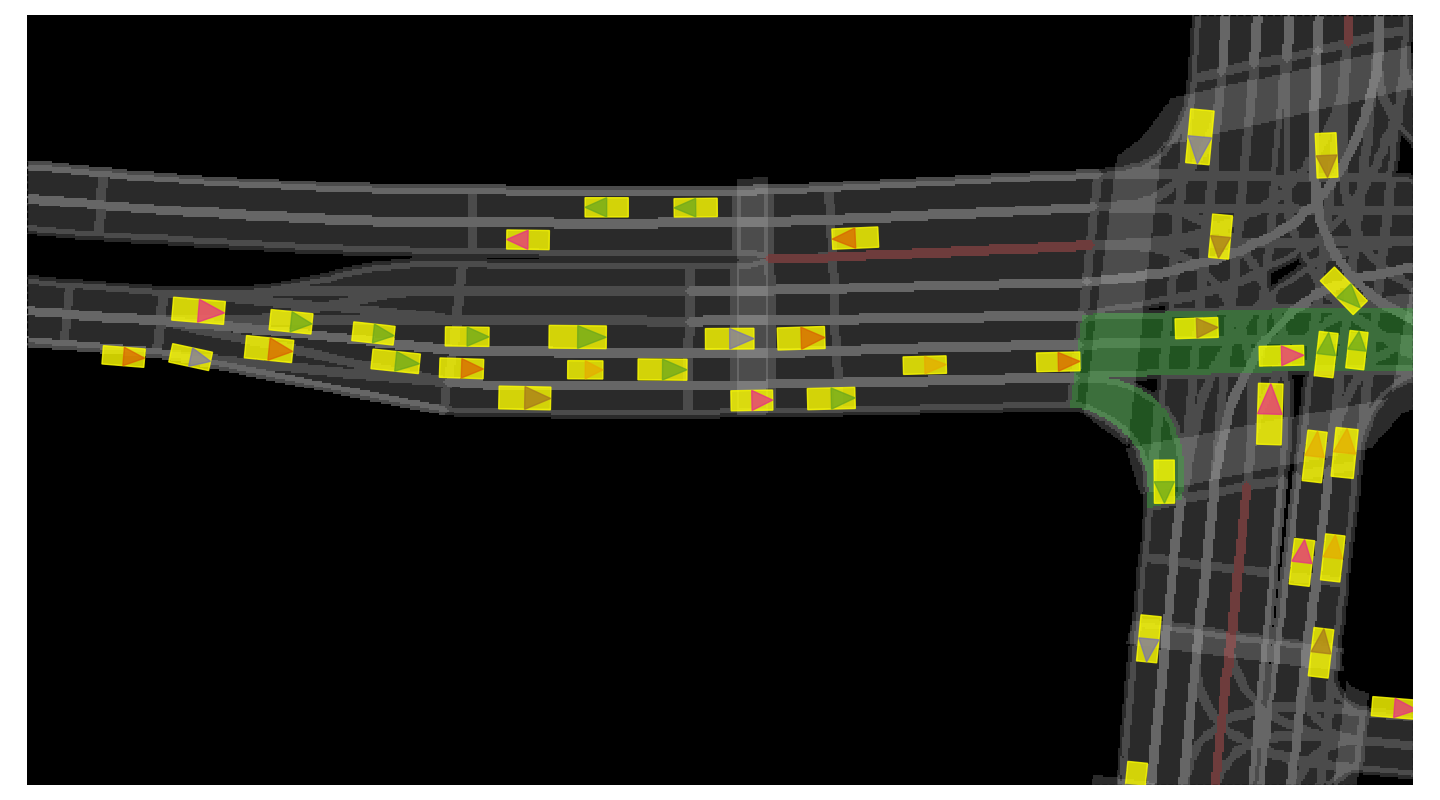}} &
        \raisebox{-0.5\height}{\includegraphics[width=0.325\linewidth, trim={3cm, 3cm, 3cm, 1cm}, clip]{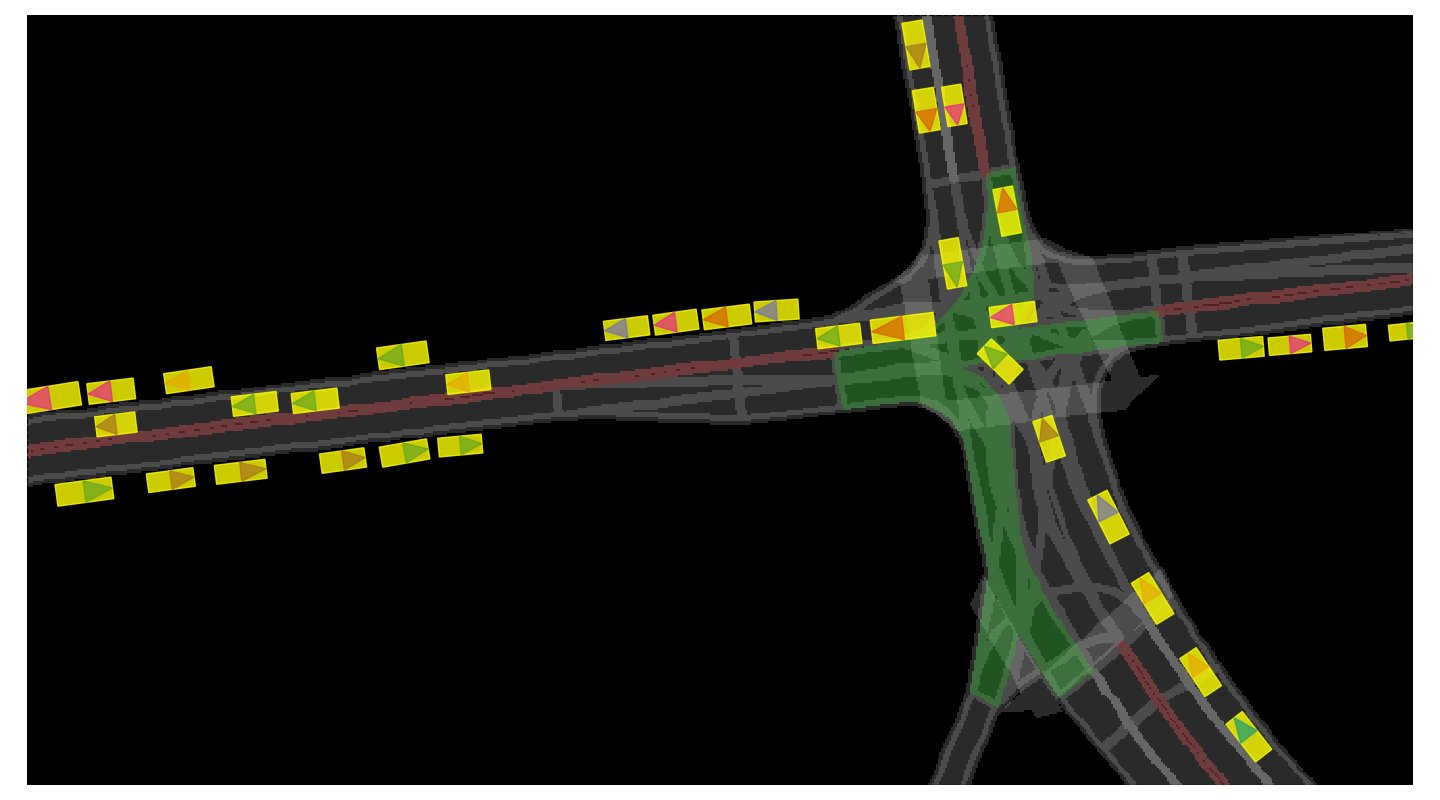}} \vspace{.1em} \\
        \rotatebox[origin=c]{90}{\textbf{MTP}} &
        \raisebox{-0.5\height}{\includegraphics[width=0.325\linewidth, trim={3cm, 3cm, 3cm, 1cm}, clip]{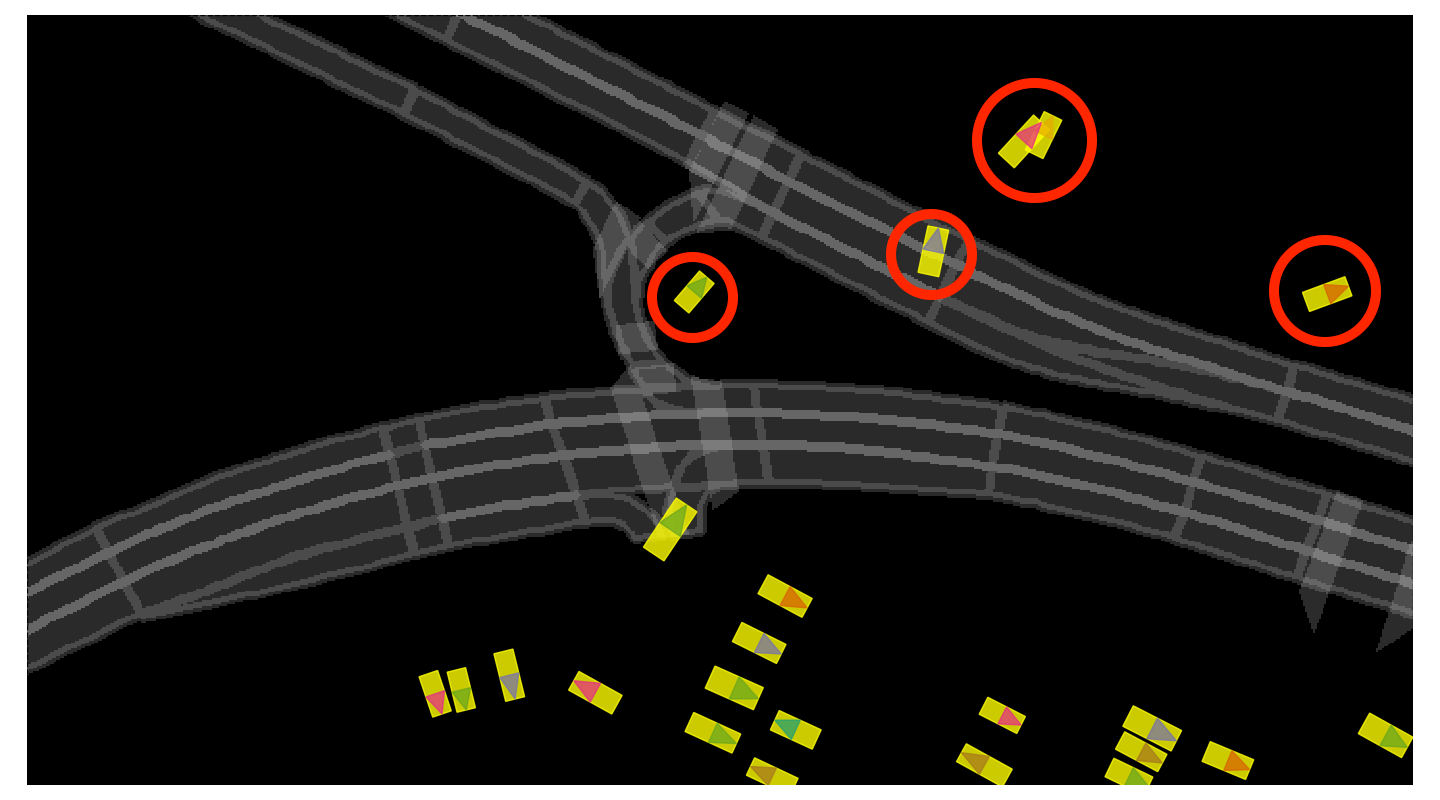}} &
        \raisebox{-0.5\height}{\includegraphics[width=0.325\linewidth, trim={3cm, 3cm, 3cm, 1cm}, clip]{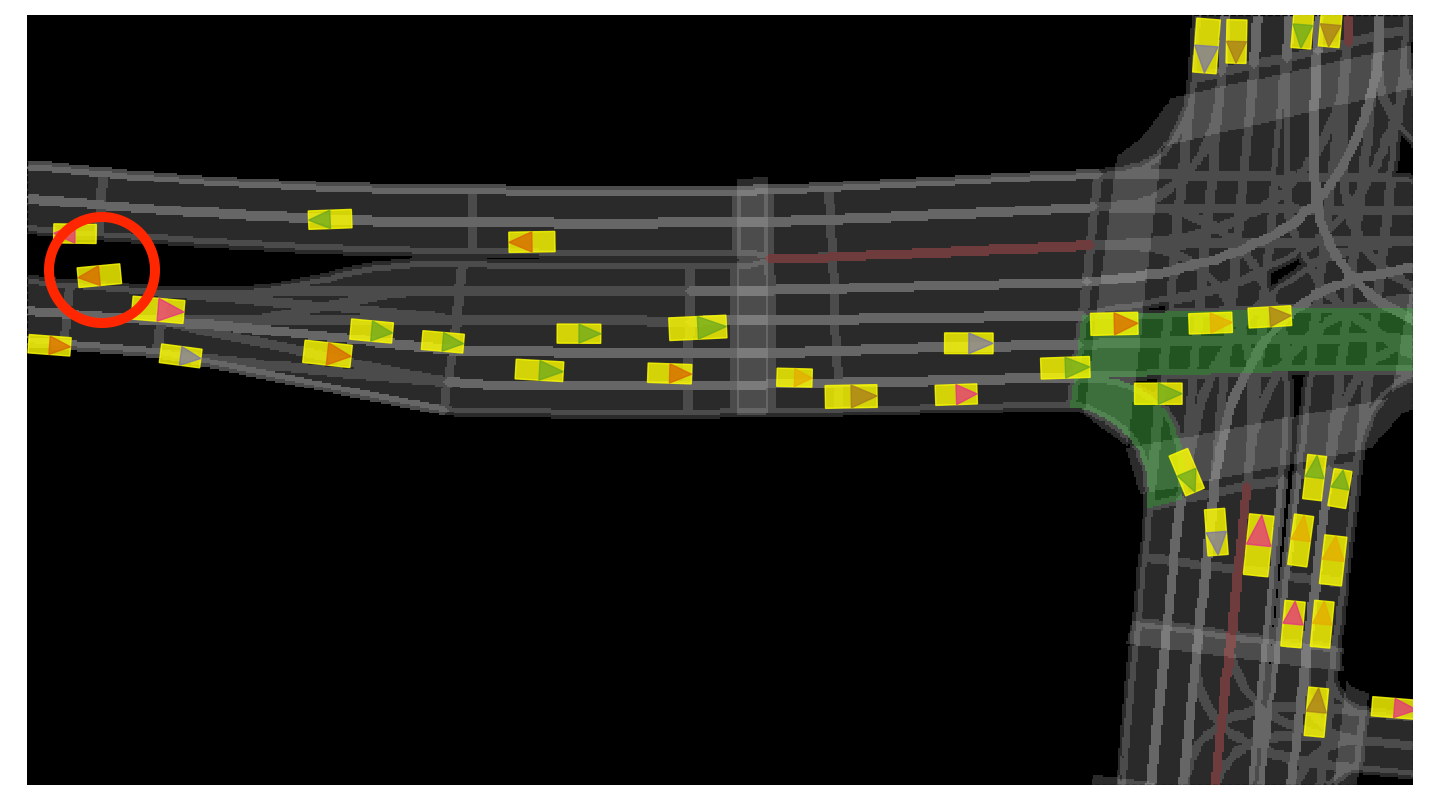}} &
        \raisebox{-0.5\height}{\includegraphics[width=0.325\linewidth, trim={3cm, 3cm, 3cm, 1cm}, clip]{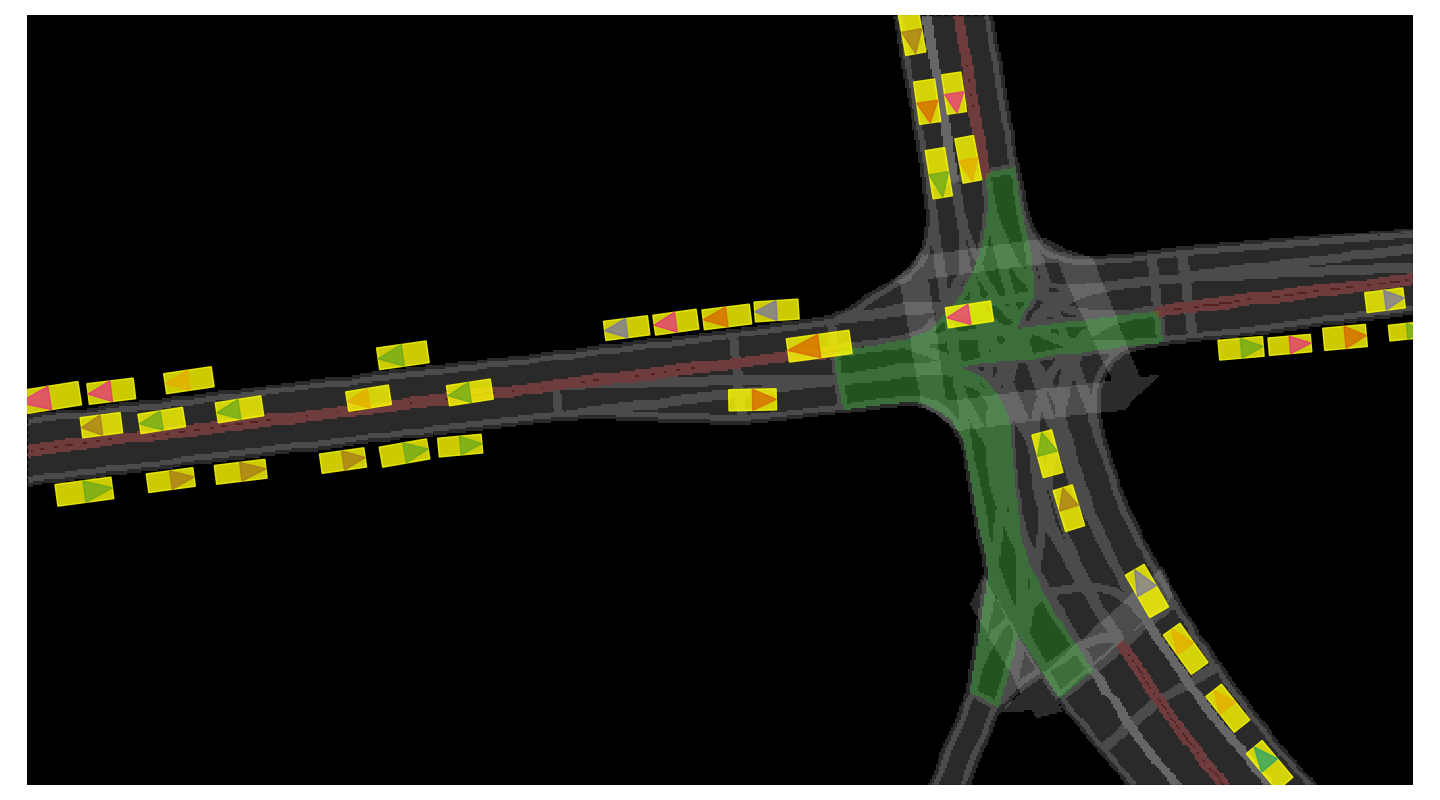}} \vspace{.1em} \\
        \rotatebox[origin=c]{90}{\textbf{ESP}} &
        \raisebox{-0.5\height}{\includegraphics[width=0.325\linewidth, trim={3cm, 3cm, 3cm, 1cm}, clip]{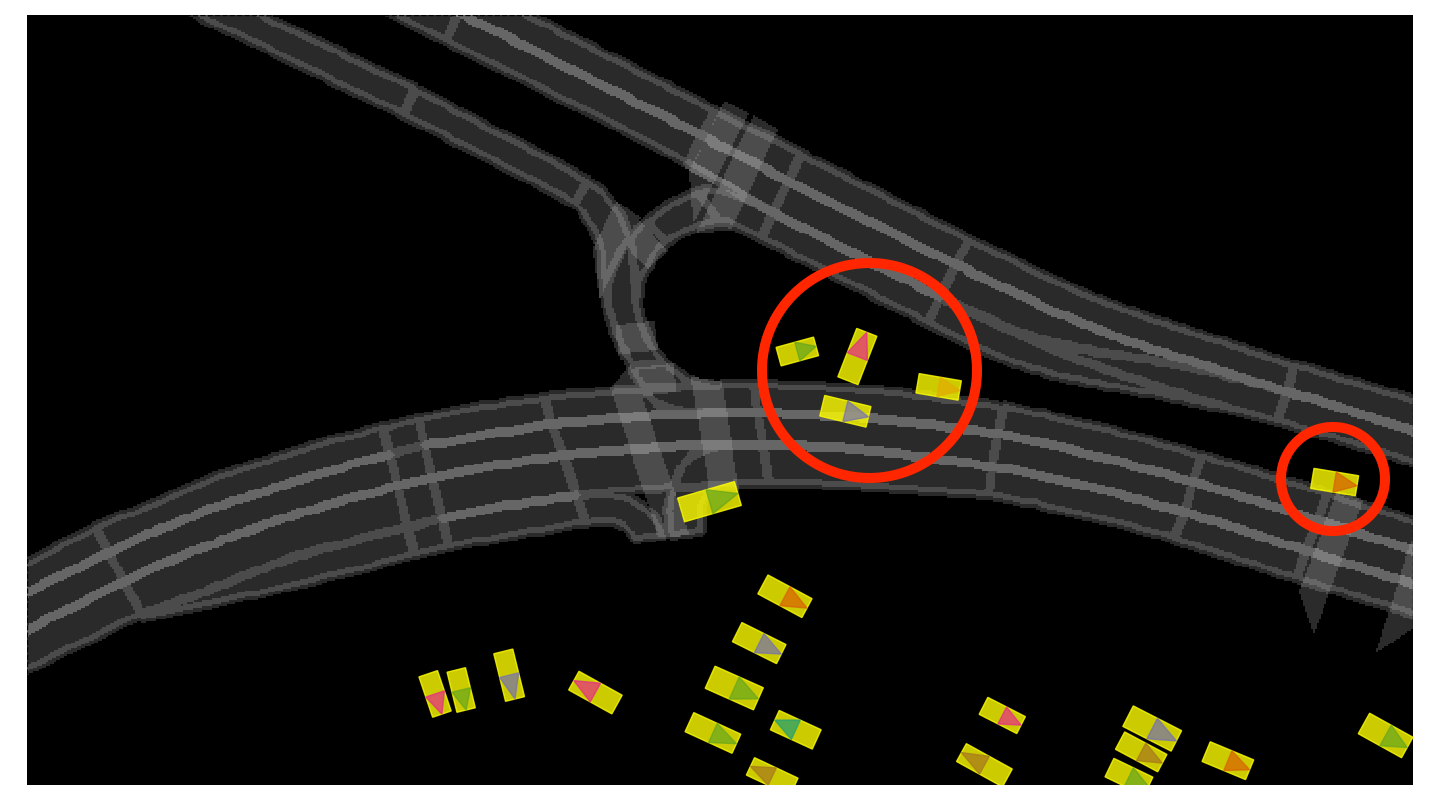}} &
        \raisebox{-0.5\height}{\includegraphics[width=0.325\linewidth, trim={3cm, 3cm, 3cm, 1cm}, clip]{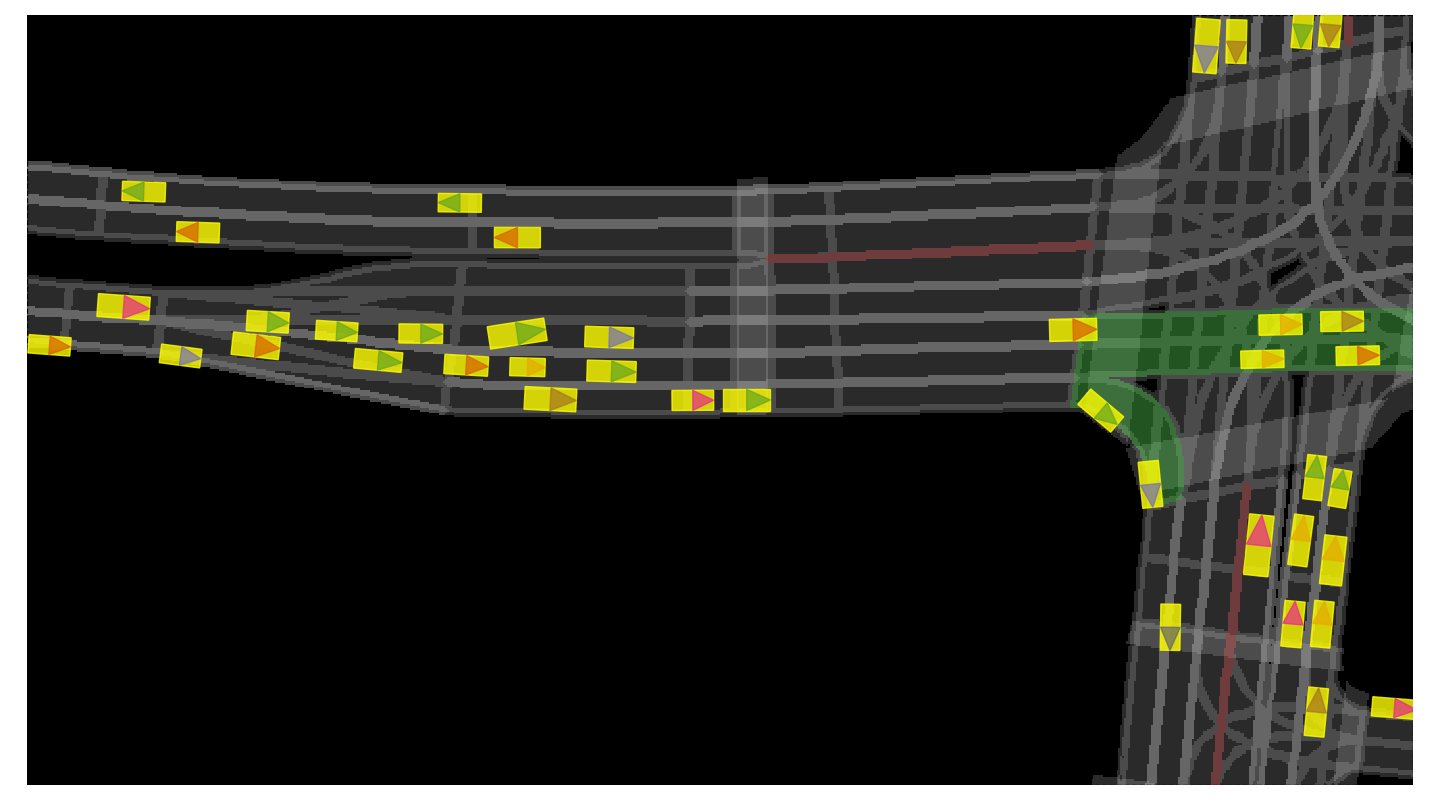}} &
        \raisebox{-0.5\height}{\includegraphics[width=0.325\linewidth, trim={3cm, 3cm, 3cm, 1cm}, clip]{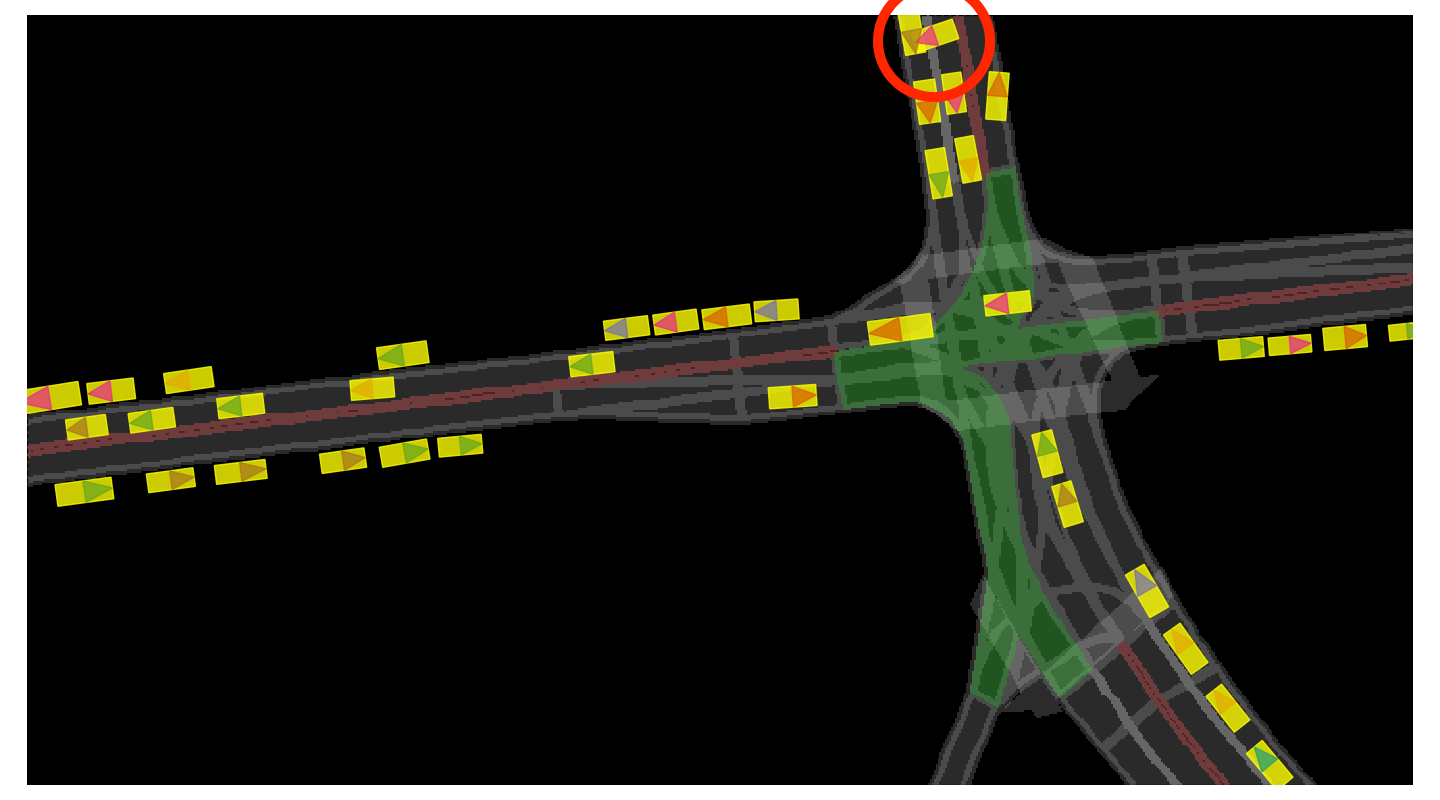}} \vspace{.1em} \\
        \rotatebox[origin=c]{90}{\textbf{ILVM}} &
        \raisebox{-0.5\height}{\includegraphics[width=0.325\linewidth, trim={3cm, 3cm, 3cm, 1cm}, clip]{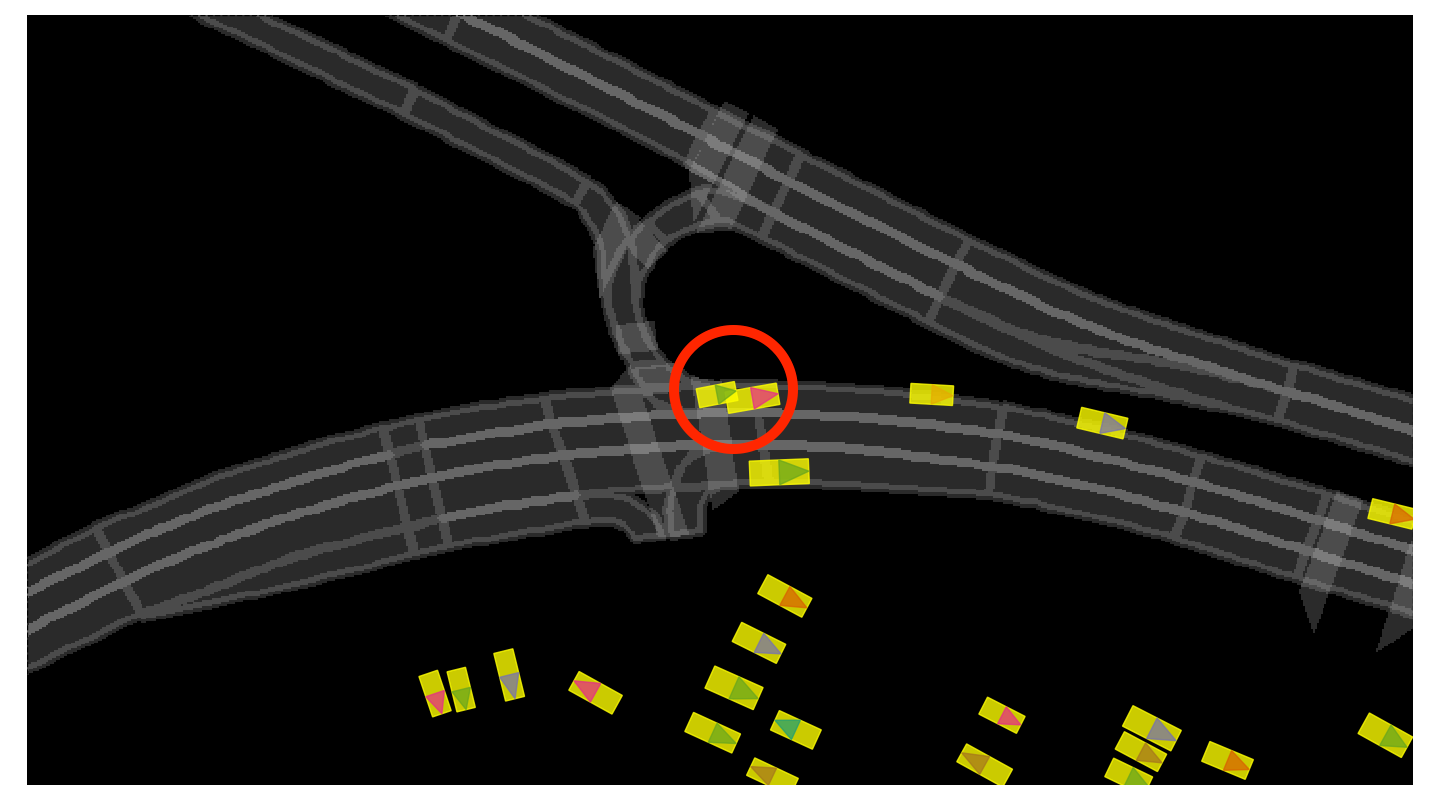}} &
        \raisebox{-0.5\height}{\includegraphics[width=0.325\linewidth, trim={3cm, 3cm, 3cm, 1cm}, clip]{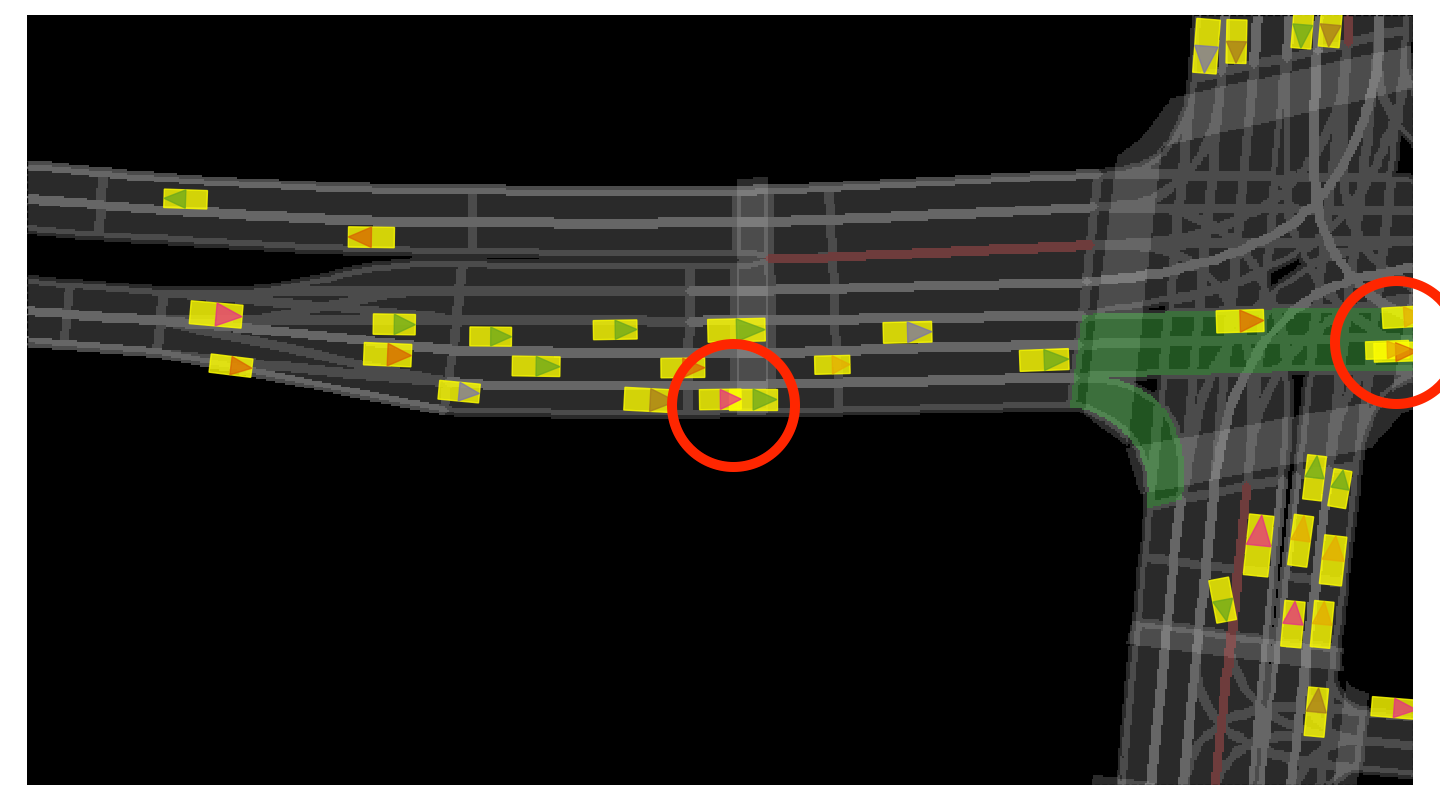}} &
        \raisebox{-0.5\height}{\includegraphics[width=0.325\linewidth, trim={3cm, 3cm, 3cm, 1cm}, clip]{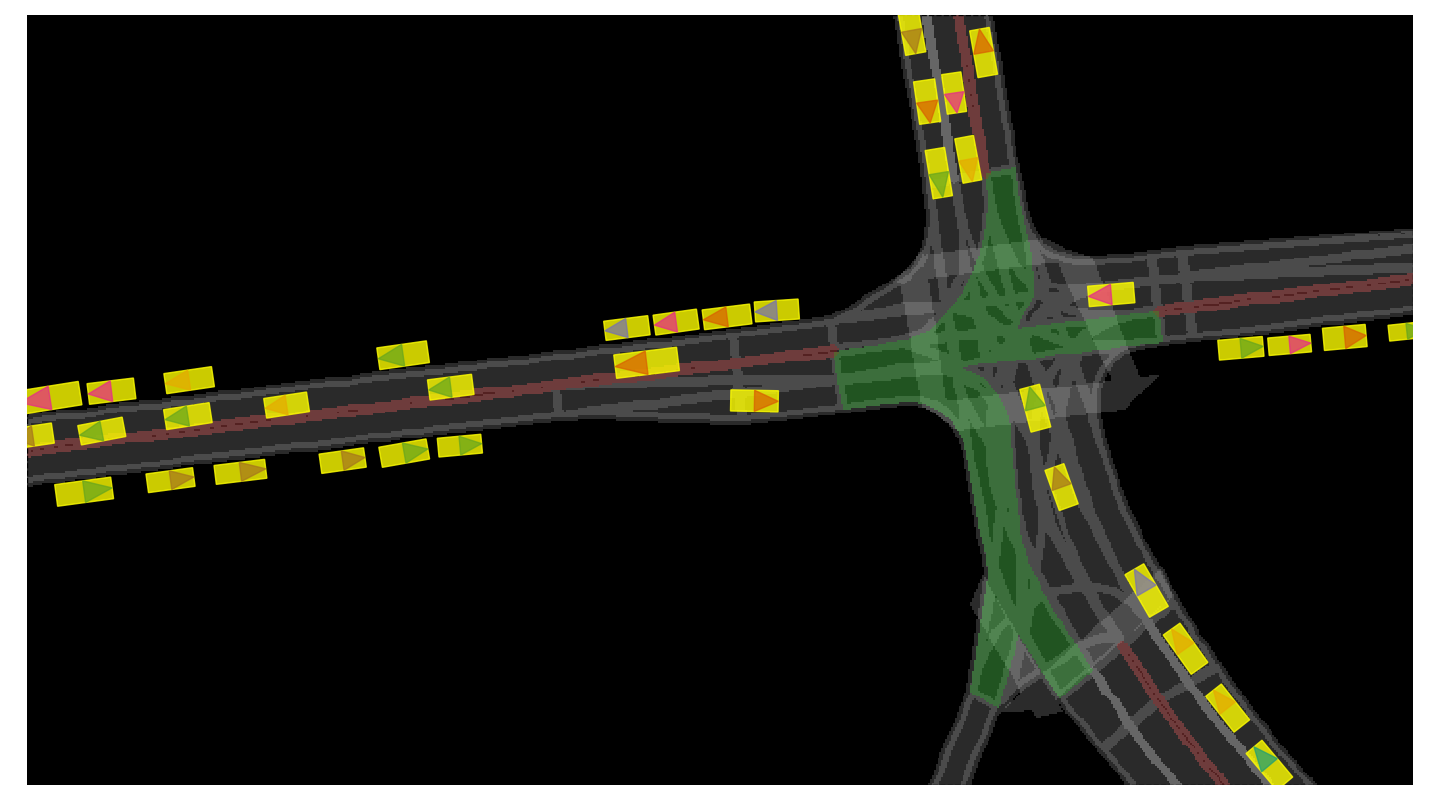}} \vspace{.1em} \\
        \rotatebox[origin=c]{90}{\textbf{DataAug}} &
        \raisebox{-0.5\height}{\includegraphics[width=0.325\linewidth, trim={3cm, 3cm, 3cm, 1cm}, clip]{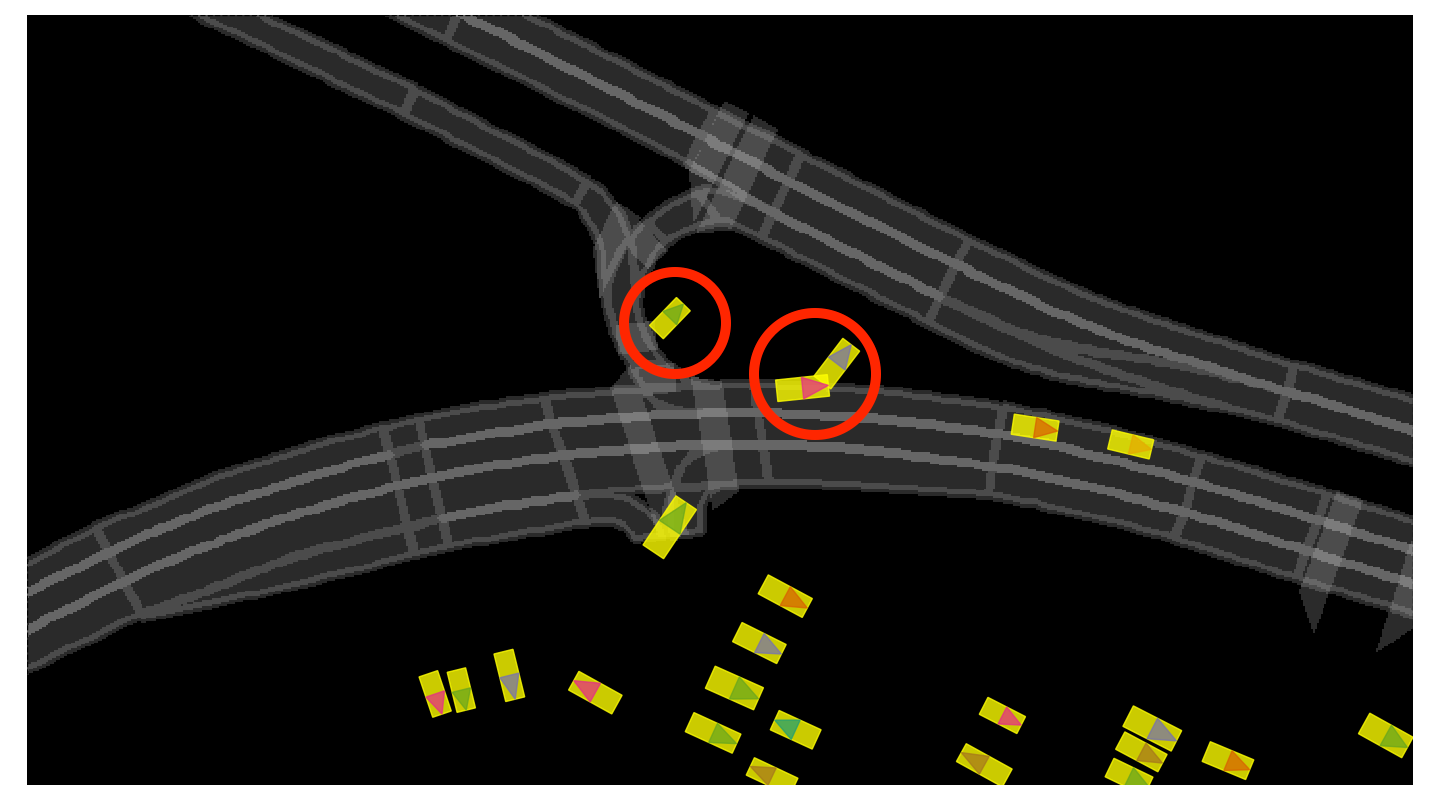}} &
        \raisebox{-0.5\height}{\includegraphics[width=0.325\linewidth, trim={3cm, 3cm, 3cm, 1cm}, clip]{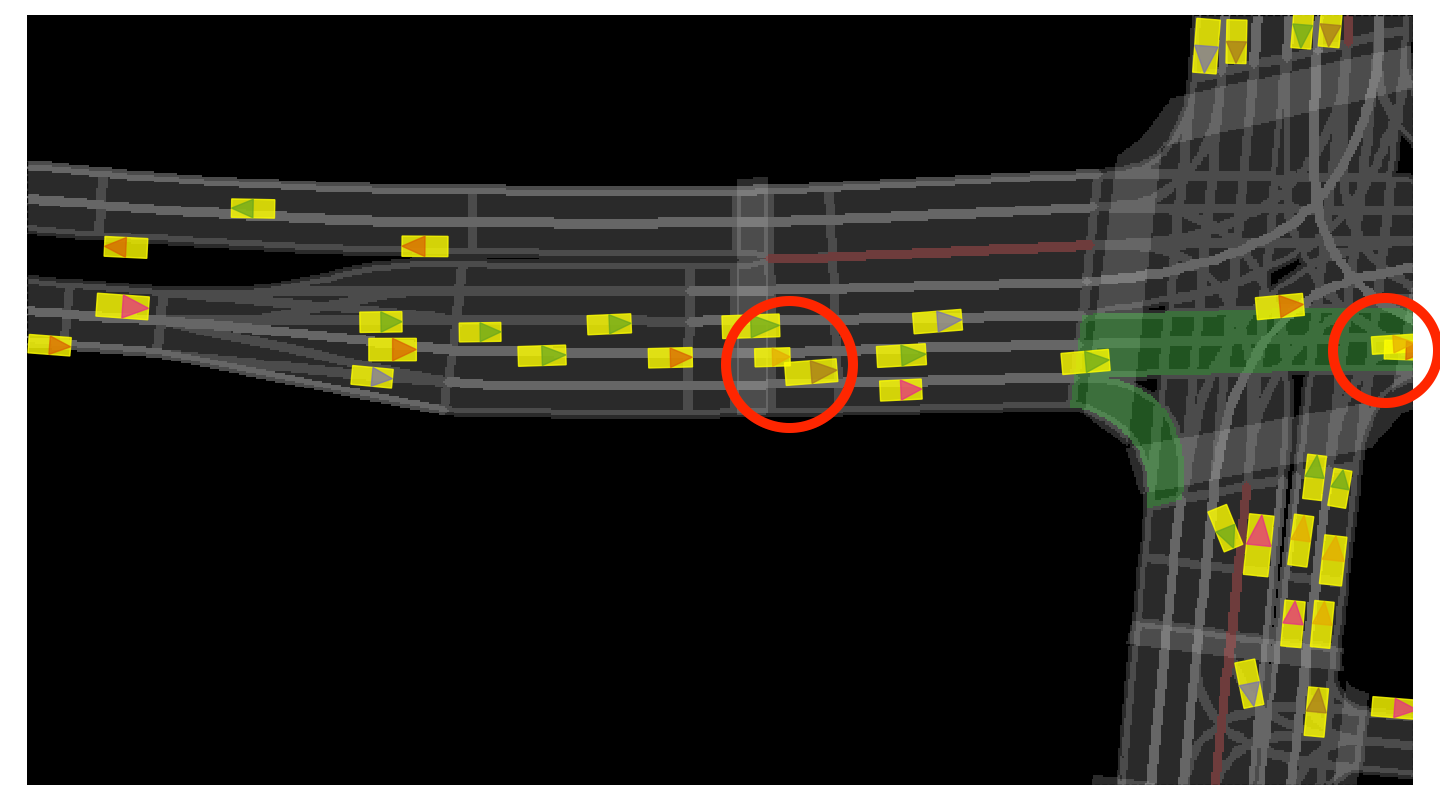}} &
        \raisebox{-0.5\height}{\includegraphics[width=0.325\linewidth, trim={3cm, 3cm, 3cm, 1cm}, clip]{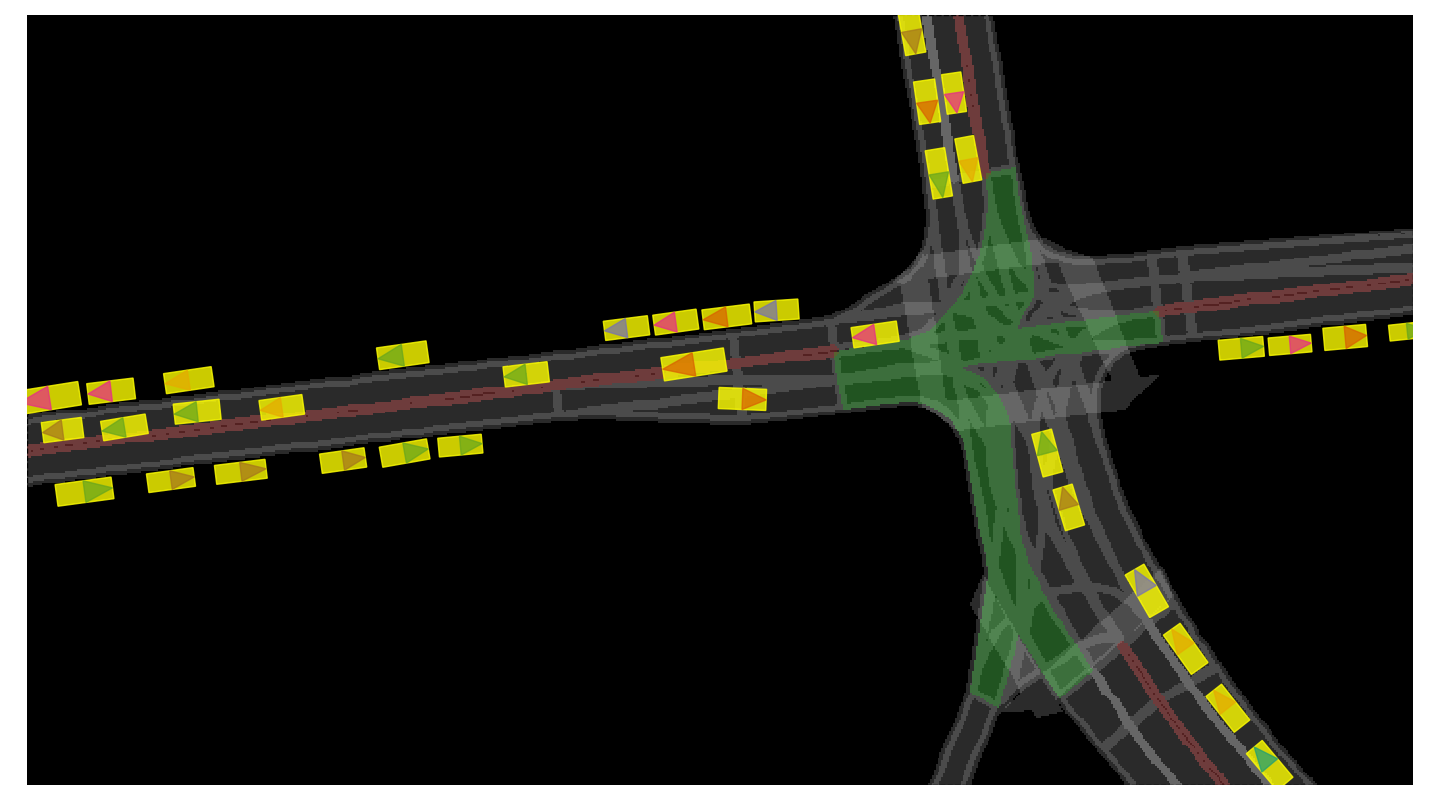}} \vspace{.1em} \\
        \rotatebox[origin=c]{90}{\textbf{AdversarialIL}} &
        \raisebox{-0.5\height}{\includegraphics[width=0.325\linewidth, trim={3cm, 3cm, 3cm, 1cm}, clip]{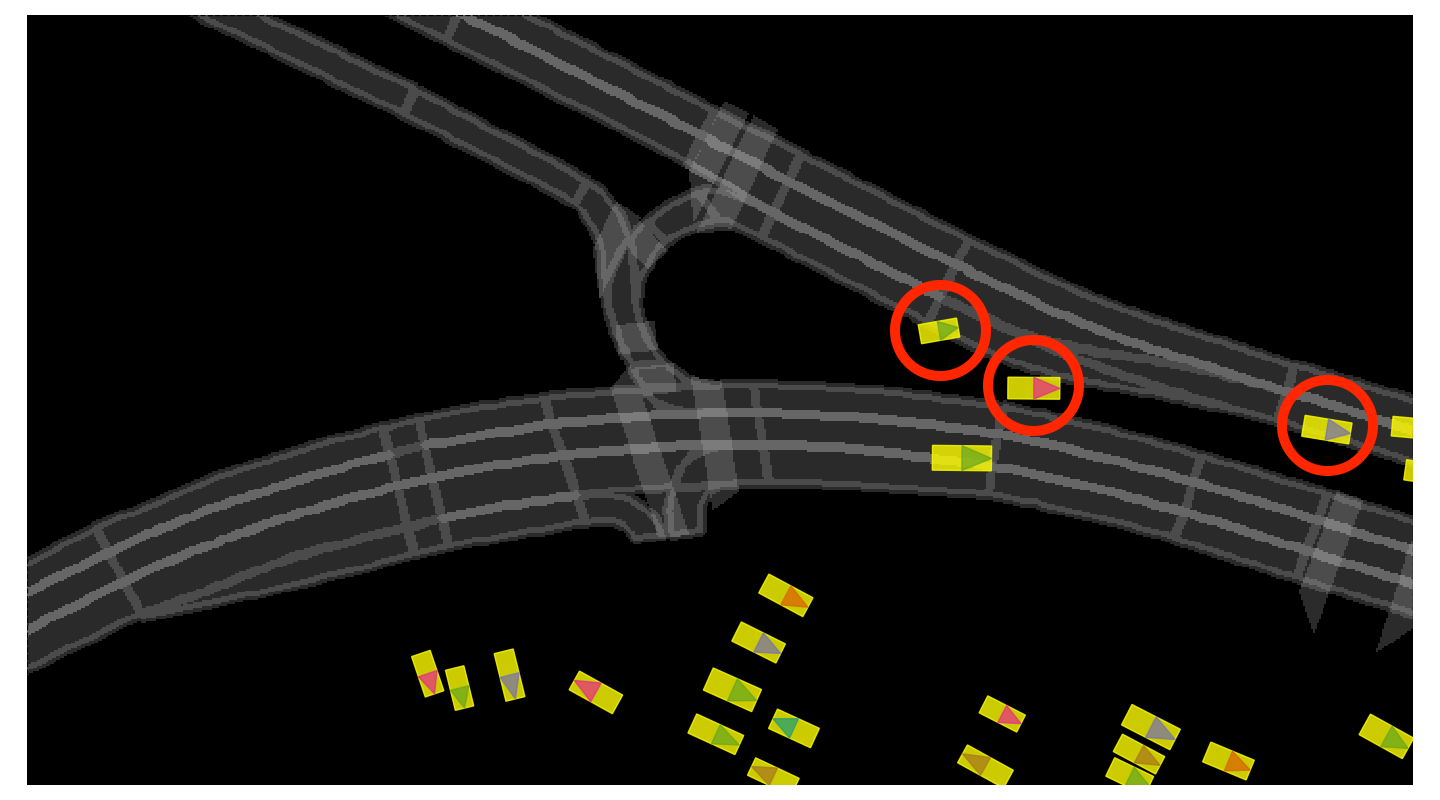}} &
        \raisebox{-0.5\height}{\includegraphics[width=0.325\linewidth, trim={3cm, 3cm, 3cm, 1cm}, clip]{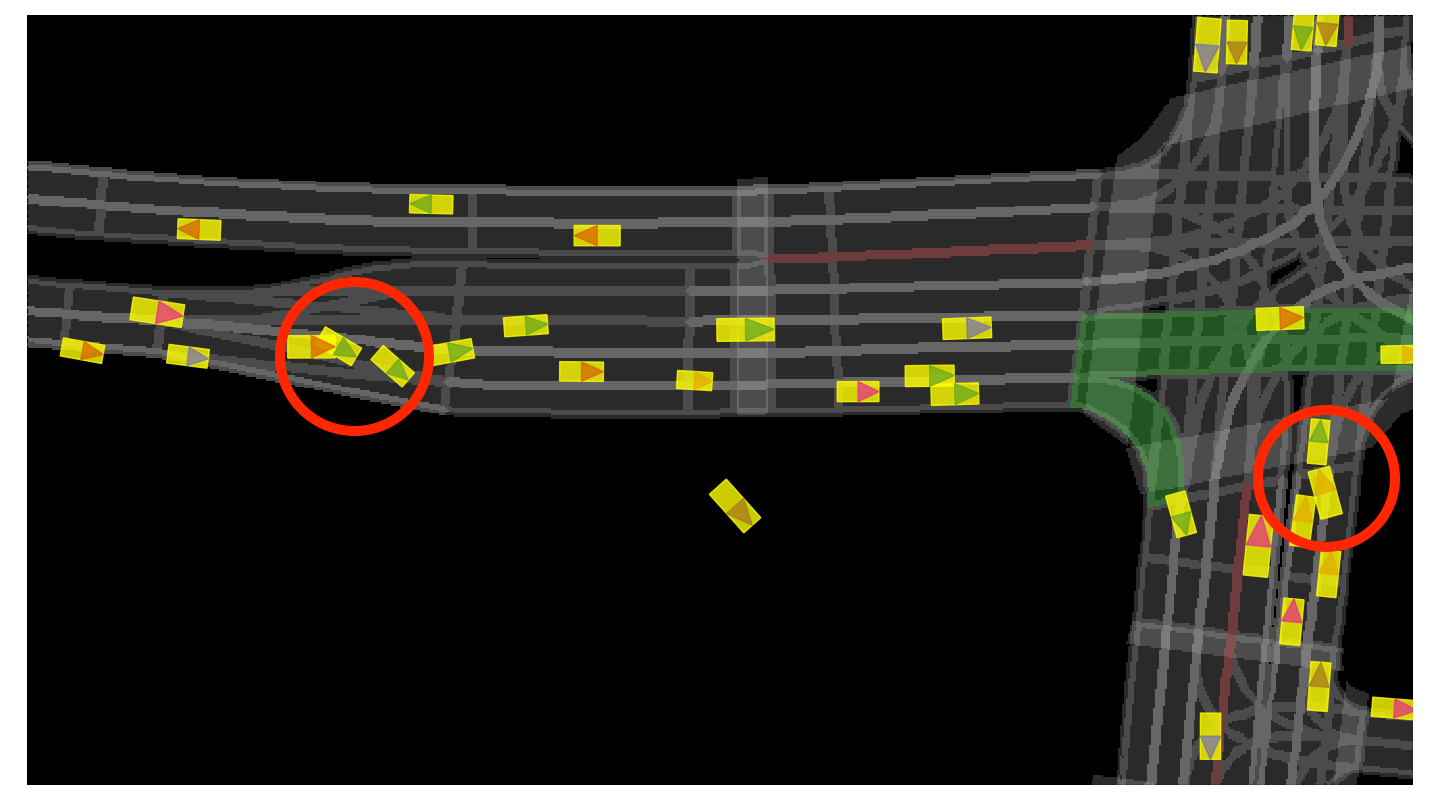}} &
        \raisebox{-0.5\height}{\includegraphics[width=0.325\linewidth, trim={3cm, 3cm, 3cm, 1cm}, clip]{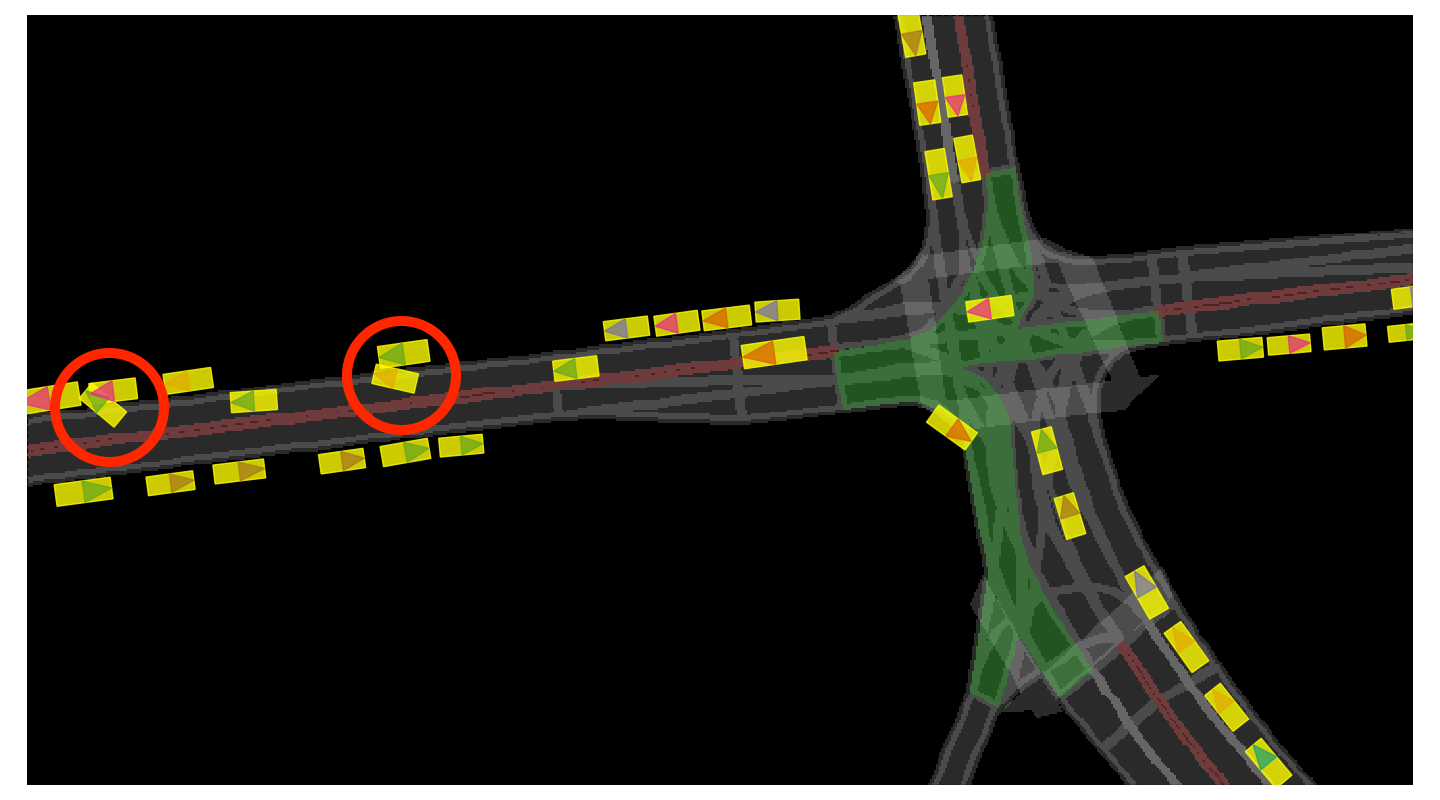}} \vspace{.1em} \\
        \rotatebox[origin=c]{90}{\textbf{Our \ourmodelshort{}}} &
        \raisebox{-0.5\height}{\includegraphics[width=0.325\linewidth, trim={3cm, 3cm, 3cm, 1cm}, clip]{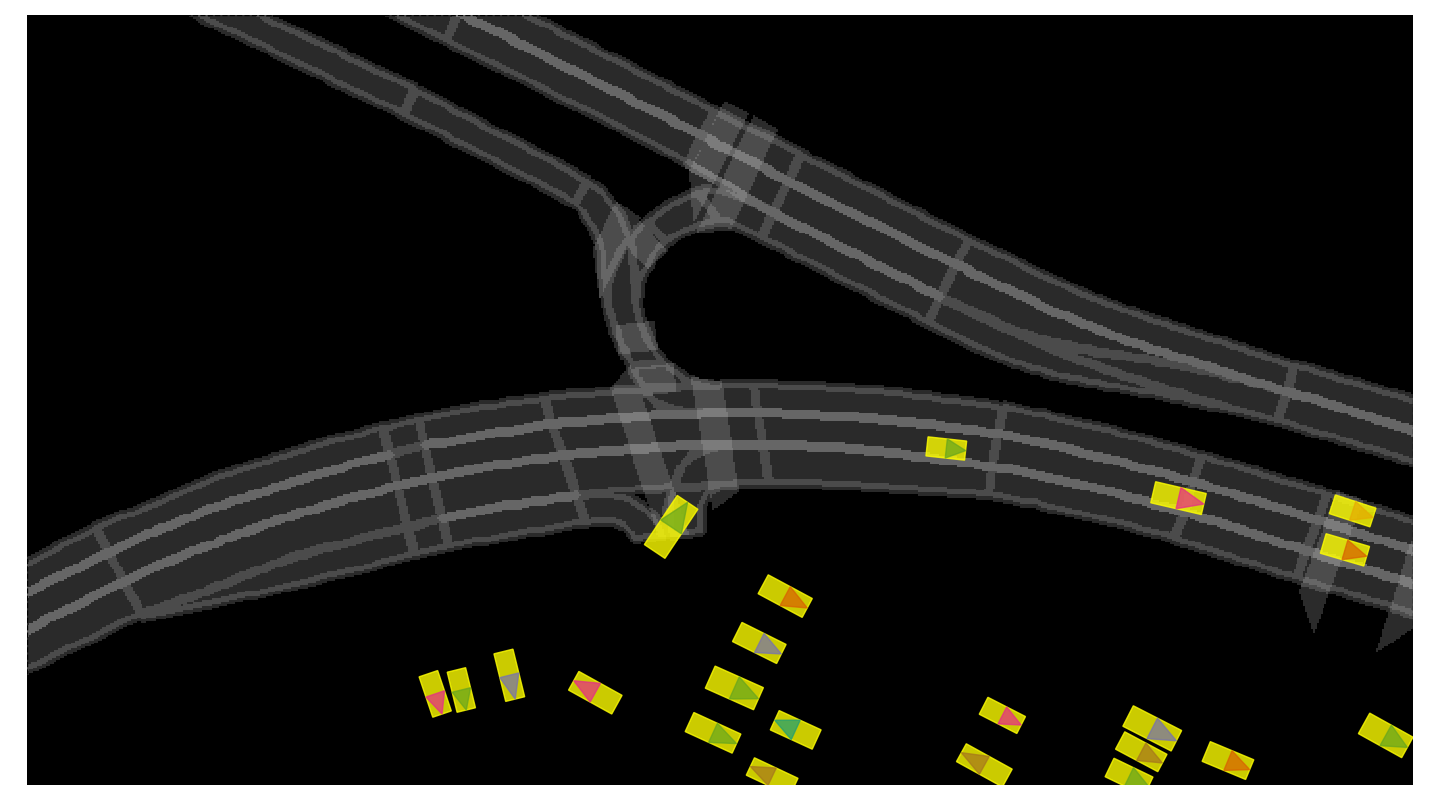}} &
        \raisebox{-0.5\height}{\includegraphics[width=0.325\linewidth, trim={3cm, 3cm, 3cm, 1cm}, clip]{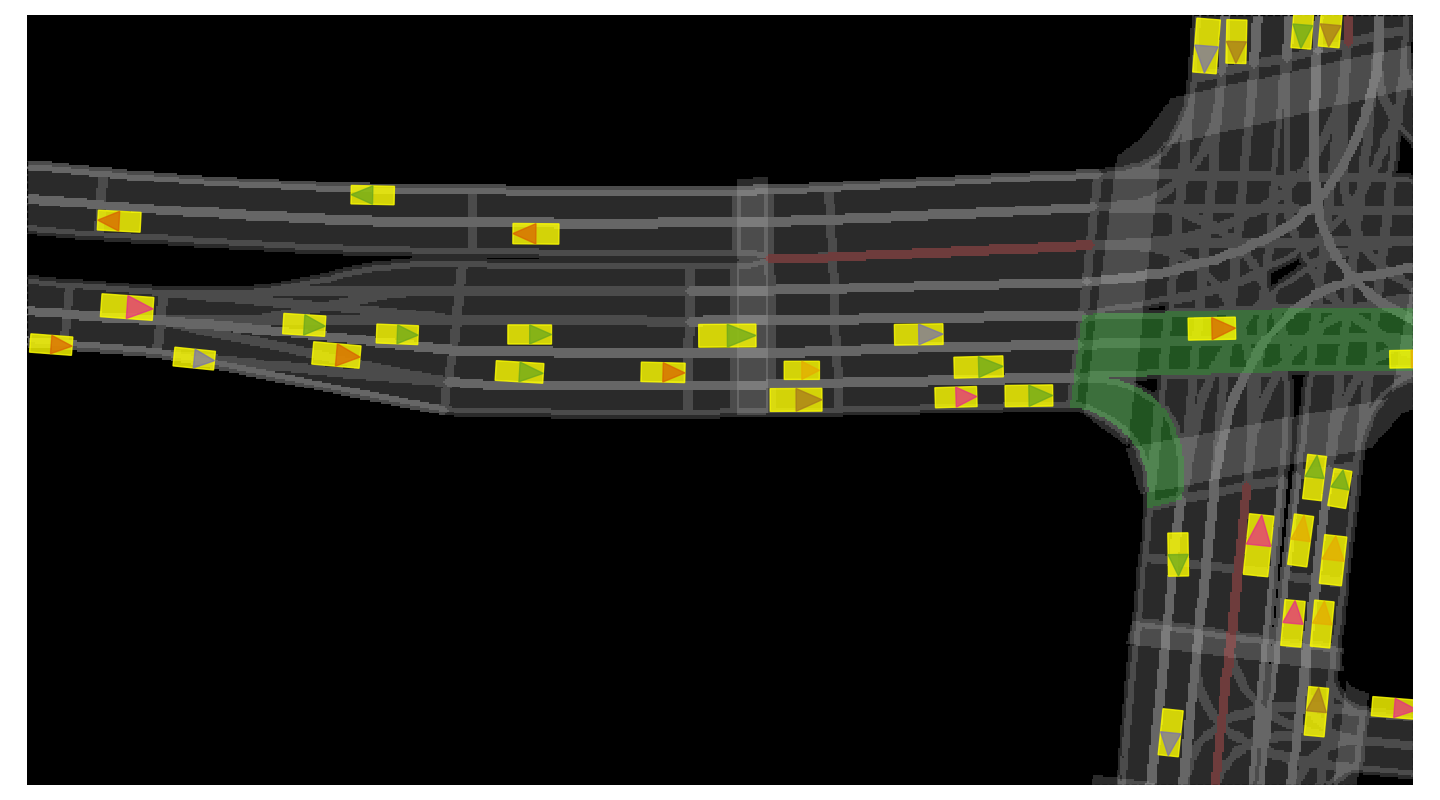}} &
        \raisebox{-0.5\height}{\includegraphics[width=0.325\linewidth, trim={3cm, 3cm, 3cm, 1cm}, clip]{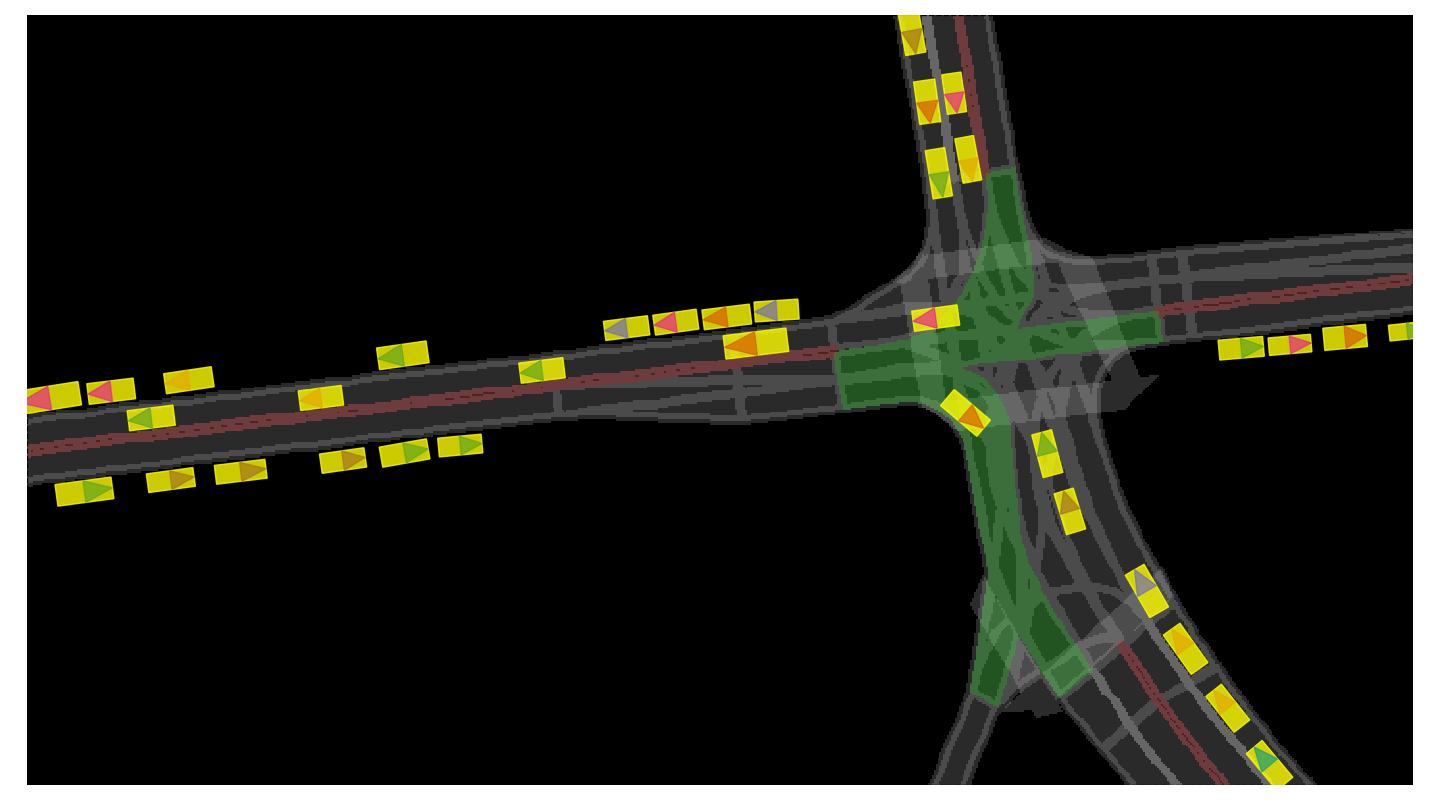}} \\
    \end{tabular}
    \caption{Comparison between traffic scenarios simulated by the baselines and our model. We show a snapshot at 6s after the start of the simulation. Red circles highlight collisions and traffic rule violations}
    \label{fig:supp_qualitative_comparison}
\end{figure*}

\noindent
In this section, we showcase additional qualitative results.
Please refer to the supplementary video for animated sequences.
Figure~\ref{fig:supp_qualitative_our} and \ref{fig:supp_qualitative_our_2} show additional traffic scenarios sampled from our model.
Figure~\ref{fig:supp_qualitative_comparison} shows comparison between baselines and our model.


\typeout{get arXiv to do 4 passes: Label(s) may have changed. Rerun}
\end{document}